\documentclass[10pt,twocolumn,letterpaper]{article}

\usepackage[pagenumbers]{cvpr} % To force page numbers, e.g. for an arXiv version

\usepackage{graphicx}
\usepackage{amsmath}
\usepackage{amssymb}
\usepackage{amsthm}
\usepackage{tabularx}
\usepackage{booktabs}
\usepackage{multirow}
\usepackage{multicol}
\usepackage{enumitem}
\usepackage{array}
\usepackage{color, colortbl}
\usepackage{makecell}
\usepackage{threeparttable}
\usepackage{lipsum}
\usepackage{pifont}
\usepackage{bbm}

\usepackage[pagebackref,breaklinks,colorlinks]{hyperref}

\usepackage[capitalize]{cleveref}
\crefname{section}{Sec.}{Secs.}
\Crefname{section}{Section}{Sections}
\Crefname{table}{Table}{Tables}
\crefname{table}{Tab.}{Tabs.}

\newcommand{\bz}{\mathbf{z}}
\newcommand{\bw}{\mathbf{w}}
\newcommand{\bm}{\mathbf{m}}
\newcommand{\bff}{\mathbf{f}}
\newcommand{\bd}{\mathbf{d}}

\newcommand{\cX}{\mathcal{X}}
\newcommand{\cY}{\mathcal{Y}}
\newcommand{\cZ}{\mathcal{Z}}
\newcommand{\cW}{\mathcal{W}}
\newcommand{\cWp}{\mathcal{W}^{+}}
\newcommand{\IR}{\mathbb{R}}
\newcommand{\norm}[1]{\left\lVert#1\right\rVert}

\def\cvprPaperID{7332} % *** Enter the CVPR Paper ID here
\def\confName{CVPR}
\def\confYear{2022}

\begin{document}

\title{SemanticStyleGAN: Learning Compositional Generative Priors \\for Controllable Image Synthesis and Editing\vspace{-1.0em}}

\author{Yichun Shi \quad\quad  Xiao Yang \quad\quad Yangyue Wan \quad\quad Xiaohui Shen\\[5pt]
{\tt\small \{yichun.shi,yangxiao.0,wanyangyue,shenxiaohui.kevin\}@bytedance.com}\\[5pt]
ByteDance Inc., USA \\[7pt]
{\small\href{https://SemanticStyleGAN.github.io}{https://SemanticStyleGAN.github.io}}
}
% \maketitle

\twocolumn[{%
\captionsetup{font=small}
\renewcommand\twocolumn[1][]{#1}%
\renewcommand{\arraystretch}{0.2}
\newcommand{\mmc}[1]{\multicolumn{1}{c}{#1}}
\maketitle
\thispagestyle{empty}
\vspace{-3.0em}

\begin{center}
\captionsetup{font=small}
\centering
\footnotesize
\setlength\tabcolsep{1px}
\newcommand{\www}{0.121\linewidth}
\renewcommand{\arraystretch}{0.2}
\newcolumntype{Y}{>{\centering\arraybackslash}X}
\begin{tabularx}{\linewidth}{ccccccccc}
    & Coarse Structure & Background & Face & Eyes & Eyebrows & Mouth & Hair \\
    & \includegraphics[width=\www]{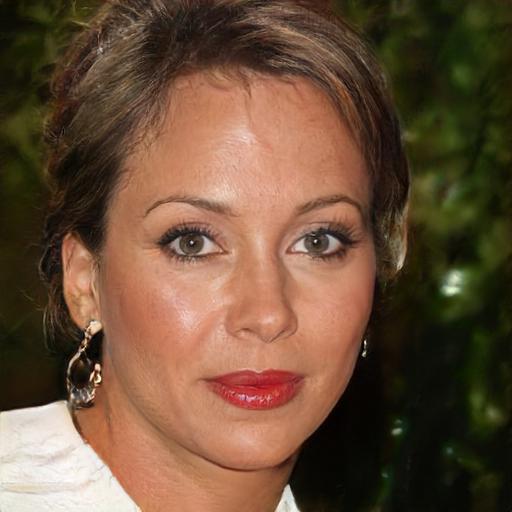}
    & \includegraphics[width=\www]{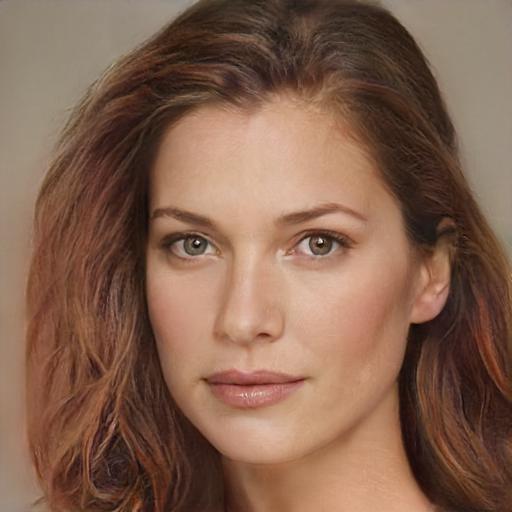}
    & \includegraphics[width=\www]{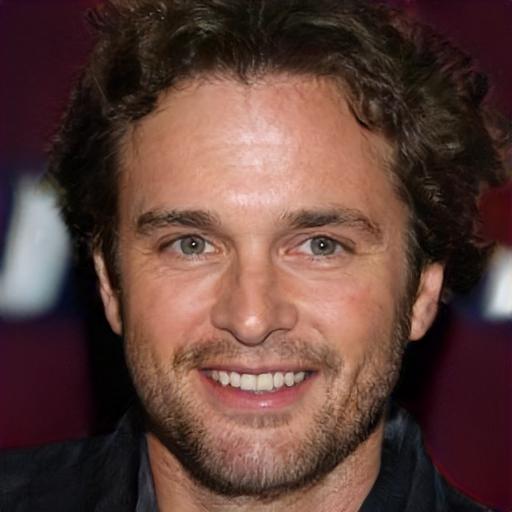}
    & \includegraphics[width=\www]{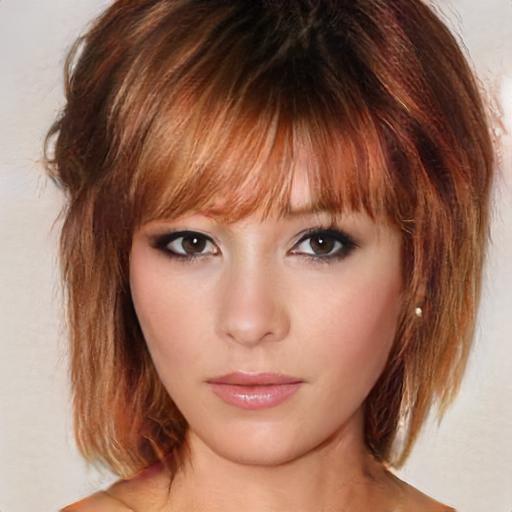}
    & \includegraphics[width=\www]{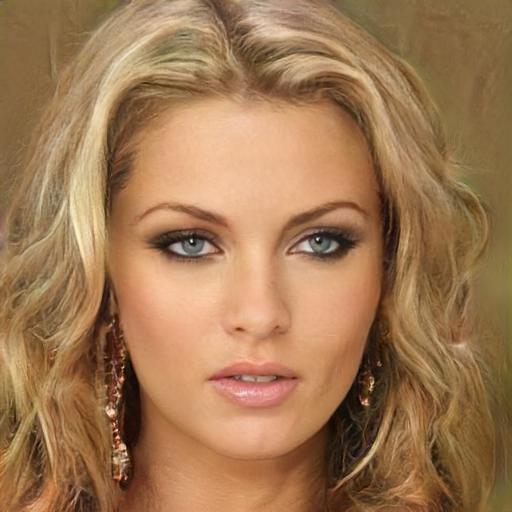}
    & \includegraphics[width=\www]{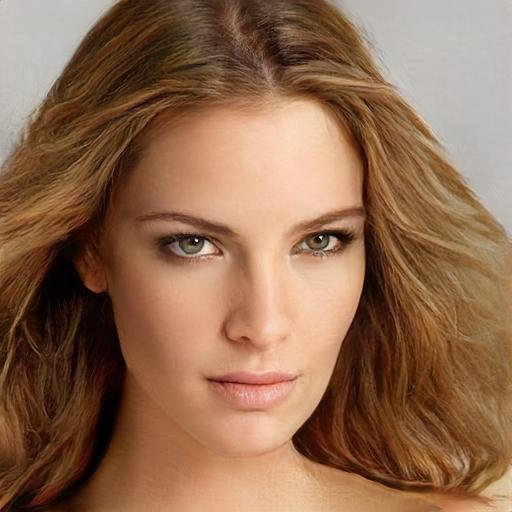}
    & \includegraphics[width=\www]{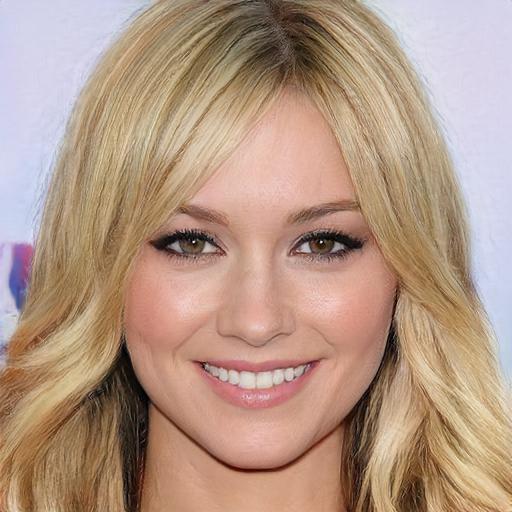} \\
    \includegraphics[width=\www]{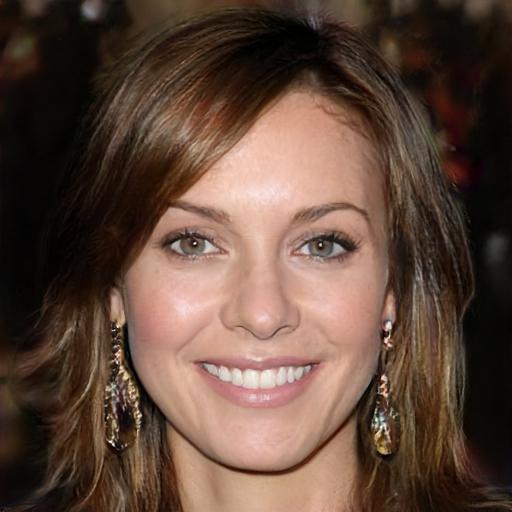}
    & \includegraphics[width=\www]{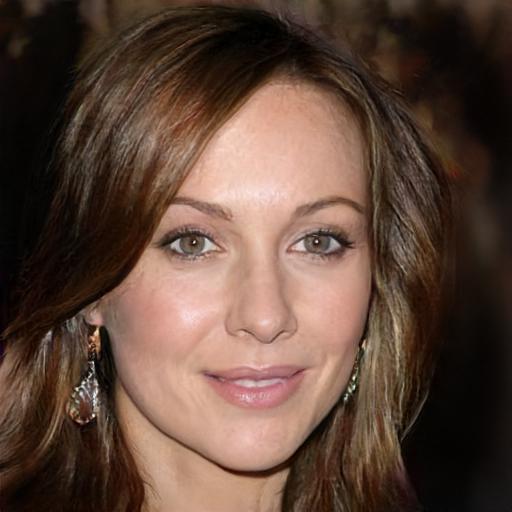}
    & \includegraphics[width=\www]{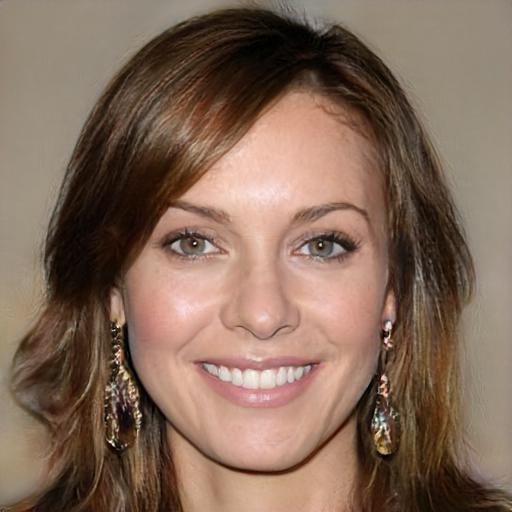}
    & \includegraphics[width=\www]{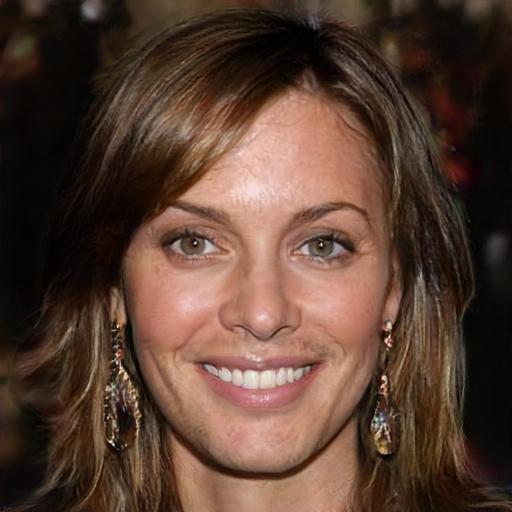}
    & \includegraphics[width=\www]{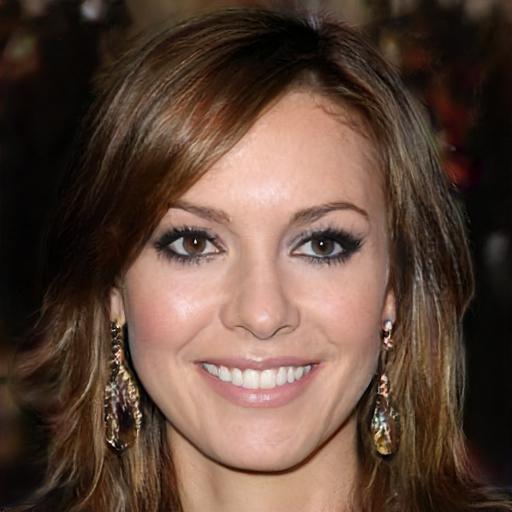}
    & \includegraphics[width=\www]{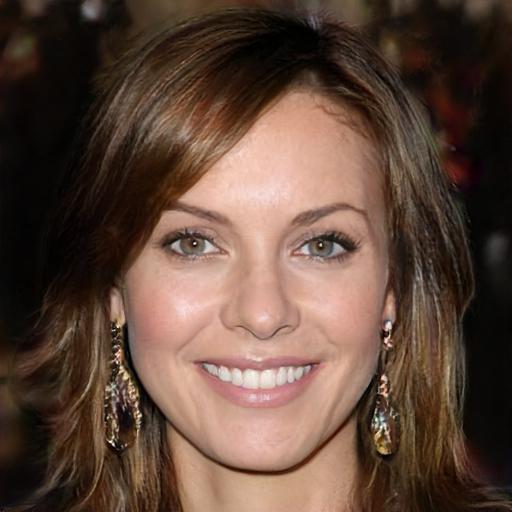}
    & \includegraphics[width=\www]{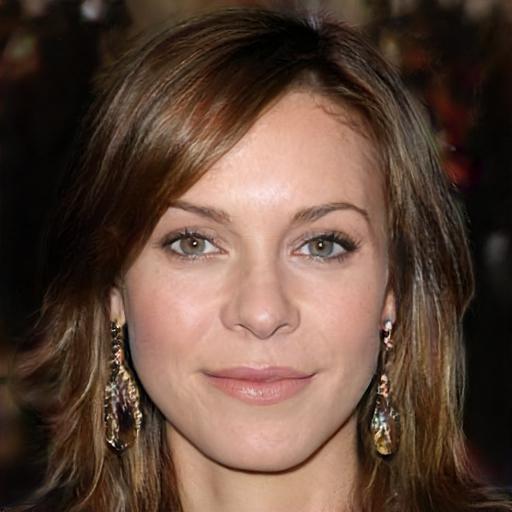}
    & \includegraphics[width=\www]{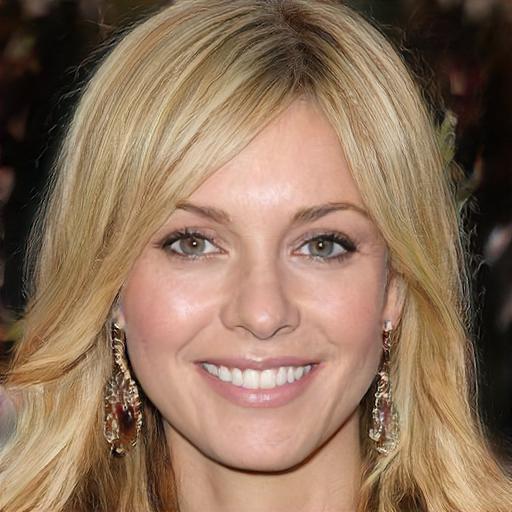} \\
    \includegraphics[width=\www]{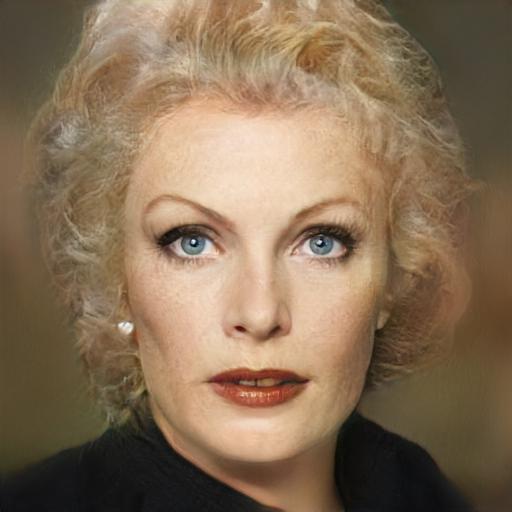}
    & \includegraphics[width=\www]{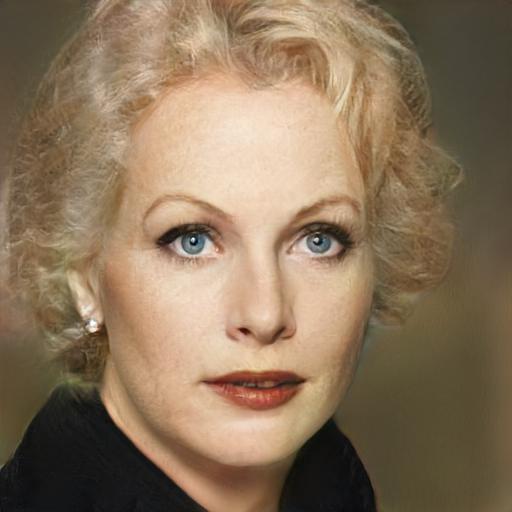}
    & \includegraphics[width=\www]{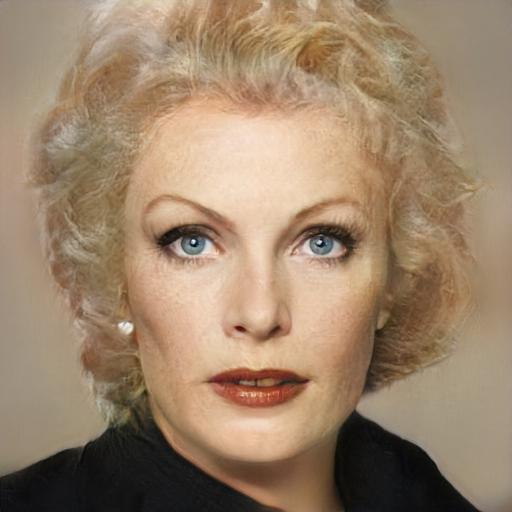}
    & \includegraphics[width=\www]{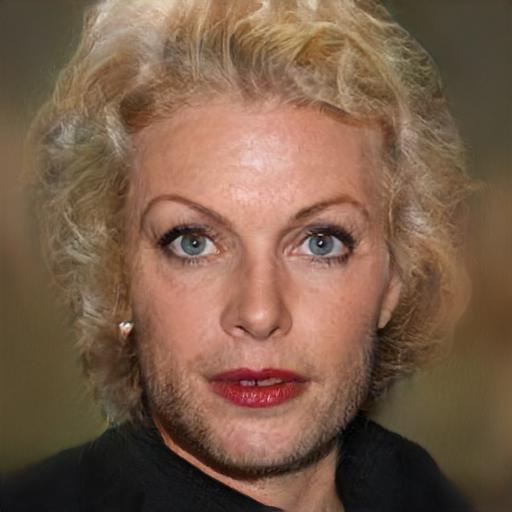}
    & \includegraphics[width=\www]{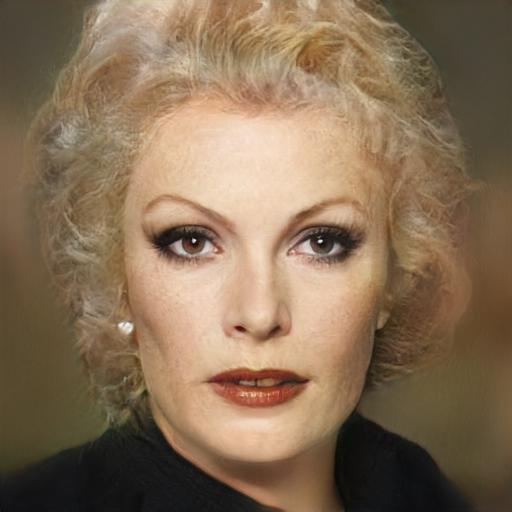}
    & \includegraphics[width=\www]{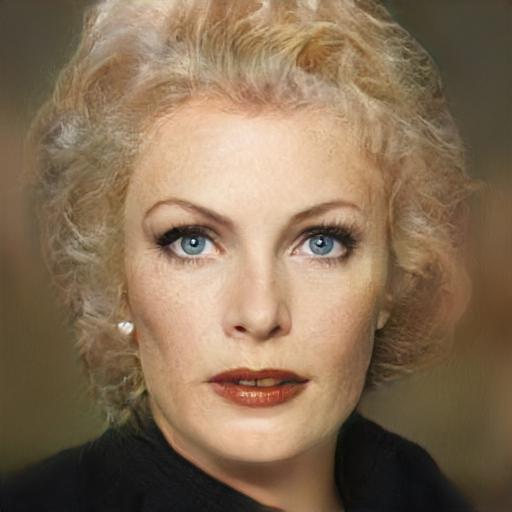}
    & \includegraphics[width=\www]{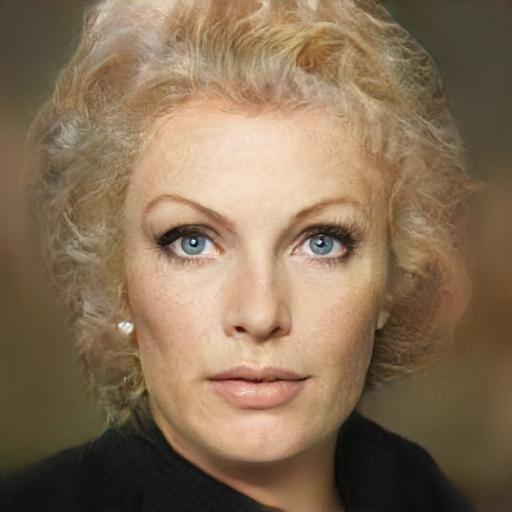}
    & \includegraphics[width=\www]{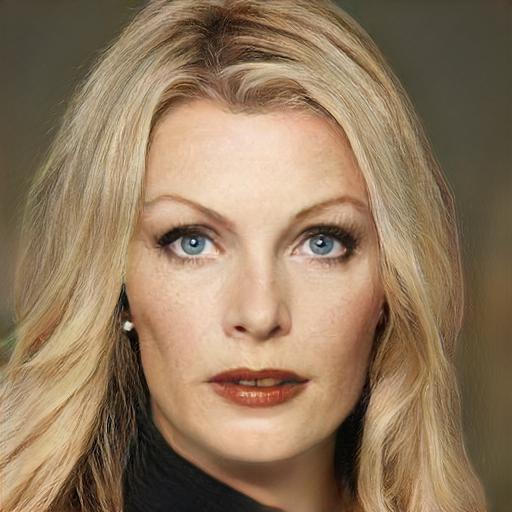} \\
    \includegraphics[width=\www]{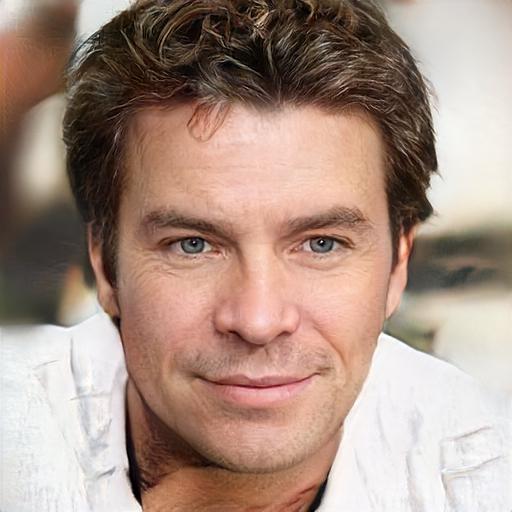}
    & \includegraphics[width=\www]{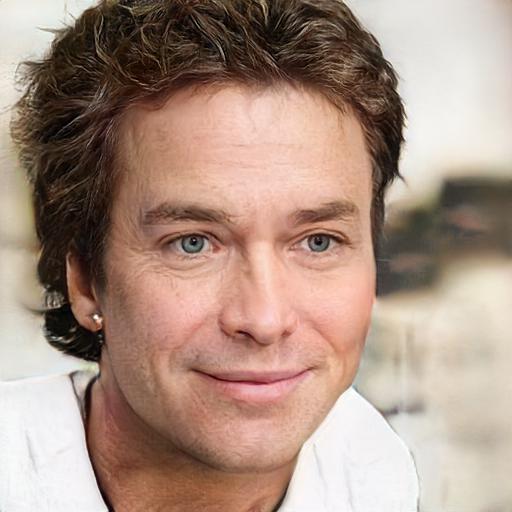}
    & \includegraphics[width=\www]{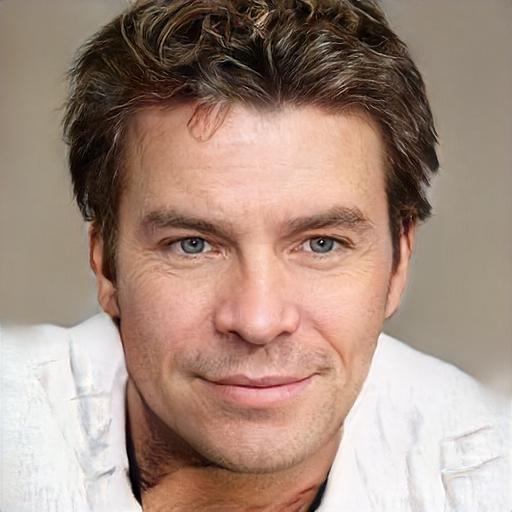}
    & \includegraphics[width=\www]{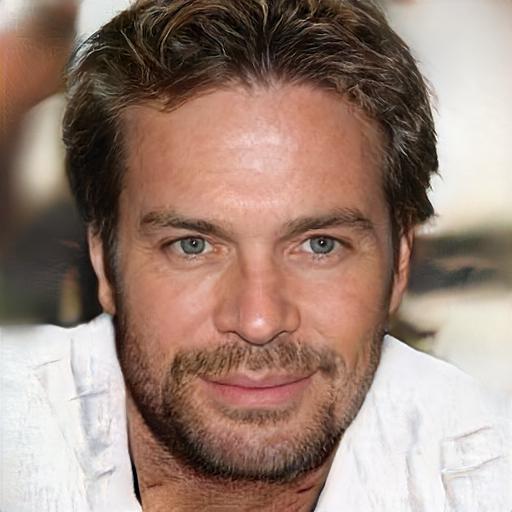}
    & \includegraphics[width=\www]{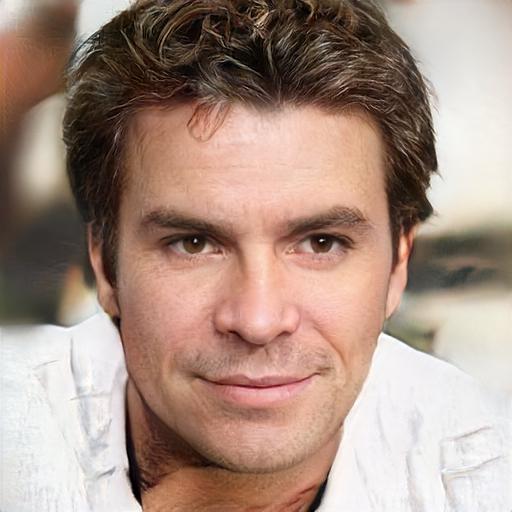}
    & \includegraphics[width=\www]{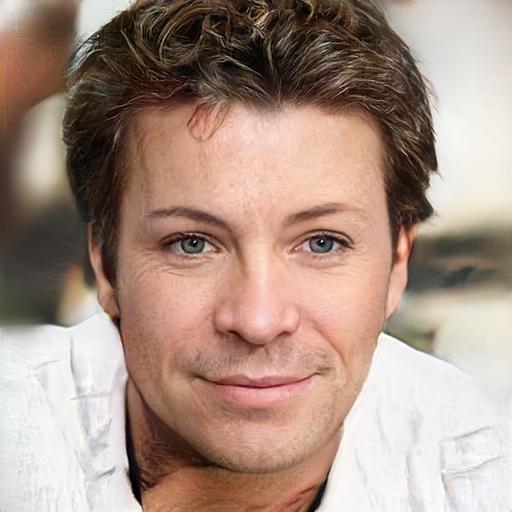}
    & \includegraphics[width=\www]{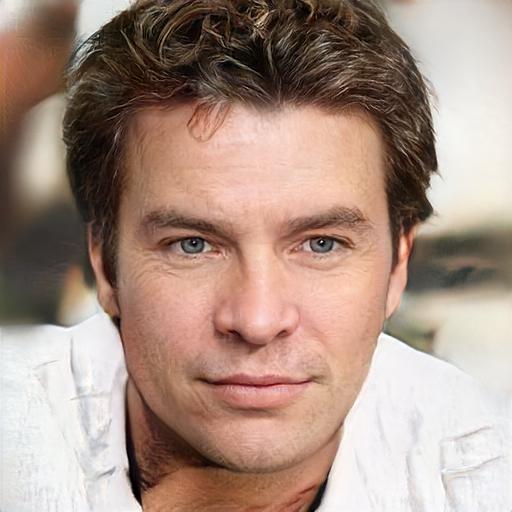}
    & \includegraphics[width=\www]{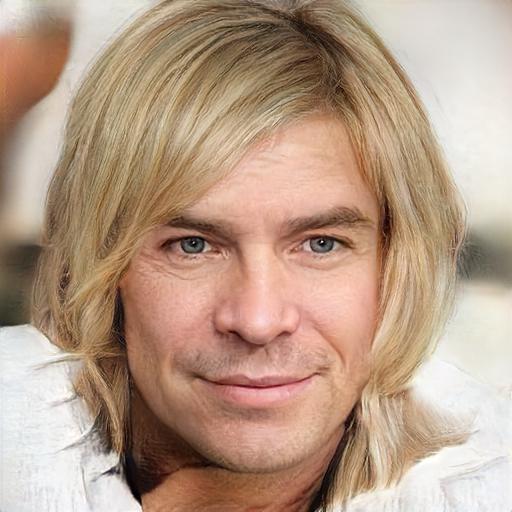} \\
    % \includegraphics[width=\www]{fig_appendix/stylemix/probes/probe_09.jpg}
    % & \includegraphics[width=\www]{fig_appendix/stylemix/coarse/09.jpg}
    % & \includegraphics[width=\www]{fig_appendix/stylemix/background/09.jpg}
    % & \includegraphics[width=\www]{fig_appendix/stylemix/face/09.jpg}
    % & \includegraphics[width=\www]{fig_appendix/stylemix/eye/09.jpg}
    % & \includegraphics[width=\www]{fig_appendix/stylemix/eyebrow/09.jpg}
    % & \includegraphics[width=\www]{fig_appendix/stylemix/mouth/09.jpg}
    % & \includegraphics[width=\www]{fig_appendix/stylemix/hair/09.jpg} \\
\end{tabularx}
    \vspace{-1.0em}\captionof{figure}{SemanticStyleGAN factorizes its latent space based on semantic regions. Here, we show the results of style mixing by swapping local latent codes. Note that our models also disentangles shape and texture but we are simultaneously changing both here. }\vspace{-1.0em}
    \label{fig:stylemix_frontal}
\end{center}
}]

%%%%%%%%% ABSTRACT
\begin{abstract}
\vspace{-1.5em}
Recent studies have shown that StyleGANs provide promising prior models for downstream tasks on image synthesis and editing. However, since the latent codes of StyleGANs are designed to control global styles, it is hard to achieve a fine-grained control over synthesized images. We present SemanticStyleGAN, where a generator is trained to model local semantic parts separately and synthesizes images in a compositional way. The structure and texture of different local parts are controlled by corresponding latent codes. Experimental results demonstrate that our model provides a strong disentanglement between different spatial areas. When combined with editing methods designed for StyleGANs, it can achieve a more fine-grained control to edit synthesized or real images. The model can also be extended to other domains via transfer learning. Thus, as a generic prior model with built-in disentanglement, it could facilitate the development of GAN-based applications and enable more potential downstream tasks.
% \footnote{Visit \href{https://SemanticStyleGAN.github.io}{https://SemanticStyleGAN.github.io} for more information.}
\end{abstract}
\vspace{-1.3em}

%%%%%%%%% BODY TEXT

\section{Introduction}
\label{sec:intro}
\vspace{-0.2em}
Recent studies on Generative Adversarial Networks (GANs) have made impressive progress on image synthesis, where photo-realistic images can be generated from random codes in a latent space~\cite{BigGAN,stylegan,stylegan2,karras2021alias}. These models provide powerful generative priors for downstream tasks by serving as neural renderers. However, their synthesis procedure is usually stochastic and no user control is naturally promised. Thus, it is still a challenging problem to achieve controllable image synthesis and editing utilizing generative priors.

\begin{figure}
    \centering
    \subfloat[StyleGAN]{\includegraphics[width=0.49\linewidth]{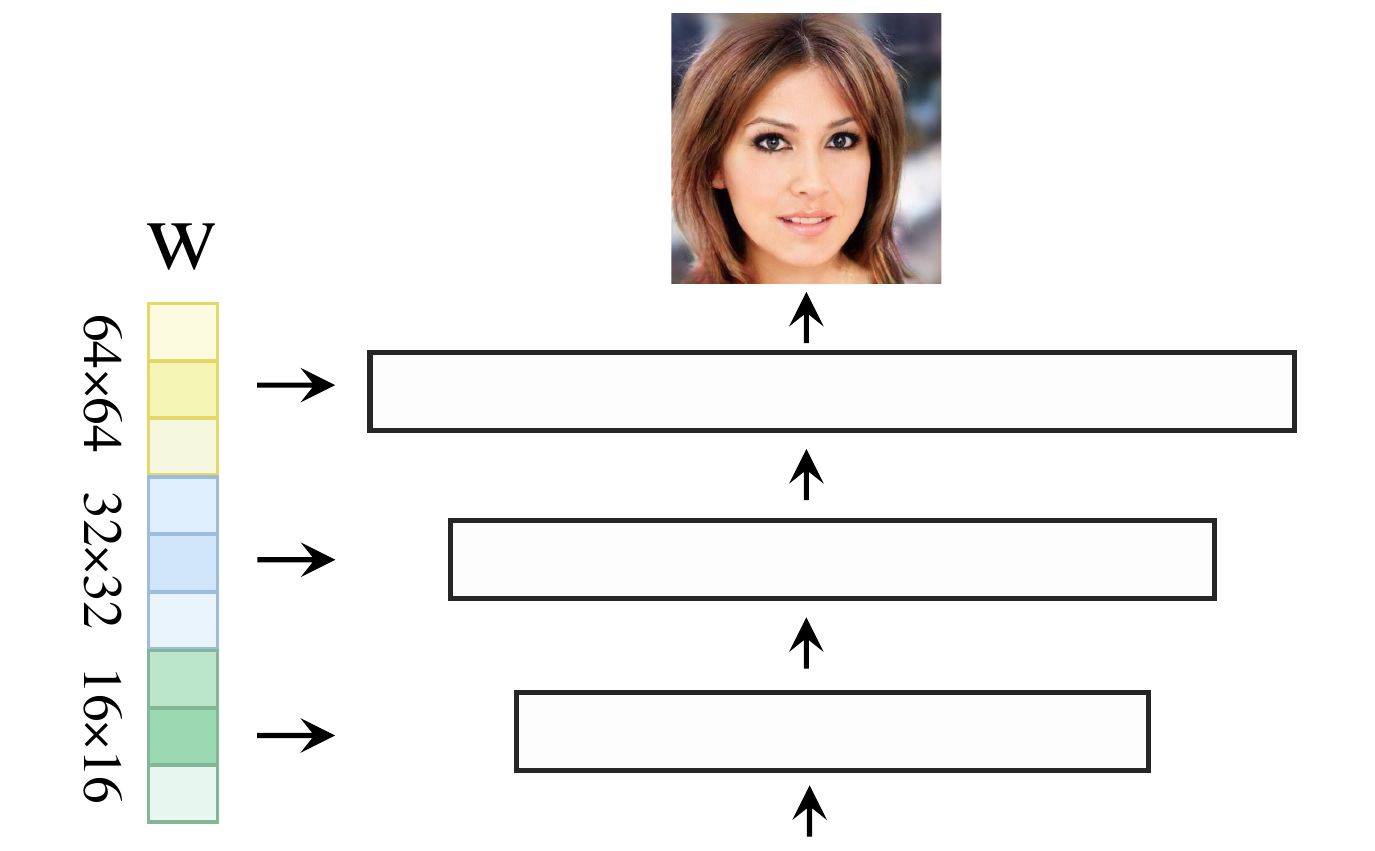}}\hfill
    \subfloat[SemanticStyleGAN]{\includegraphics[width=0.49\linewidth]{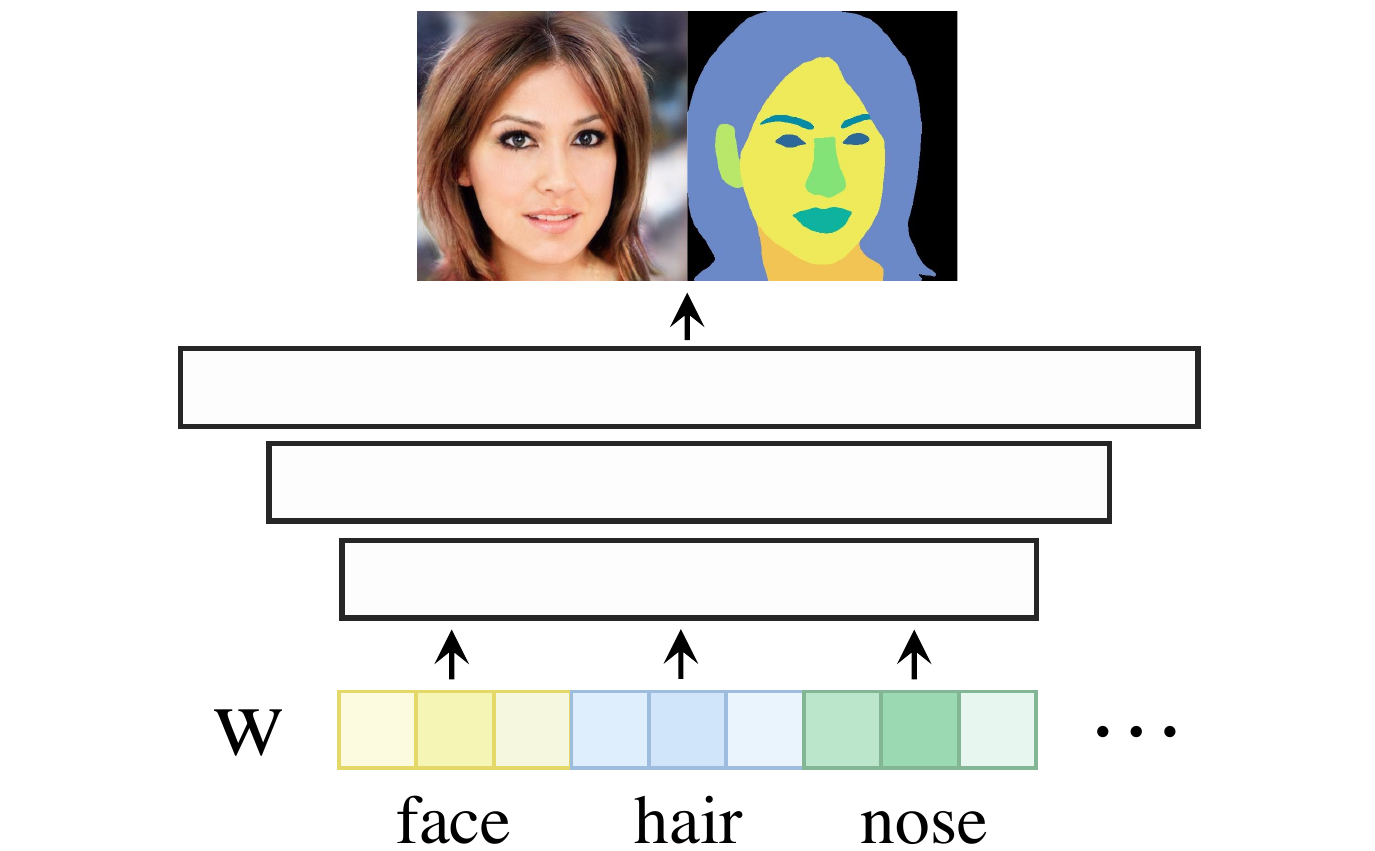}}\\
    \vspace{-0.8em}\caption{Abstract illustration of our method. Unlike StyleGAN, whose latent codes are associated with different granularity. The latent space of SemanticStyleGAN is factorized over different regions, which controls both local shape and texture.}
    \label{fig:abstract_idea}\vspace{-1.5em}
\end{figure}

One of the most famous work among such generative priors is the StyleGAN series~\cite{stylegan,stylegan2,karras2021alias}, where each generated image is conditioned on a set of coarse-to-fine latent codes (See \cref{fig:abstract_idea}). However, the meanings of these latent codes are still relatively ambiguous. Thus, a plethora of studies have attempted to further investigate into the latent space of StyleGAN to improve controllability. It is shown that by learning a linear boundary or a neural network in the latent space of StyleGAN, one could control the global attributes~\cite{shen2020interpreting,shen2020closed,harkonen2020ganspace,abdal2021styleflow} or 3D structure~\cite{tewari2020stylerig} of the generated images. Furthermore, by using an optimization/encoder-based method, real images can also be embedded into the latent space to create a unified synthesis/editing model~\cite{abdal2019image2stylegan,abdal2020image2stylegan++,zhu2020indomain,tov2021e4e,richardson2021psp,alaluf2021restyle,wang2021high}. However, as pure learning-based methods, these solutions inevitability suffer from the biases in the StyleGAN latent space. For example, since different attributes could be correlated in StyleGAN, it often happens that unexpected attributes or local parts are changed while one wants to edit a certain attribute or area. 

To obtain a more precise control, another solution is to train a new GAN model from scratch by introducing additional supervision or inductive biases. For example, by using 3D rendered faces, CONFIG~\cite{kowalski2020config} and DiscoFaceGAN~\cite{deng2020discofacegan} aim to build a GAN where pose, 3D information are factorized in the latent space. GAN-Control~\cite{shoshan2021gancontrol} disentangles the latent space by incorporating pre-trained attribute models for contrastive learning. Given the recent progress on neural rendering, it has also been shown that 3D-controllable GANs can be trained from images by injecting volumetric rendering into the synthesis procedure~\cite{nguyen2019hologan,schwarz2020graf,chan2021pigan,gu2021stylenerf,zhou2021cips3d}. However, a major limitation of above-mentioned models is that they are designed for holistic attributes and there is no fine-grained local editability.

% Built upon layout-to-image translation models~\cite{wang2018Pix2PixHD,park2019SPADE}, SEAN~\cite{zhu2020SEAN} proposed to use a style-based generator to synthesize images from semantic maps. The resulting model is able to edit local texture of different semantic parts without affecting irrelevant areas. However, in order to change the local structure of each area, one has to manually edit the input segmentation map, which is intractable for automatic systems.

In this work, we propose SemanticStyleGAN, which introduces a new type of generative prior for controllable image synthesis. Unlike prior work, the latent space of SemanticStyleGAN is factorized based on semantic parts defined by semantic segmentation masks (\cref{fig:abstract_idea} (b)). Each semantic part is modulated individually with corresponding local latent codes and an image is synthesized by composing local feature maps. Different from layout-to-image translation methods~\cite{zhu2020SEAN,wang2021image,chen2021sofgan}, our local latent codes are able to control both the structure and texture of semantic parts (See~\cref{fig:stylemix_frontal}). Compared to attribute-conditional GANs~\cite{kowalski2020config,deng2020discofacegan,shoshan2021gancontrol}, our model is not designed for any specific task and can serve as a generic prior like StyleGAN.
% Inspired by recent work on compositional generative models~\cite{greff2019IODINE,burgess2019monet,niemeyer2021giraffe}, we introduce semantic segmentations as a supervision during the training of generator. The generator then synthesizes an image by composing feature maps of different semantic parts, whose latent styles are defined independently. Thus, our model provides full controllability on each local area. As shown in \cref{fig:cover},  one could could traverse through the structure code of hair to have different hairstyles or change the texture code to arrive at a different color without changing other parts of the face. Similar to StyleGAN itself, our model does not directly serve a specific target application. 
Thus, it can be combined with latent manipulation methods designed for StyleGAN to edit output images while providing more precise local controls. The contributions of this work can be summarized as follows:
\begin{itemize}\vspace{-0.5em}
    \item A compositional generator architecture that disentangles the latent space into different semantic areas to control the structure and texture of local parts.\vspace{-0.5em}
    \item A GAN training framework that learns the joint modeling of image and semantic segmentation masks.\vspace{-0.5em}
    \item Experiments showing that our generator can be combined with existing latent manipulation methods to edit images in a more controllable fashion.\vspace{-0.5em}
    \item Experiments showing that our generator can be adapted to other domains with only limited images while preserving spatial disentanglement.\vspace{-0.5em}
\end{itemize}

\section{Related Work}
\subsection{GAN Latent Space for Image Editing}
Given the success of GANs on synthesizing high quality images~\cite{stylegan,stylegan2,BigGAN}, many studies have attempted to utilize GANs as a image prior to achieve controllable image synthesis and editing. These studies can be categorized into two types. The first type aims to learn a model to manipulate the latent space of a pre-trained GAN network to achieve editability. For example, InterFaceGAN~\cite{shen2020interpreting}, GANSpace~\cite{harkonen2020ganspace} and StyleFlow~\cite{abdal2021styleflow} trains a attribute model in the StyleGAN latent space to control binary attributes. StyleRig~\cite{tewari2020stylerig} learns a set of latent space networks to change the pose and lighting. Similarly, StyleFusion~\cite{kafri2021stylefusion} learns to fuse semantic parts from different images in the latent space. The second type aims to learn a GAN with more disentangled latent space using additional supervision. For example, CONFIG~\cite{kowalski2020config} and DiscoFaceGAN~\cite{deng2020discofacegan} uses 3D-rendered data to disentangle pose, identity, expression from other information. GAN-Control~\cite{shoshan2021gancontrol} separates attributes like identity and age in the latent space by utilizing pre-trained attribute models. Besides these, StyleMapGAN~\cite{kim2021stylemapgan} propose to use style maps to modulate a synthesis network, but the meaning of each style pixel is unclear. Different from prior works, we propose a new type of factorization in the GAN latent space according to semantic labels. Our disentangled latent codes could independently control the shape and texture of each semantic part in the output image.

\begin{figure*}
    \centering
    \includegraphics[width=\linewidth]{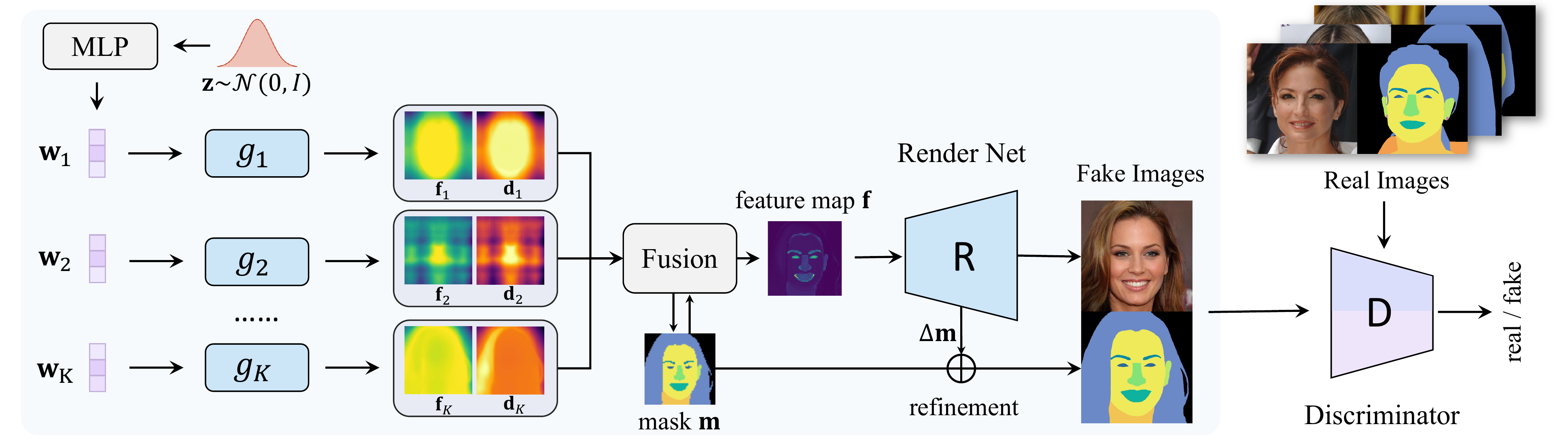}
    \vspace{-1.8em}\caption{Overview of our training framework. A MLP first maps randomly sampled codes into $\cW$ space. The $\bw$ code is used to modulate the weights of local generators. Each local generator $g_k$ outputs a feature map $\bff_k$ and a pseudo-depth map $\bd_k$, which are fused into a coarse segmentation mask $\bm$ and a global feature map $\bff$ for image synthesis. The render network $R$, which is only conditioned on the feature map, refines upsampled $\bm$ into a high-resolution segmentation mask by learning a residual $\Delta \bm$ and generates the fake image. A dual-branch discriminator models the joint distribution of RGB images and semantic segmentation masks.}
    \label{fig:overview}\vspace{-1.0em}
\end{figure*}

\subsection{Compositional Image Synthesis}
A plethora of studies have investigated how to build generative models to mimic the compositional nature of the world. 
% These work generally focus on object-centric generation, where object-level instances can be discovered, generated and manipulated independently in the generation process. 
To achieve compositionality, some studies propose to take images as input and compose a complicated scene with elements from real images~\cite{arandjelovic2019object,azadi2020compositional,sbai2021surprising}. 
%These methods cannot generate fake images from scratch. 
On the other side, the majority studies aim to build a generative model that unsupervisedly discovered different objects in the training images and then synthesize them from independent latent codes. Most of these methods assume that objects are positioned independently in the scene and a compositional generative model is designed to discover such objects~\cite{gregor2015draw,eslami2016attend,yang2017lrgan,greff2019IODINE,burgess2019monet,anciukevicius2020object,yang2020learning,ehrhardt2020relate,van2020investigating,hudson2021compositional}. Some other methods approach the compositional synthesis from a 3D perspective and disentangles objects and background by leraning multi-view datasets~\cite{nanbo2020learning,henderson2020unsupervised,nguyen2020blockgan,niemeyer2021giraffe}. Similar to these work, we inject composition as an inductive-bias to encourage disentanglement. However, we focus on semantic parts that are defined by humans. This allows us to decompose highly correlated local parts below object level (e.g. hair and face) and enables more fine-grained control during synthesis.

\subsection{Layout-based Generators for Local Editing}
In the layout-to-image translation problem, a layout image is provided as the condition for controllable image synthesis. The layout image can be a semantic segmentation mask~\cite{chen2017CRN,qi2018SIMS,wang2018Pix2PixHD,park2019SPADE,liu2019learning,zhu2020SEAN,zhu2020semantically,wang2021image,chen2021sofgan}, a sketch image~\cite{wang2018Pix2PixHD,chen2020deepfacedrawing,richardson2021psp}, etc. Among these, some studies have attempted to represent different semantic parts with latent codes~\cite{zhu2020SEAN,zhu2020semantically,chen2021sofgan}. But since the layout is controlled by the input segmentation mask, they are only able to control the local texture. Our method also shares similarity with prior research that utilizes semantic masks as intermediate representations for generation~\cite{hong2018inferring,johnson2018image,ashual2019specifying}, but they are engineered to serve conditional generation tasks and not able to generate images from scratch. Recently, some researchers have also analyzed the correlation between StyleGAN style space and semantic masks~\cite{collins2020editing,wu2021stylespace,kafri2021stylefusion} or supervise the latent manipulation with segmentation masks ~\cite{futschik2021real,ling2021editgan,pernuvs2021high} to achieve local editing. In contrast to these methods, we build a semantic-aware generator that directly associates different local areas with latent codes, these codes can then be used to edit both local structure and texture.

\section{Methodology}
A typical GAN framework learns a generator that maps a vector $\bz\sim \cZ$ to an image, where $\cZ$ is usually a standard normal distribution. In StyleGANs~\cite{stylegan,stylegan2}, to handle the non-linearity of data distribution, $\bz$ is first mapped into a latent code $\bw~\sim \cW$ with an MLP. This $\cW$ space is then extended into a $\cWp$ space that controls the output styles at different resolutions~\cite{stylegan}. However, these latent codes do not have a strictly defined meaning and can hardly be used individually. 

We propose to build a generator whose $\cWp$ space is disentangled for different semantic areas. Formally, given a labeled dataset $D=\{(x_1,y_1),(x_2,y_2)...,(x_n,y_n)\}$, where $y_i\in\{0,1\}^{H\times W\times K}$ is the semantic segmentation mask of image $x_i$ and $K$ is the number of semantic classes, our generator gives a factorized $\cWp$ such that:
\vspace{-0.7em}\begin{equation}\vspace{-0.3em}
    \cWp = \cW^{\text{base}} \times \cW^{1} \times \cW^{2} \times ... \times \cW^{K}.
\end{equation}
Here each local latent code $\bw^k\in\cW^k$ controls the shape and texture of $k_{th}$ semantic area defined in segmentation labels while $\bw^{\text{base}}\in\cW^{\text{base}}$ is a shared code that controls the coarse structure, such as pose. Each $\bw^k$ is further decomposed into a shape code $\bw^k_s$ and a texture code $\bw^k_t$. The generator $G:\cWp\rightarrow \cX\times \cY$ maps the latent codes to an RGB image and a semantic segmentation mask. To this end, we identify two major challenges:
\begin{enumerate}[leftmargin=15pt]\vspace{-0.4em}
    \item How to decouple different local areas? \vspace{-0.4em}
    \item How to ensure the semantic meanings of these areas? \vspace{-0.4em}
\end{enumerate}
For the first problem, inspired by compositional generative models~\cite{greff2019IODINE,burgess2019monet,niemeyer2021giraffe}, we introduce local generators and a compositional synthesis procedure as the inductive bias. For the second problem, we use a dual-branch discriminator $D:\cX\times\cY\rightarrow\mathbb{R}$ that models the joint distribution $p(x,y)$ to supervise the shapes of local parts after composition.

\begin{figure}\vspace{-0.5em}
    \centering
    \includegraphics[width=\linewidth]{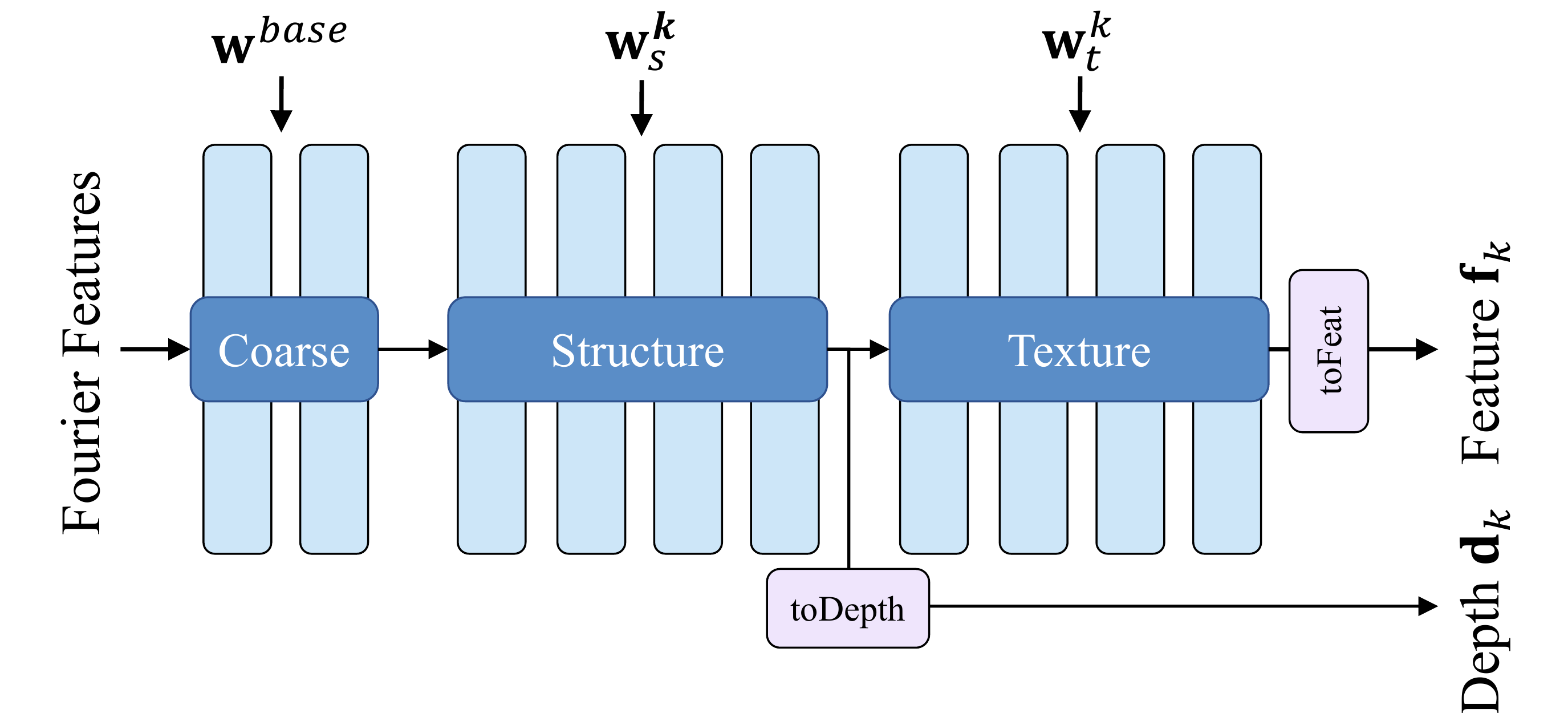}
    \vspace{-2.0em}\caption{The architecture of local generator. Blue blocks are modulated 1$\times$1 convolution layers whose weights are conditioned on input latent codes. Purple blocks are linear transformation layers.}\vspace{-1.5em}
    \label{fig:local_generator}
\end{figure}

\subsection{Generator}
The overall structure of our generator is shown in~\Cref{fig:overview}. Similar to StyleGAN2~\cite{stylegan,stylegan2}, an 8-layer MLP first maps $\bz$ to the intermediate code $\bw$. Then, $K$ local generators are introduced to model different semantic parts using $\bw$. A render net $R$ takes in the fused results from local generators and outputs an RGB image and a corresponding semantic segmentation mask. 

\vspace{-1.0em}\paragraph{Local Generator} Following recent work on continuous image rendering 
~\cite{anokhin2021CIPS,skorokhodov2021INRGAN,zhou2021cips3d}, we use modulated MLPs for local generators (\cref{fig:local_generator}), which allows explicit spatial control over synthesized output. Given Fourier features~\cite{tancik2020fourier} (position encoding) $\mathbf{p}$ and latent codes as inputs, a local generator $g_k$ outputs a feature map $\bff_k$ and a pseudo-depth map $\bd_k$:
\vspace{-0.8em}\begin{equation}
    g_k: (\mathbf{p}, \bw^{\text{base}}, \bw^k_s, \bw^k_t) \mapsto (\bff_k, \bd_k).
\end{equation}
Here, to reduce the computation cost, the input Fourier feature map as well as the outputs are of a reduced size $H^c\times W^c$, smaller than the final output image. In practice, we choose it to be $64\times 64$ to balance the efficiency and quality. During training, style mixing~\cite{stylegan} is conducted independently within each local generator between $\bw^{\text{base}}$, $\bw^k_s$ and $\bw^k_t$ such that different local parts and different shapes and textures could work collaboratively for synthesis. We note that the pseudo-depth maps here are not strictly depth maps, we call them ``depth'' because the they are used for a composition strategy that mimics the z-buffering process.

\vspace{-1.0em}\paragraph{Fusion} In the fusion step, we first generate a coarse segmentation mask $\bm\in \IR^{K\times H^c\times W^c}$ from pseudo-depth maps. Following prior work on compositional generation~\cite{greff2019IODINE,burgess2019monet}, the pseudo-depth maps are used as logits for softmax function:
\vspace{-0.4em}\begin{equation}\vspace{-0.2em}
    \bm_k(i,j) = \frac{\exp(\bd_k(i,j))}{\sum_{k'}^K{\exp(\bd_{k'}(i,j))}},
\end{equation}
where $\bm_k(i,j)$ denotes the pixel $(i,j)$ in the $k_{th}$ class of mask $\bm$ and similarly for $\bd_k(i,j)$. The feature maps are then aggregated by:
\vspace{-0.4em}\begin{equation}\vspace{-0.2em}
    \bff = \sum_{k=1}^{K} \bm_k \odot \bff_k.
\end{equation}
Here $\odot$ denotes element-wise multiplication. The aggregated feature map $\bff$ contains all the information about the output image and is sent into $R$ for rendering. We note that directly using $\bm$ for feature aggregation could be problematic when some classes are transparent. Thus, we use a modified version $\tilde{\bm}$ for feature aggregation in case of transparent classes, \eg glasses (See appendix for details). 

\vspace{-1.0em}\paragraph{Render Net}
The render net $R$ is similar to the original StyleGAN2 generator with a few modifications. First, it does not use modulated convolution layers and the output is purely conditioned on the input feature map $\bf$. Second, we input the feature map at both $16\times 16$ and $64\times 64$ resolutions, where feature concatenation is conducted at $64\times64$. The additional input of low-resolution feature map allows a better blending between different parts. Last, we find that directly training with $\bm$ is difficult due to the intrinsic gap between softmax outputs and real segmentation masks. Thus,  besides the ToRGB branch after each convolution layer, we have an additional ToSeg branch as in SemanticGAN~\cite{li2021SemanticGAN} to output residuals to refine the coarse segmentation mask $\bm$ into the final mask $\hat{y}=upsample(\bm)+\Delta \bm$ that has the same size as output image. Here a regularization loss is needed such that the final mask would not deviate too much from the coarse mask:
\vspace{-0.4em}\begin{equation}\vspace{-0.2em}
    \mathcal{L}_{mask} = \norm{\Delta \bm}^2.
\end{equation}

\subsection{Discriminator and Learning Framework}
In order to model the joint distribution $p(x,y)$, the discriminator needs to take both RGB images and segmentation masks as input. 
% We initially followed SemanticGAN~\cite{li2021SemanticGAN} to use a two-discriminator framework but we found it leads to poor synthesis quality. On the other side, simply concatenating images and masks leads to unstable training.
We found that a simple conconcatenation does not work due to the large gradient magnitude on segmentation masks. Thus, we propose to use a dual-branch discriminator $D(x,y)$ that has two convolution branches for $x$ and $y$, respectively. The outputs are then summed up for fully connected layers. Such a design allows us to separately regularize the gradient norm of the segmentation branch with an additional R1 regularization loss $\mathcal{L}_{R1_{seg}}$. The resulting training framework is similar to StyleGAN2 with the loss function:
\vspace{-0.4em}\begin{equation}\vspace{-0.2em}
    \mathcal{L}_{all} = \mathcal{L}_{StyleGAN2} + \lambda_{mask} \mathcal{L}_{mask} +\lambda_{R1_{seg}} \mathcal{L}_{R1_{seg}},
\end{equation}
where $\mathcal{L}_{StyleGAN2}$ denotes the loss functions used in the original StyleGAN2.

\section{Implementation Details}
\label{sec:implementation_details}
We implement our methods using PyTorch 1.15 library. We use the same optimizer and batch settings as in StyleGAN2. $\lambda_{R1_{img}}$, $\lambda_{R1_{seg}}$, $\lambda_{mask}$ are set to $10$, $1000$ and $100$, respectively. Style mixing probability and path regularization are reduced to $0.3$ and $0.5$, respectively. For some experiments, we fine-tune our models on image-only datasets. In such cases, we drop the segmentation branch in discriminator and use the original StyleGAN2 loss functions to fine-tune the model. \emph{Due to the space limit, more details about network architectures are given in appendix}.

\begin{figure*}[t]
\captionsetup{font=small}
\centering
\scriptsize
\setlength\tabcolsep{1px}
\newcommand{\www}{0.1\linewidth}
\renewcommand{\arraystretch}{0.1}
\newcolumntype{Y}{>{\centering\arraybackslash}X}
\begin{tabularx}{\linewidth}{p{6pt} ccc ccc ccc c}
    \raisebox{2.2\height}{\rotatebox[origin=c]{90}{Image}} & 
    \includegraphics[width=\www]{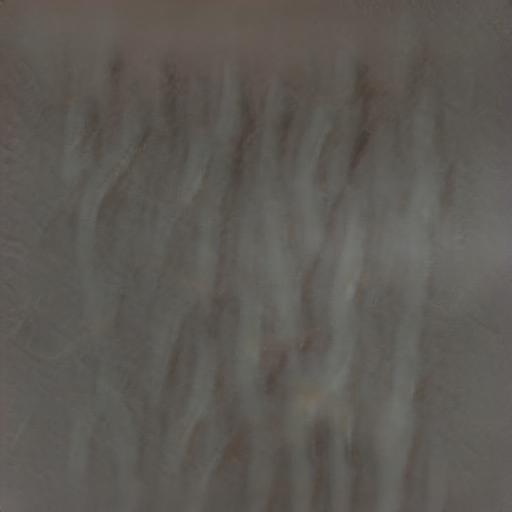}\hfill
    \includegraphics[width=\www]{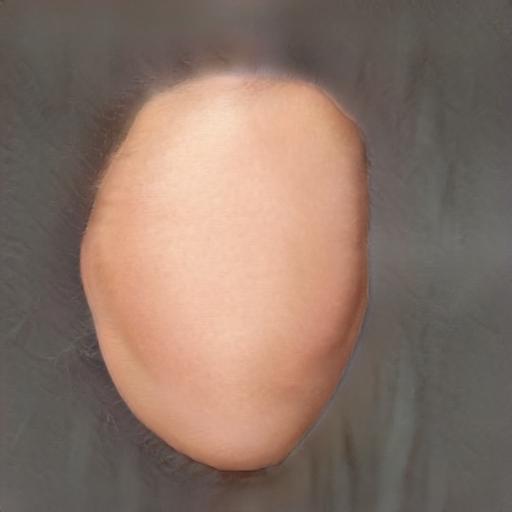}\hfill
    \includegraphics[width=\www]{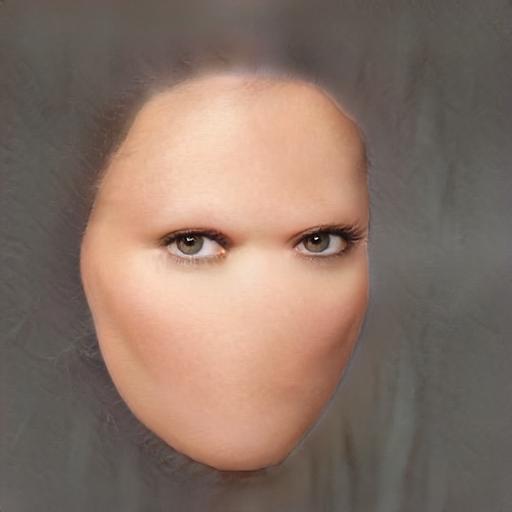}\hfill
    \includegraphics[width=\www]{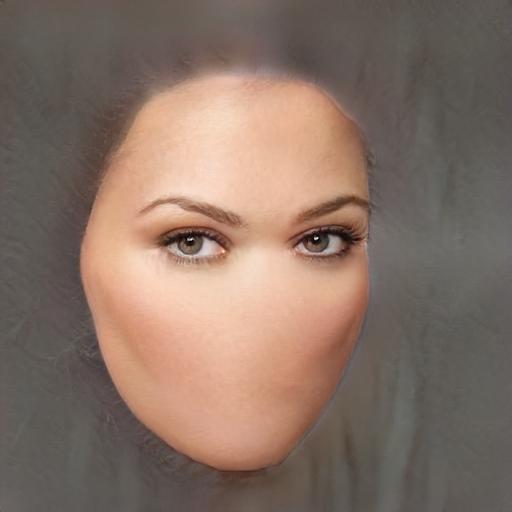}\hfill
    \includegraphics[width=\www]{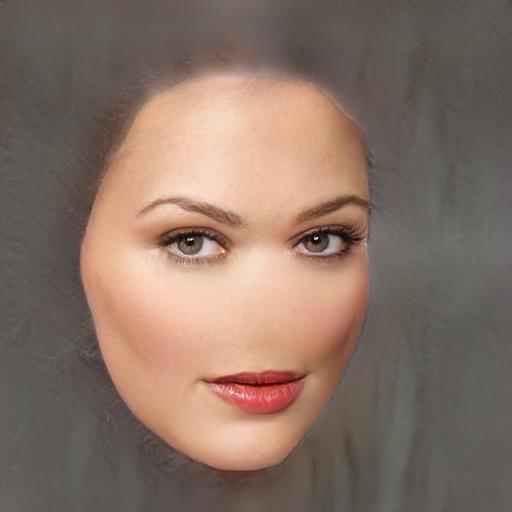}\hfill
    \includegraphics[width=\www]{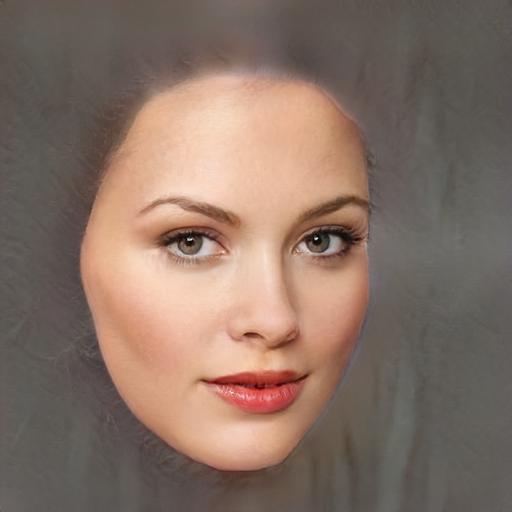}\hfill
    \includegraphics[width=\www]{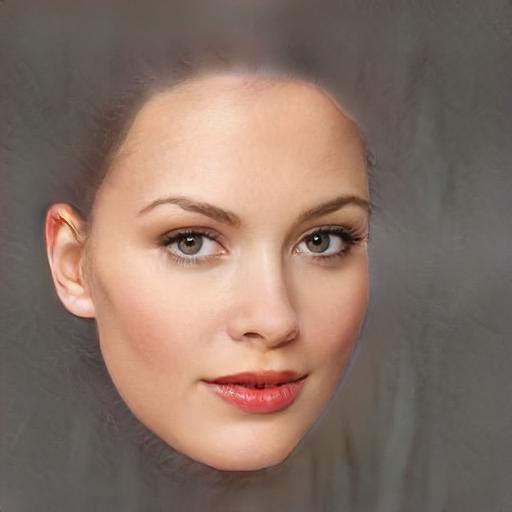}\hfill
    \includegraphics[width=\www]{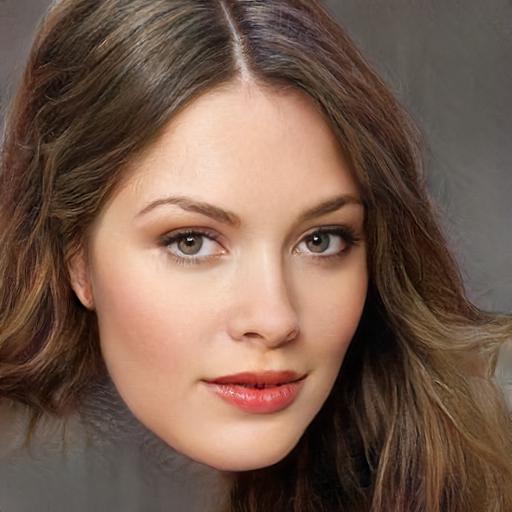}\hfill
    \includegraphics[width=\www]{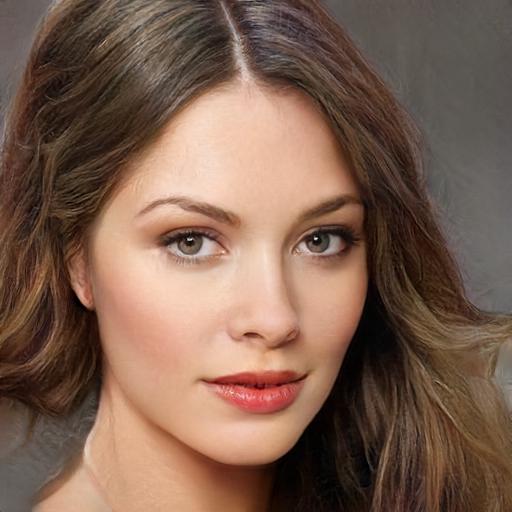}\hfill
    \includegraphics[width=\www]{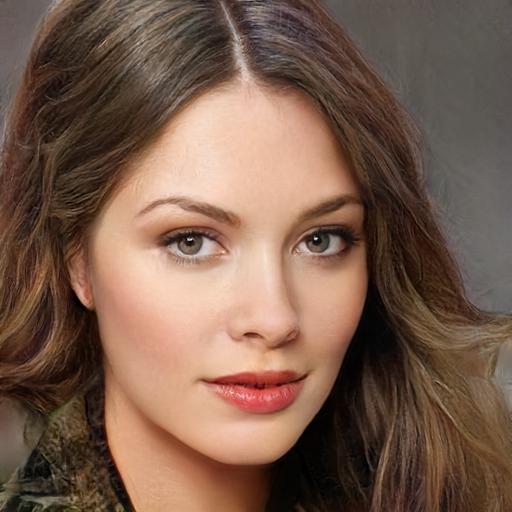}\\
    \raisebox{1.3\height}{\rotatebox[origin=c]{90}{Psuedo-depth}} & 
    \includegraphics[width=\www]{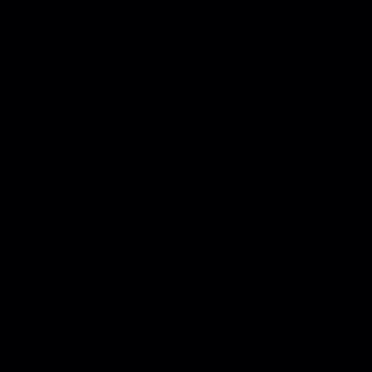}\hfill
    \includegraphics[width=\www]{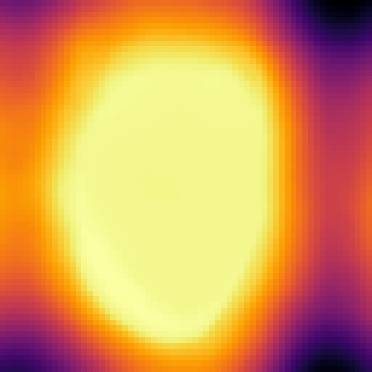}\hfill
    \includegraphics[width=\www]{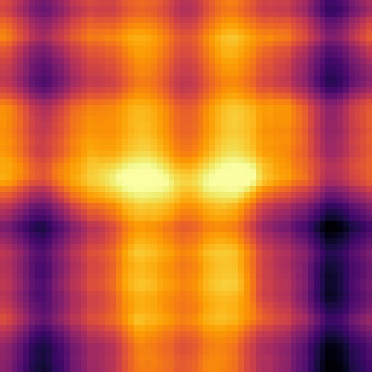}\hfill
    \includegraphics[width=\www]{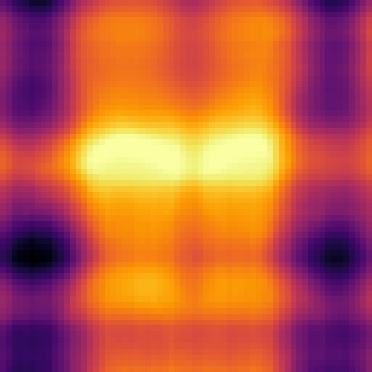}\hfill
    \includegraphics[width=\www]{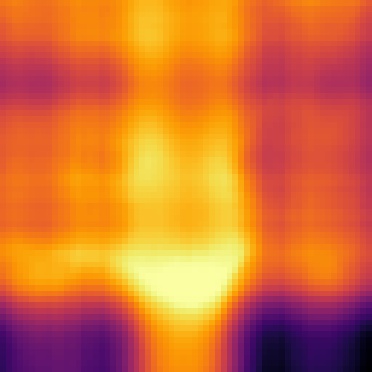}\hfill
    \includegraphics[width=\www]{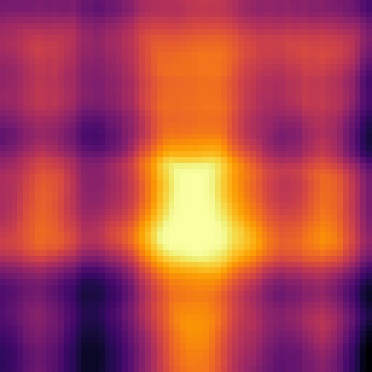}\hfill
    \includegraphics[width=\www]{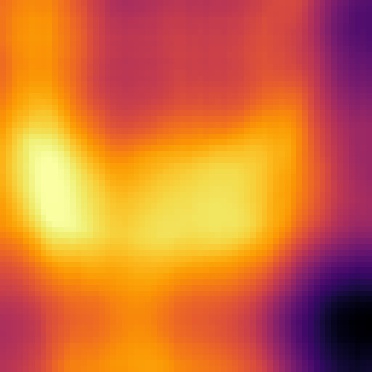}\hfill
    \includegraphics[width=\www]{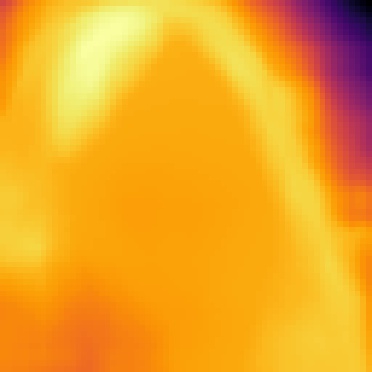}\hfill
    \includegraphics[width=\www]{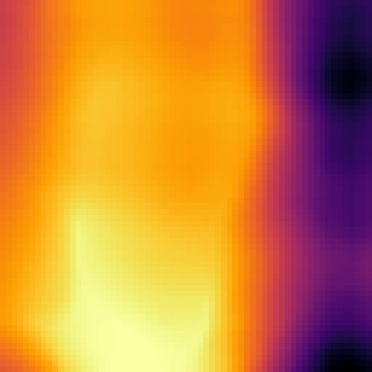}\hfill
    \includegraphics[width=\www]{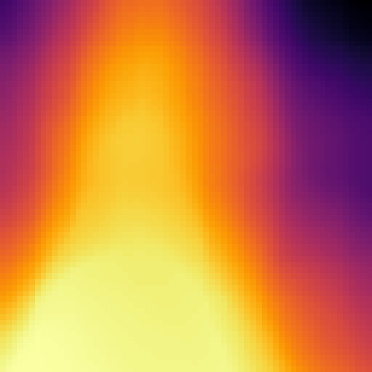}\\
    \raisebox{1.2\height}{\rotatebox[origin=c]{90}{Segmentation}} & 
    \includegraphics[width=\www]{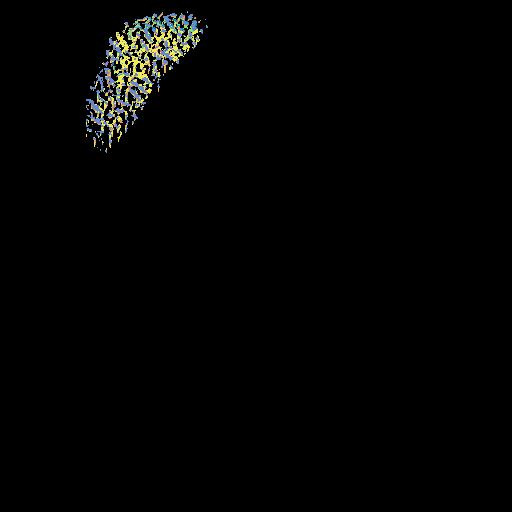}\hfill
    \includegraphics[width=\www]{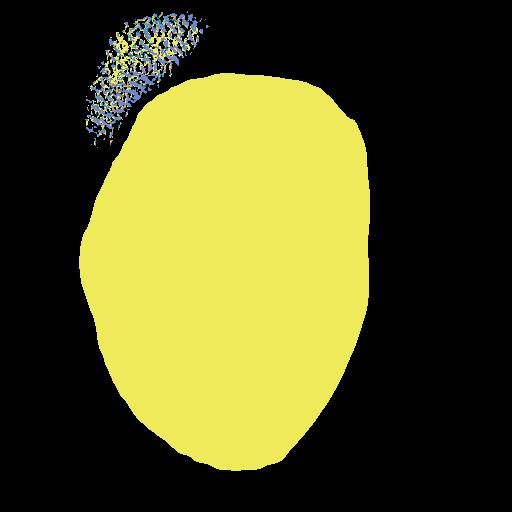}\hfill
    \includegraphics[width=\www]{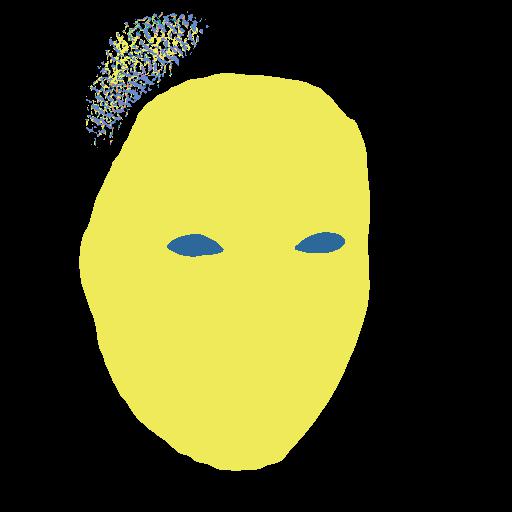}\hfill
    \includegraphics[width=\www]{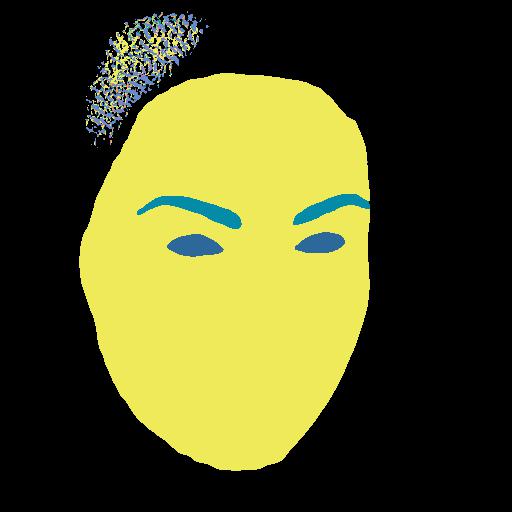}\hfill
    \includegraphics[width=\www]{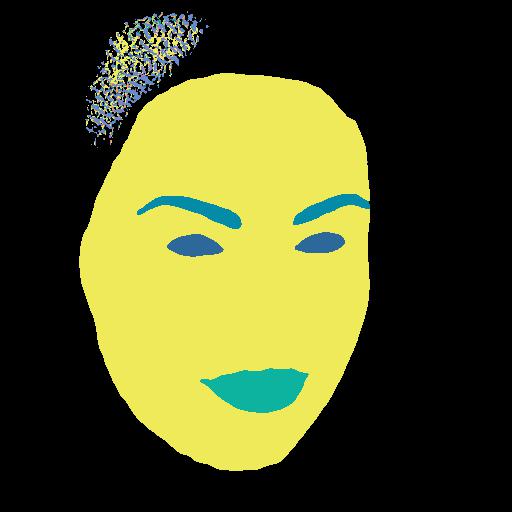}\hfill
    \includegraphics[width=\www]{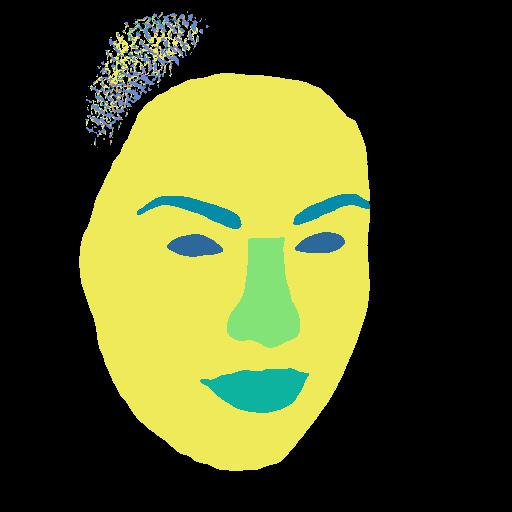}\hfill
    \includegraphics[width=\www]{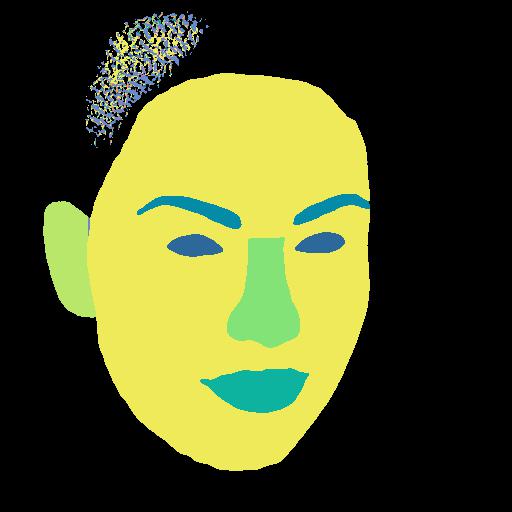}\hfill
    \includegraphics[width=\www]{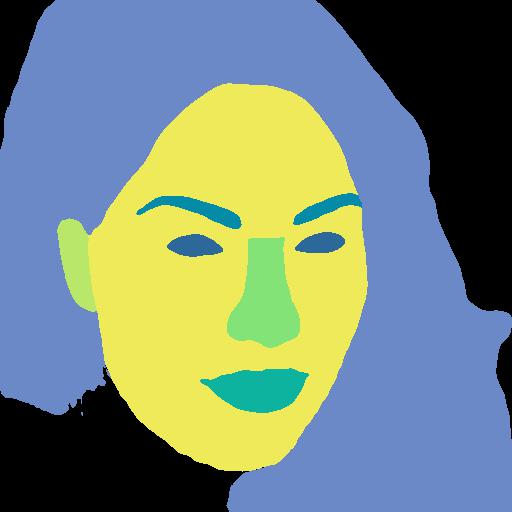}\hfill
    \includegraphics[width=\www]{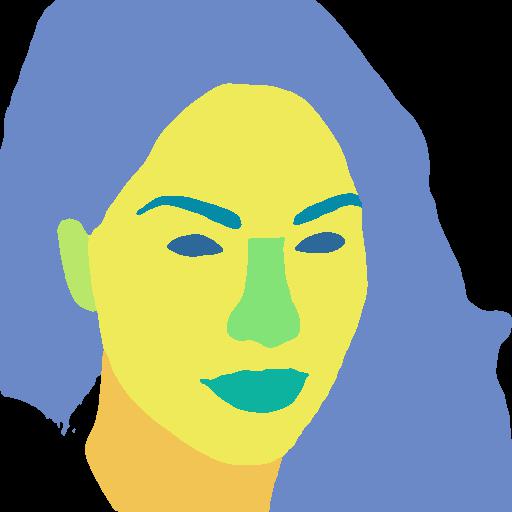}\hfill
    \includegraphics[width=\www]{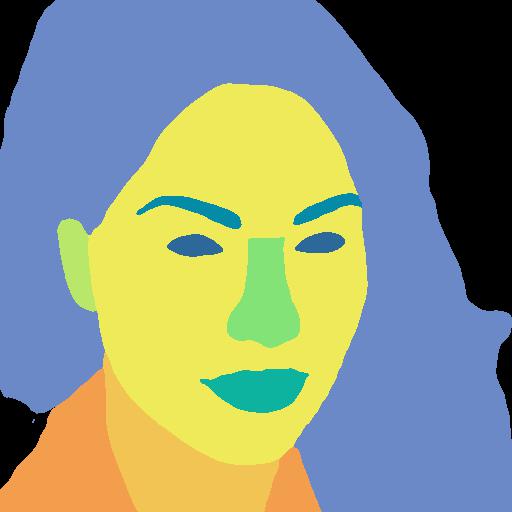}\\
\end{tabularx}
    \vspace{-1.0em}\caption{Illustration of compositional synthesis. Starting from background, we gradually add more components into the feature map. The second row shows the pseudo-depth map of each corresponding component used for fusion. Note that the ``hair'' generator outputs a complete shape even though it is covered by the face. During synthesis, all pseudo-depth maps are fused without an order.}\vspace{-1.0em}
    \label{fig:components}
\end{figure*}

\begin{table}[t]
\captionsetup{font=small}
\newcommand{\mr}[1]{\multirow{2}{*}{#1}}
\setlength{\tabcolsep}{4pt}
\footnotesize
\begin{center}
\begin{tabularx}{1.0\linewidth}{X cccc}
\toprule
Method & Data & Compositional & FID$\downarrow$ & IS$\uparrow$ \\
\midrule            
% StyleGAN2                       & img       &\ding{55}& 5.83  & 3.31 \\\hline
% SemanticGAN                     & img\&seg  &\ding{55}& 18.46 & 2.76 \\
% \; + proposed training          & img\&seg  &\ding{55}& 7.05  & 3.31 \\
% SemanticStyleGAN (ours)         & img\&seg  &\ding{51}& 6.42  & 3.21 \\
StyleGAN2                       & img       &\ding{55}& 4.45  & 3.40 \\\hline
SemanticGAN                     & img\&seg  &\ding{55}& 18.54 & 2.77 \\
\; + proposed training          & img\&seg  &\ding{55}& 7.50  & 3.51 \\
SemanticStyleGAN (ours)         & img\&seg  &\ding{51}& 6.42  & 3.21 \\
\bottomrule
\end{tabularx}
\vspace{-1.0em}\caption{Quantitative evaluation on synthesis quality. All the models are trained on CelebAMask-HQ at 256$\times$256. ``img'' and ``seg'' refer to RGB image and segmentation mask, respectively.}\vspace{-2.5em}
\label{tab:synthesis_quality}
\end{center}
\end{table}

\begin{figure}
    \centering
    \includegraphics[width=\linewidth]{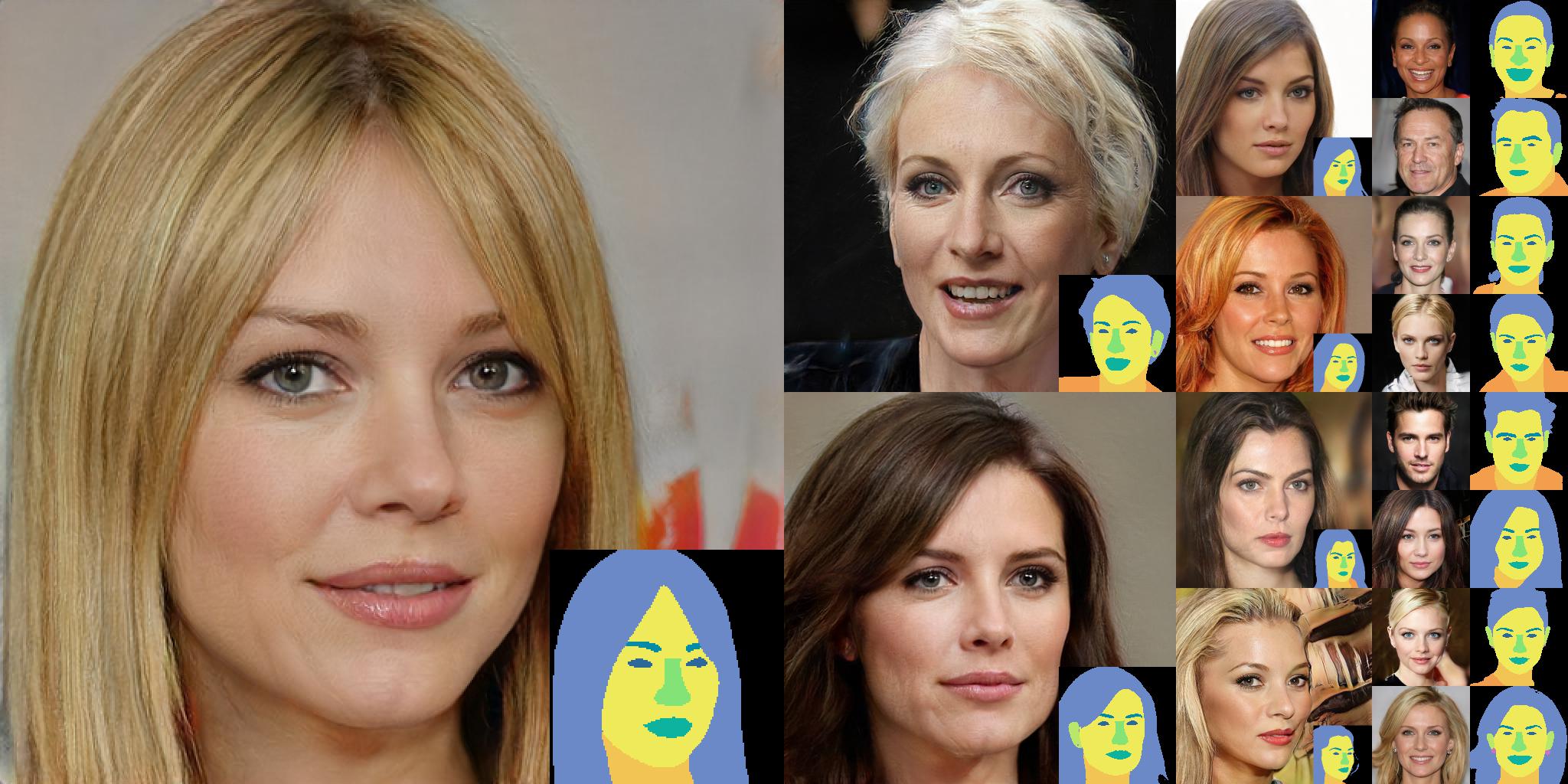}
    \vspace{-1.6em}\caption{Example generation results of our model trained on CelebAMask-HQ. The images are generated at a resolution of 512$\times$512 with a truncation of $0.7$.}
    \label{fig:collage}\vspace{-0.5em}
\end{figure}

\section{Experiments}

\begin{figure}[t]
\captionsetup{font=small}
\centering
\scriptsize
\setlength\tabcolsep{1px}
\newcommand{\www}{0.16\linewidth}
\renewcommand{\arraystretch}{0.1}
\newcolumntype{Y}{>{\centering\arraybackslash}X}
\begin{tabularx}{\linewidth}{Y>{\centering\arraybackslash}c}
    \raisebox{2.0\height}{\rotatebox[origin=c]{90}{All}} & 
    \includegraphics[width=\www]{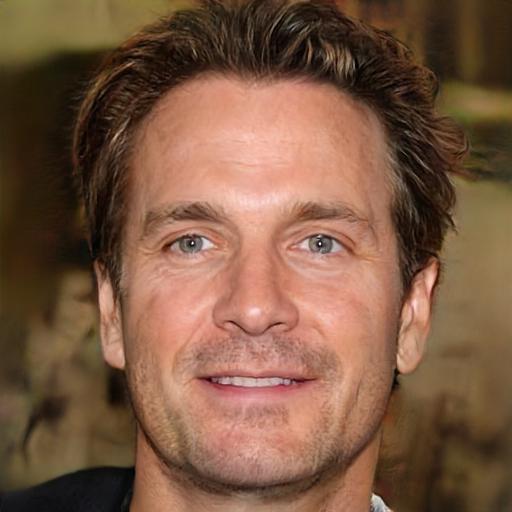}\hfill
    \includegraphics[width=\www]{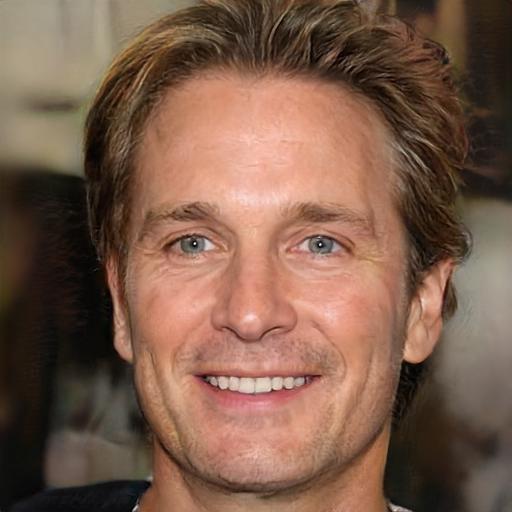}\hfill
    \includegraphics[width=\www]{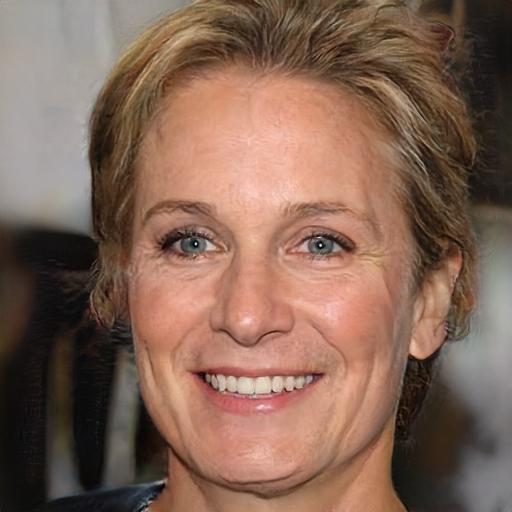}\hfill
    \includegraphics[width=\www]{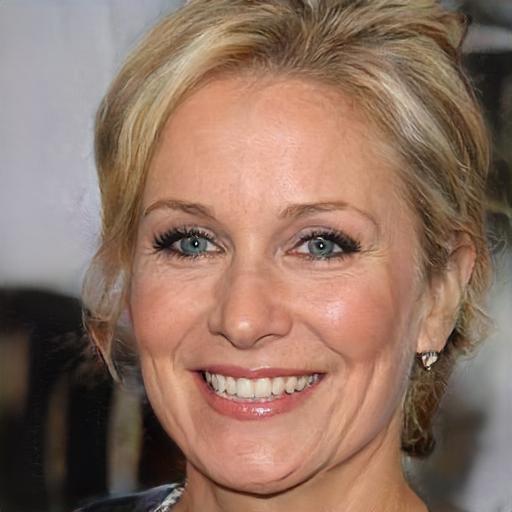}\hfill
    \includegraphics[width=\www]{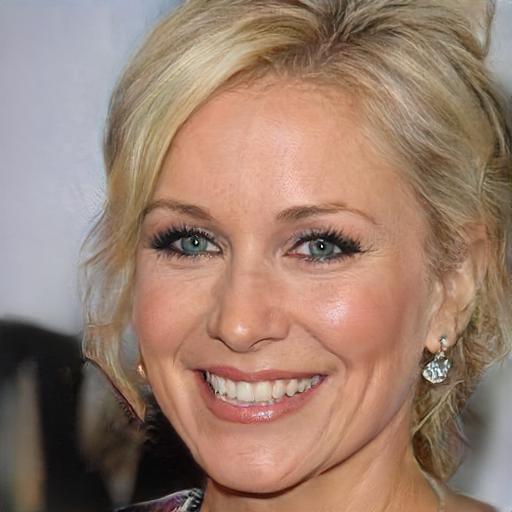}\hfill
    \includegraphics[width=\www]{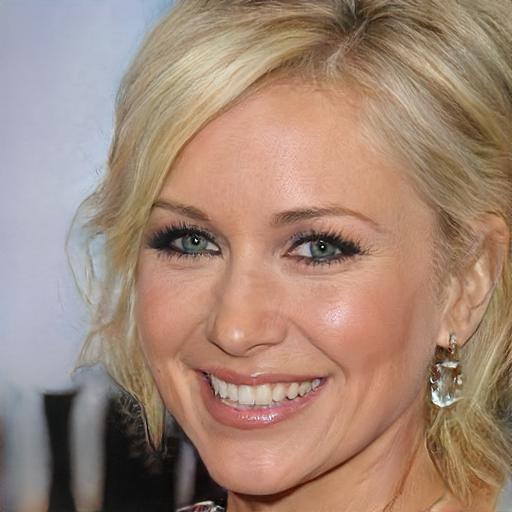}\\
    \raisebox{0.9\height}{\rotatebox[origin=c]{90}{Background}} & 
    \includegraphics[width=\www]{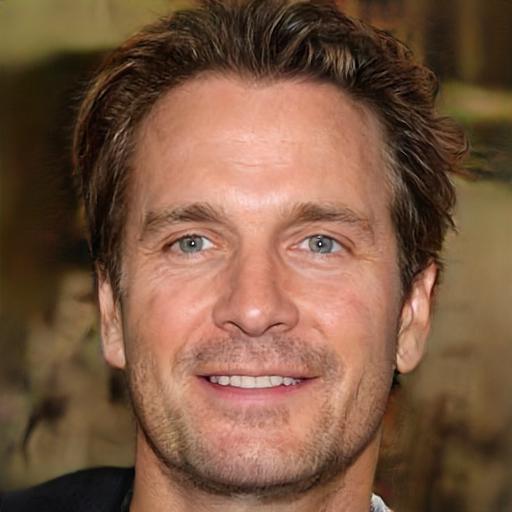}\hfill
    \includegraphics[width=\www]{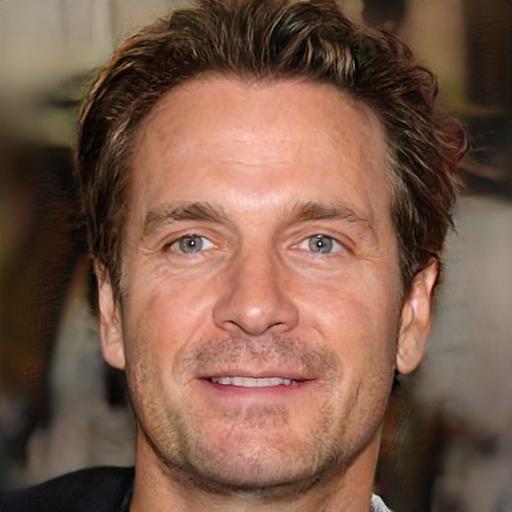}\hfill
    \includegraphics[width=\www]{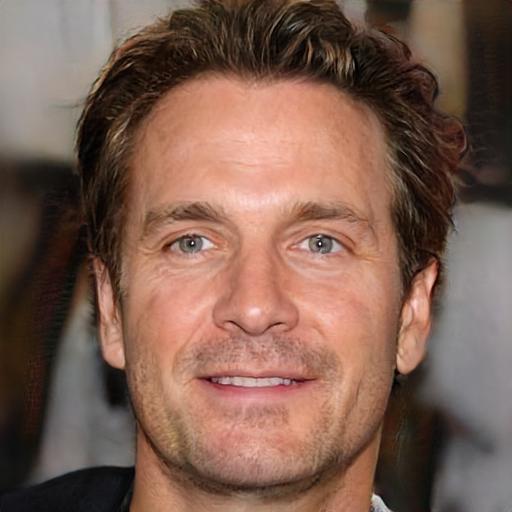}\hfill
    \includegraphics[width=\www]{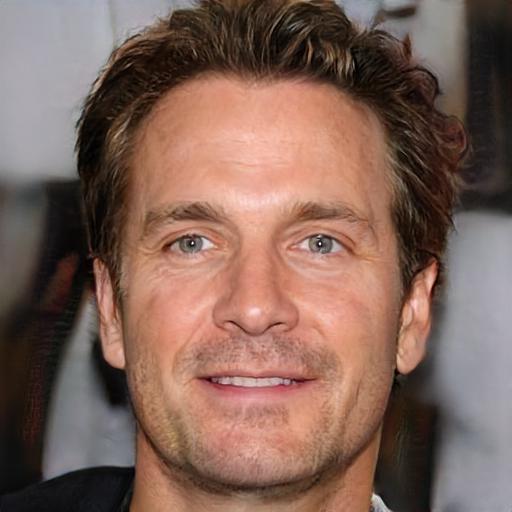}\hfill
    \includegraphics[width=\www]{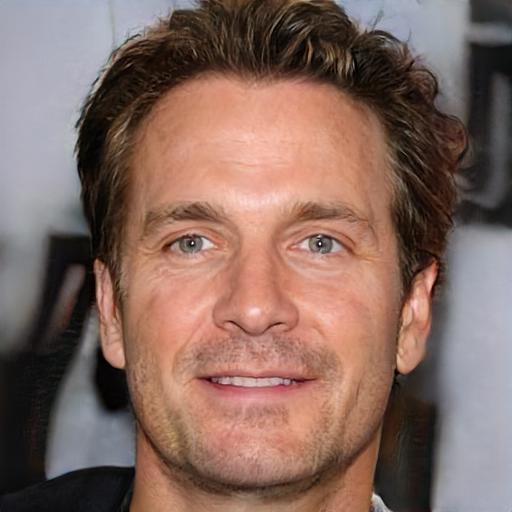}\hfill
    \includegraphics[width=\www]{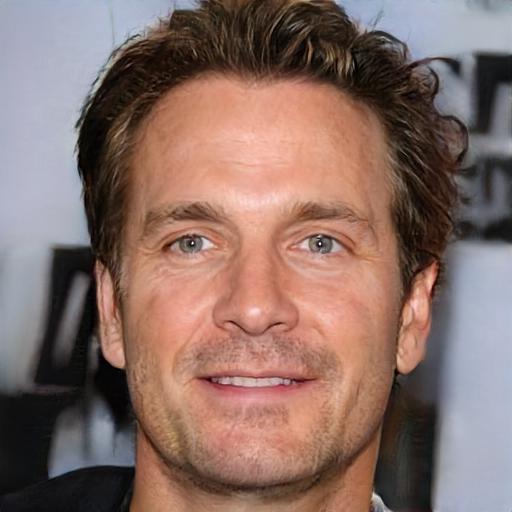}\\
    \raisebox{1.9\height}{\rotatebox[origin=c]{90}{Face}} & 
    \includegraphics[width=\www]{fig/interpolation/layer_02_01.jpeg}\hfill
    \includegraphics[width=\www]{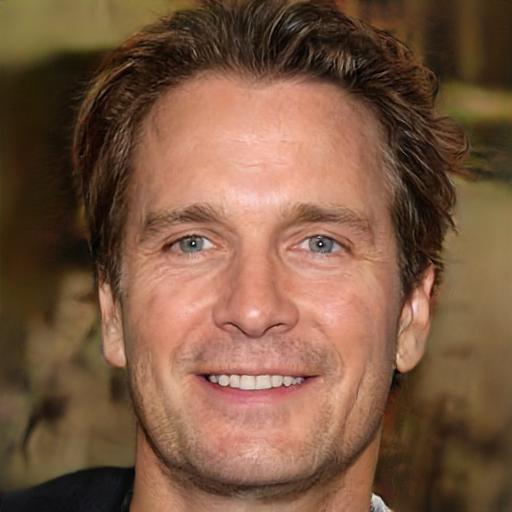}\hfill
    \includegraphics[width=\www]{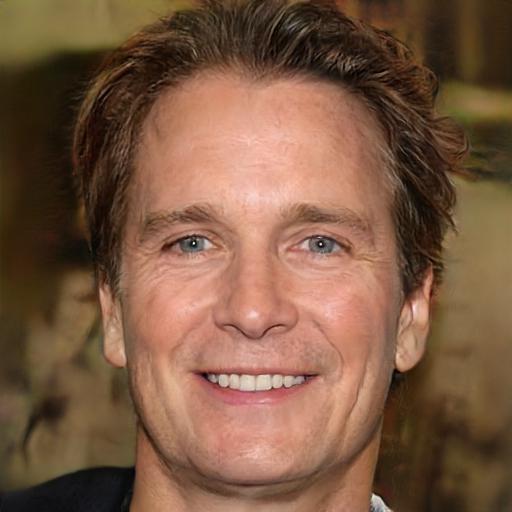}\hfill
    \includegraphics[width=\www]{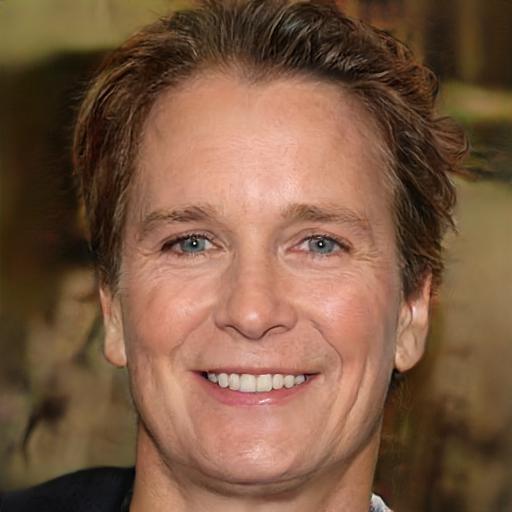}\hfill
    \includegraphics[width=\www]{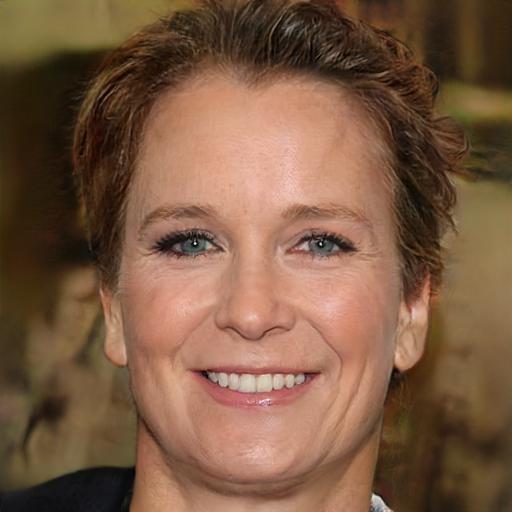}\hfill
    \includegraphics[width=\www]{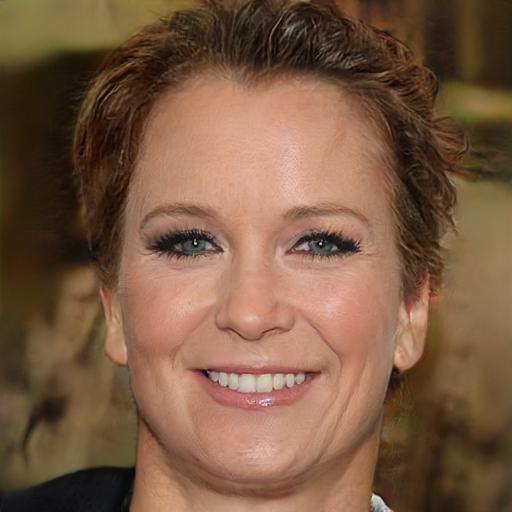}\\
    \raisebox{1.8\height}{\rotatebox[origin=c]{90}{Hair}} & 
    \includegraphics[width=\www]{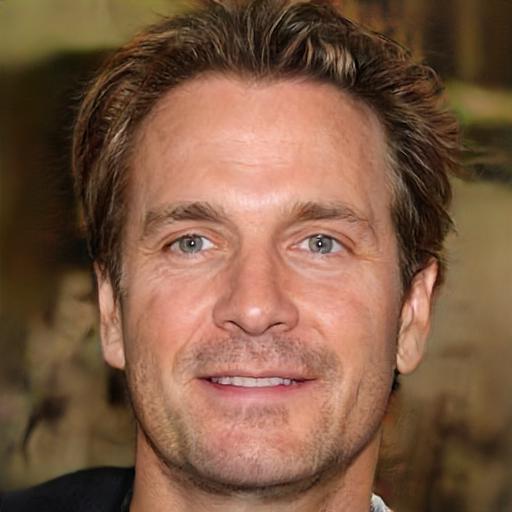}\hfill
    \includegraphics[width=\www]{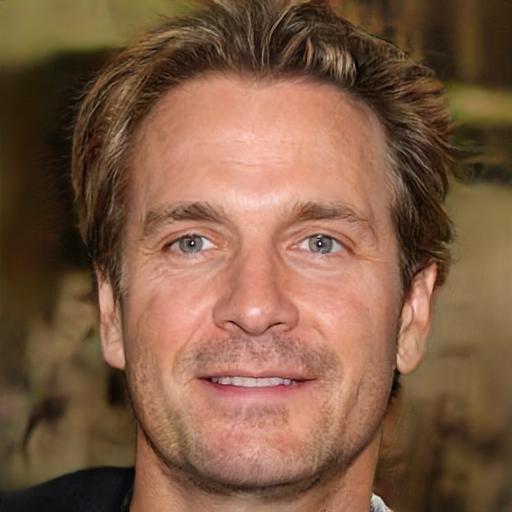}\hfill
    \includegraphics[width=\www]{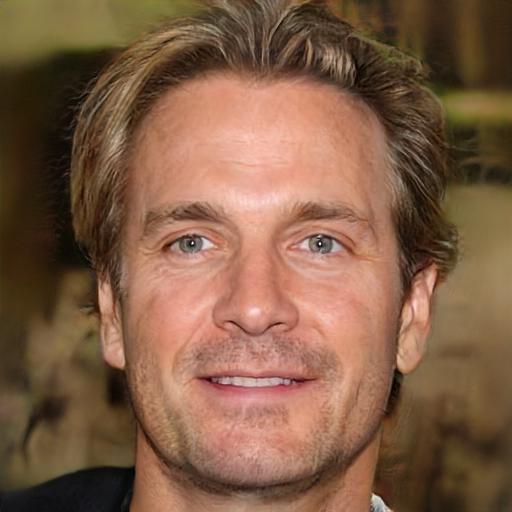}\hfill
    \includegraphics[width=\www]{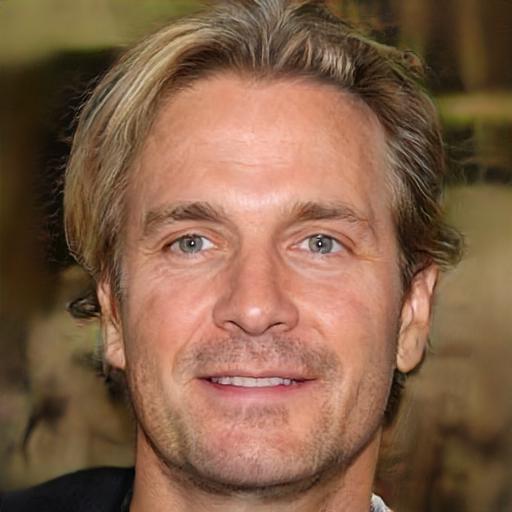}\hfill
    \includegraphics[width=\www]{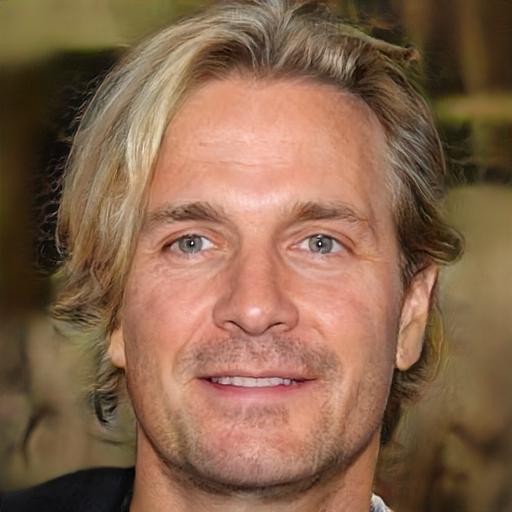}\hfill
    \includegraphics[width=\www]{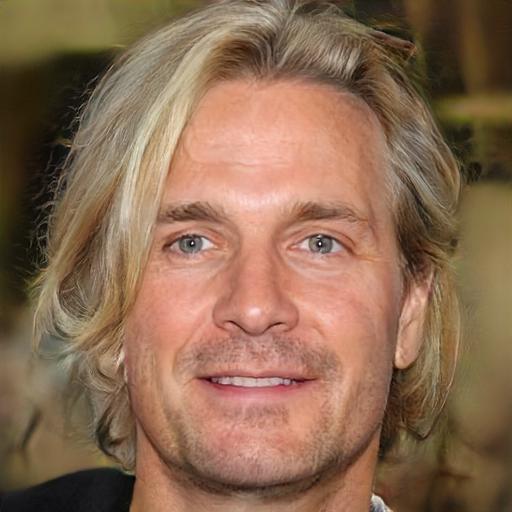}\\
\end{tabularx}
    \vspace{-1.0em}\caption{Results of latent interpolation on the whole latent space and specified subspaces. Here, ``Face'' refers to all the components relevant to face, including eyes, mouth, etc.}\vspace{-0.8em}
    \label{fig:interpolation}
\end{figure}

\subsection{Semantic-aware and Disentangled Generation}
% \lipsum[1]
% \lipsum[2]

We first evaluate our model on the synthesis quality and its local disentanglement. For synthesis quality, we compare our model with StyleGAN2~\cite{stylegan2} and SemanticGAN~\cite{li2021SemanticGAN}. The original StyleGAN2, which neither models segmentation masks nor provides local controllability, is compared against as an upper bound of synthesis quality. SemanticGAN modifies StyleGAN2 into a joint training framework to output both image and segmentation masks. Since its goal is to conduct segmentation, it does not allows local control either. All the models are trained on the the first 28,000 images of CelebAMask-HQ resized to 256$\times$256. Fréchet Inception Distance (FID)~\cite{heusel2017FID} and Inception Score (IS)~\cite{salimans2016improved} are used to measure the synthesis quality.

Our project is initially built on SemanticGAN framework for learning a semantic-aware model. The original SemanticGAN is semi-supervised and we change it to use all the training labels. As shown in \cref{tab:synthesis_quality}, SemanticGAN achieves much lower quality compared to original StyleGAN, indicating that learning a joint model of images and segmentation masks is a challenging task. Hypothesizing that the main bottleneck of SemanticGAN is the additional patch discriminator used for learning segmentation masks,  we replace it with the proposed dual-branch discriminator. The new training framework achieves much better synthesis score. We further replace the SemanticGAN generator with our SemanticStyleGAN generator. Compared to SemanticGAN generator, our model shows a similar synthesis quality while providing additional controllability on each semantic area. We then extend our model to 512$\times$512 resolution and achieve a FID and IS of $7.22$ and $3.47$, respectively. For reference, the StyleGAN2 generator achieves a FID and IS of $6.47$ and $3.55$, respectively. \cref{fig:collage} shows the synthesis results of the 512$\times$512 model.

To interpret the compositional synthesis of our model, \cref{fig:components} shows the results of synthesis with limited components. We first disable all the foreground generators and gradually add them into the forward process. It can be seen that these local generators can work independently to generate a semantic part. The pseudo-depth maps, in spite of the lack of 3D supervision, learn meaningful shapes that could be used to collaboratively compose different faces. 
% \cref{fig:background_exp} shows the results of synthesizing fake faces on a real background image. The image is first inverted into a feature map via optimization, and then we replace background generator with this feature map. The resulting synthesized face blends into the background naturally. Besides, since we are using Fourier features as input, we can also control the location and size of the generated foreground in the image.

\begin{figure}[t]
\captionsetup{font=small}
\centering
\footnotesize
\setlength\tabcolsep{0.3pt}
\newcommand{\www}{0.198\linewidth}
\renewcommand{\arraystretch}{0.2}
\newcolumntype{Y}{>{\centering\arraybackslash}X}
\begin{tabularx}{\linewidth}{ccccc}
    \makecell{real image} & \makecell{reconstruction} & \makecell{synthesized} & \makecell{translation} & \makecell{zoom out}  \\ 
    \includegraphics[width=\www]{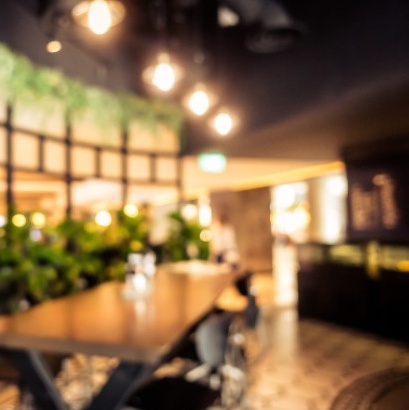} & 
    \includegraphics[width=\www]{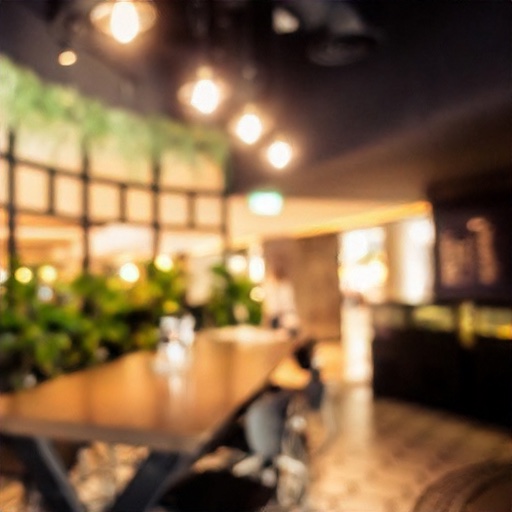} & 
    \includegraphics[width=\www]{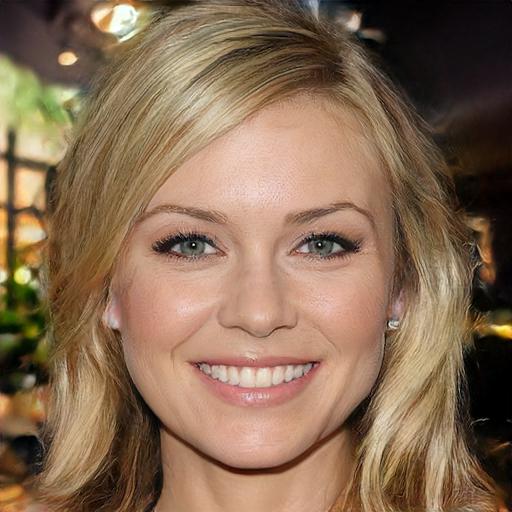} &
    \includegraphics[width=\www]{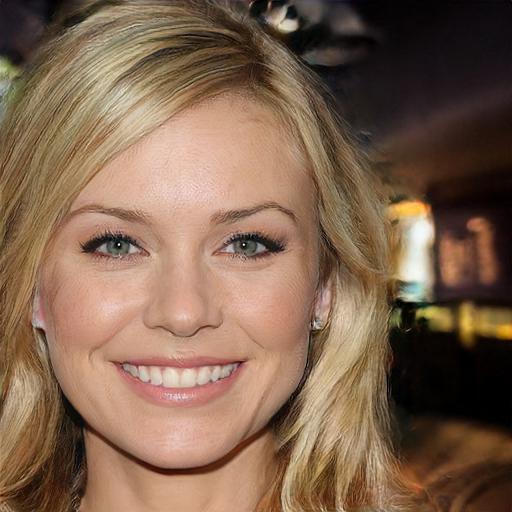} &
    \includegraphics[width=\www]{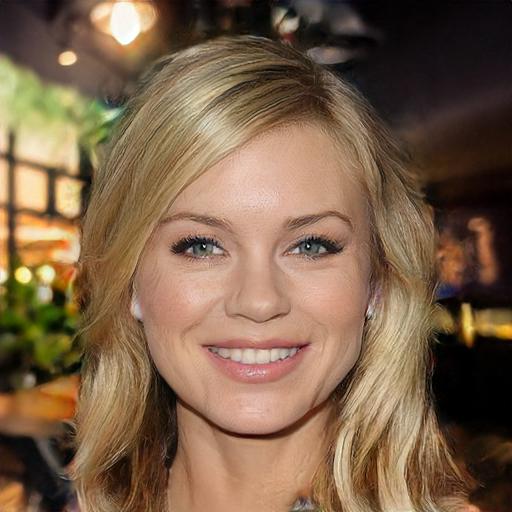}\\
\end{tabularx}
    \vspace{-1.0em}\caption{Image composition and transformation via Fourier feature manipulation. Real images are used as background for synthesis and inverted into feature maps. Then foreground can be synthesized on this real image in the feature space. The location and size of foreground can be controlled via Fourier features.}\vspace{-0.8em}
    \label{fig:background_exp}
\end{figure}

\cref{fig:interpolation} shows the results of latent interpolation of our generator model. The first row shows that our model could interpolate smoothly between two randomly sampled images. Besides, we can interpolate on a specific semantic area by changing the corresponding latent codes, e.g. face or hair, while fixing irrelevant parts. The results indicate that our model has learned a smooth and disentangled latent space for semantic editing.
% \cref{fig:style_mix} further shows the results of individually changing shape and texture codes. 
Overall, even though 
there is no explicit constraint during training, we observe that our model could disentangle most local shapes and textures. We also refer the readers to the appendix for more results on semantic local style mixing.
We note that unlike traditional GANs that generates a complete image, such a compositional process also allows our model to generate the foreground only and control it by manipulating the Fourier features (See~\cref{fig:background_exp}).

% \begin{figure}[t]
% \captionsetup{font=small}
% \centering
% \scriptsize
% \setlength\tabcolsep{1px}
% \newcommand{\www}{0.192\linewidth}
% \renewcommand{\arraystretch}{0.1}
% \newcolumntype{Y}{>{\centering\arraybackslash}X}
% \begin{tabularx}{\linewidth}{p{7pt}c}
%     \raisebox{1.6\height}{\rotatebox[origin=c]{90}{Face}} & 
%     \includegraphics[width=\www]{fig/style_mix_new/layer_01_03.jpeg}\hfill
%     \includegraphics[width=\www]{fig/style_mix_new/layer_01_06.jpeg}\hfill
%     \includegraphics[width=\www]{fig/style_mix_new/image.jpeg}\hfill
%     \includegraphics[width=\www]{fig/style_mix_new/layer_02_02.jpeg}\hfill
%     \includegraphics[width=\www]{fig/style_mix_new/layer_02_09.jpeg}\\
%     \raisebox{1.8\height}{\rotatebox[origin=c]{90}{Hair}} & 
%     \includegraphics[width=\www]{fig/style_mix_new/layer_03_00.jpeg}\hfill
%     \includegraphics[width=\www]{fig/style_mix_new/layer_03_04.jpeg}\hfill
%     \includegraphics[width=\www]{fig/style_mix_new/image.jpeg}\hfill
%     \includegraphics[width=\www]{fig/style_mix_new/layer_04_01.jpeg}\hfill
%     \includegraphics[width=\www]{fig/style_mix_new/layer_04_05.jpeg}\\
% \end{tabularx}
%     \vspace{-1.0em}\caption{Random sampling in local latent spaces. The middle row shows a randomly sampled face. The first two and last two columns show the results of changing shape and texture codes, respectively.  Here, ``Face'' refers to all the components relevant to face, including eyes, mouth, etc.}\vspace{-1.0em}
%     \label{fig:style_mix}
% \end{figure}

\subsection{Controlled Synthesis and Image Editing}
With the semantic decomposition in the latent space, our model provides a more disentangled generative prior for image editing. Here, we evaluate our model on downstream editing tasks and compares it to StyleGAN2. We use the pytorch conversion of official StyleGAN2 (config-F on FFHQ 1024x1024) as our baseline, which is widely used in relevant studies on image editing.  The 512$\times$512 model is used for our method.

\begin{table}[t]
\captionsetup{font=small}
\newcommand{\mr}[1]{\multirow{2}{*}{#1}}
\setlength{\tabcolsep}{2.2pt}
\footnotesize
\begin{center}
\begin{tabularx}{1.0\linewidth}{X ccc}
\toprule
Method & MSE$\downarrow$ & ID$\uparrow$  & LPIPS$\downarrow$ \\
\midrule            
StyleGAN2 (FFHQ)        & 0.031$\pm$ 0.015  & 0.654$\pm$ 0.097 & 0.309$\pm$0.046 \\\hline
StyleGAN2               & 0.029$\pm$ 0.016  & 0.575$\pm$ 0.119 & 0.330$\pm$0.052 \\
SemanticStyleGAN        & 0.031$\pm$ 0.017  & 0.602$\pm$ 0.122 & 0.335$\pm$0.051 \\
% Ours (w/ RenderNet)     & 0.025$\pm$ 0.014  & 0.648$\pm$ 0.111 & 0.319$\pm$0.050 \\
\bottomrule
\end{tabularx}
\vspace{-1.0em}\caption{Quantitative evaluation of reconstruction performance using Restyle (psp) encoder. The bottom two rows (StyleGAN2 and Ours) are trained on the same split of CelebAMask-HQ.}
\label{tab:reconstruction}
\end{center}\vspace{-2.5em}
\end{table}

\begin{figure*}[t]
\captionsetup{font=small}
\centering
\footnotesize
\setlength\tabcolsep{0.0pt}
\newcommand{\www}{0.087\linewidth}
\renewcommand{\arraystretch}{0.4}
\newcolumntype{Y}{>{\centering\arraybackslash}X}
\begin{tabularx}{\linewidth}{p{8pt} c @{\hskip 2pt}|@{\hskip 2pt} c @{\hskip 1.5pt} cc @{\hskip 1.5pt} cc @{\hskip 2pt}|@{\hskip 2pt} c @{\hskip 1.5pt} cc @{\hskip 1.5pt} cc}
    \raisebox{1.0\height}{\rotatebox[origin=c]{90}{Expression}} & 
    \includegraphics[width=\www]{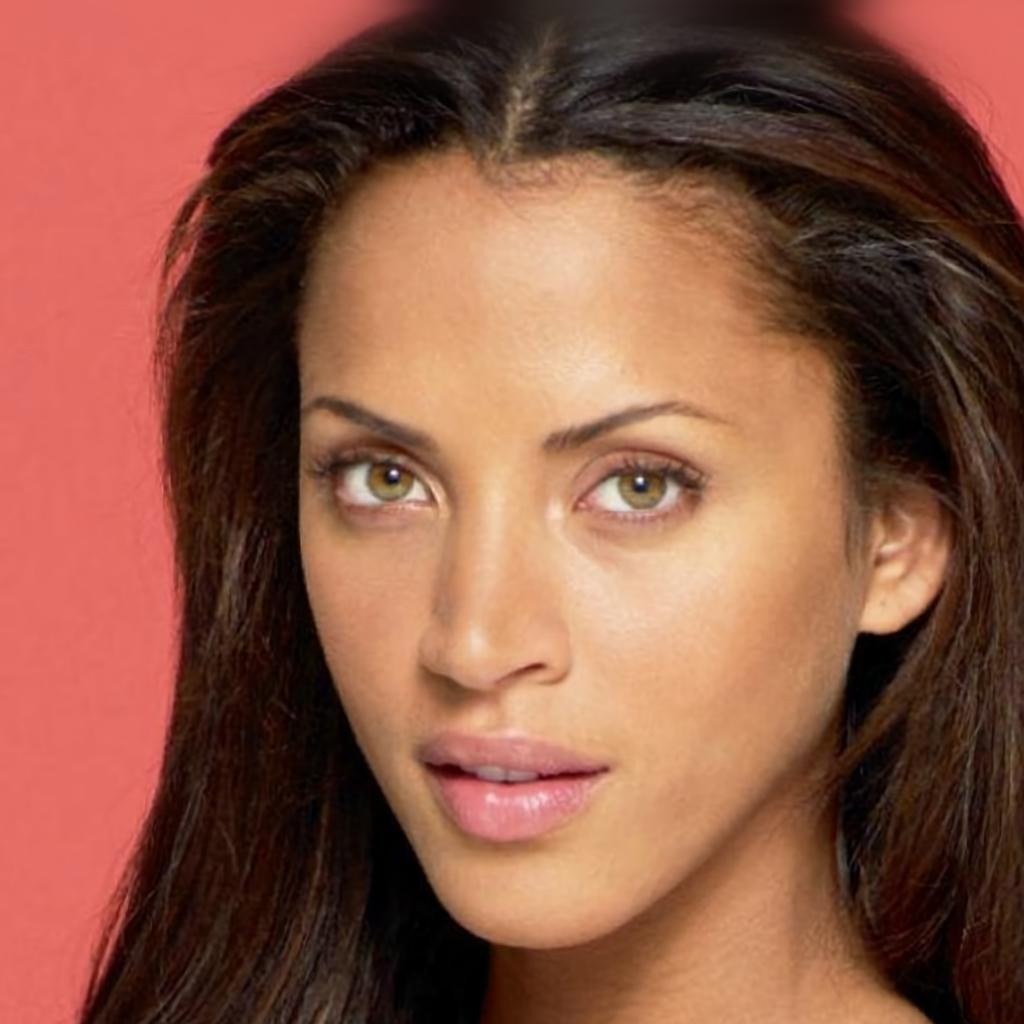} &
    \includegraphics[width=\www]{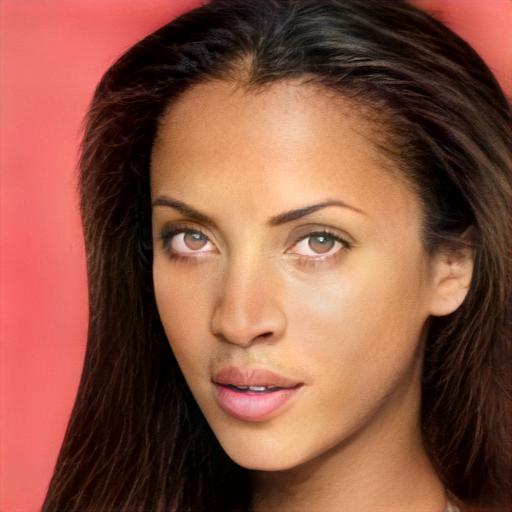} &
    \includegraphics[width=\www]{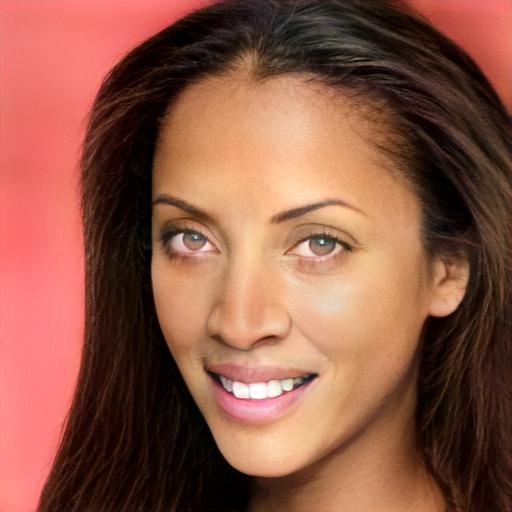} &
    \includegraphics[width=\www]{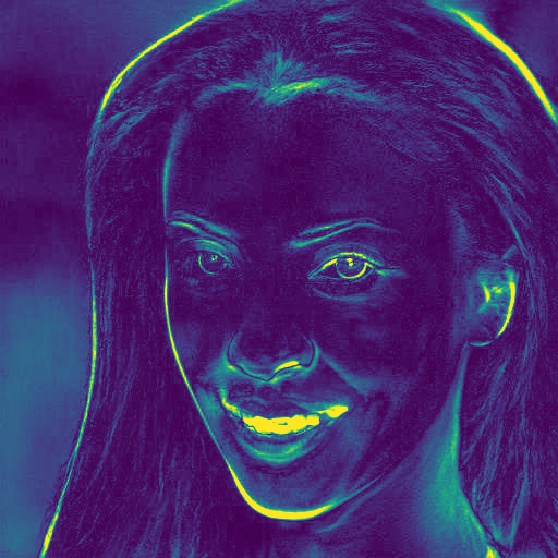} &
    \includegraphics[width=\www]{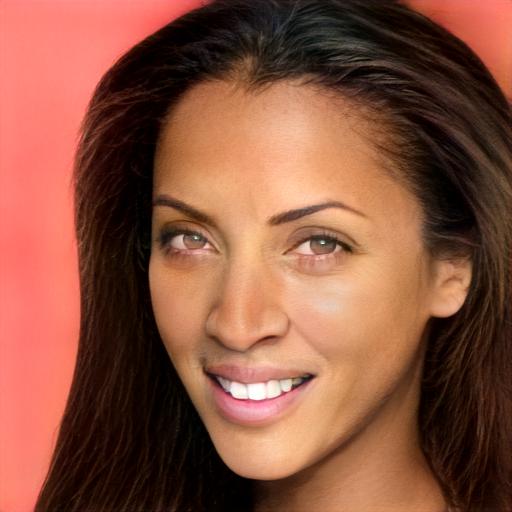} &
    \includegraphics[width=\www]{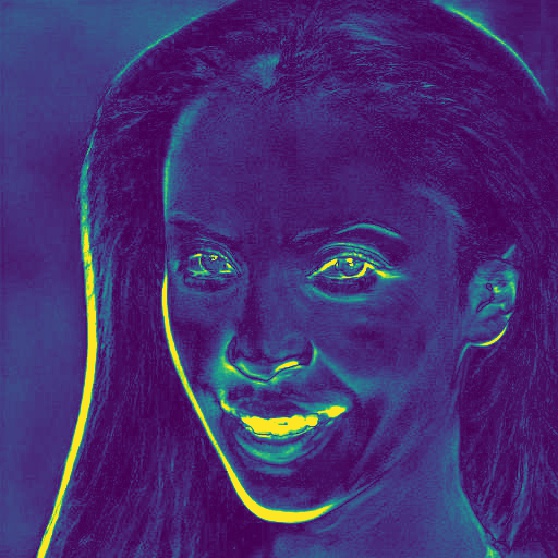} &
    \includegraphics[width=\www]{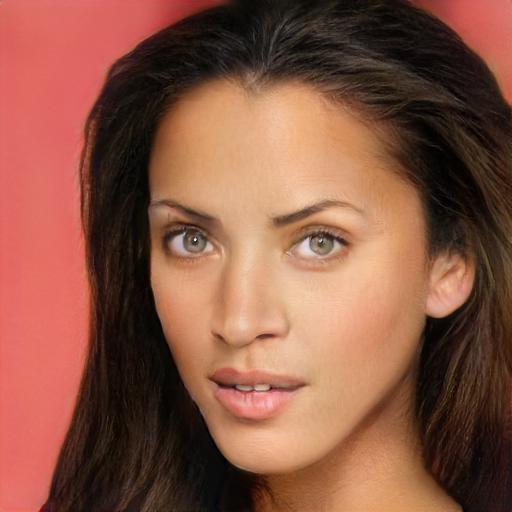} &
    \includegraphics[width=\www]{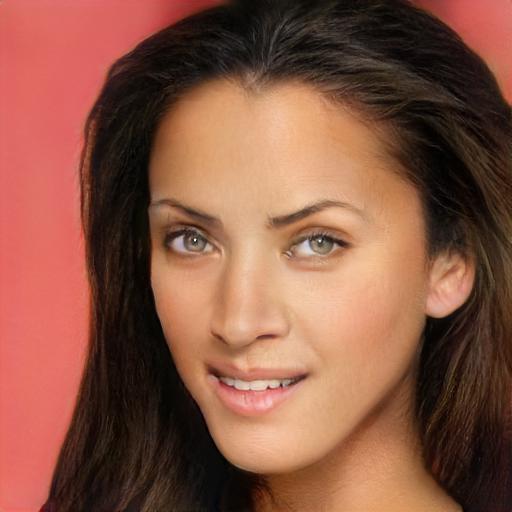} &
    \includegraphics[width=\www]{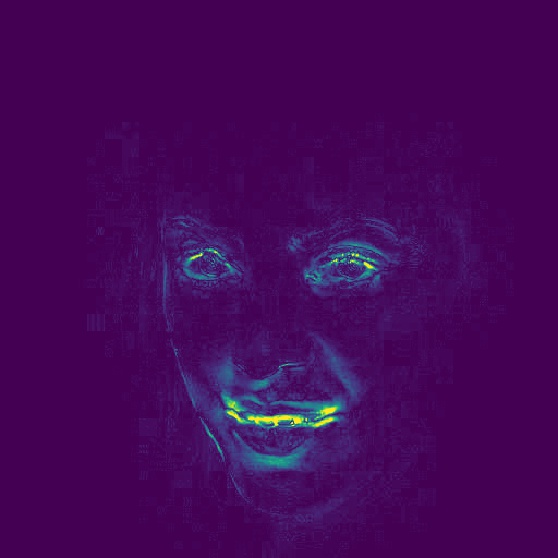} &
    \includegraphics[width=\www]{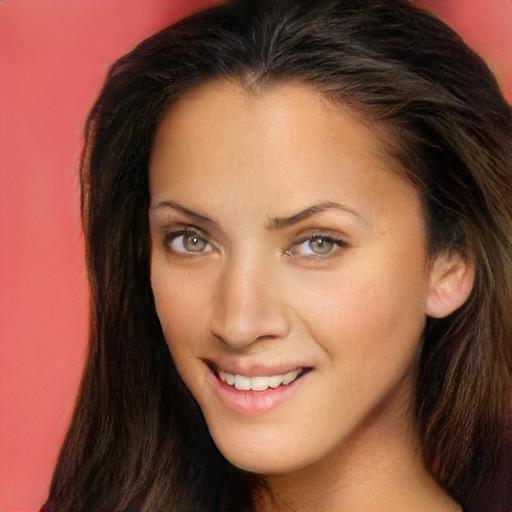} &
    \includegraphics[width=\www]{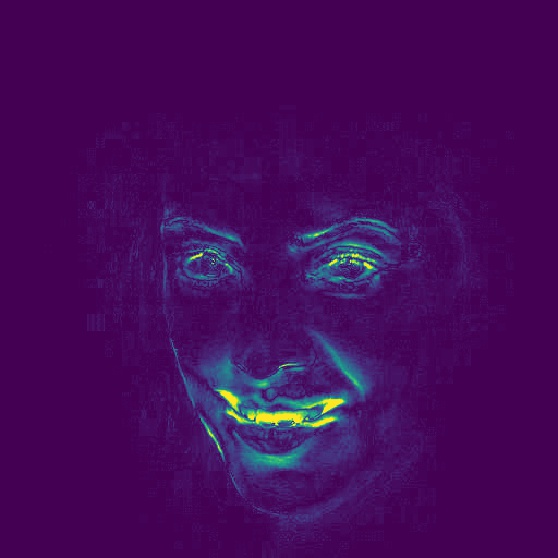}\\
    \raisebox{1.8\height}{\rotatebox[origin=c]{90}{Bald}} & 
    \includegraphics[width=\www]{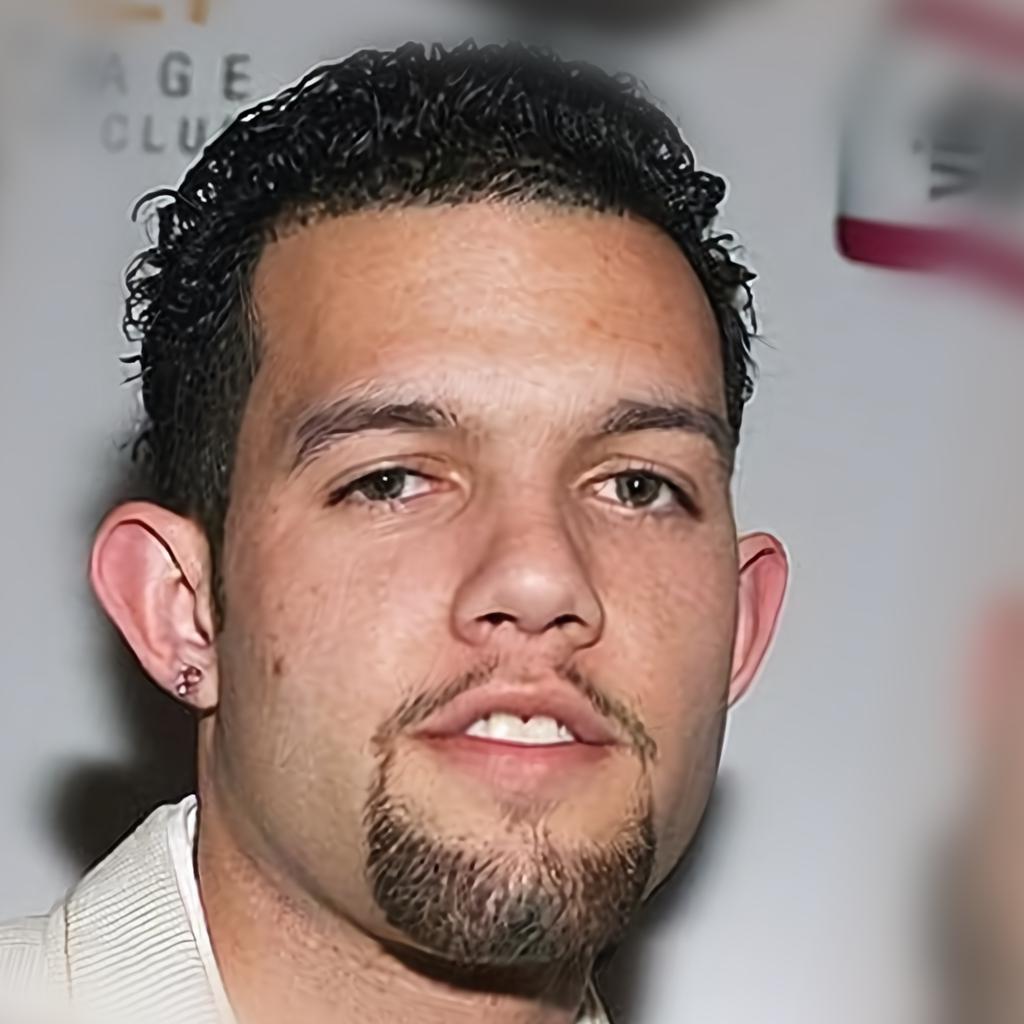} &
    \includegraphics[width=\www]{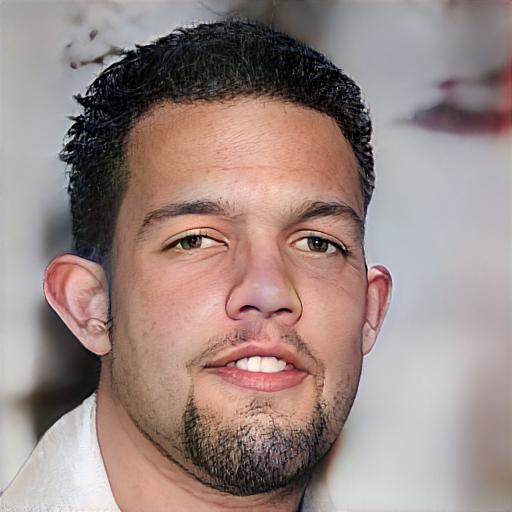} &
    \includegraphics[width=\www]{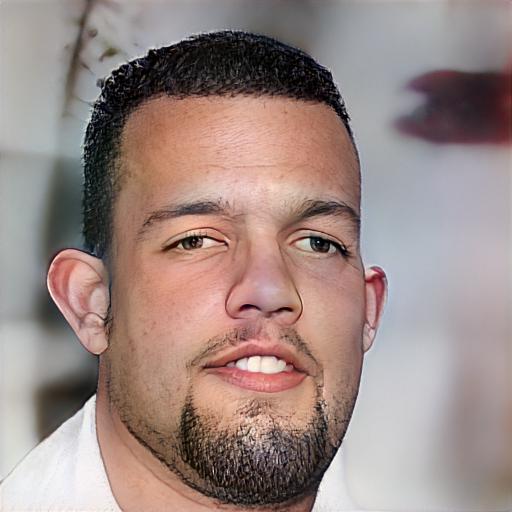} &
    \includegraphics[width=\www]{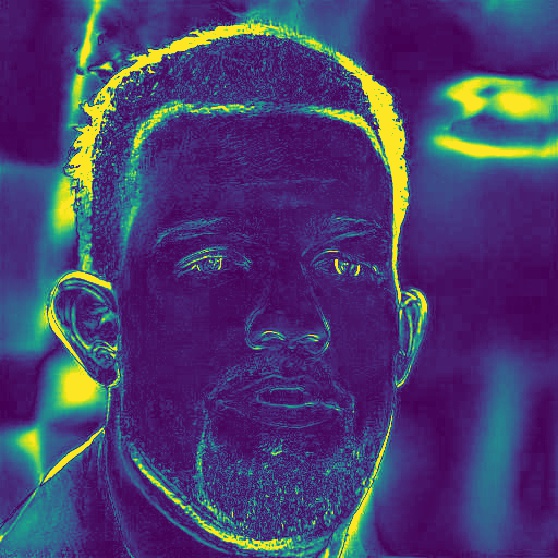} &
    \includegraphics[width=\www]{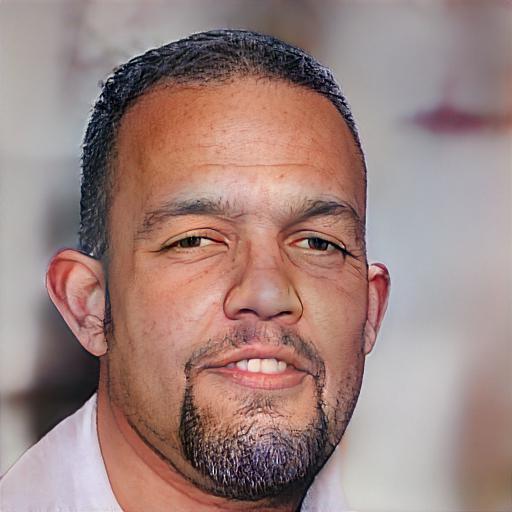} &
    \includegraphics[width=\www]{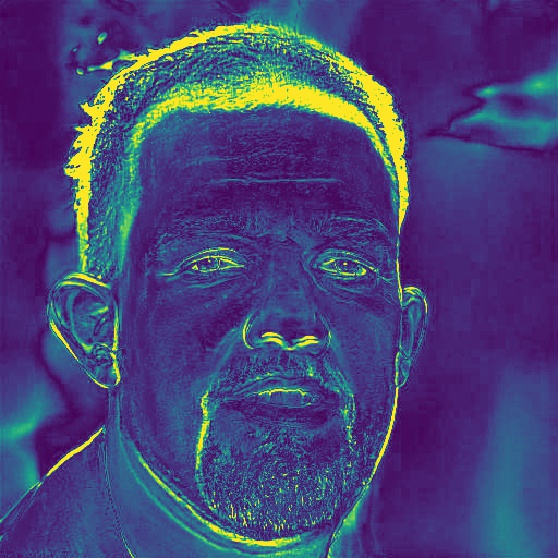} &
    \includegraphics[width=\www]{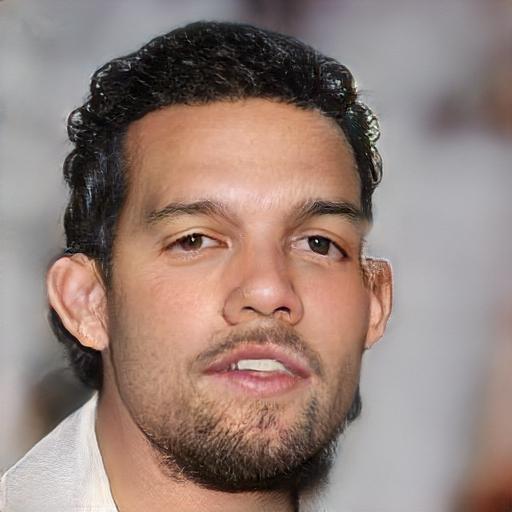} &
    \includegraphics[width=\www]{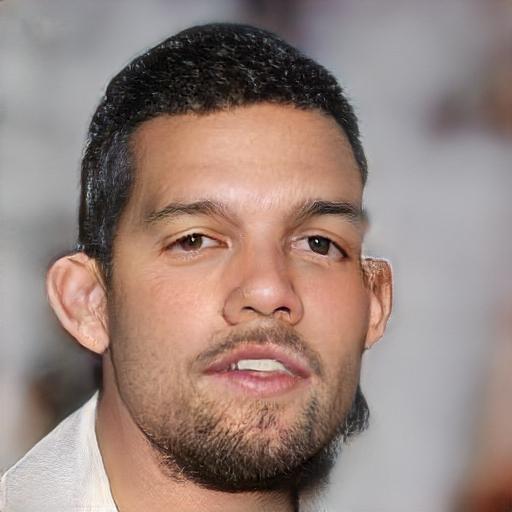} &
    \includegraphics[width=\www]{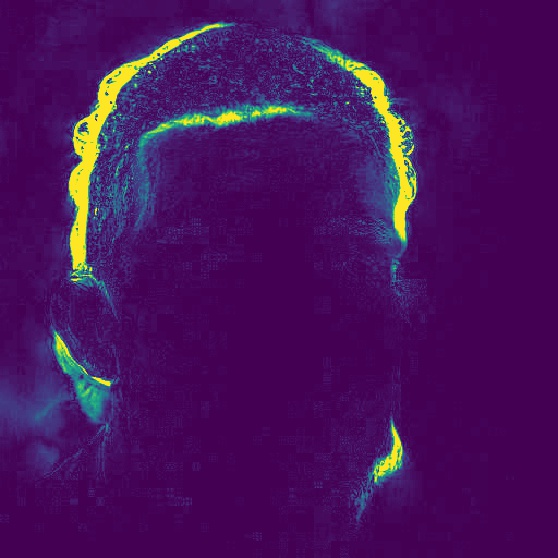} &
    \includegraphics[width=\www]{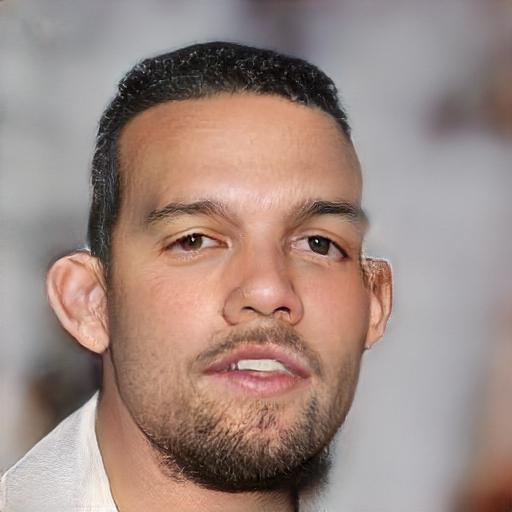} &
    \includegraphics[width=\www]{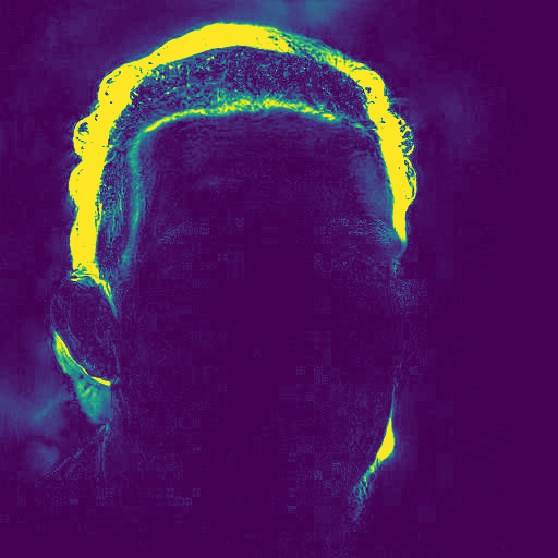}\\
    \raisebox{1.8\height}{\rotatebox[origin=c]{90}{Bangs}} & 
    \includegraphics[width=\www]{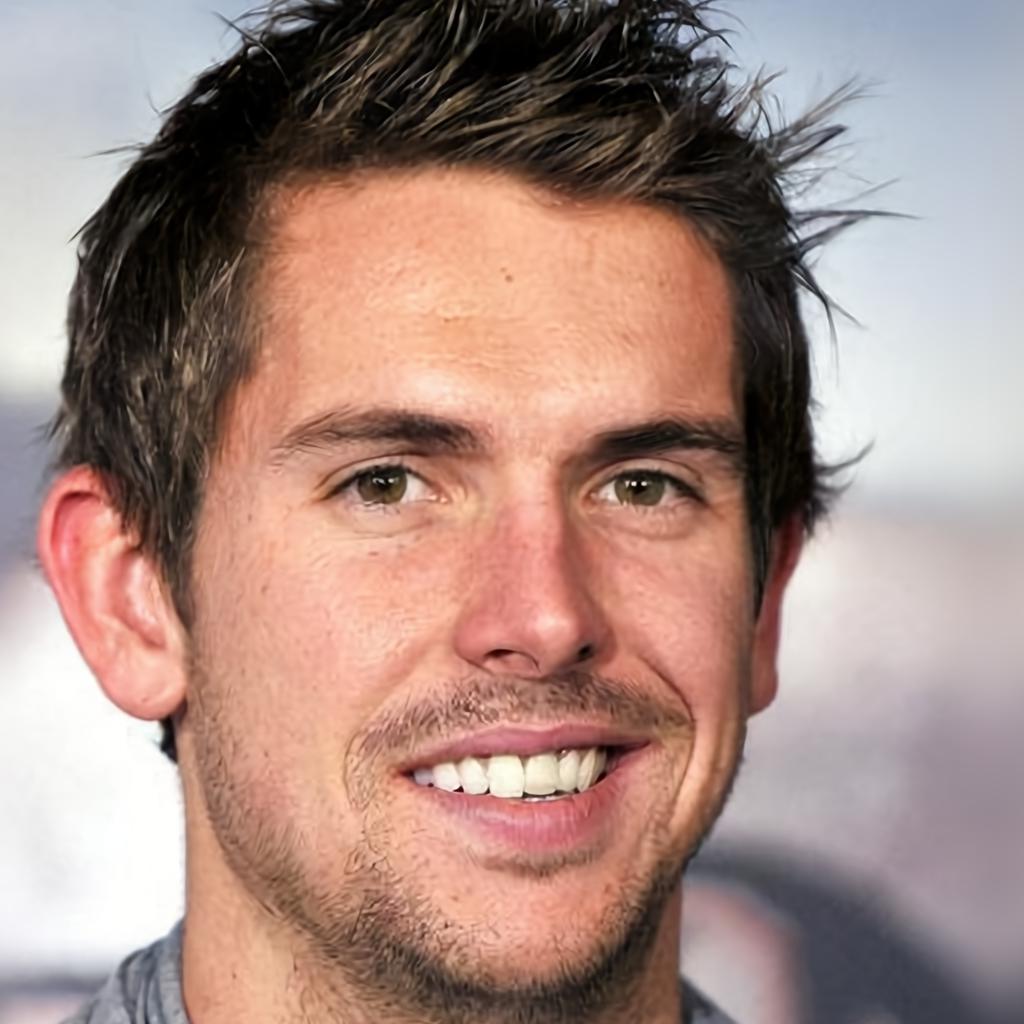} &
    \includegraphics[width=\www]{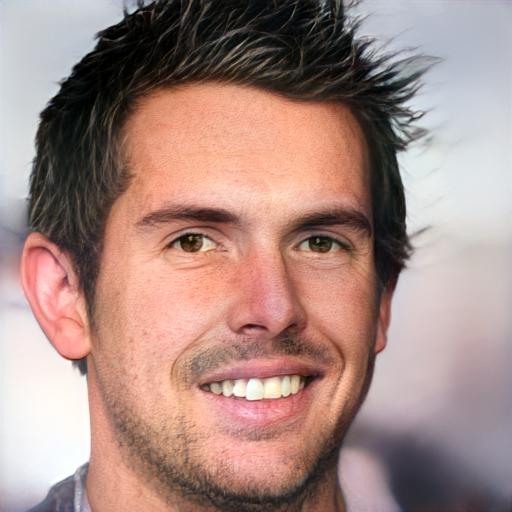} &
    \includegraphics[width=\www]{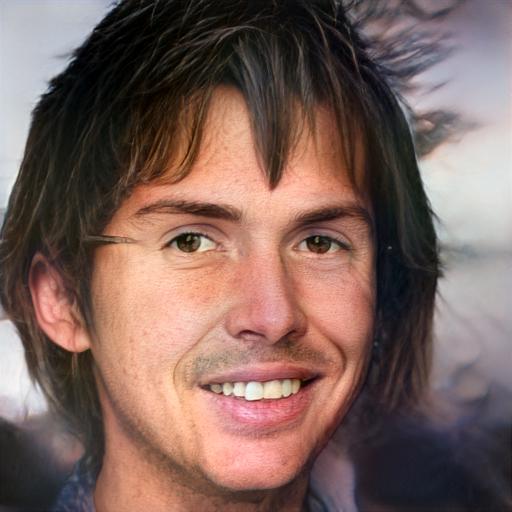} &
    \includegraphics[width=\www]{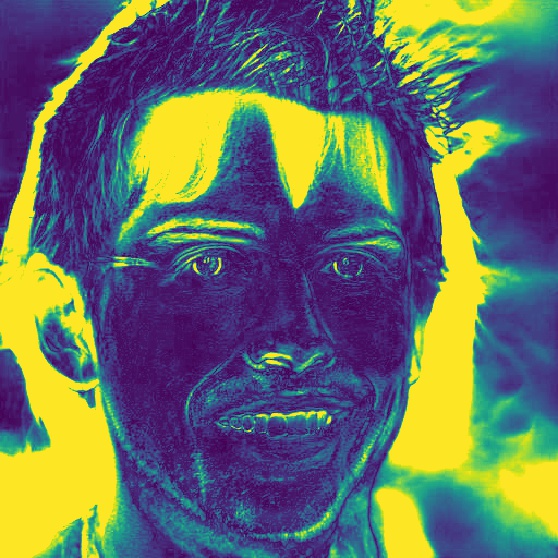} &
    \includegraphics[width=\www]{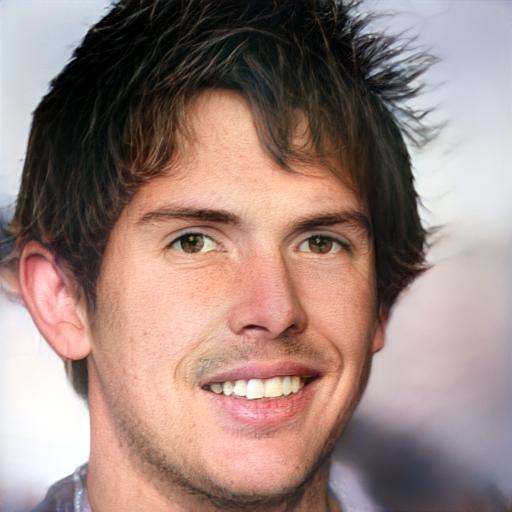} &
    \includegraphics[width=\www]{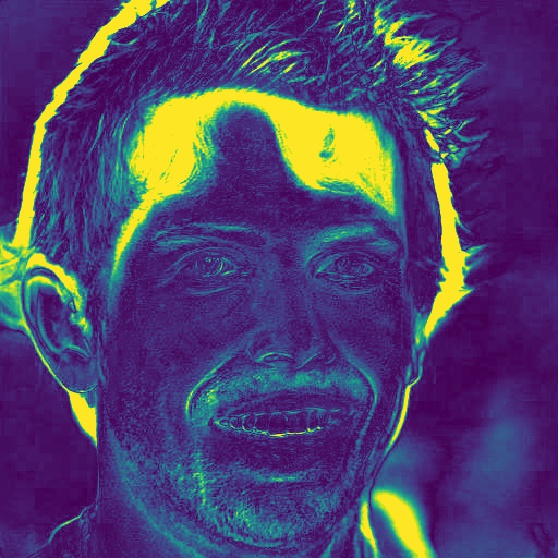} &
    \includegraphics[width=\www]{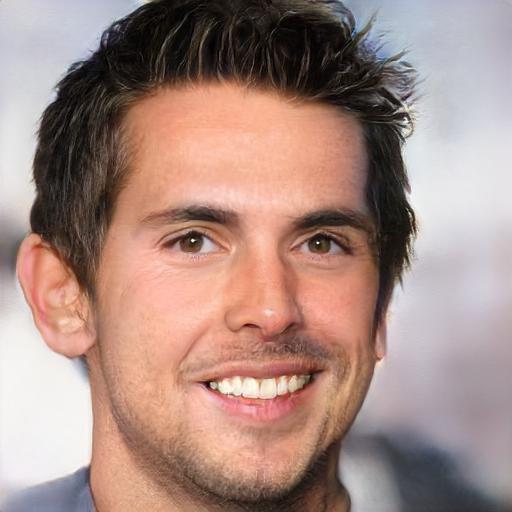} &
    \includegraphics[width=\www]{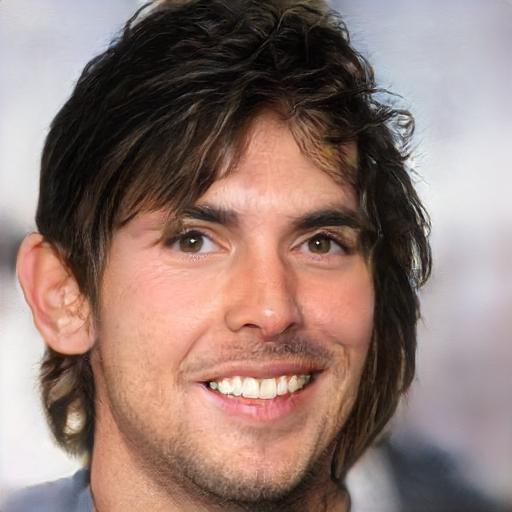} &
    \includegraphics[width=\www]{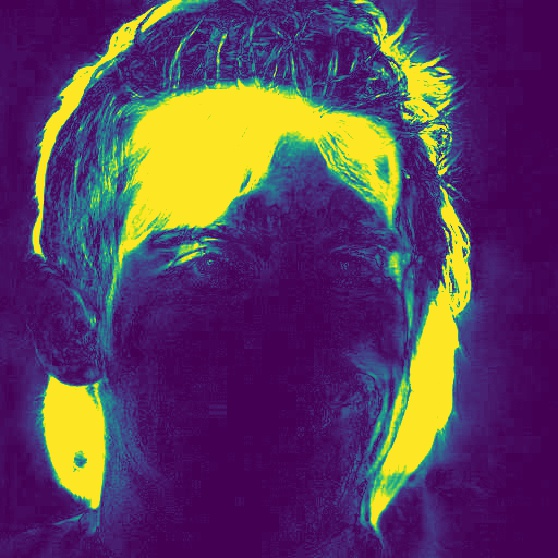} &
    \includegraphics[width=\www]{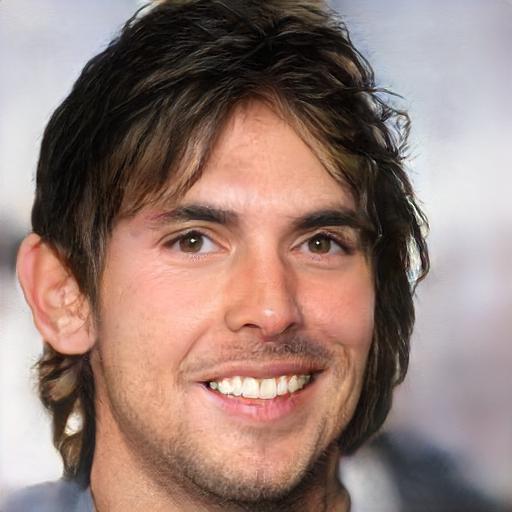} &
    \includegraphics[width=\www]{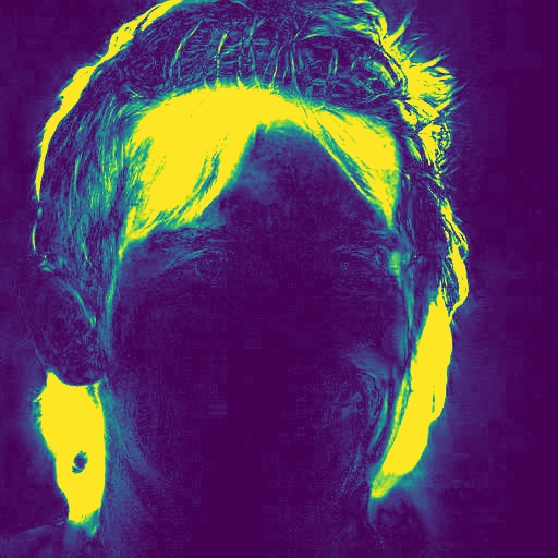}\\
    \raisebox{1.4\height}{\rotatebox[origin=c]{90}{Beard}} & 
    \includegraphics[width=\www]{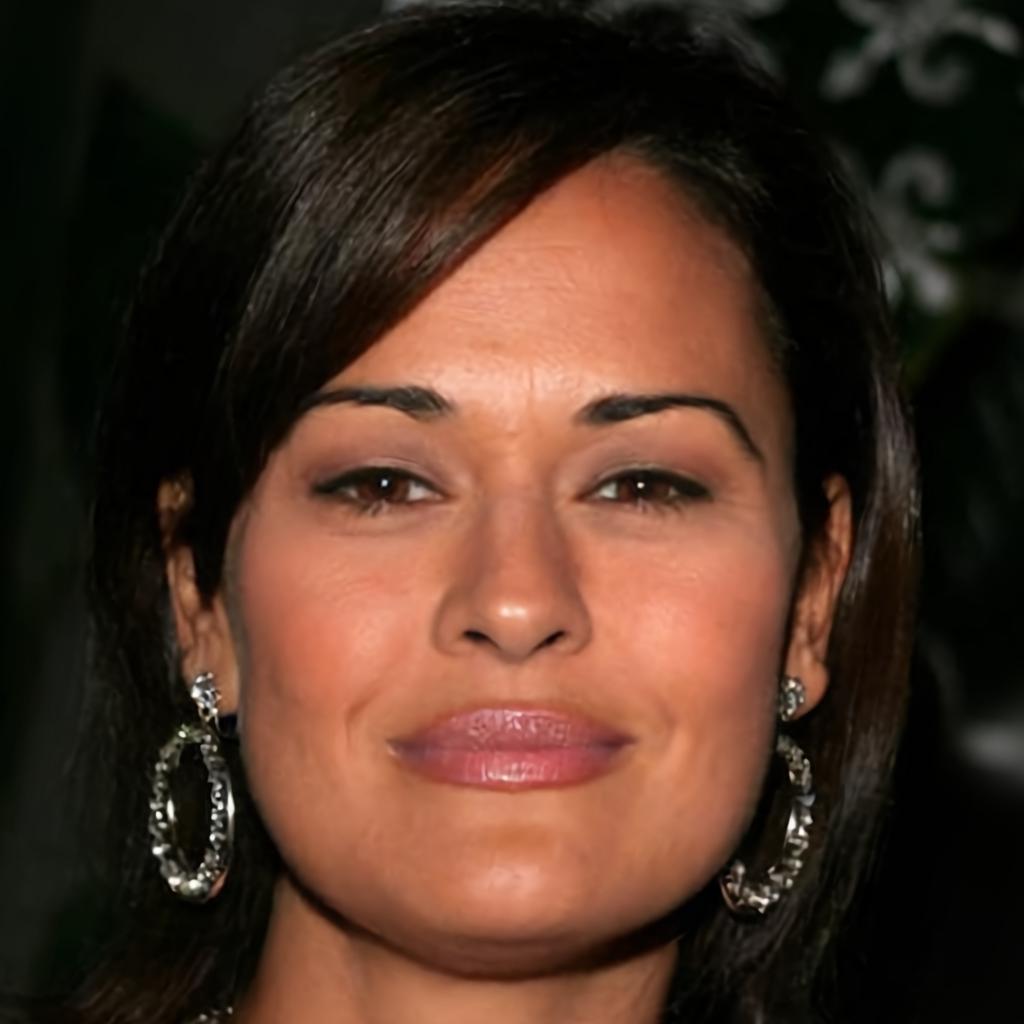} &
    \includegraphics[width=\www]{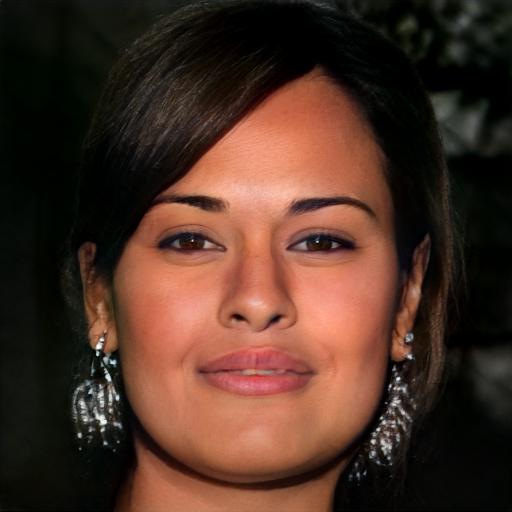} &
    \includegraphics[width=\www]{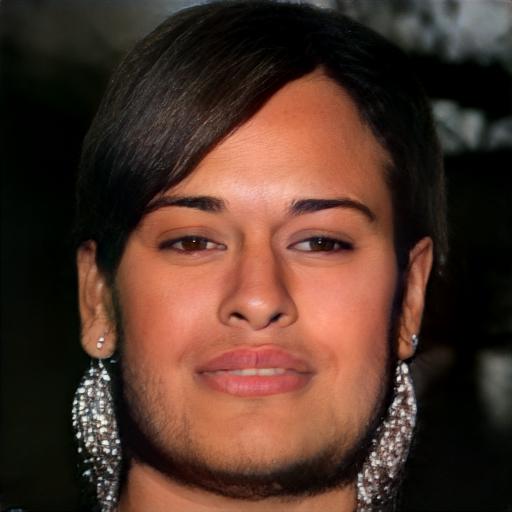} &
    \includegraphics[width=\www]{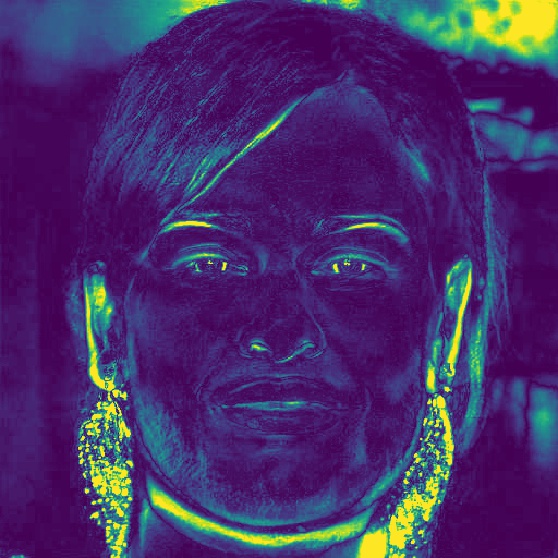} &
    \includegraphics[width=\www]{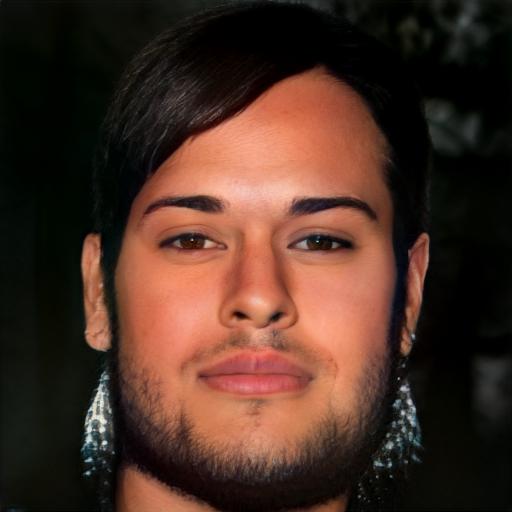} &
    \includegraphics[width=\www]{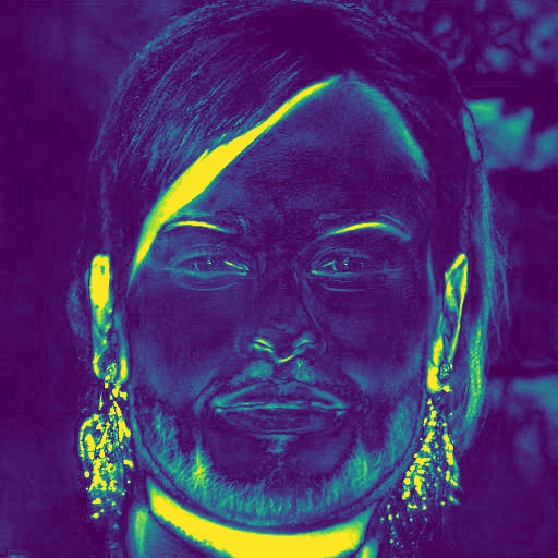} &
    \includegraphics[width=\www]{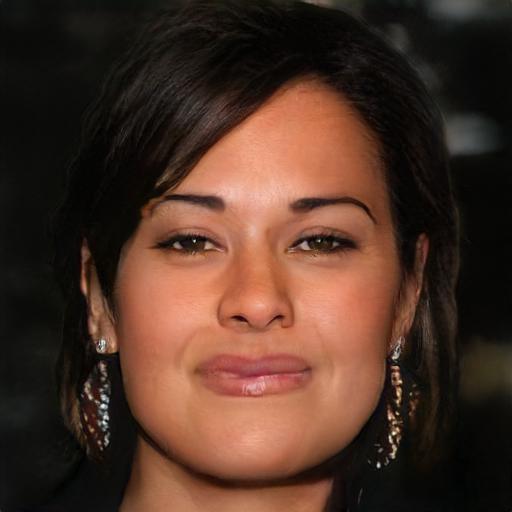} &
    \includegraphics[width=\www]{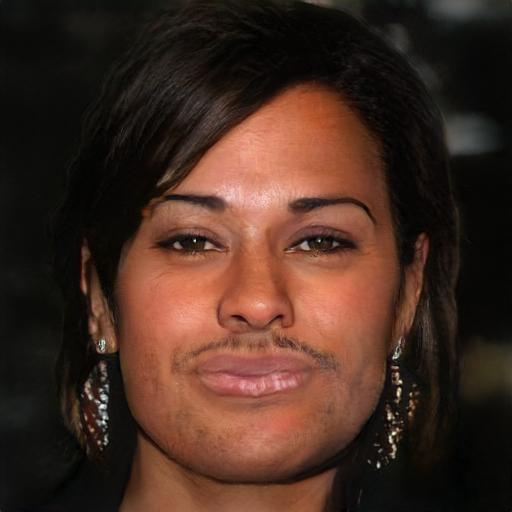} &
    \includegraphics[width=\www]{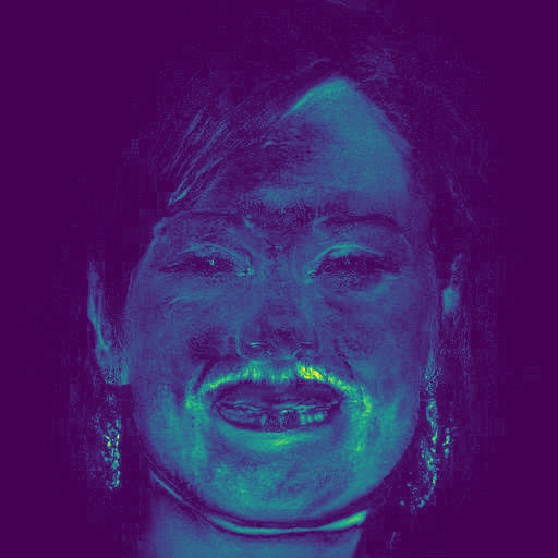} &
    \includegraphics[width=\www]{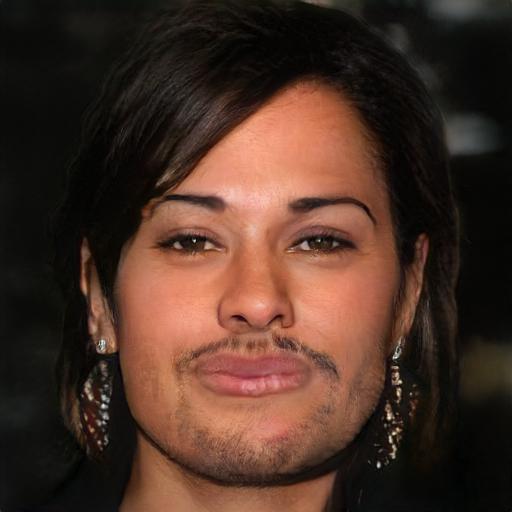} &
    \includegraphics[width=\www]{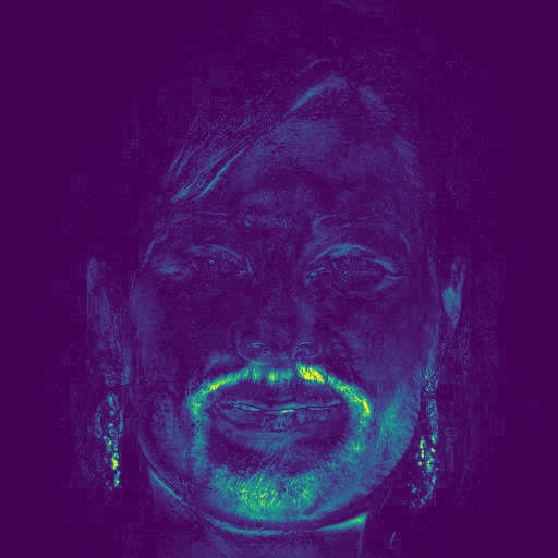}\\
    & \multicolumn{1}{c}{Input} & \makecell{Inversion\\(StyleGAN2)} & \multicolumn{2}{c}{StyleFlow} & \multicolumn{2}{c}{InterFaceGAN}  & \makecell{Inversion\\(Ours)} & \multicolumn{2}{c}{StyleFlow+Ours} & \multicolumn{2}{c}{InterFaceGAN+Ours} \\
\end{tabularx}
    \vspace{-1.0em}\caption{Results of GAN inversion and editing. For each attribute and method, we show the inversion result of Restyle encoder, the edited image and the difference map between them.}\vspace{-0.4em}
    \label{fig:celeba_editing}\vspace{-1.0em}
\end{figure*}

\begin{figure}
    \centering
    \includegraphics[width=1.0\linewidth]{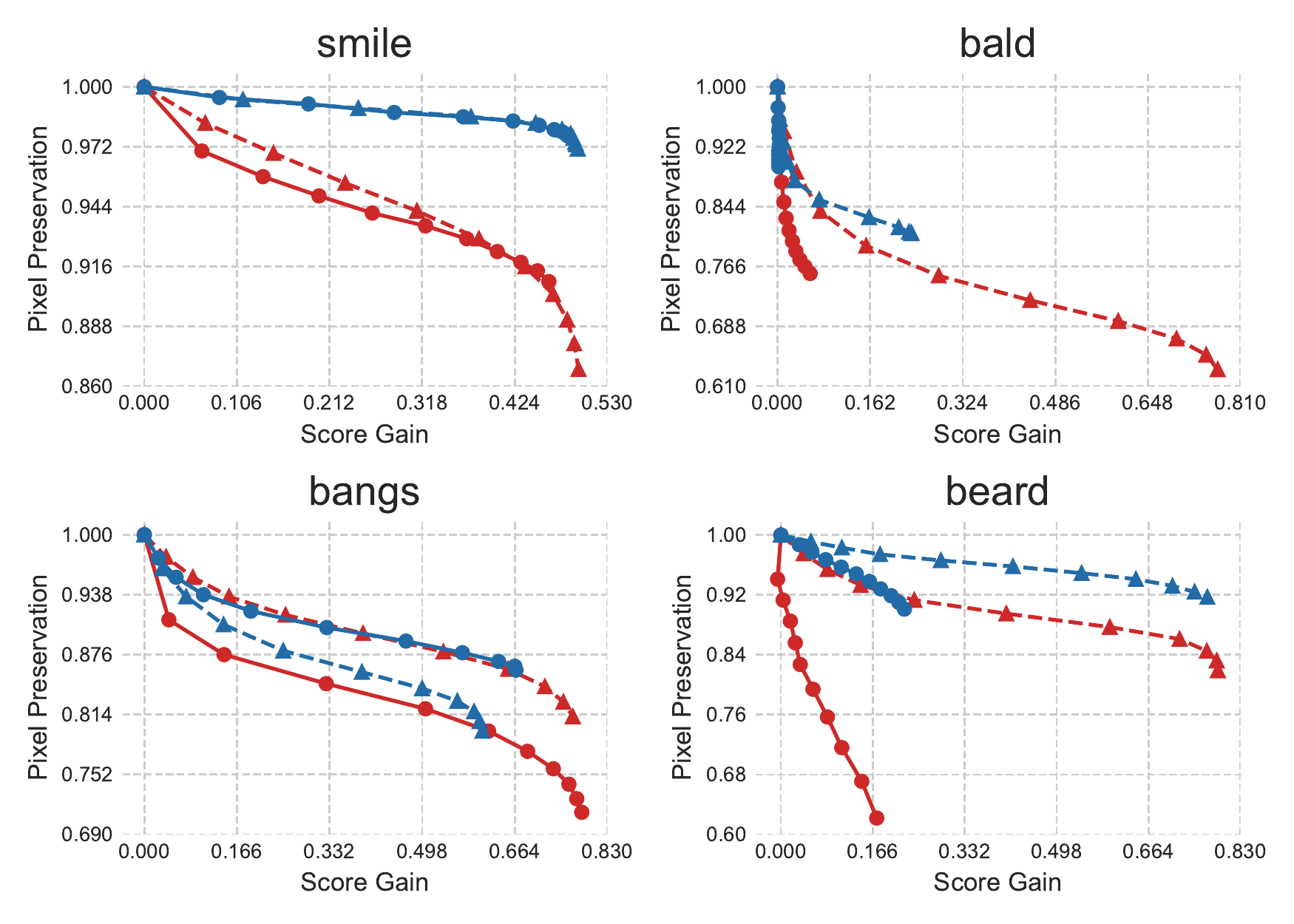}\\[-0.5em]
    \includegraphics[width=1.0\linewidth]{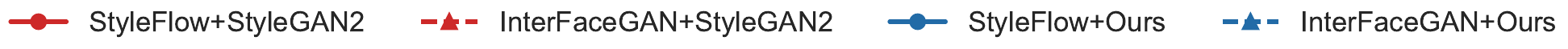}\\
    \vspace{-0.8em}\caption{Quantitative comparison of local attribute editing using StyleGAN2 and our model when combined with StyleFlow and InterFaceGAN.}\vspace{-0.8em}
    \label{fig:disentanglement}
\end{figure}

\subsubsection{Encoding and Editing Real Images}

To evaluate the editing results on real images, we first need to embed such images into the GAN latent space. Here, we adopt a state-of-the-art GAN encoder, i.e. Restyle-psp~\cite{alaluf2021restyle}, for both StyleGAN2 and our model. We use the official model from Restyle authors for StyleGAN2 while a new encoder is trained for our model with default hyper-parameters. For reference, we also train a encoder for our StyleGAN2 that is trained on CelebAMask-HQ. \cref{tab:reconstruction} shows the quantitative results of image reconstruction using restyle encoders. Overall, our model achieves a comparable performance in terms of reconstruction.

The next question is whether our model can be applied to local editing on these reconstructed images. Here, we adopt two popular editing methods that were proposed for StyleGAN2: InterFaceGAN~\cite{shen2020interpreting} and StyleFlow~\cite{abdal2021styleflow}. Both methods need to generate a set of fake images and label their attributes to train a latent manipulation model. In particular, InterFaceGAN learns a linear SVM while StyleFlow uses a conditional continuous normalizing flow~\cite{grathwohl2018ffjord} to model the latent attribute manipulation. For both generators, we randomly synthesize 50,000 images for labeling. Following InterFaceGAN, a ResNet-50~\cite{he2016deep} is trained on CelebA dataset~\cite{liu2015celeba} to label these images. 
During the experiments, we found that our model trained on CelebAMask-HQ exhibits a much lower diversity compared to FFHQ-based StyleGAN2. Thus, we fine-tune our model on FFHQ for $1,000$ steps (See \cref{sec:implementation_details}), for which we observe a sufficient improvement of diversity without loss of controllability.

We choose 4 local attributes covering different parts of the face image for editing experiments, namely smile, baldness, beard and bangs, and test on the last $1,000$ images of CelebAMask-HQ, which were not used for training. For StyleGAN2, we keep the original selection of latent dimensions in these methods for content preservation. For ours, we manually choose relevant areas for editing, \eg hair for baldness and face for beard, which can be regarded as a trivial step during deployment. \cref{fig:celeba_editing} shows the qualitative results of applying InterFaceGAN to StyleGAN2 and our model. Although InterFaceGAN successfully edits the attributes on StyleGAN2, irrelevant parts are inevitably altered due to the entanglement in the latent space. In comparison, our model focuses only on specified semantic areas. We also conduct a quantitative evaluation of the editing task. For each image, we control the degree of manipulation to generate 10 images. Then a ``preservation-score'' curve is plotted using the attribute classifier. Here, $score\;gain$ refers to the average gain in classification score of the target attribute. $pixel\;preservation$ refers to $1$ minus the $\ell1$ loss between the two images. The $\ell1$ loss is an approximation of $\ell0$ loss, which computes the number of pixels that has been altered. In our experiments, we found this simple metric best correlates with the spatial difference between images. From \cref{fig:disentanglement}, it can be seen that our model achieves a better overall performance. Note that for baldness, our model stops when it removes all hairs, but InterFaceGAN+StyleGAN2 keeps increasing the score by adapting into correlated attributes (such as aging). For bangs, our model tends to increase the overall length of hairs, which could be an inherited bias from original training data. Besides, we found that StyleFlow is more sensitive to label imbalance. Thus, given the small number of bald examples, it fails to learn the baldness attribute for both generators.

\begin{figure}[t]
\captionsetup{font=small}
\centering
\footnotesize
\setlength\tabcolsep{0pt}
\newcommand{\www}{0.244\linewidth}
\renewcommand{\arraystretch}{0.2}
\newcolumntype{Y}{>{\centering\arraybackslash}X}
\begin{tabularx}{\linewidth}{p{6pt} cccc}
    & \makecell{initial\\ face} & \makecell{``a person with \\brown skin''} & \makecell{``a person with \\purple long hair''} & \makecell{``a person with \\blue eyes''}  \\
    \raisebox{1.5\height}{\rotatebox[origin=c]{90}{\scriptsize StyleCLIP}} & 
    \includegraphics[width=\www]{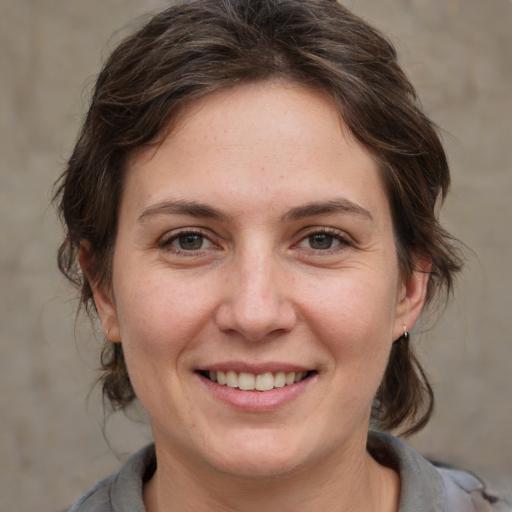} & 
    \includegraphics[width=\www]{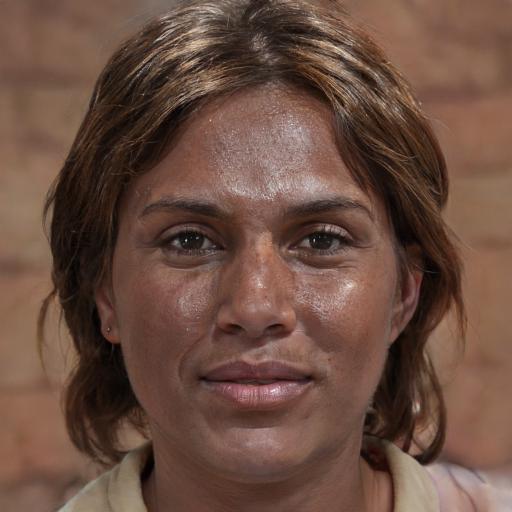} &
    \includegraphics[width=\www]{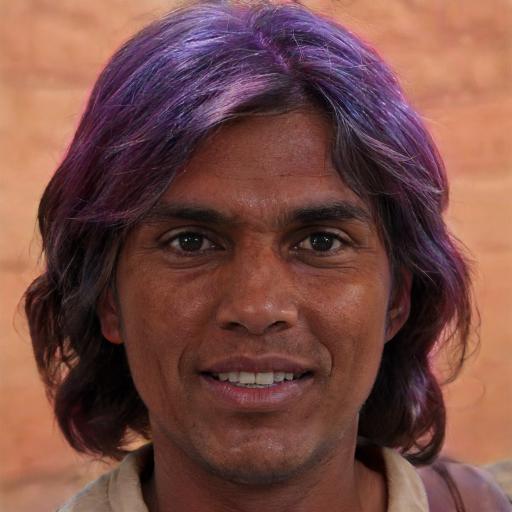} &
    \includegraphics[width=\www]{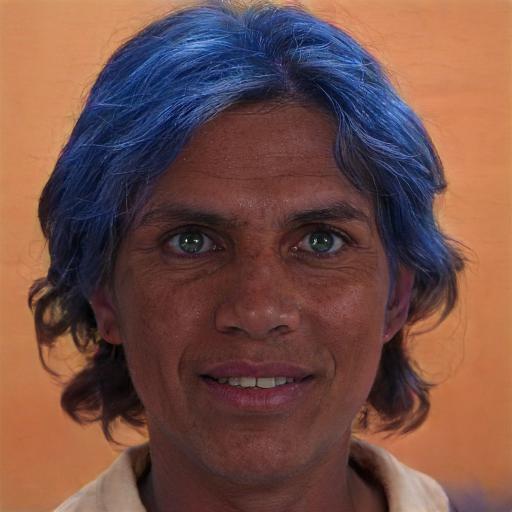}\\
    \raisebox{1.0\height}{\rotatebox[origin=c]{90}{\scriptsize  StyleCLIP + Ours}} & 
    \includegraphics[width=\www]{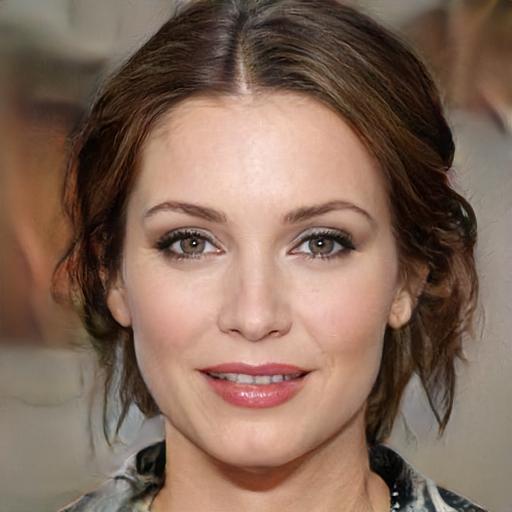} & 
    \includegraphics[width=\www]{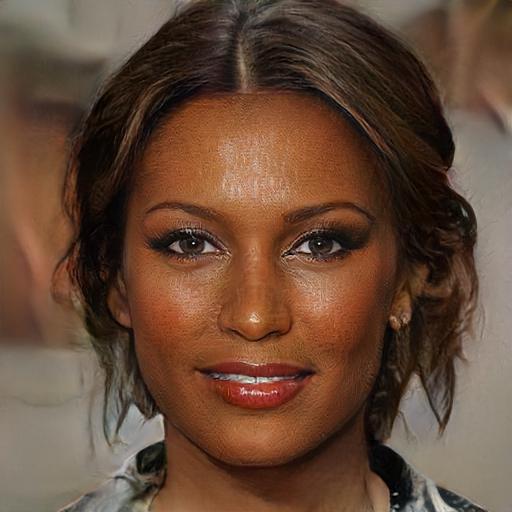} &
    \includegraphics[width=\www]{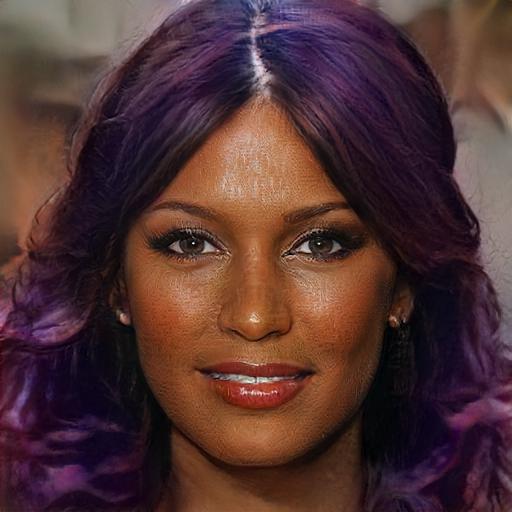} &
    \includegraphics[width=\www]{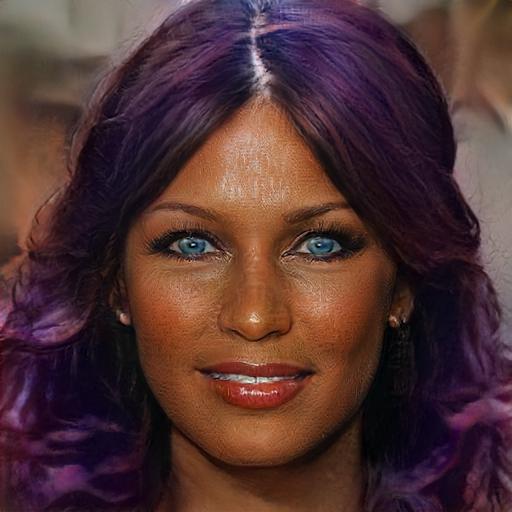}\\
\end{tabularx}
    \vspace{-1.0em}\caption{Results of text-guided image synthesis under sequential editing. Starting from an average fake face, the first row (from left to right) shows the results of sequentially applying optimization-based StyleCLIP~\cite{patashnik2021styleclip} with StyleGAN2 while the second row shows the results of our model with the same input texts. }\vspace{-1.0em}
    \label{fig:styleclip}
\end{figure}

\subsubsection{Text-guided Synthesis}
Recent work have shown that one could use a text-image embedding, such as CLIP~\cite{radford2021CLIP}, to guide the synthesis of StyleGAN2 for controlled synthesis~\cite{patashnik2021styleclip}. Similar to attribute editing, StyleGAN2 suffers from the local disentanglement problem on. \cref{fig:styleclip} shows a few examples of using StyleCLIP~\cite{patashnik2021styleclip} to manipulate a synthesized image with a sequence of text prompts. Here, we use the optimization-based version of StyleCLIP as it is flexible for any input text. It can be seen that the original StyleCLIP often modifies the whole image while the text is trying to change only a specified area. Our model, by additionally let the user to choose relevant areas, can faithfully constrain the editing to local parts. The results indicate that our model could be a more suitable tool for text-guided portrait synthesis where detailed descriptions are provided.

\begin{figure}[t]
\captionsetup{font=small}
\centering
\scriptsize
\setlength\tabcolsep{1px}
\newcommand{\www}{0.165\linewidth}
\renewcommand{\arraystretch}{0.1}
\newcolumntype{Y}{>{\centering\arraybackslash}X}
\begin{tabularx}{\linewidth}{p{5pt}c}
    \raisebox{1.4\height}{\rotatebox[origin=c]{90}{Photos}} & 
    \includegraphics[width=\www]{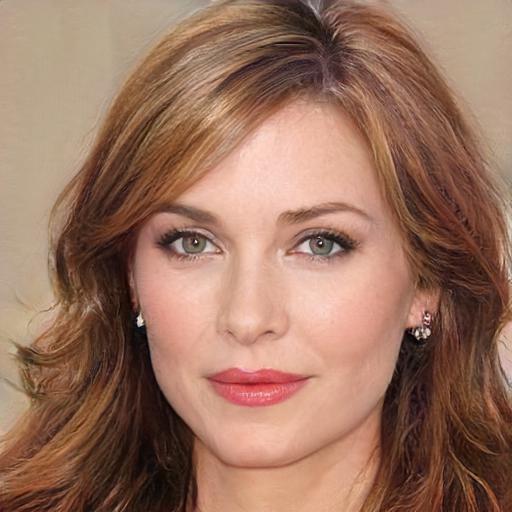}\hfill
    \includegraphics[width=\www]{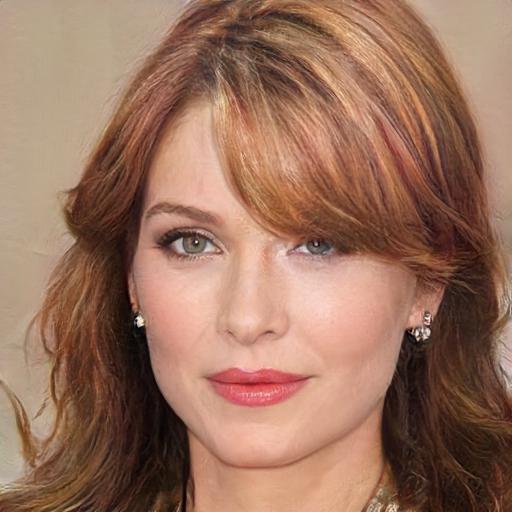}\hfill
    \includegraphics[width=\www]{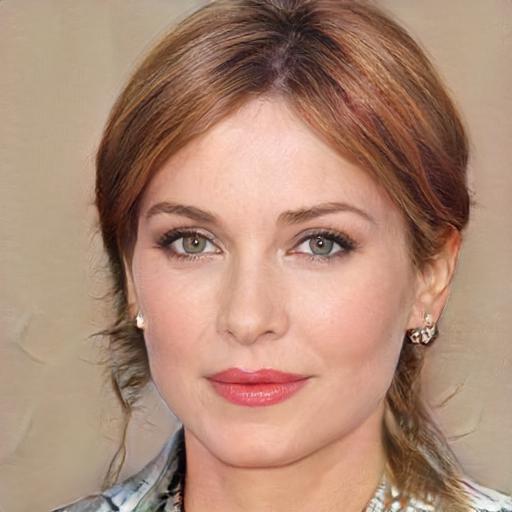}\hfill
    \includegraphics[width=\www]{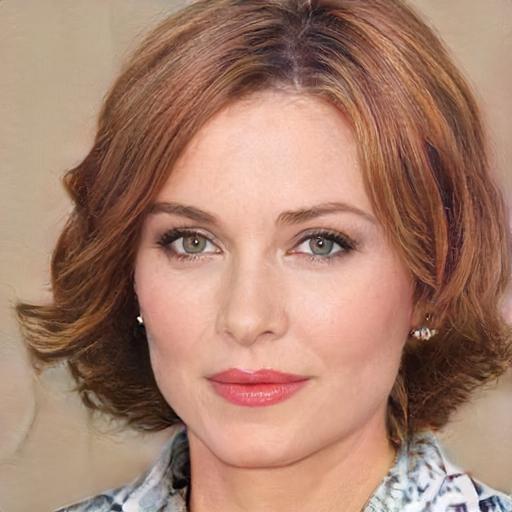}\hfill
    \includegraphics[width=\www]{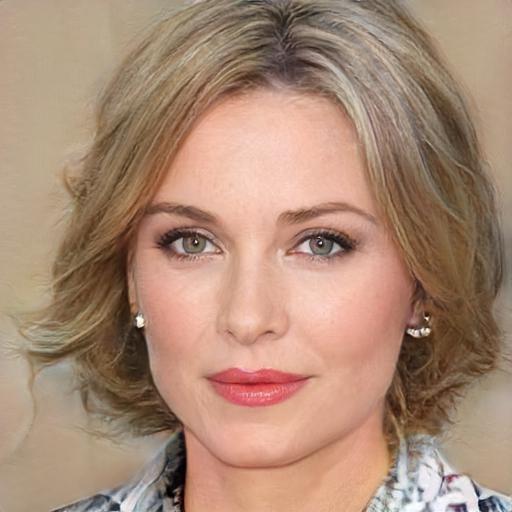}\hfill
    \includegraphics[width=\www]{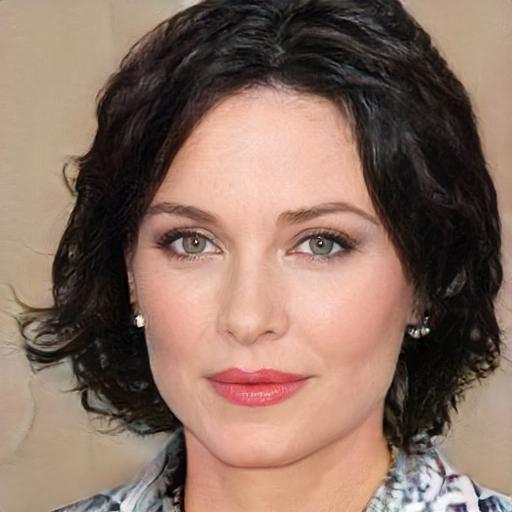}\\
    \raisebox{1.3\height}{\rotatebox[origin=c]{90}{Toonify}} &
    \includegraphics[width=\www]{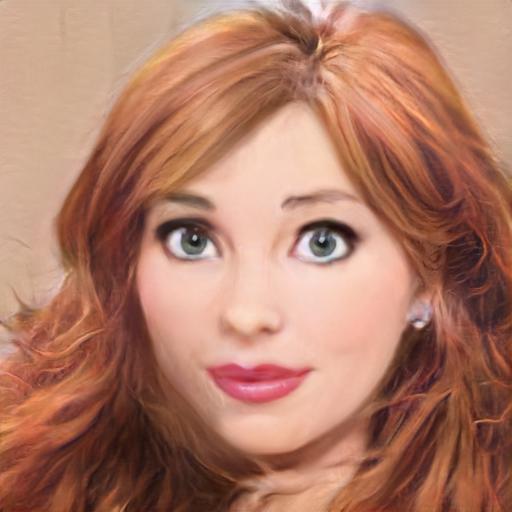}\hfill
    \includegraphics[width=\www]{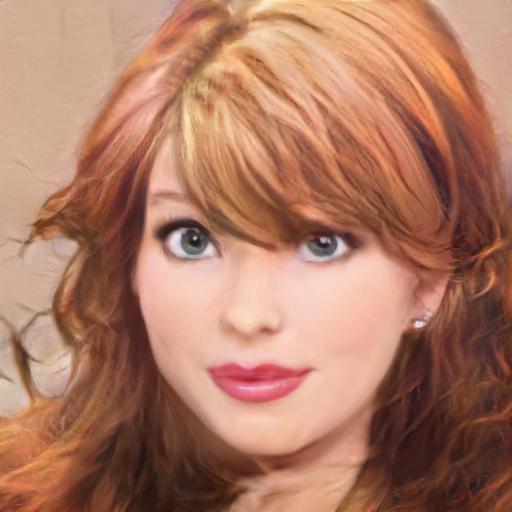}\hfill
    \includegraphics[width=\www]{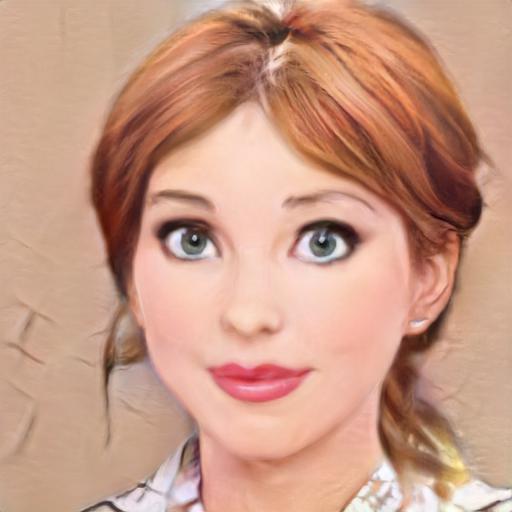}\hfill
    \includegraphics[width=\www]{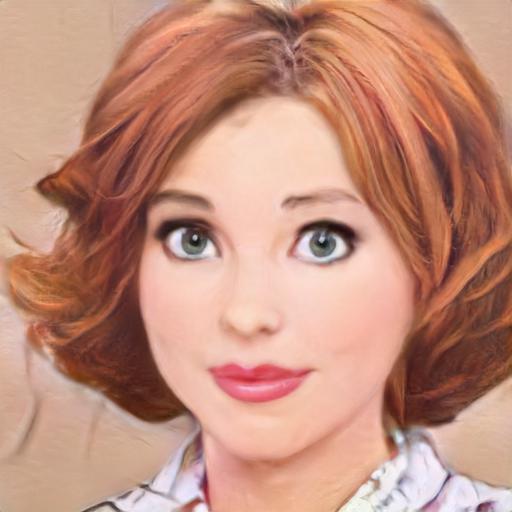}\hfill
    \includegraphics[width=\www]{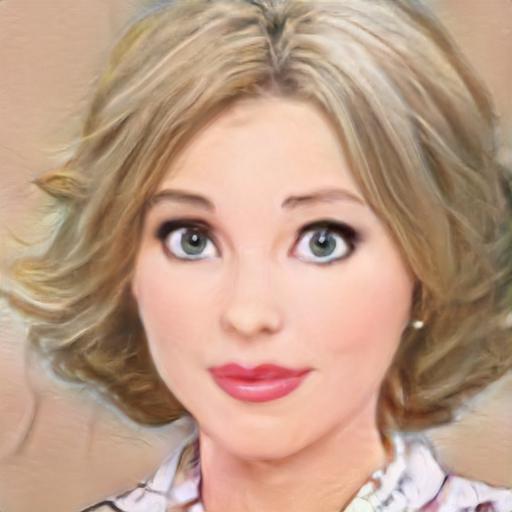}\hfill
    \includegraphics[width=\www]{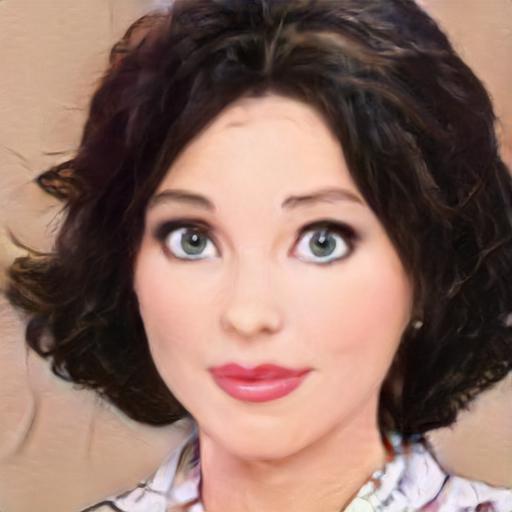}\\
    \raisebox{1.1\height}{\rotatebox[origin=c]{90}{MetFaces}} &
    \includegraphics[width=\www]{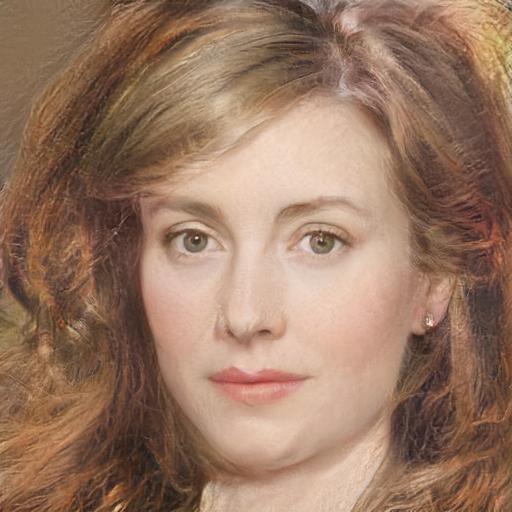}\hfill
    \includegraphics[width=\www]{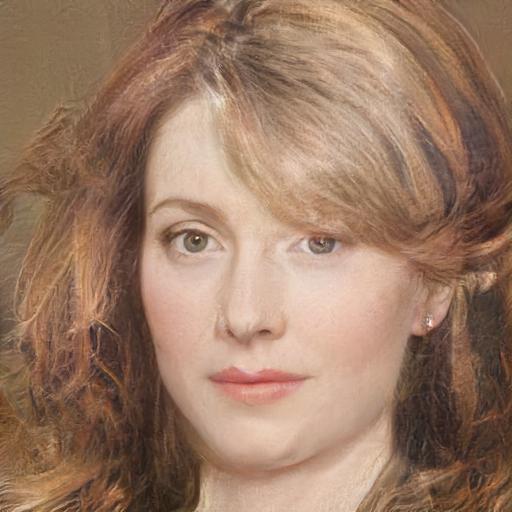}\hfill
    \includegraphics[width=\www]{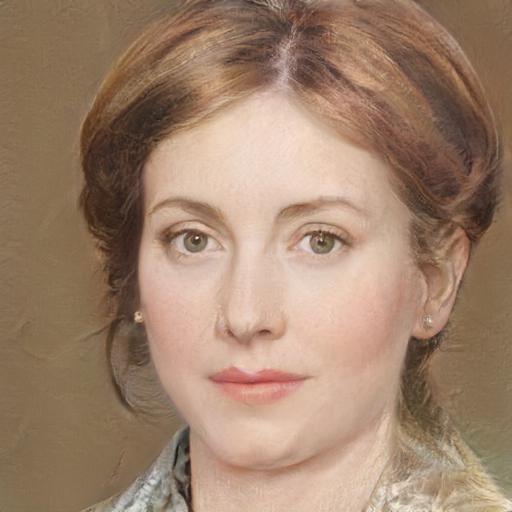}\hfill
    \includegraphics[width=\www]{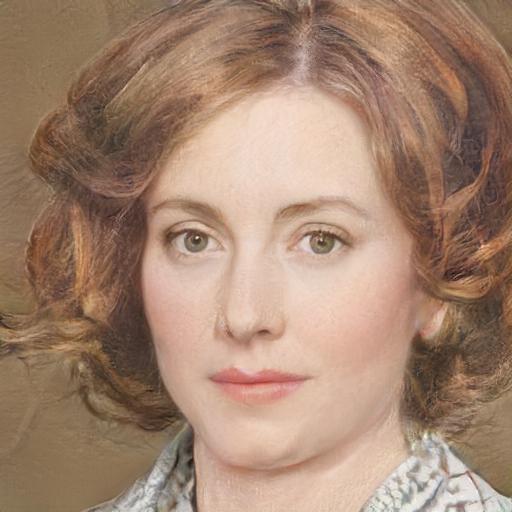}\hfill
    \includegraphics[width=\www]{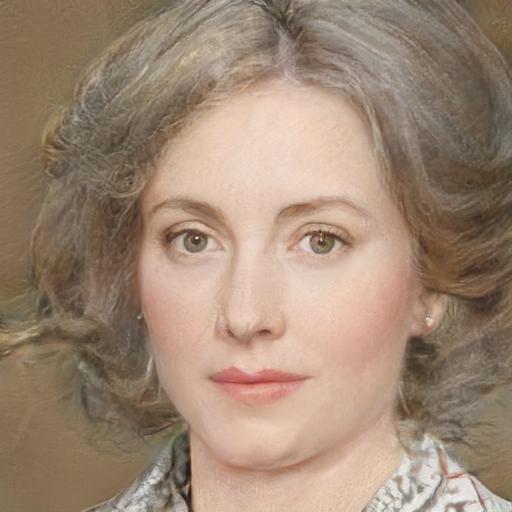}\hfill
    \includegraphics[width=\www]{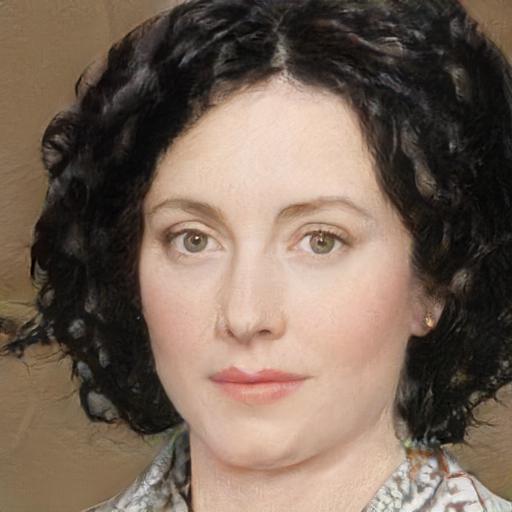}\\
    \raisebox{1.3\height}{\rotatebox[origin=c]{90}{Bitmoji}} &
    \includegraphics[width=\www]{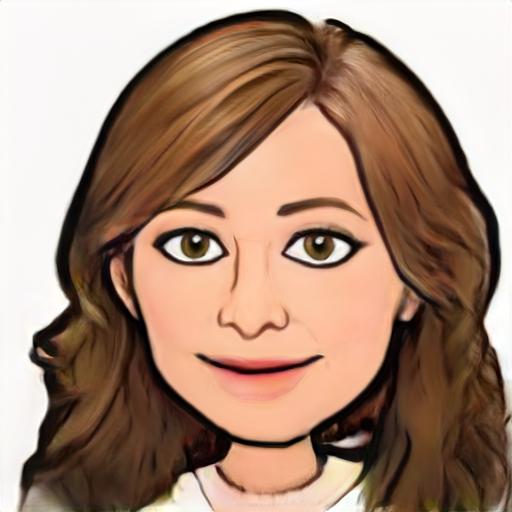}\hfill
    \includegraphics[width=\www]{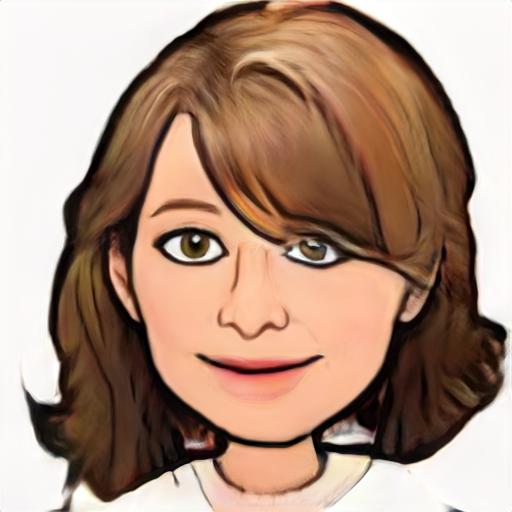}\hfill
    \includegraphics[width=\www]{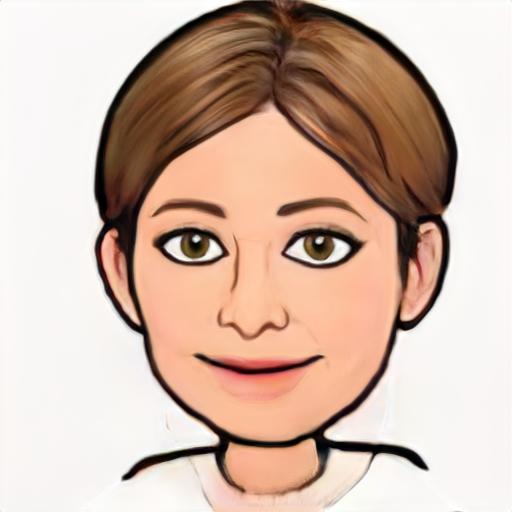}\hfill
    \includegraphics[width=\www]{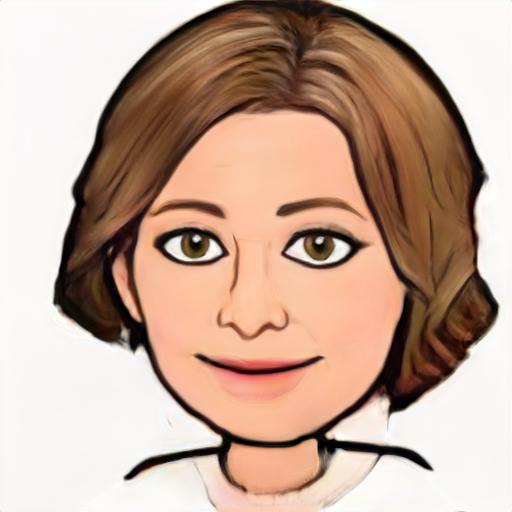}\hfill
    \includegraphics[width=\www]{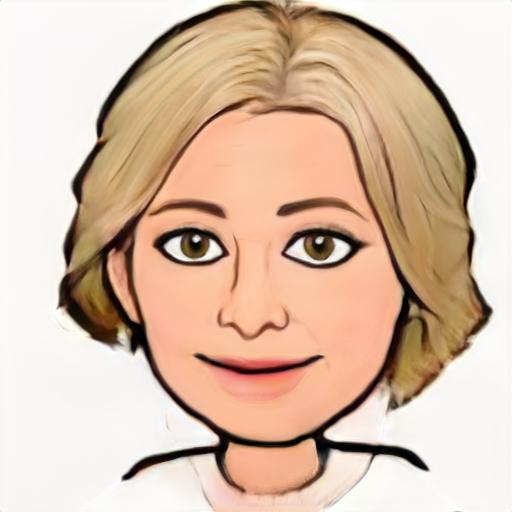}\hfill
    \includegraphics[width=\www]{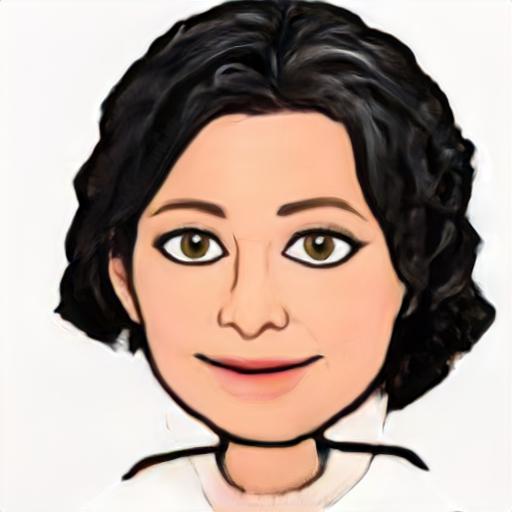}\\
\end{tabularx}
    \vspace{-0.2em}\caption{Examples of changing hair styles on adapted new domains. The first four and last three columns show the results of different latent codes for hair shapes and textures, respectively.}\vspace{-0.3em}
    \label{fig:multidomain}
\end{figure}

\begin{figure}[t]
\captionsetup{font=small}
\centering
\scriptsize
\setlength\tabcolsep{1px}
\newcommand{\www}{0.192\linewidth}
\renewcommand{\arraystretch}{0.1}
\newcolumntype{Y}{>{\centering\arraybackslash}X}
\begin{tabularx}{\linewidth}{Y>{\centering\arraybackslash}c}
    \raisebox{2.4\height}{\rotatebox[origin=c]{90}{Hair}} & 
    \includegraphics[width=\www]{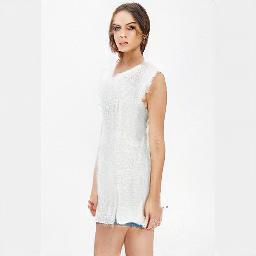}\hfill
    \includegraphics[width=\www]{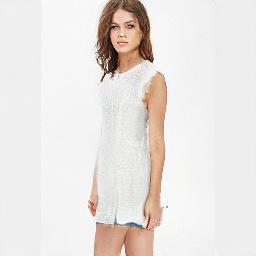}\hfill
    \includegraphics[width=\www]{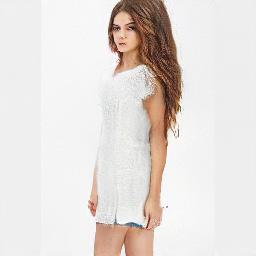}\hfill
    \includegraphics[width=\www]{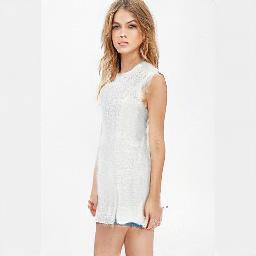}\hfill
    \includegraphics[width=\www]{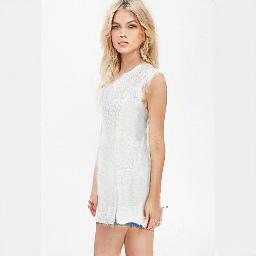}\\
    \raisebox{2.6\height}{\rotatebox[origin=c]{90}{Top}} & 
    \includegraphics[width=\www]{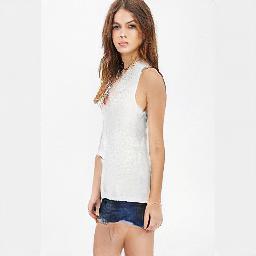}\hfill
    \includegraphics[width=\www]{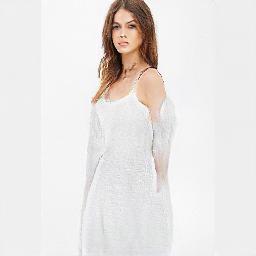}\hfill
    \includegraphics[width=\www]{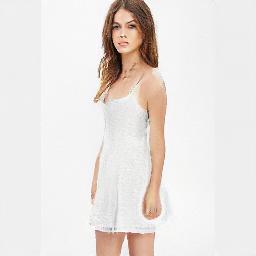}\hfill
    \includegraphics[width=\www]{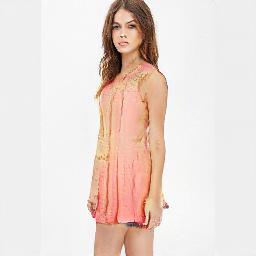}\hfill
    \includegraphics[width=\www]{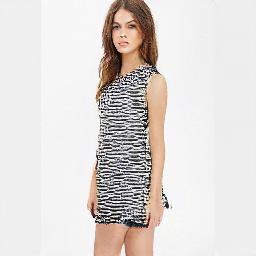}\\
    \raisebox{1.5\height}{\rotatebox[origin=c]{90}{Bottom}} & 
    \includegraphics[width=\www]{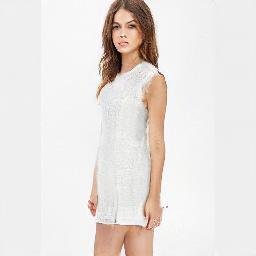}\hfill
    \includegraphics[width=\www]{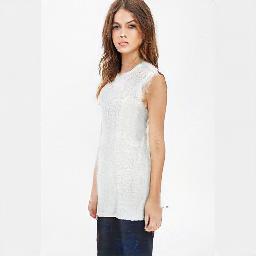}\hfill
    \includegraphics[width=\www]{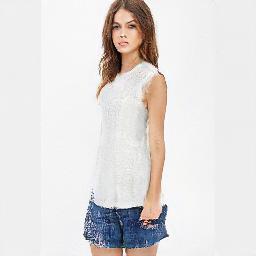}\hfill
    \includegraphics[width=\www]{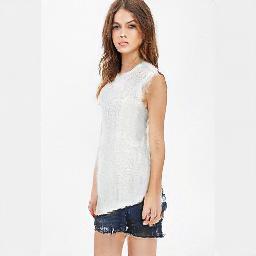}\hfill
    \includegraphics[width=\www]{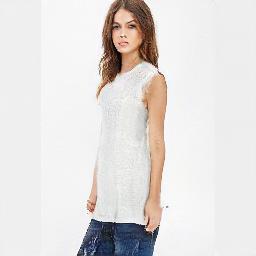}\\
\end{tabularx}
    \vspace{-0.2em}\caption{Controlled generation results on the DeepFashion dataset. Our model can generate various style for different parts.}\vspace{-0.5em}
    \label{fig:traverse_deepfashion}
\end{figure}

\subsection{Results on Other Domains}
\label{sec:exp:other_domains}
Training our model from scratch requires access to images and segmentation masks at the same time, which might not be feasible in some cases. Thus, we would like to ask whether the model can be fine-tuned on image-only datasets while preserving the local disentanglement (See~\cref{sec:implementation_details} for fine-tuning). \cref{fig:multidomain} shows the results after fine-tuning our model on the Toonify~\cite{pinkney2020toonify}, MetFaces dataset~\cite{karras2020stylegan_ada} and BitMoji~\cite{BitMoji}. All of these datasets have a much smaller number of images compared to CelebAMask-HQ and no segmentation masks. We train our model for hundreds of steps until perceptually good results are generated. It can be seen that, for datasets with a limited domain gap, our model is able to maintain local controllability even after fine-tuning. 

In spite of the experiments on face datasets so far, our method indeed does not include any module that is designed for face only and hence can be applied to other objects as well. \cref{fig:traverse_deepfashion} shows the results of training our model on the DeepFashion dataset~\cite{liu2016deepfashion}, for which we obtain the labels from~\cite{zhu2020semantically}. With the default hyper-parameters, we find that our model can be successfully trained on fashion datasets and we can similarly control the structure and texture of different semantic parts in the latent space.

\section{Limitations and Discussion}
\paragraph{Applicable Datasets} Although we have shown that our method can be applied to other domains beyond face photos, we still see a limitation caused by the design and supervision. Since we need to build a local generator for each class, the method would not scale to datasets that have too many semantic classes, such as scenes~\cite{zhou2017ADE20k}. Besides, for the purpose of synthesis quality, we change the semi-supervised framework of SemanticGAN~\cite{li2021SemanticGAN} into fully-supervised, which limits our model from training on image-only datasets from scratch. It would be beneficial to develop a semi-supervised version of our method in the future.

\vspace{-0.5em}\paragraph{Disentanglement}
As the disentanglement between pose, shape and texture is only enforced by the design of layer separation in local generators, we see that the boundary between them is still sometimes ambiguous. For example, the shared coarse structure code could encode some information about expression and the shape code could affect the beard. However, in this work, we mainly focus on the spatial disentanglement between different semantic parts and we believe additional regularization losses or architecture tuning could be incorporated in the future to better decouple those information.

% The disentanglement of structure and texture in our method is achieved by the inductive bias in the local generator, where additional layer are modulated to output the feature map. Such a design is indeed borrowed from prior work on controllable generation~\cite{schwarz2020graf,niemeyer2021giraffe}. However, we observe that such a design still cannot guarantee the complete disentanglement between structure and texture. In the future, we would like to explore how to achieve a better disentanglement between these two such that we may animate a face by controlling its structure. Here, we note that a simple solution is to cut off the connection between structure layers and texture layers in the local generator such that structure layer would only controls the semantic segmentation. However, although it is achievable, we found it slightly hurts the synthesis quality and therefore keep the current architecture to guarantee the prior of image quality.

\vspace{-0.5em}\paragraph{Societal Impact}
Our work focuses on the technical problem of improving controllability of GANs and is not specifically designed for any malicious uses. This being said, we do see that the method could be potentially extended into controversial applications such as generating fake profiles. Therefore, we believe that the images synthesized using our approach should present itself as synthetic.

\section{Conclusion}
In this paper, we present a new type of GAN method that synthesizes images in a controllable way. Through the design of local generators, masked feature aggregation and joint modeling of images and segmentation masks, we are able to model the structure and texture of different semantic areas separately. Experiments show that our method is able to synthesize high-quality images while disentangling different local parts. By combining our model with other editing methods, we can edit synthesized images with a more fine-grained control. Experiments also show that our model can be adapted to image-only datasets while preserving disentanglement capability. We believe the proposed method presents a new and interesting direction of GAN priors for controllable image synthesis, which could shed light on many potential downstream tasks.
\clearpage

%%%%%%%%% REFERENCES
{\small
\bibliographystyle{ieee_fullname}
\bibliography{egbib}
}

\appendix

\section{Implementation Details}
\subsection{Fusion with Transparent Classes}
Following Sec 3.1 in the main paper, a coarse semantic mask $\bm$ is fused from pseudo-depth maps, which is further used to aggregate local feature maps. In general cases, the aggregation can be simply achieved by computing $\bff=\sum_{k}^{K}\bm_k\odot \bff_i$, where the frontal class in the semantic mask will be chosen for the output feature. However, in the case of transparent classes, this formulation could be problematic. For example, although the whole eye area could be labeled as glasses in the semantic masks, we are still able to see the skin behind it. Thus, we treat such transparent classes separately during feature aggregation. In particular, we use a modified mask:
\begin{equation}
\label{eq:transparent_fusion}
\begin{aligned}
    \tilde{\bm}_{k}(i,j) = & \frac{\mathbbm{1}_{NT}(k)\exp(\bd_k(i,j))}{\sum_{k'}^K{\mathbbm{1}_{NT}(k')\exp(\bd_{k'}(i,j))}} + \mathbbm{1}_{T}(k)\bm_k(i,j),
\end{aligned}
\end{equation}
where $\mathbbm{1}_T(k)$ is an indicator function that equals 1 if $k$ is a transparent class and $0$ otherwise. $\mathbbm{1}_{NT}(k)$ is the opposite indicator function for non-transparent classes. The first part of  \cref{eq:transparent_fusion} here means that we first aggregate the features without the transparent classes. Then in the second part of \cref{eq:transparent_fusion}, we add the transparent features using their their original weights in mask $m$. In this way, the feature map will not be affected if there are no transparent classes. If there are, they would be added onto the feature map as additional residuals. Note that this formulation assumes that transparent classes do not overlap with each other. In our experiments, we set glasses and earings as transparent glasses. In fact, the model can also be trained stably by simply using the original mask $\bm$ for fusion, but the texture behind transparent classes could be distorted.

\begin{figure}
    \centering
    \includegraphics[width=\linewidth]{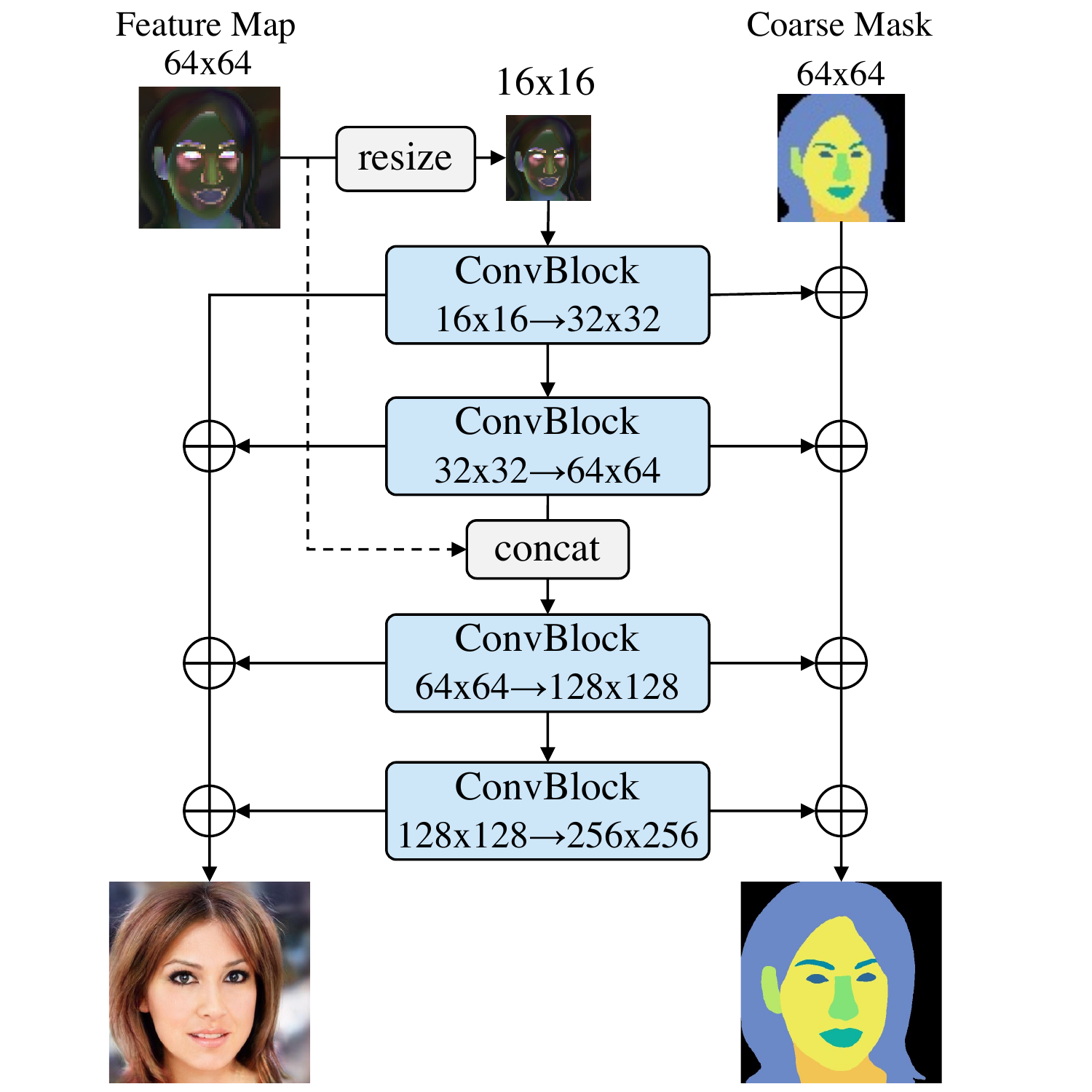}
    \caption{Details of the render net. Here, we take 256$\times$256 model as the example. A ``ConvBlock'' is a StyleGAN2 convolution block that have 2 convolution layers. We remove the style modulation and add a linear segmentation output branch in each convolution layer. $\oplus$ indicates upsampling and summation.}
    \label{fig:render_net}
\end{figure}

\begin{figure}
    \centering
    \includegraphics[width=\linewidth]{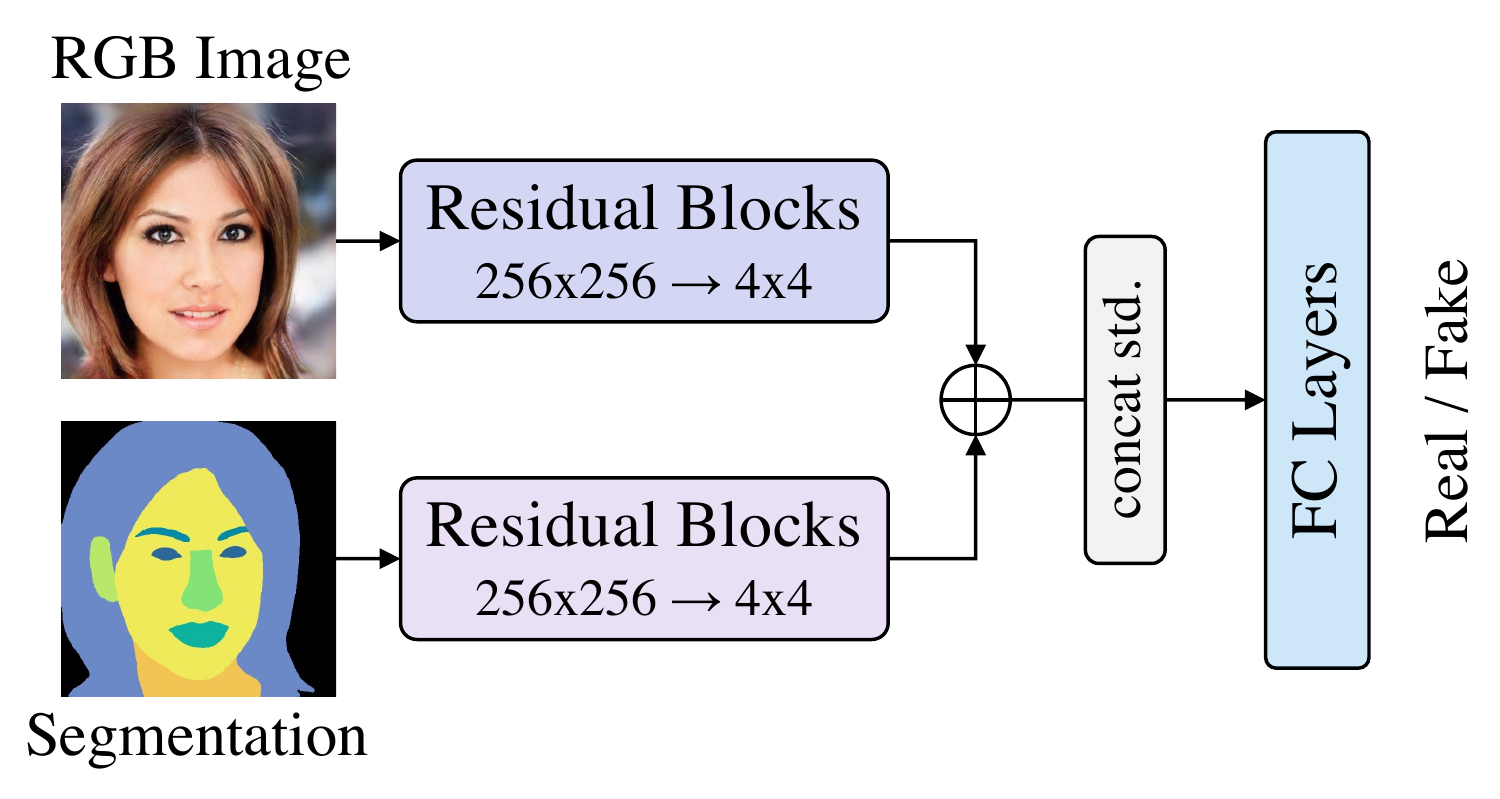}
    \caption{Details of the dual-branch discriminator. The ``Residual Blocks" are the convolution layers. The image branch and segmentation are symmetric except the input channels. ``concat std.'' is the step to of calculating standard deviation. The discriminator would be equivalent to StyleGAN2 discriminator if we remove the segmentation branch.}
    \label{fig:discriminator}
\end{figure}

\subsection{Architecture Details}
As shown in Fig.~3 in the main paper, each \textbf{local generator} is a modulated MLP (implemented by 1$\times$1 convolution) that has 10 layers. The input and output feature maps are both of size $64\times 64$. All the hidden layers has 64 channels. The Fourier feature at the input is first transformed into the hidden feature map with a linear fully connected (FC) layer. The ``toDepth" layer is a FC layer that outputs a 1-channel pseudo-depth map. The ``toFeat'' is a FC layer that outputs a 512-channel feature map. To encourage the disentanglement between shape and texture, we stop the gradient between shape and texture layers except for the background generator. We also fix the pseudo-depth map of background generator to be all $0$s.

The detailed architecture of \textbf{render network} is shown in \cref{fig:render_net}. Note that there is an upsampling and residual operation every layer for the segmentation mask, so $\Delta \bm$ is not explicitly computed. Instead, we calculate $\mathcal{L}_{mask}$ by the difference between downsampled output segmentation and coarse mask $\bm$.

The detailed architecture of \textbf{discriminator} is shown in \cref{fig:discriminator}. It is similar to StyleGAN2 discriminator except that we add an additional segmentation branch that is symmetric to image branch. During fine-tuning, we remove this branch and the discriminator reduces to an image discriminator.

\subsection{Efficiency}
The 256$\times$256 and 512$\times$512 models are trained on 4 and 8 32GB Tesla V100 GPUs, respectively. For the 512$\times$512 model, our model takes about two and a half day to train 150,000 steps with a batch size of 32 on 8 32GB Nvidia Tesla V100 GPUs, where the best model is then selected. For inference, it takes $0.137$s for our model to generate an image on a single GPU without parallelizing local generators.

\section{Additional Discussion}
\begin{figure}[t]
\captionsetup{font=small}
\centering
\scriptsize
\setlength\tabcolsep{1px}
\newcommand{\www}{0.155\linewidth}
\renewcommand{\arraystretch}{0.1}
\newcolumntype{Y}{>{\centering\arraybackslash}X}
\begin{tabularx}{\linewidth}{p{5pt}c}
    \raisebox{1.6\height}{\rotatebox[origin=c]{90}{Bald}} & 
    \includegraphics[width=\www]{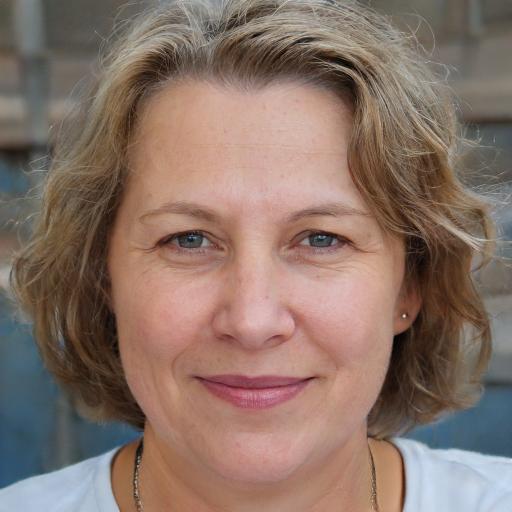}\hfill
    \includegraphics[width=\www]{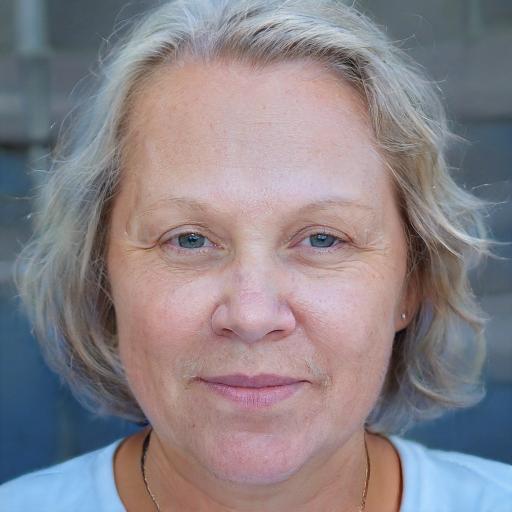}\hfill
    \includegraphics[width=\www]{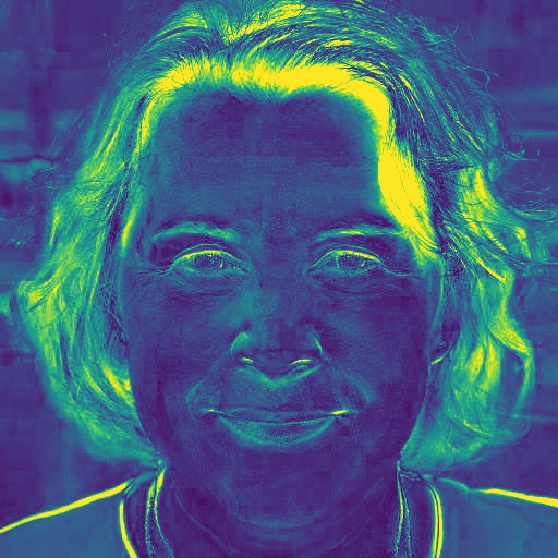}\quad
    \includegraphics[width=\www]{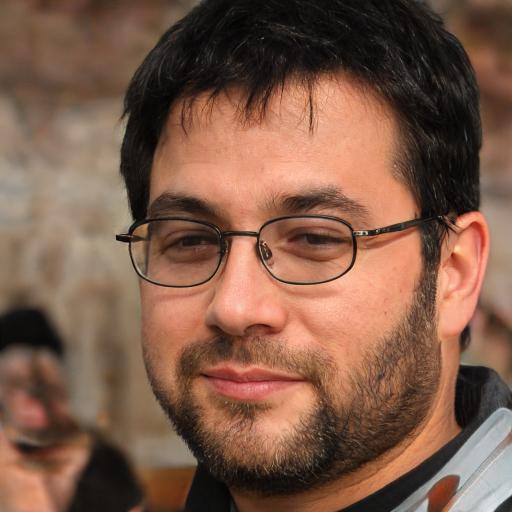}\hfill
    \includegraphics[width=\www]{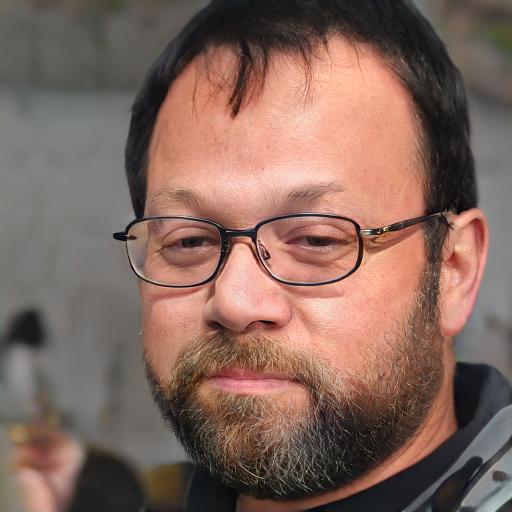}\hfill
    \includegraphics[width=\www]{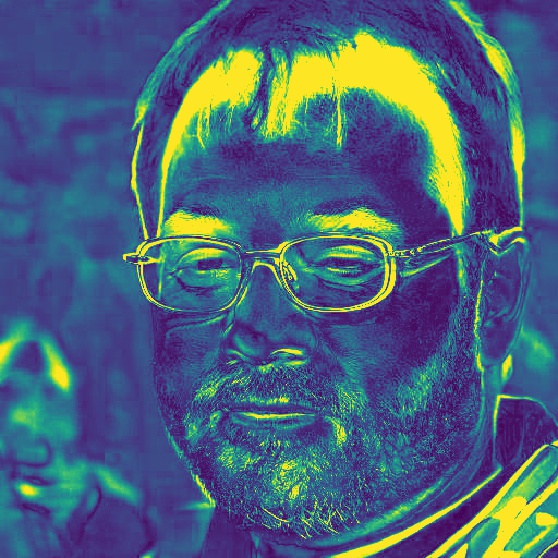}\\
    \raisebox{1.6\height}{\rotatebox[origin=c]{90}{Smile}} & 
    \includegraphics[width=\www]{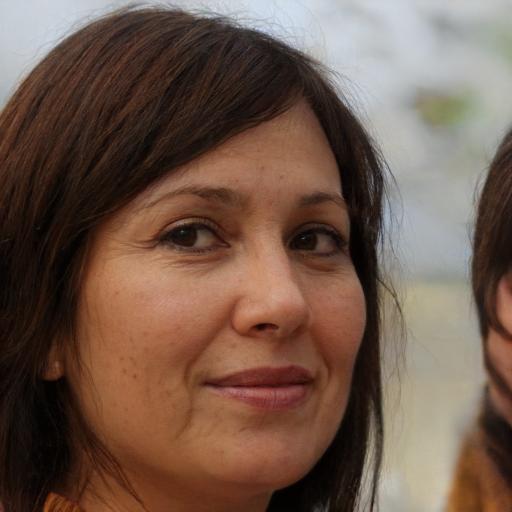}\hfill
    \includegraphics[width=\www]{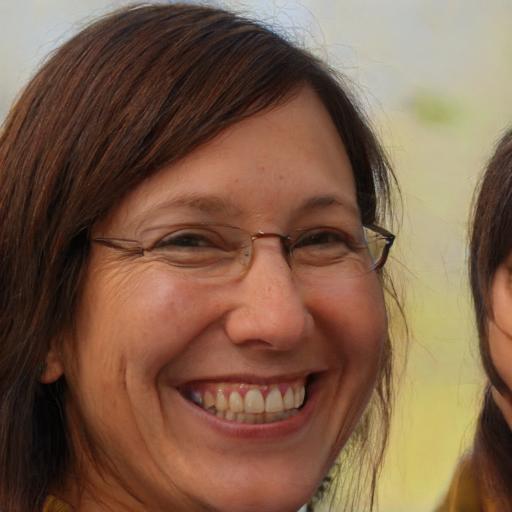}\hfill
    \includegraphics[width=\www]{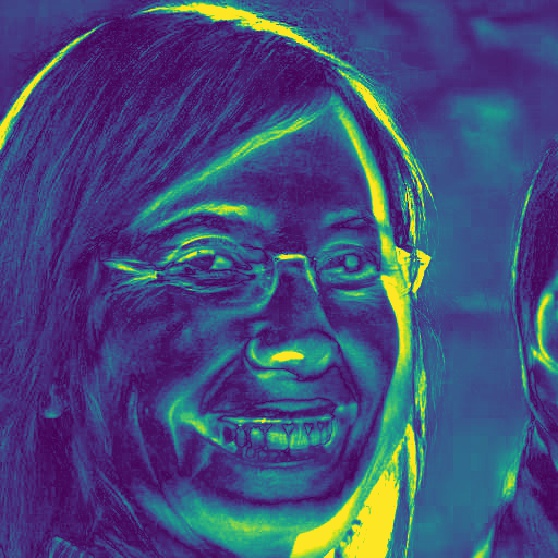}\quad
    \includegraphics[width=\www]{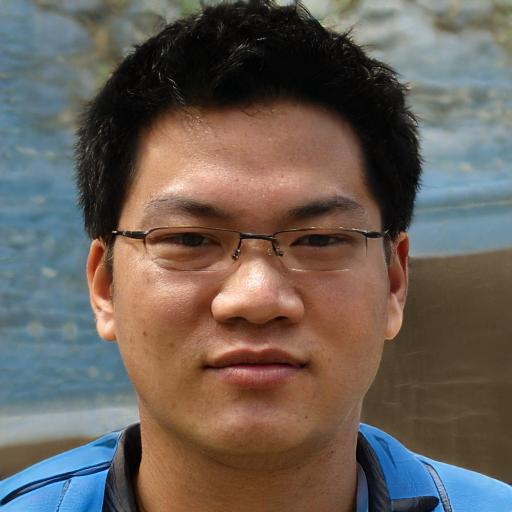}\hfill
    \includegraphics[width=\www]{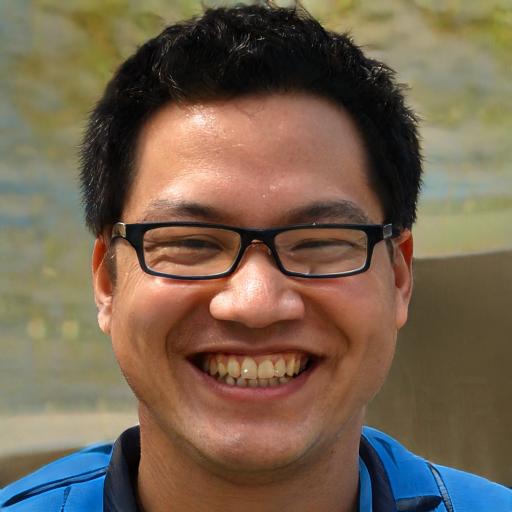}\hfill
    \includegraphics[width=\www]{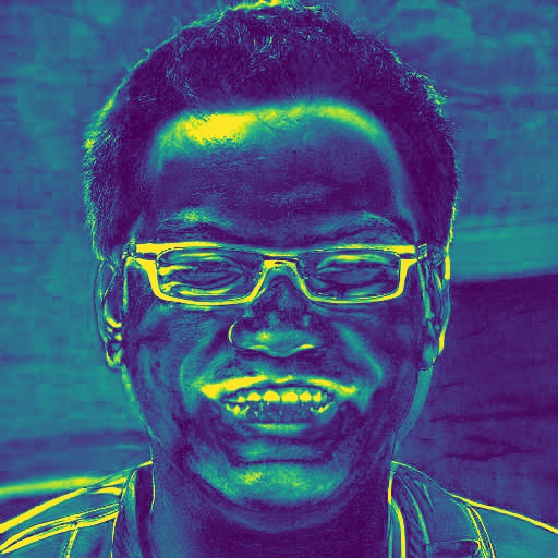}\\
\end{tabularx}
    \vspace{-0.5em}\caption{Results of InterFaceGAN on the $\mathcal{S}$ space of StyleGAN2.}
    \label{fig:interfacegan_sspace}
\end{figure}

Thanks to the reviewers, we provide additional discussions here to address some potentially shared concerns.
\vspace{-1.0em}\paragraph{StyleSpace} Wu~\etal\cite{wu2021stylespace} showed that there exists an $\mathcal{S}$ space in the StyleGAN which is more locally disentangled than the $W+$ space that we used in the main paper. However, in spite of good editing results, their method can only control limited attributes by tuning individual feature channels. It does not learn additional attributes from given labels. It is not proved yet that one can achieve the same degree of local disentanglement if latent editing methods are applied to the whole $\mathcal{S}$ space. Here, we briefly conduct such an experiment by applying InterFaceGAN to the  $\mathcal{S}$ space of the original StyleGAN2. As shown in \cref{fig:interfacegan_sspace}, $\mathcal{S}$ space still could suffer from spatial entanglement.

\begin{figure}[t]
\captionsetup{font=small}
\centering
\footnotesize
\setlength\tabcolsep{0pt}
\newcommand{\www}{0.14\linewidth}
\renewcommand{\arraystretch}{0.2}
\newcolumntype{Y}{>{\centering\arraybackslash}X}
\begin{tabularx}{\linewidth}{ccccccc}
    original & hair 1 & hair 2 & hair 3 & hair 4 & hair 5 & hair 6\\
    \includegraphics[width=\www]{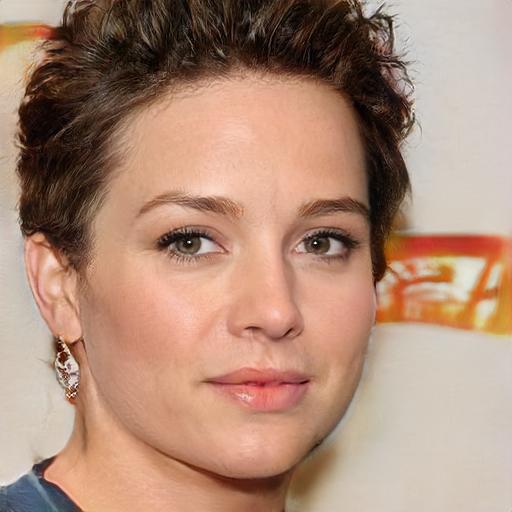} & 
    \includegraphics[width=\www]{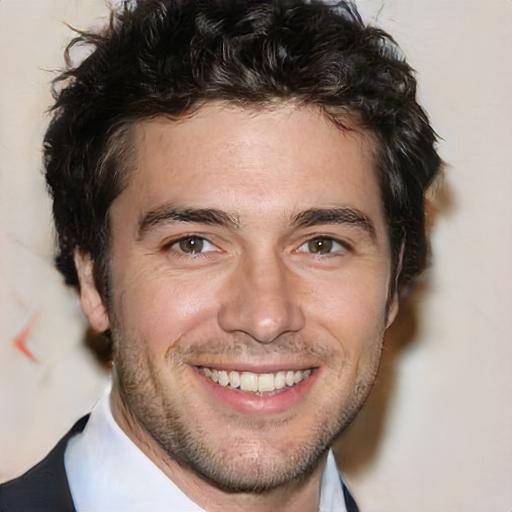} &
    \includegraphics[width=\www]{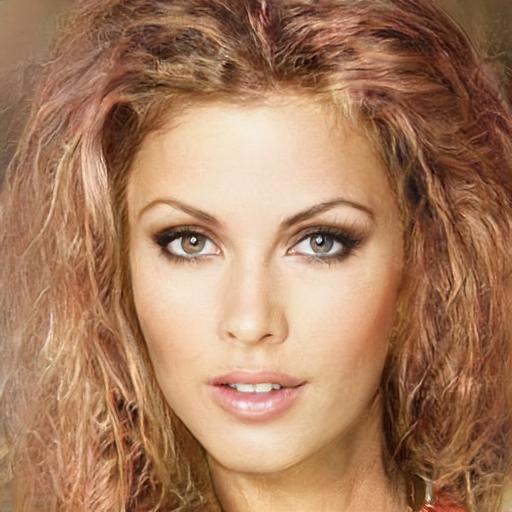} &
    \includegraphics[width=\www]{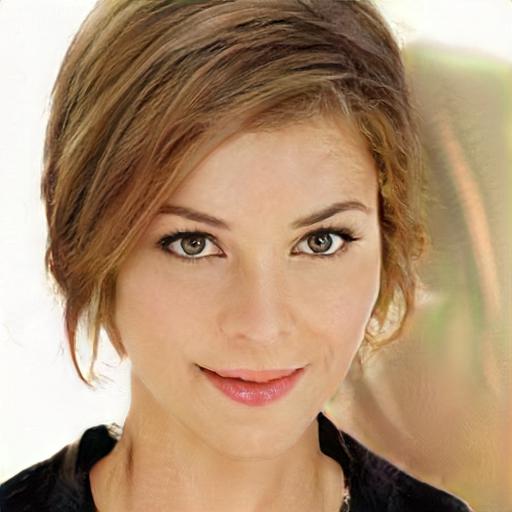} &
    \includegraphics[width=\www]{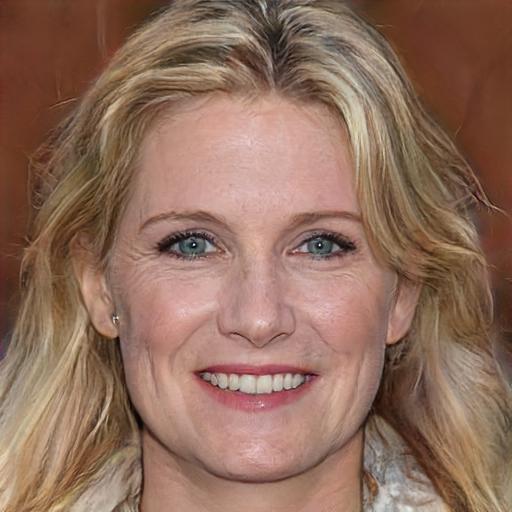} &
    \includegraphics[width=\www]{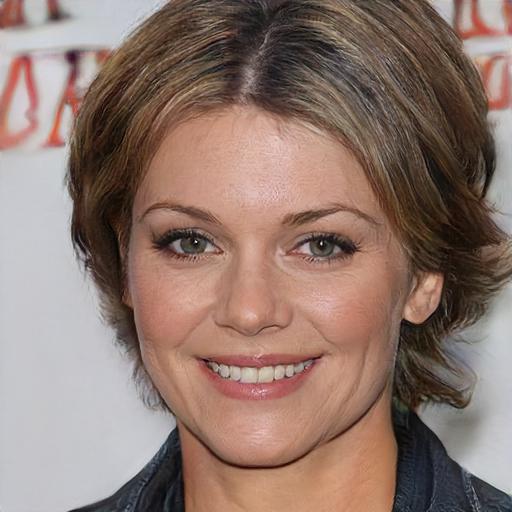} &
    \includegraphics[width=\www]{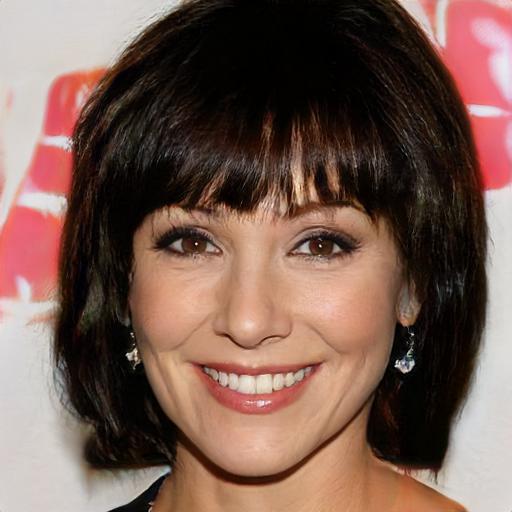} \\
    \raisebox{2em}{SEAN[69]} & 
    \includegraphics[width=\www]{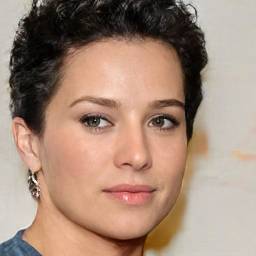} &
    \includegraphics[width=\www]{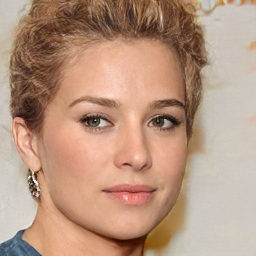} &
    \includegraphics[width=\www]{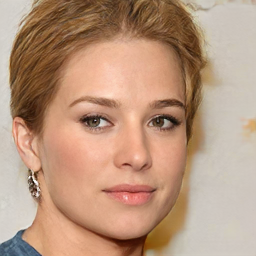} &
    \includegraphics[width=\www]{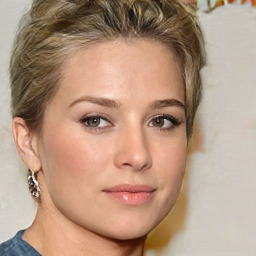} &
    \includegraphics[width=\www]{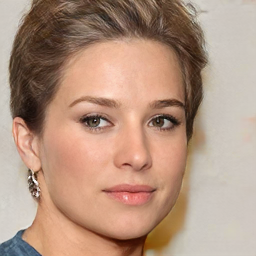} &
    \includegraphics[width=\www]{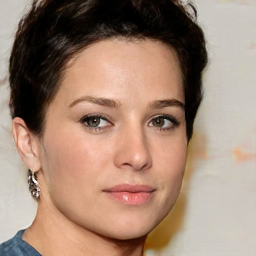} \\
    \raisebox{2em}{\makecell{Ours\\(Texture)}} & 
    \includegraphics[width=\www]{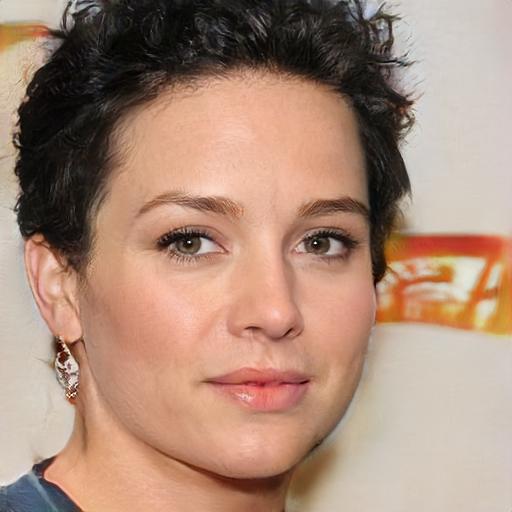} &
    \includegraphics[width=\www]{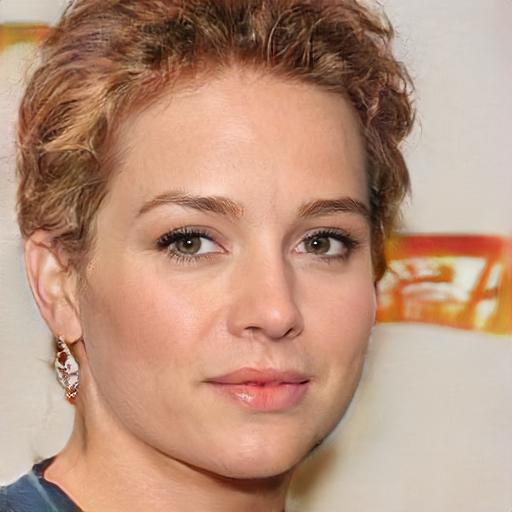} &
    \includegraphics[width=\www]{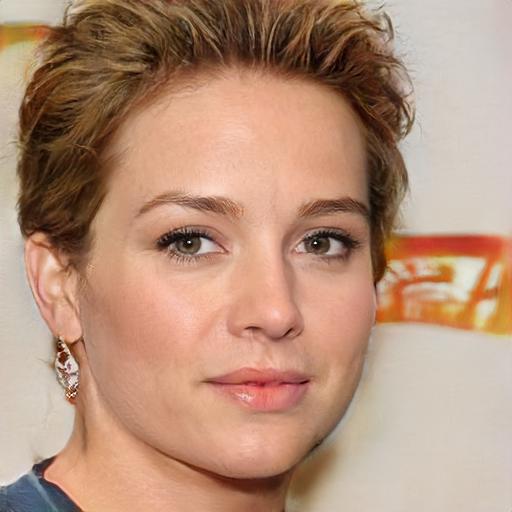} &
    \includegraphics[width=\www]{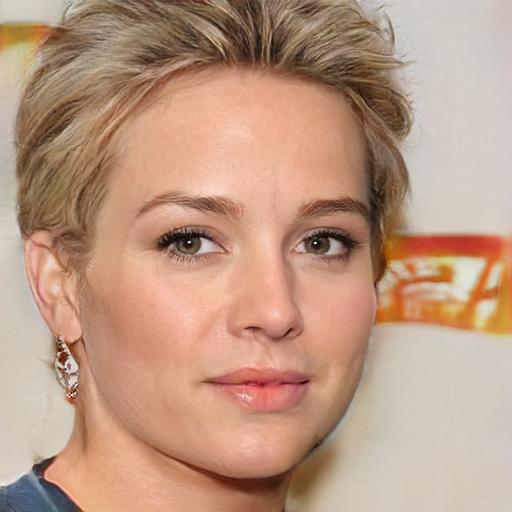} &
    \includegraphics[width=\www]{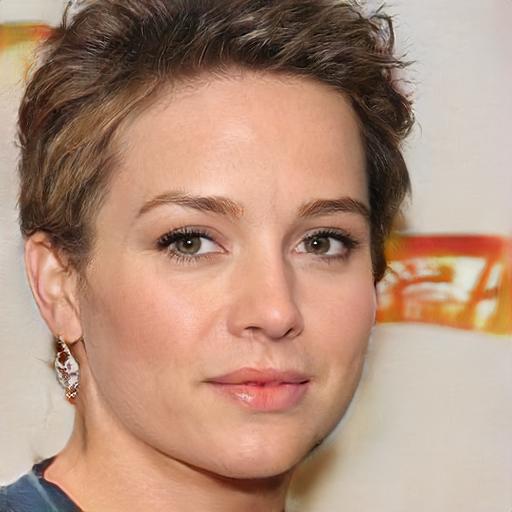} &
    \includegraphics[width=\www]{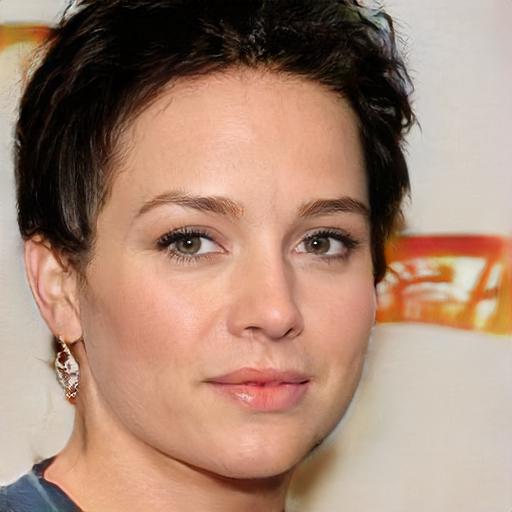} \\
    \raisebox{2em}{\makecell{Ours\\(Texture\\+Shape)}} & 
    \includegraphics[width=\www]{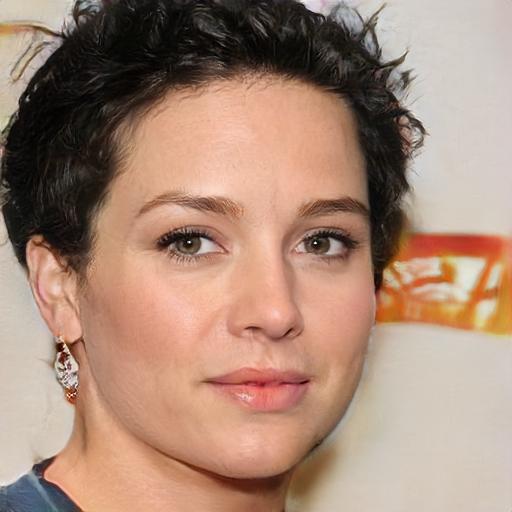} &
    \includegraphics[width=\www]{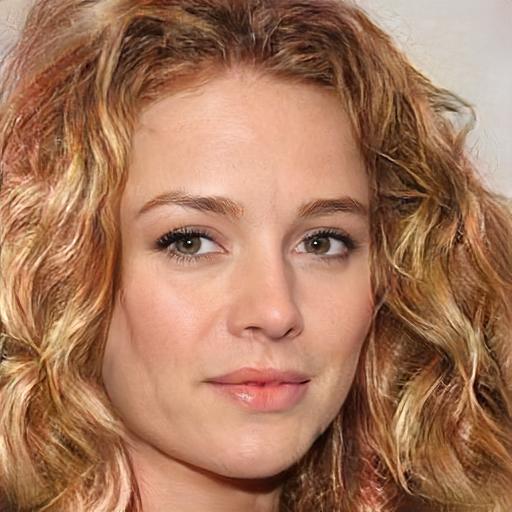} &
    \includegraphics[width=\www]{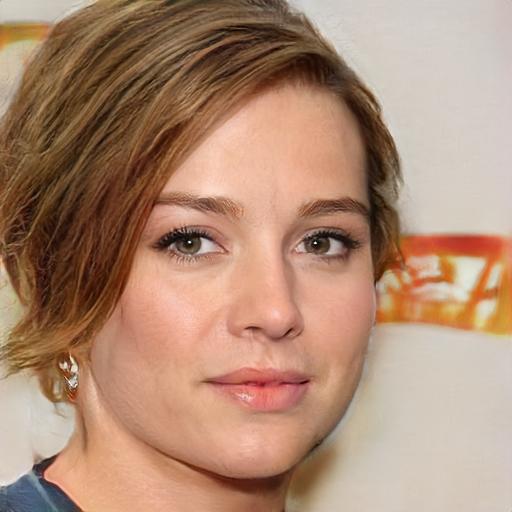} &
    \includegraphics[width=\www]{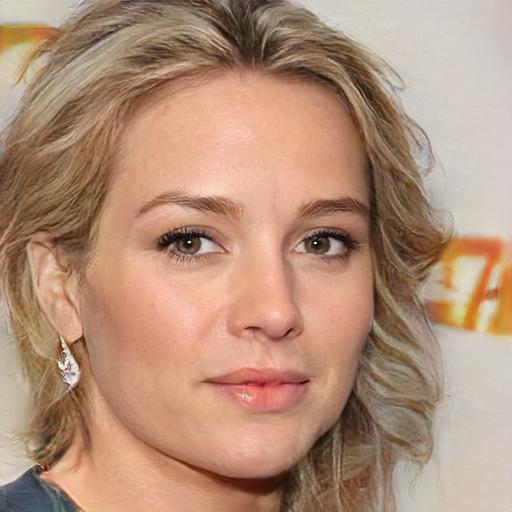} &
    \includegraphics[width=\www]{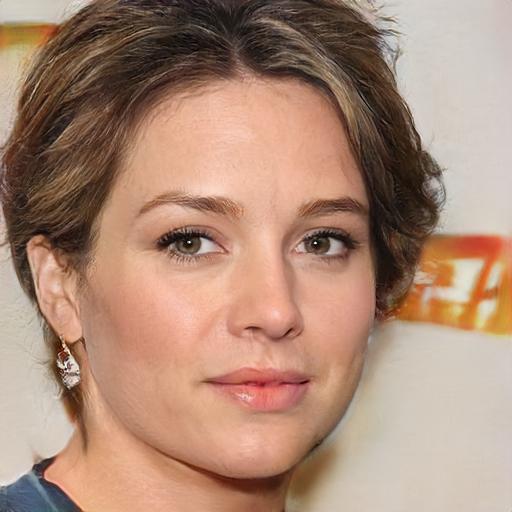} &
    \includegraphics[width=\www]{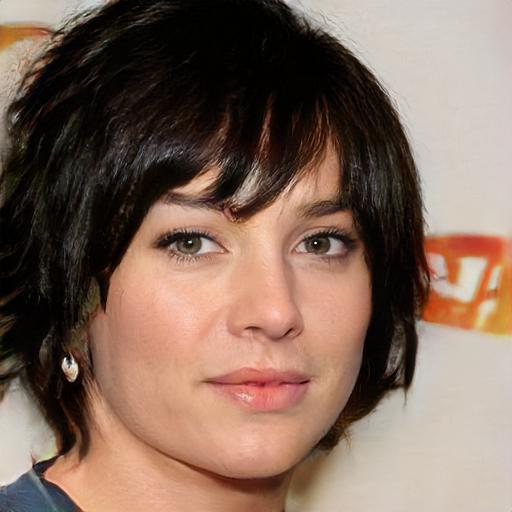} \\
\end{tabularx}
    \vspace{-0.5em}\caption{Mask-conditioned models~[69] can only transfer texture. }
    \label{fig:sean}
\end{figure}

\noindent\textbf{Mask-conditioned Models and StyleMapGAN} Mask-conditioned image translation models, such as SEAN~\cite{zhu2020SEAN} can be applied to local editing since they also learn a latent space for each semantic area. However, conditioned on a fixed semantic mask, their editing is restricted to the texture of each area (See \cref{fig:sean}). Changing the shape would require manual effort. In contrast, our unconditional model can control \textbf{both the shape and texture} with latent codes. We also note that our model, without using segmentation inputs, achieves a much lower FID (6.43) on CelebA-HQ than \cite{zhu2020SEAN} (17.66) and \cite{park2019SPADE} (22.43). StyleMapGAN~\cite{kim2021stylemapgan} has also shown the ability of local editing on synthesized images. However, by using a stylemap pyramid, it requires the editing area as an input, and the editing only happens to the texture in that area, a similar problem as SEAN. In contrast, our method is automatic and not restricted to fixed pixels.

\section{Style Mixing and Additional Results}
\begin{figure*}[t]
\captionsetup{font=small}
\centering
\footnotesize
\setlength\tabcolsep{1px}
\newcommand{\www}{0.123\linewidth}
\renewcommand{\arraystretch}{0.5}
\newcolumntype{Y}{>{\centering\arraybackslash}X}
\begin{tabularx}{\linewidth}{ccccccccc}
    & Coarse Structure & Background & Face (skin) & Eyes & Eyebrows & Mouth & Hair \\
    & \includegraphics[width=\www]{fig_appendix/stylemix/coarse/gallery.jpg}
    & \includegraphics[width=\www]{fig_appendix/stylemix/background/gallery.jpg}
    & \includegraphics[width=\www]{fig_appendix/stylemix/face/gallery.jpg}
    & \includegraphics[width=\www]{fig_appendix/stylemix/eye/gallery.jpg}
    & \includegraphics[width=\www]{fig_appendix/stylemix/eyebrow/gallery.jpg}
    & \includegraphics[width=\www]{fig_appendix/stylemix/mouth/gallery.jpg}
    & \includegraphics[width=\www]{fig_appendix/stylemix/hair/gallery.jpg} \\[0.2em]
    \includegraphics[width=\www]{fig_appendix/stylemix/probes/probe_00.jpg}
    & \includegraphics[width=\www]{fig_appendix/stylemix/coarse/00.jpg}
    & \includegraphics[width=\www]{fig_appendix/stylemix/background/00.jpg}
    & \includegraphics[width=\www]{fig_appendix/stylemix/face/00.jpg}
    & \includegraphics[width=\www]{fig_appendix/stylemix/eye/00.jpg}
    & \includegraphics[width=\www]{fig_appendix/stylemix/eyebrow/00.jpg}
    & \includegraphics[width=\www]{fig_appendix/stylemix/mouth/00.jpg}
    & \includegraphics[width=\www]{fig_appendix/stylemix/hair/00.jpg} \\[0.2em]
    \includegraphics[width=\www]{fig_appendix/stylemix/probes/probe_02.jpg}
    & \includegraphics[width=\www]{fig_appendix/stylemix/coarse/02.jpg}
    & \includegraphics[width=\www]{fig_appendix/stylemix/background/02.jpg}
    & \includegraphics[width=\www]{fig_appendix/stylemix/face/02.jpg}
    & \includegraphics[width=\www]{fig_appendix/stylemix/eye/02.jpg}
    & \includegraphics[width=\www]{fig_appendix/stylemix/eyebrow/02.jpg}
    & \includegraphics[width=\www]{fig_appendix/stylemix/mouth/02.jpg}
    & \includegraphics[width=\www]{fig_appendix/stylemix/hair/02.jpg} \\[0.2em]
    \includegraphics[width=\www]{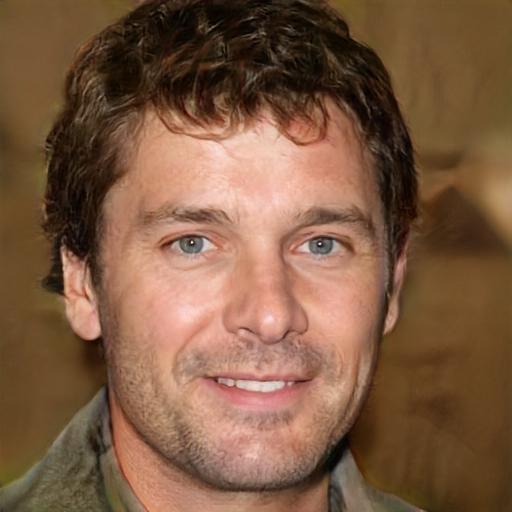}
    & \includegraphics[width=\www]{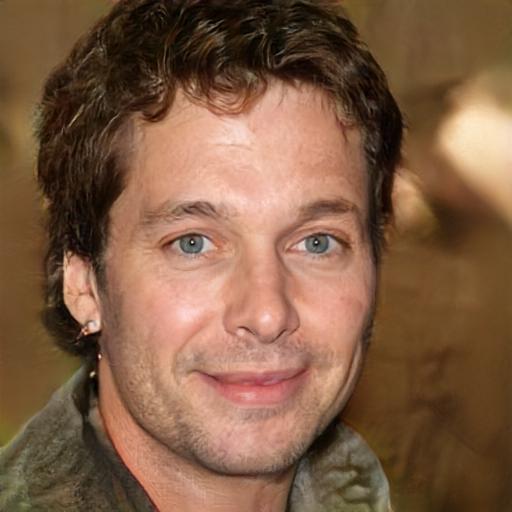}
    & \includegraphics[width=\www]{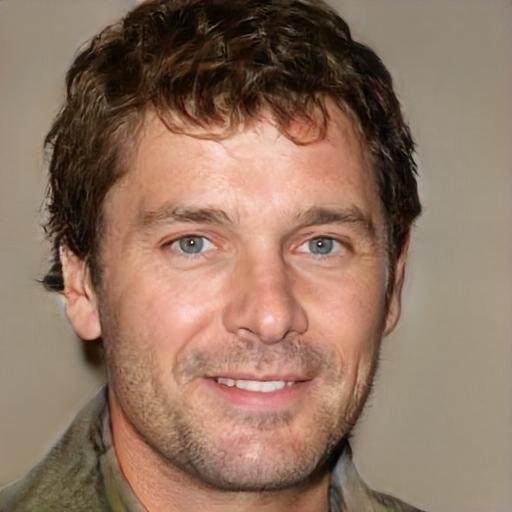}
    & \includegraphics[width=\www]{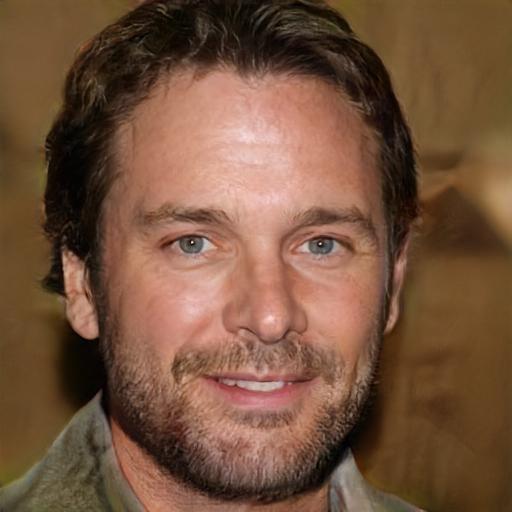}
    & \includegraphics[width=\www]{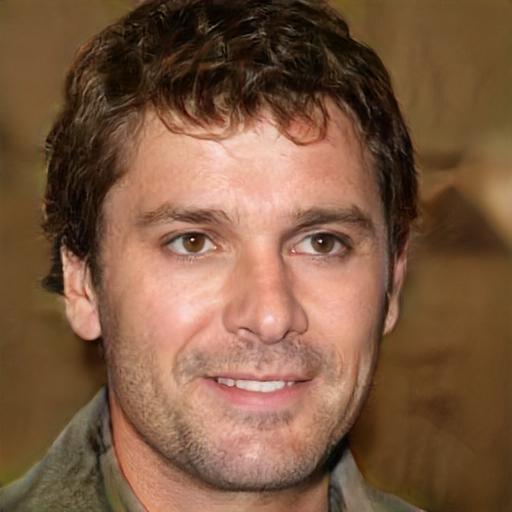}
    & \includegraphics[width=\www]{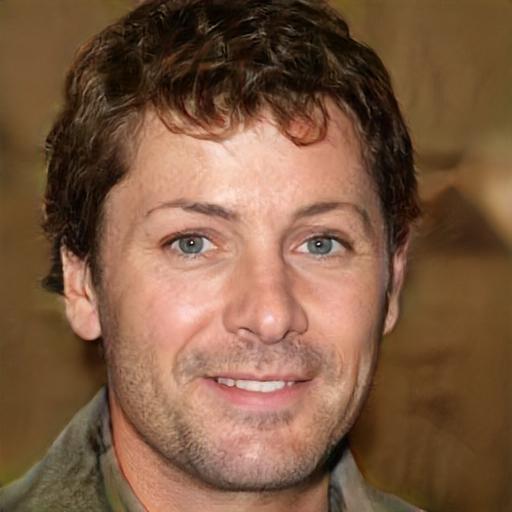}
    & \includegraphics[width=\www]{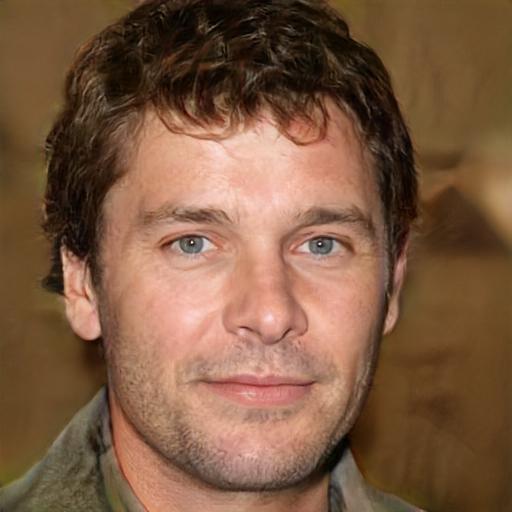}
    & \includegraphics[width=\www]{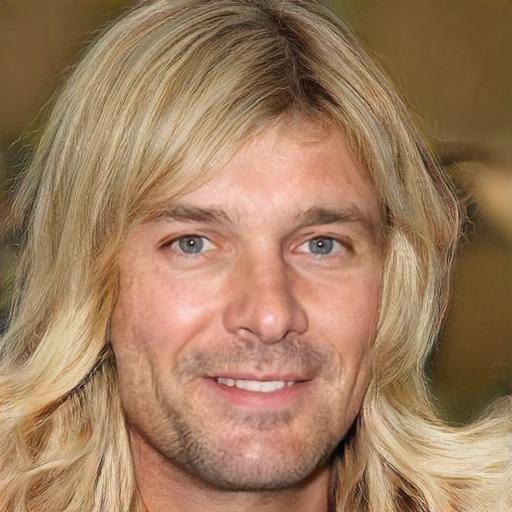} \\[0.2em]
    \includegraphics[width=\www]{fig_appendix/stylemix/probes/probe_06.jpg}
    & \includegraphics[width=\www]{fig_appendix/stylemix/coarse/06.jpg}
    & \includegraphics[width=\www]{fig_appendix/stylemix/background/06.jpg}
    & \includegraphics[width=\www]{fig_appendix/stylemix/face/06.jpg}
    & \includegraphics[width=\www]{fig_appendix/stylemix/eye/06.jpg}
    & \includegraphics[width=\www]{fig_appendix/stylemix/eyebrow/06.jpg}
    & \includegraphics[width=\www]{fig_appendix/stylemix/mouth/06.jpg}
    & \includegraphics[width=\www]{fig_appendix/stylemix/hair/06.jpg} \\[0.2em]
    \includegraphics[width=\www]{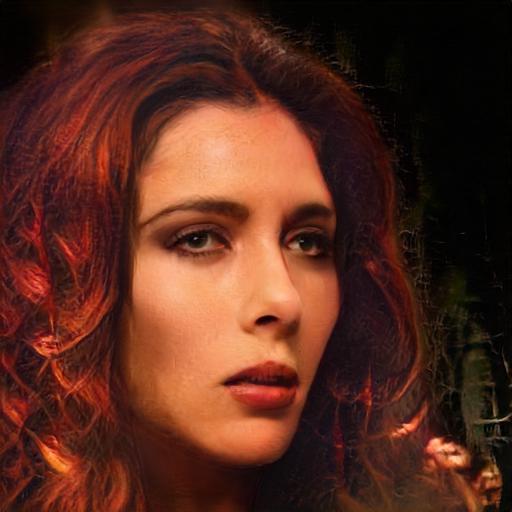}
    & \includegraphics[width=\www]{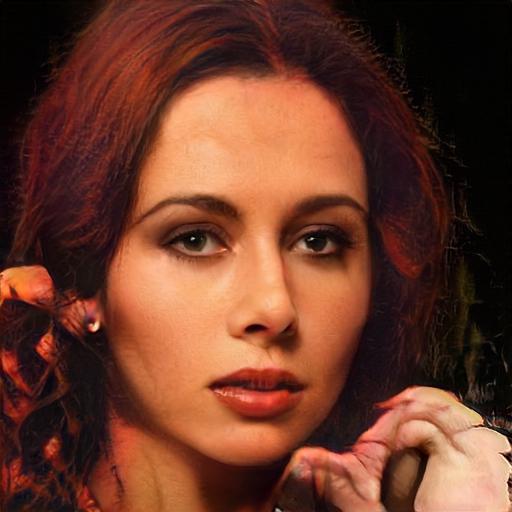}
    & \includegraphics[width=\www]{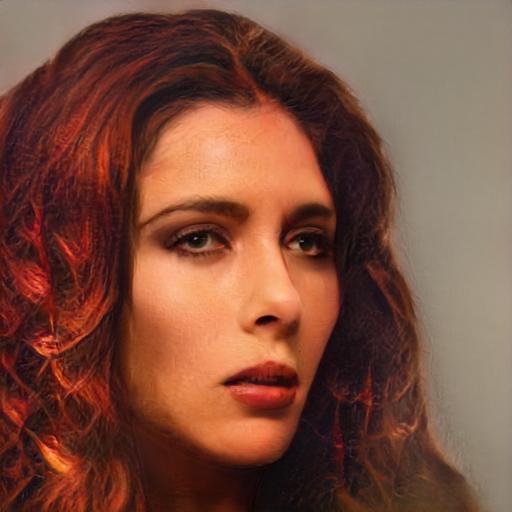}
    & \includegraphics[width=\www]{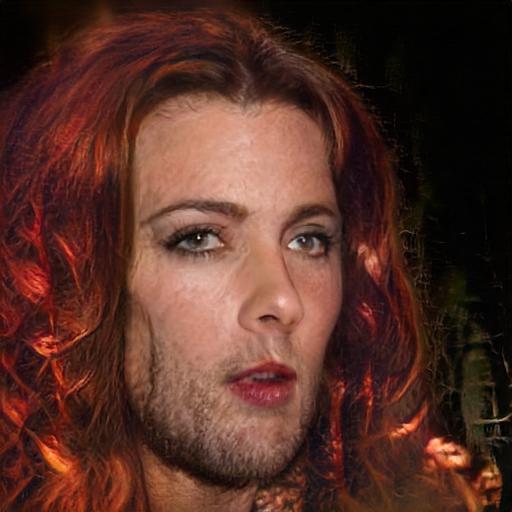}
    & \includegraphics[width=\www]{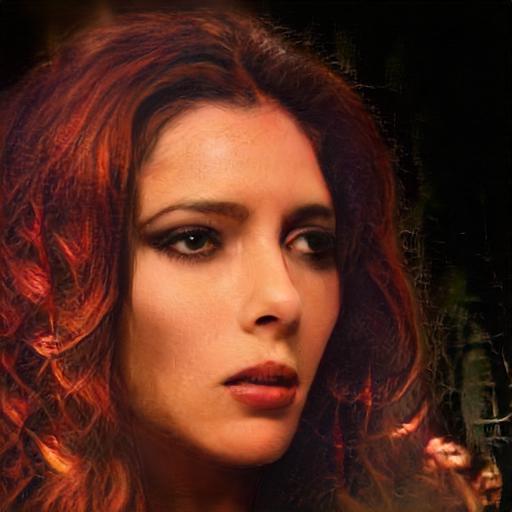}
    & \includegraphics[width=\www]{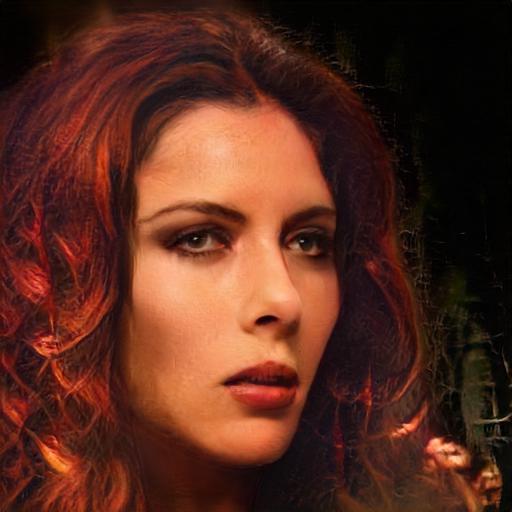}
    & \includegraphics[width=\www]{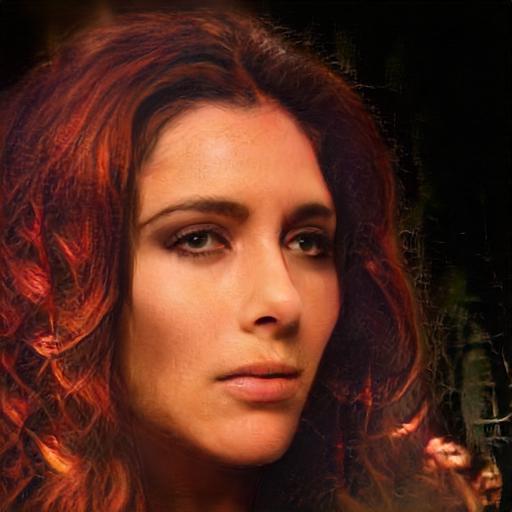}
    & \includegraphics[width=\www]{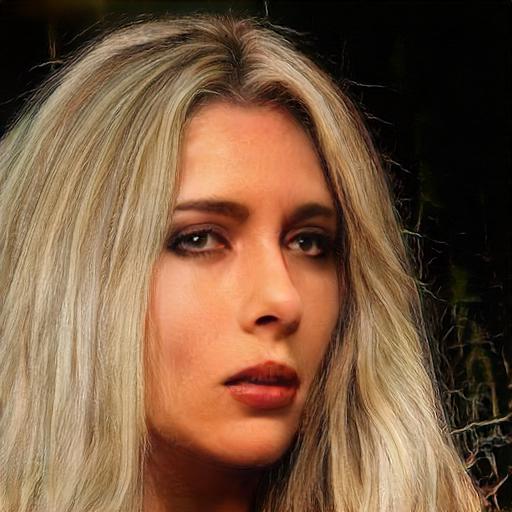} \\[0.2em]
    \includegraphics[width=\www]{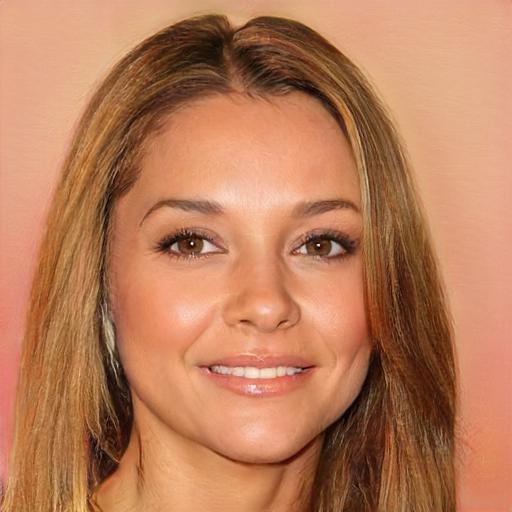}
    & \includegraphics[width=\www]{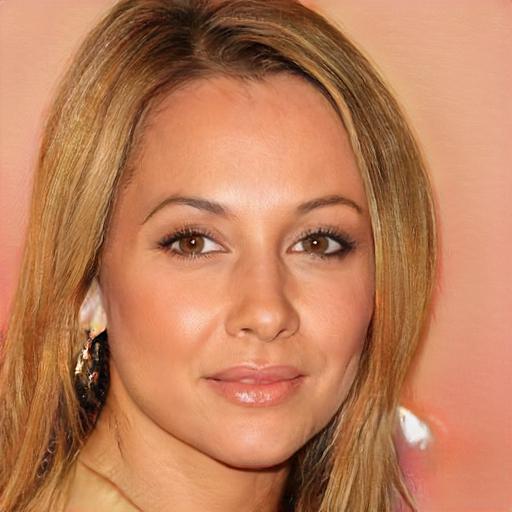}
    & \includegraphics[width=\www]{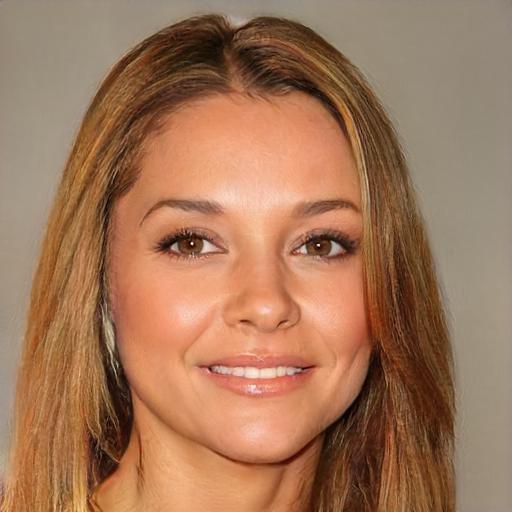}
    & \includegraphics[width=\www]{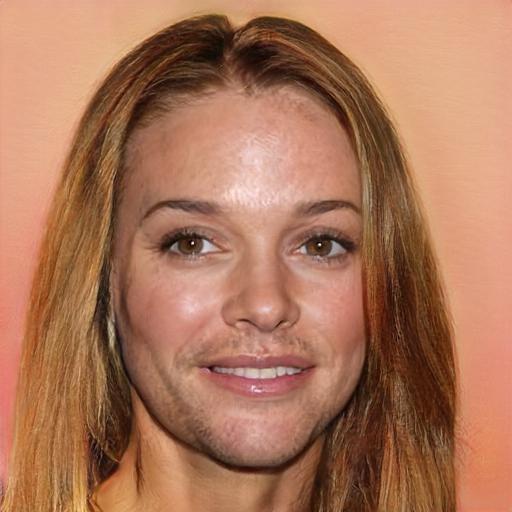}
    & \includegraphics[width=\www]{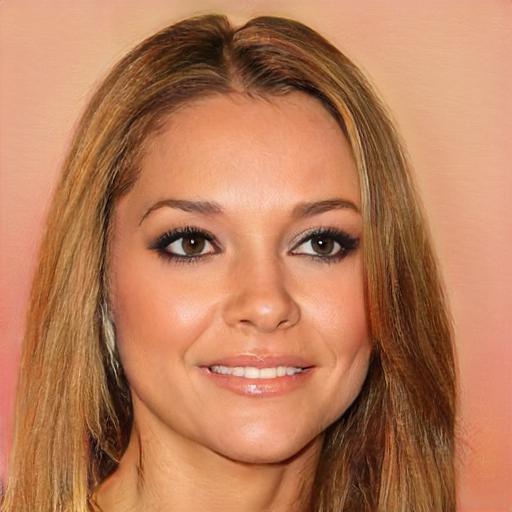}
    & \includegraphics[width=\www]{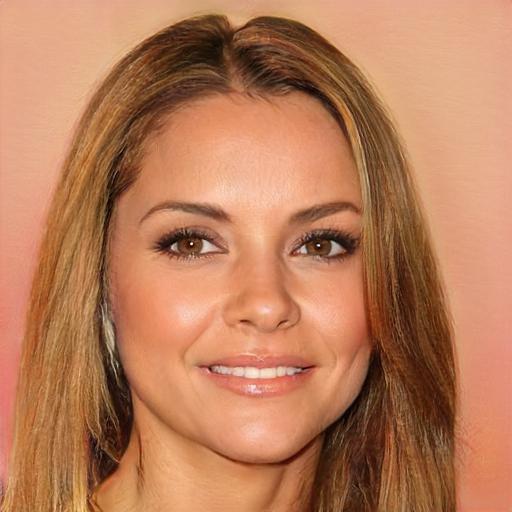}
    & \includegraphics[width=\www]{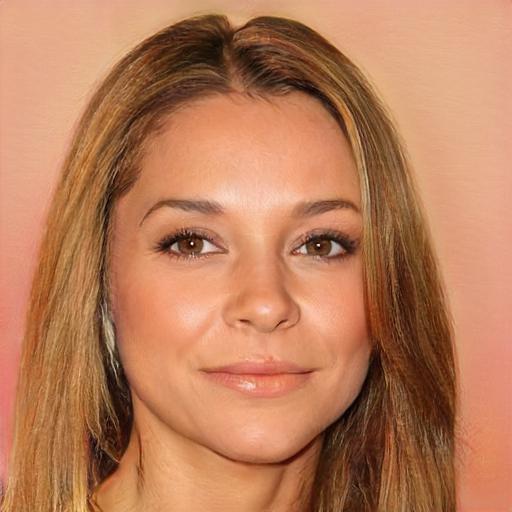}
    & \includegraphics[width=\www]{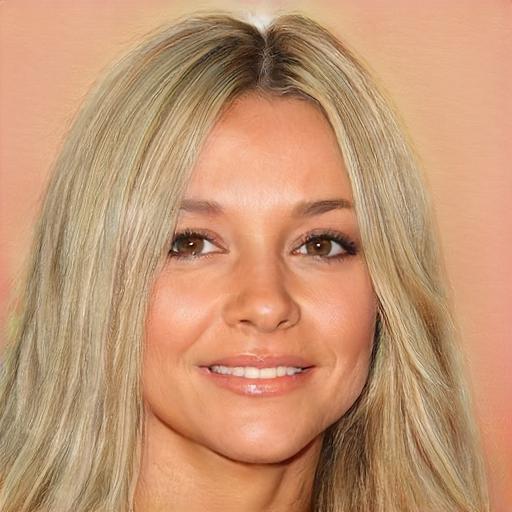} \\[0.2em]
    \includegraphics[width=\www]{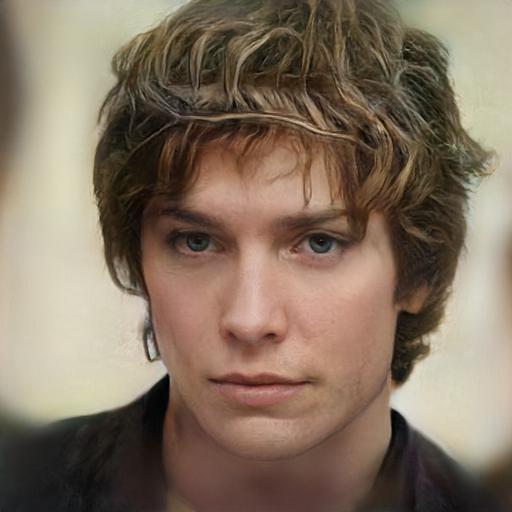}
    & \includegraphics[width=\www]{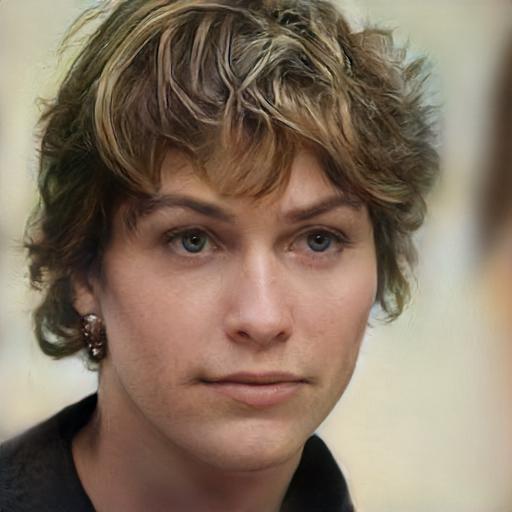}
    & \includegraphics[width=\www]{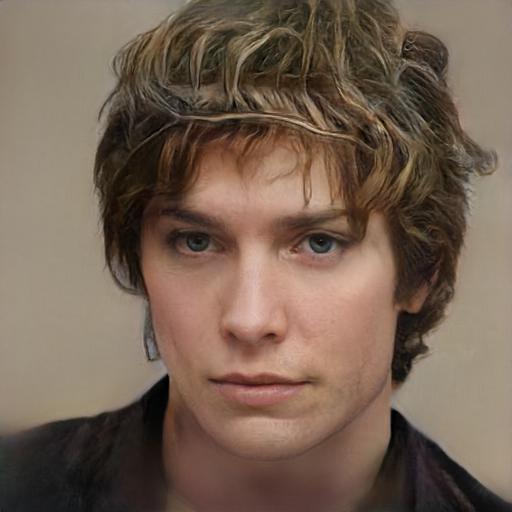}
    & \includegraphics[width=\www]{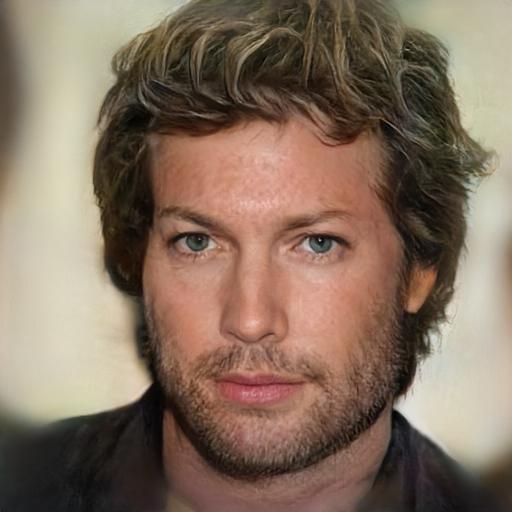}
    & \includegraphics[width=\www]{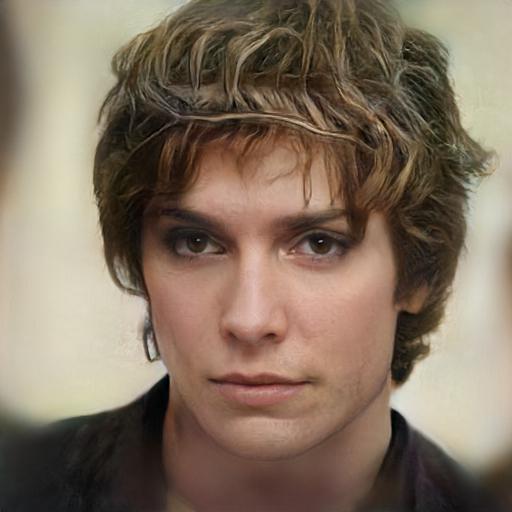}
    & \includegraphics[width=\www]{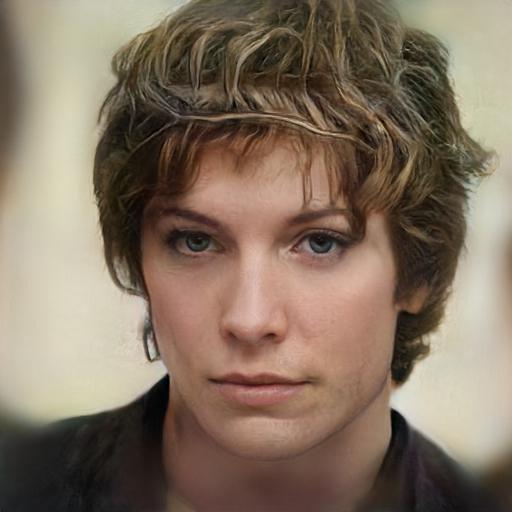}
    & \includegraphics[width=\www]{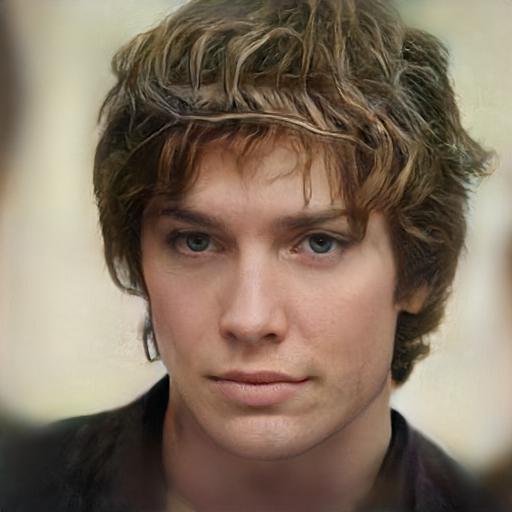}
    & \includegraphics[width=\www]{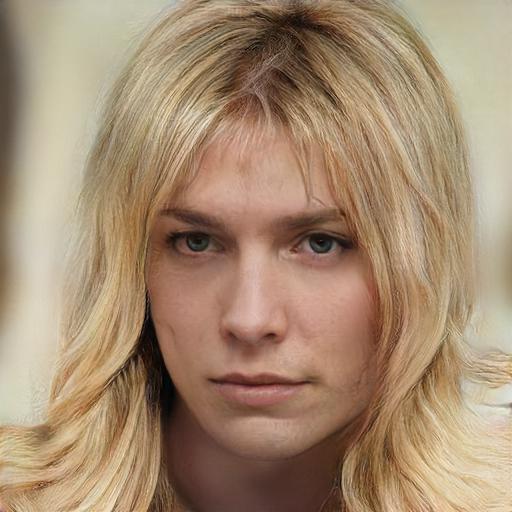} \\[0.2em]
    \includegraphics[width=\www]{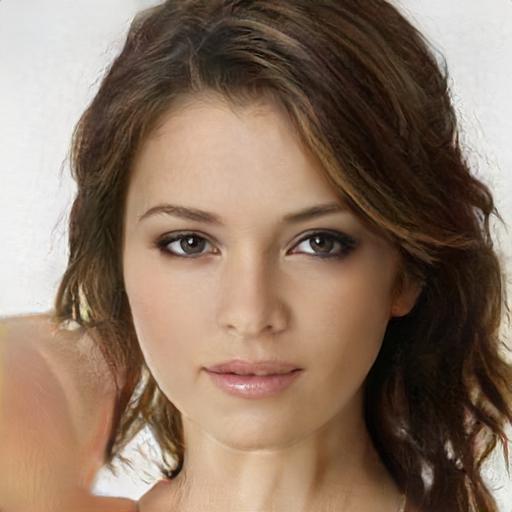}
    & \includegraphics[width=\www]{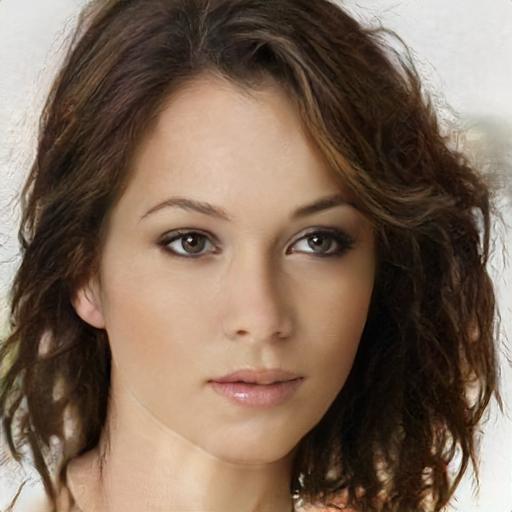}
    & \includegraphics[width=\www]{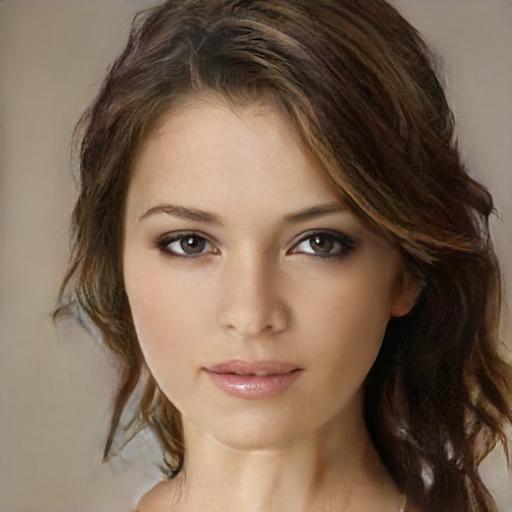}
    & \includegraphics[width=\www]{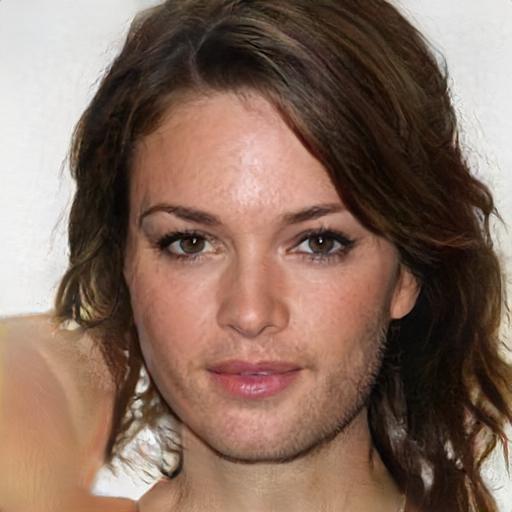}
    & \includegraphics[width=\www]{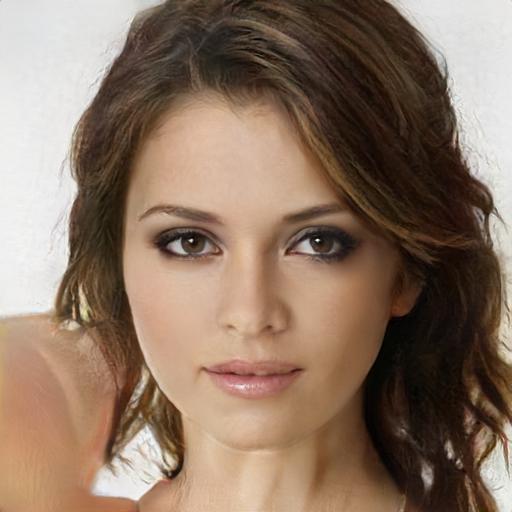}
    & \includegraphics[width=\www]{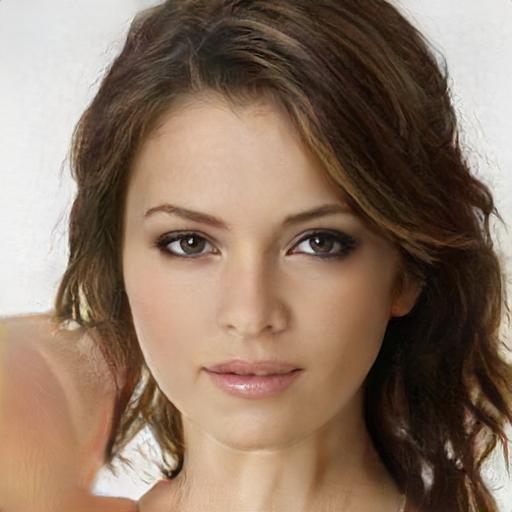}
    & \includegraphics[width=\www]{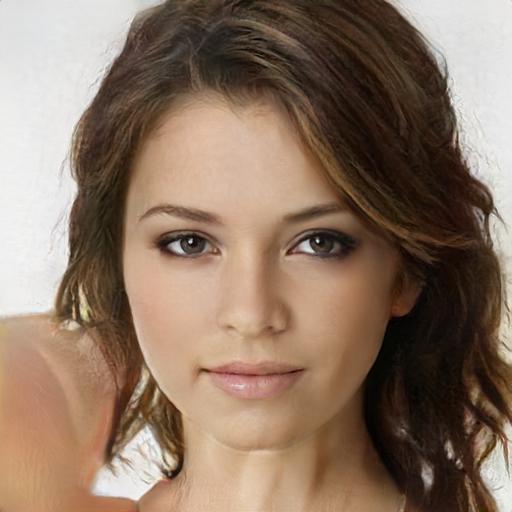}
    & \includegraphics[width=\www]{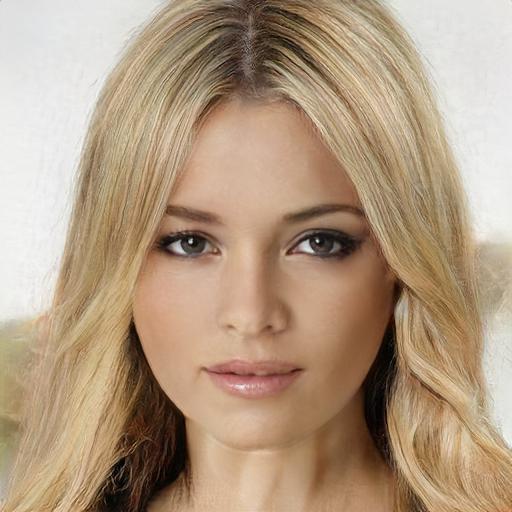} \\[0.2em]
    % \includegraphics[width=\www]{fig_appendix/stylemix/probes/probe_16.jpg}
    % & \includegraphics[width=\www]{fig_appendix/stylemix/coarse/16.jpg}
    % & \includegraphics[width=\www]{fig_appendix/stylemix/background/16.jpg}
    % & \includegraphics[width=\www]{fig_appendix/stylemix/face/16.jpg}
    % & \includegraphics[width=\www]{fig_appendix/stylemix/eye/16.jpg}
    % & \includegraphics[width=\www]{fig_appendix/stylemix/eyebrow/16.jpg}
    % & \includegraphics[width=\www]{fig_appendix/stylemix/mouth/16.jpg}
    % & \includegraphics[width=\www]{fig_appendix/stylemix/hair/16.jpg} \\[0.2em]
\end{tabularx}
    \vspace{-1.0em}\caption{Local style mixing of the model trained on CelebAMask-HQ. The first column shows randomly sampled images for editing. The remaining columns show the results of mixing local styles using the reference images in the first row. }\vspace{-1.0em}
    \label{appendix:fig:stylemix}
\end{figure*}

\begin{figure*}[t]
\captionsetup{font=small}
\centering
\footnotesize
\setlength\tabcolsep{1px}
\newcommand{\www}{0.14\linewidth}
\renewcommand{\arraystretch}{0.5}
\newcolumntype{Y}{>{\centering\arraybackslash}X}
\begin{tabularx}{\linewidth}{ccccccccc}
    & Coarse Structure & Face (skin) & Eyes & Eyebrows & Mouth & Hair \\
    & \includegraphics[width=\www]{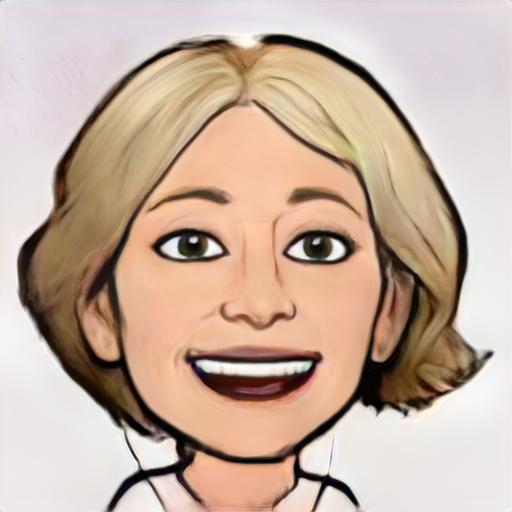}
    & \includegraphics[width=\www]{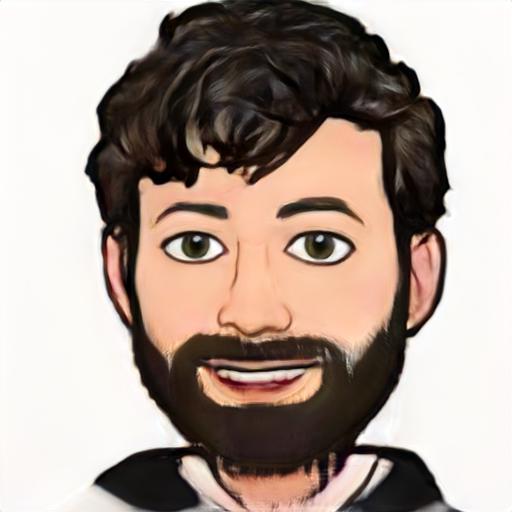}
    & \includegraphics[width=\www]{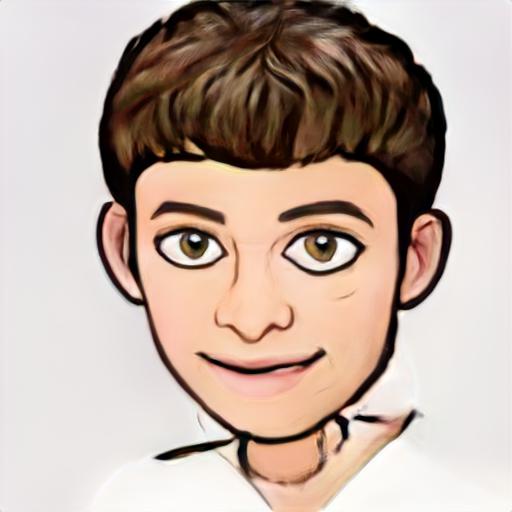}
    & \includegraphics[width=\www]{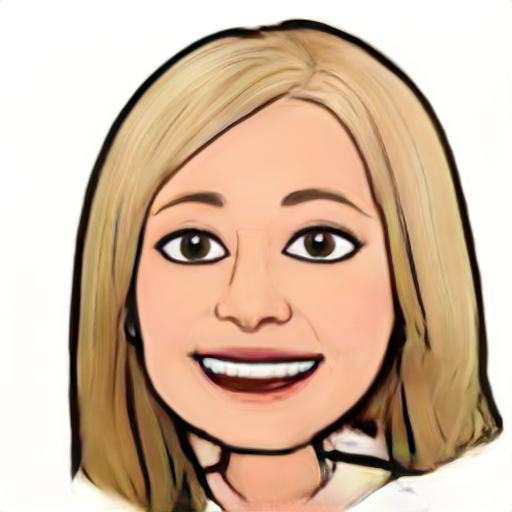}
    & \includegraphics[width=\www]{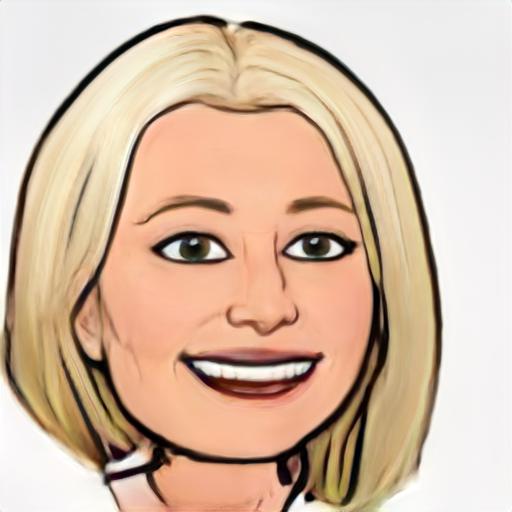}
    % & \includegraphics[width=\www]{fig_appendix/stylemix_bitmoji/ear/gallery.jpg}
    & \includegraphics[width=\www]{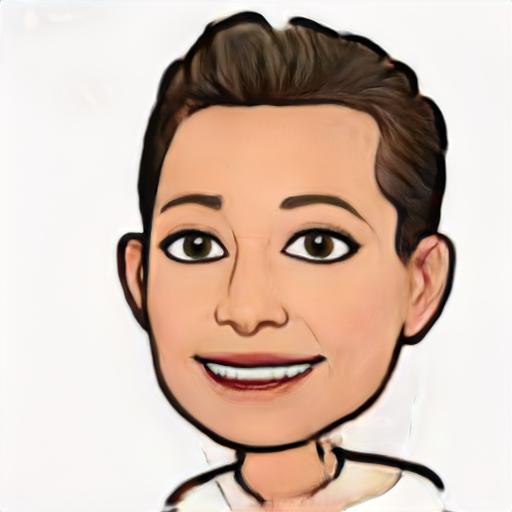} \\[0.2em]
    \includegraphics[width=\www]{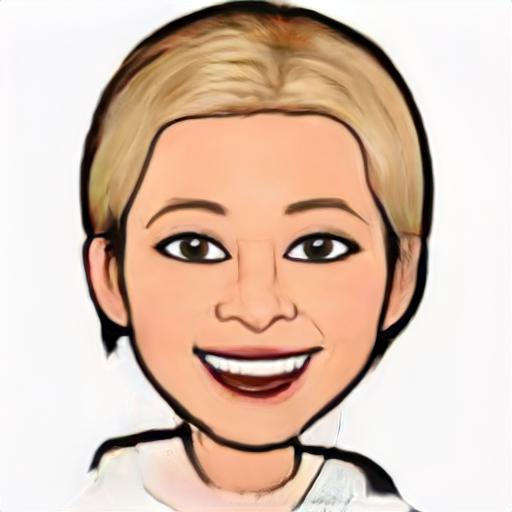}
    & \includegraphics[width=\www]{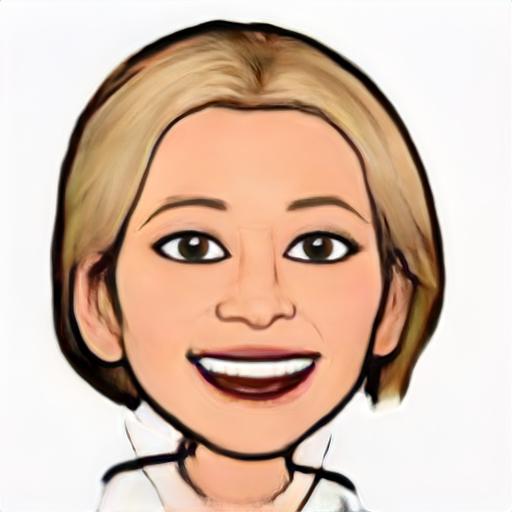}
    & \includegraphics[width=\www]{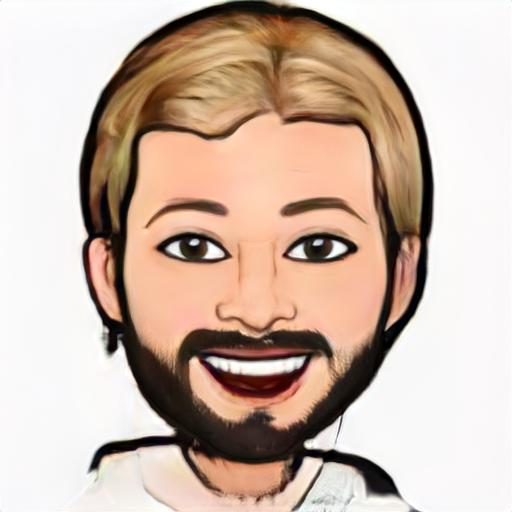}
    & \includegraphics[width=\www]{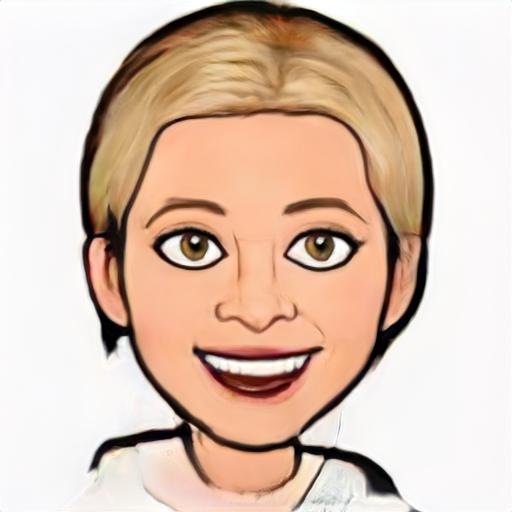}
    & \includegraphics[width=\www]{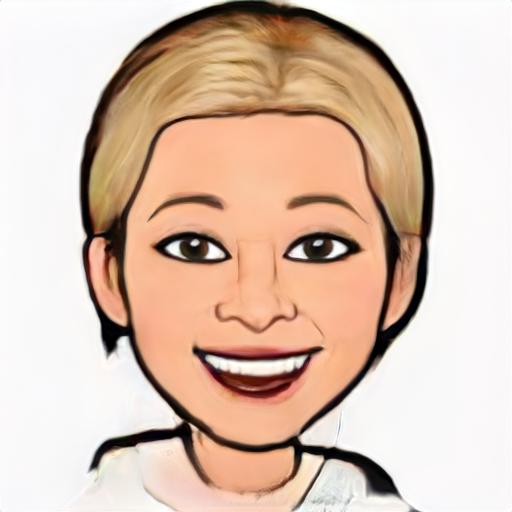}
    & \includegraphics[width=\www]{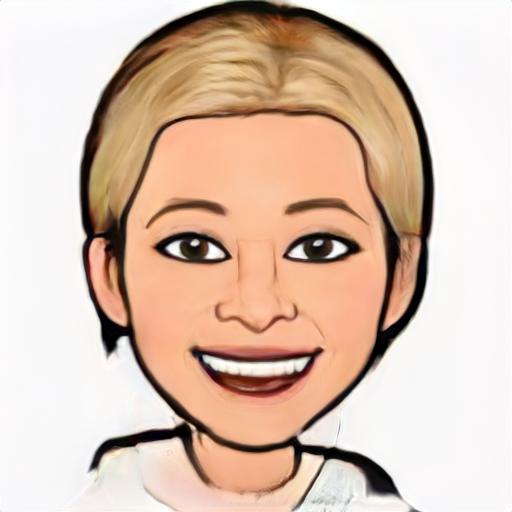}
    % & \includegraphics[width=\www]{fig_appendix/stylemix_bitmoji/ear/00.jpg}
    & \includegraphics[width=\www]{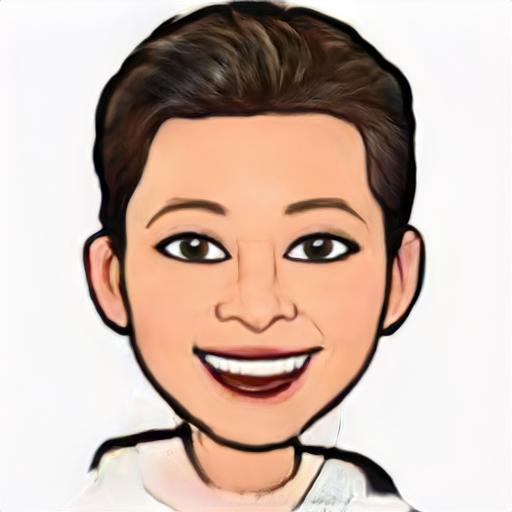} \\[0.2em]
    \includegraphics[width=\www]{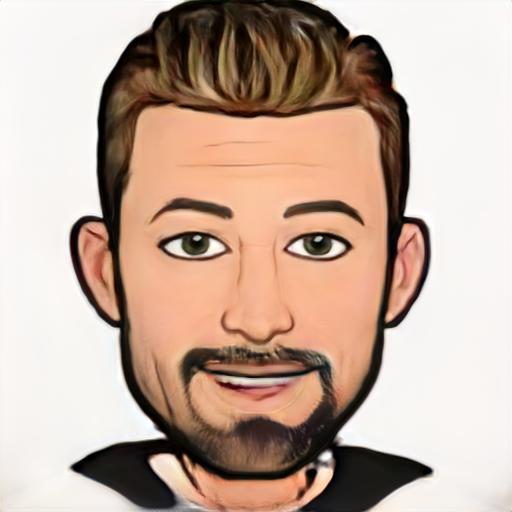}
    & \includegraphics[width=\www]{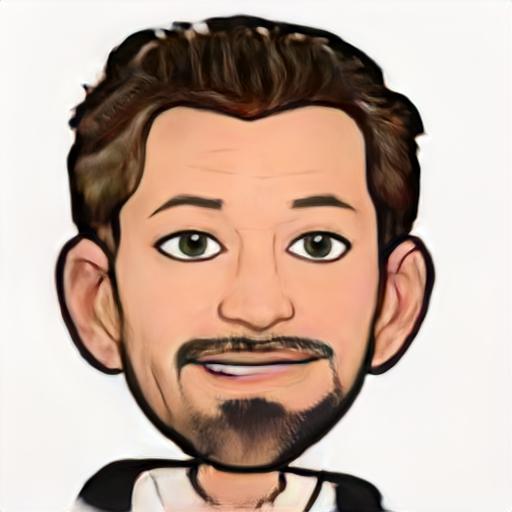}
    & \includegraphics[width=\www]{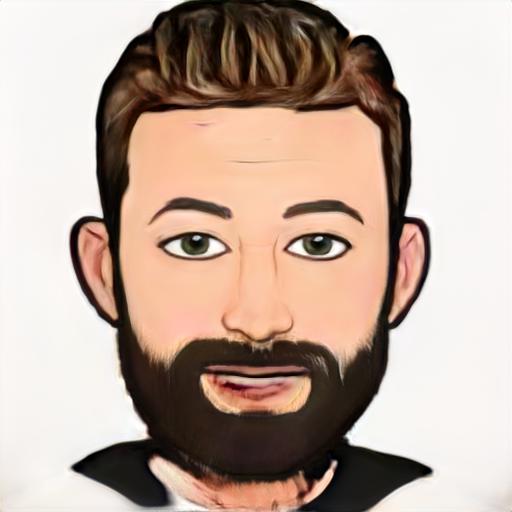}
    & \includegraphics[width=\www]{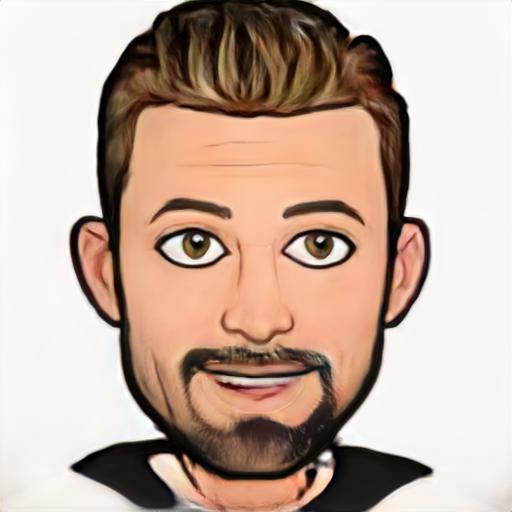}
    & \includegraphics[width=\www]{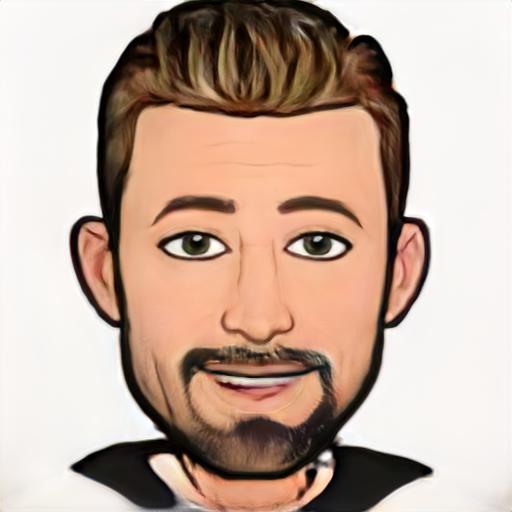}
    & \includegraphics[width=\www]{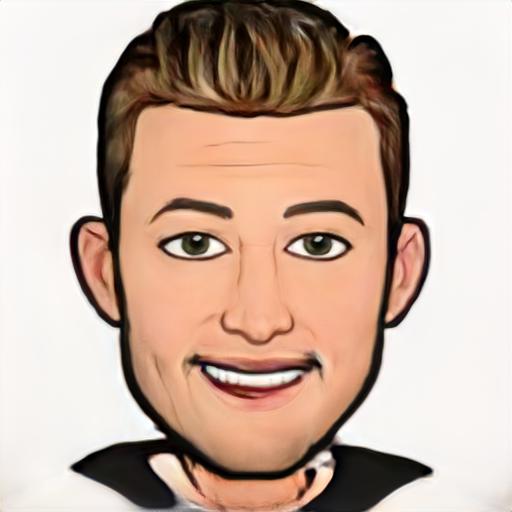}
    % & \includegraphics[width=\www]{fig_appendix/stylemix_bitmoji/ear/01.jpg}
    & \includegraphics[width=\www]{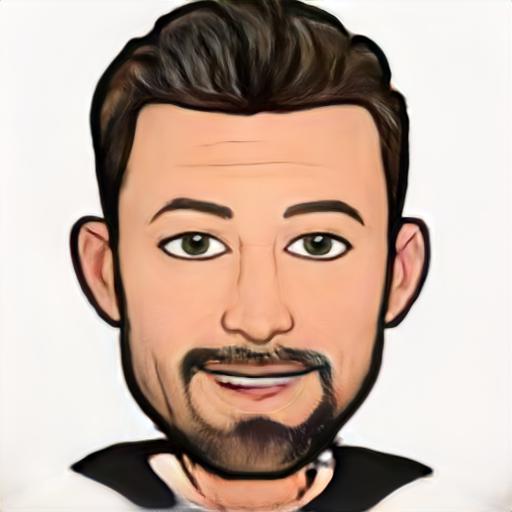} \\[0.2em]
    \includegraphics[width=\www]{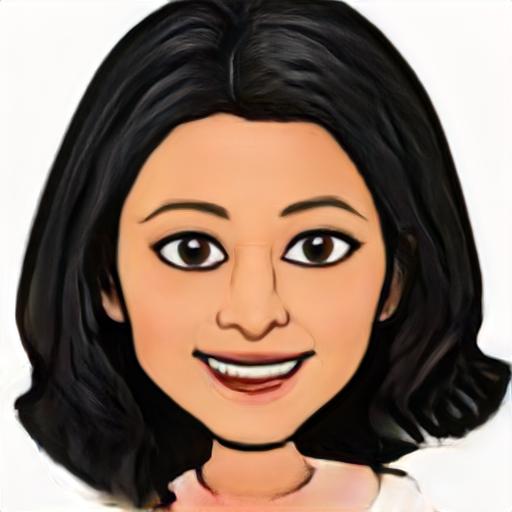}
    & \includegraphics[width=\www]{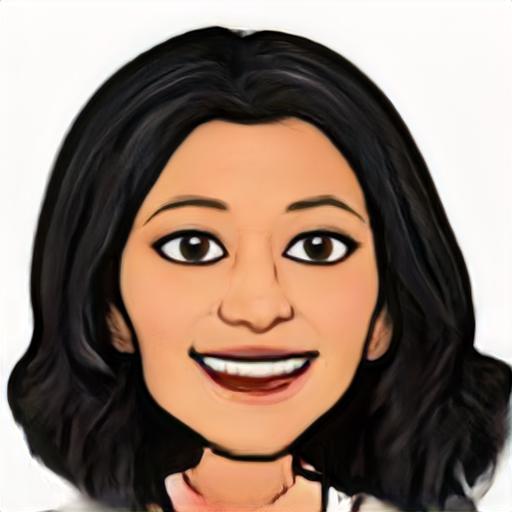}
    & \includegraphics[width=\www]{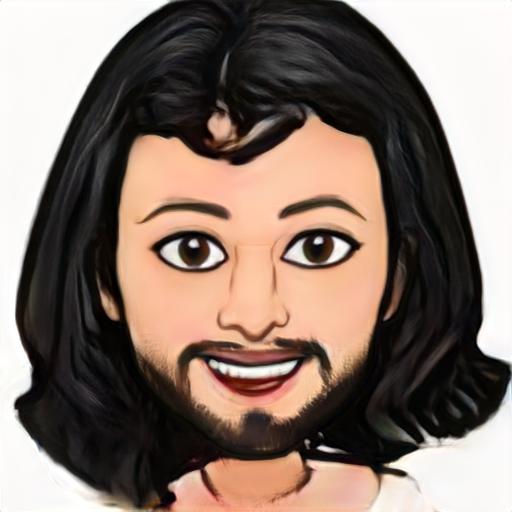}
    & \includegraphics[width=\www]{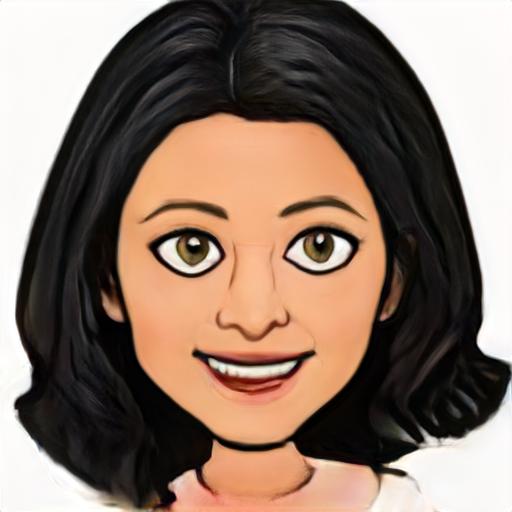}
    & \includegraphics[width=\www]{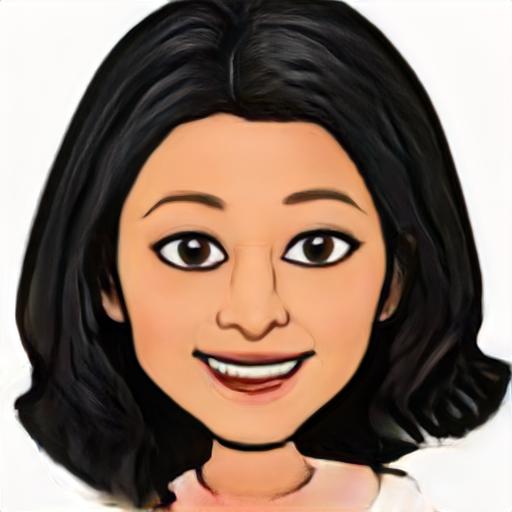}
    & \includegraphics[width=\www]{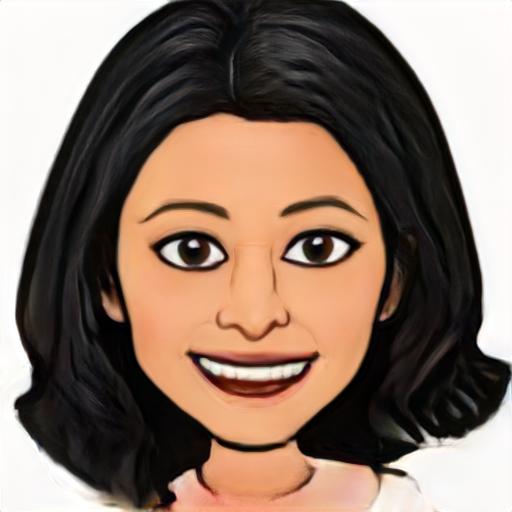}
    % & \includegraphics[width=\www]{fig_appendix/stylemix_bitmoji/ear/04.jpg}
    & \includegraphics[width=\www]{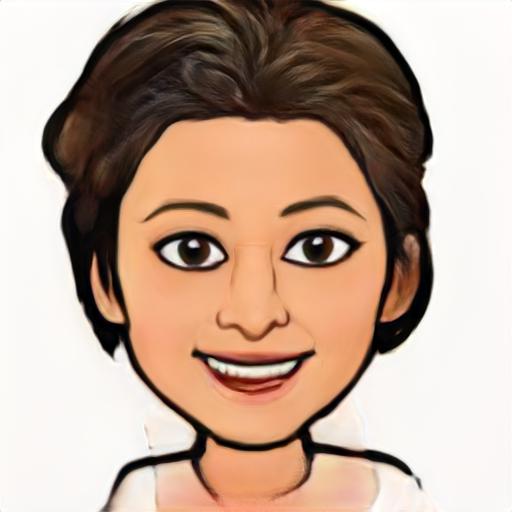} \\[0.2em]
    \includegraphics[width=\www]{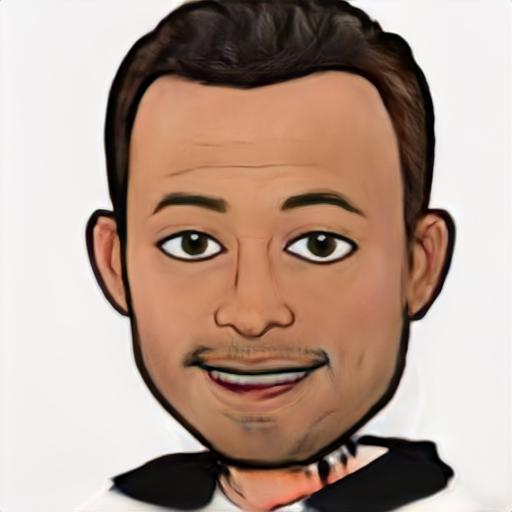}
    & \includegraphics[width=\www]{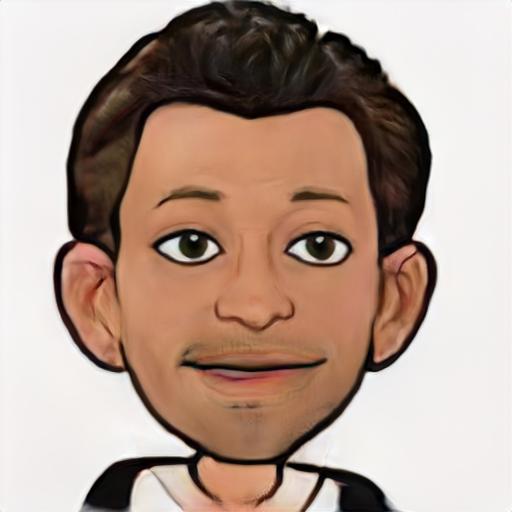}
    & \includegraphics[width=\www]{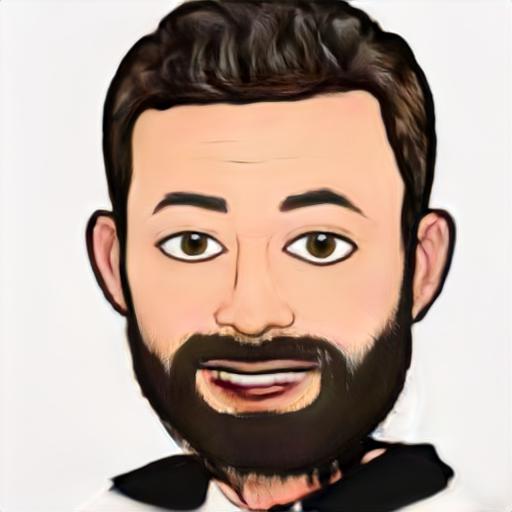}
    & \includegraphics[width=\www]{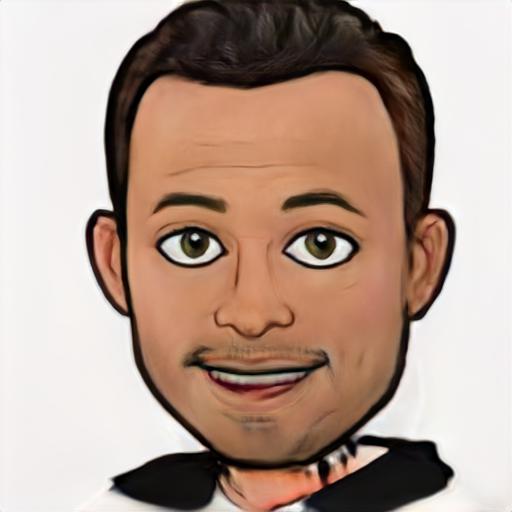}
    & \includegraphics[width=\www]{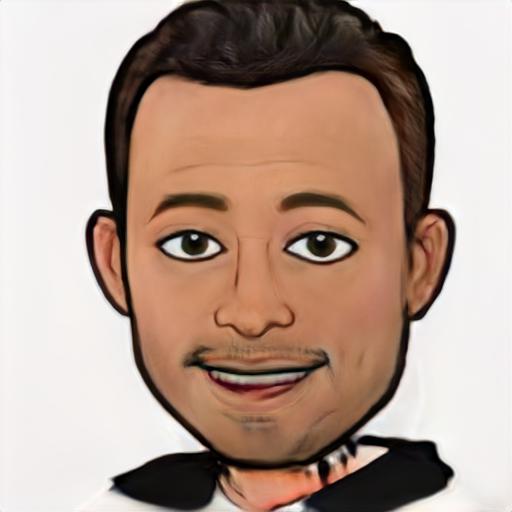}
    & \includegraphics[width=\www]{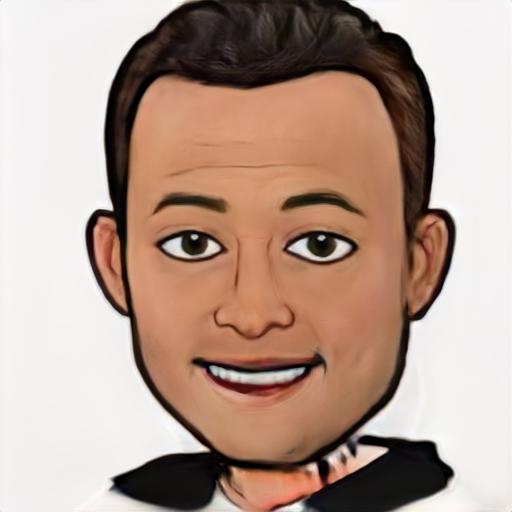}
    % & \includegraphics[width=\www]{fig_appendix/stylemix_bitmoji/ear/05.jpg}
    & \includegraphics[width=\www]{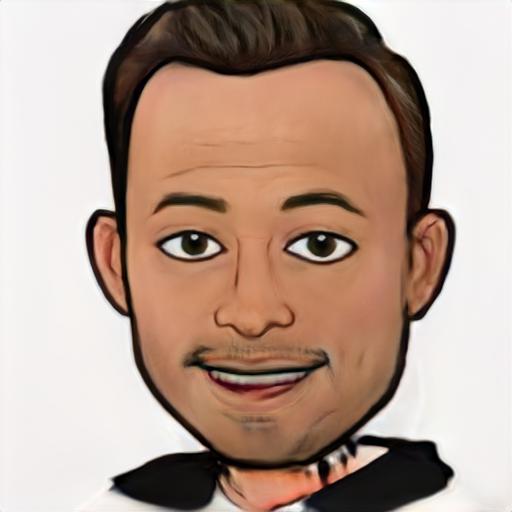} \\[0.2em]
    \includegraphics[width=\www]{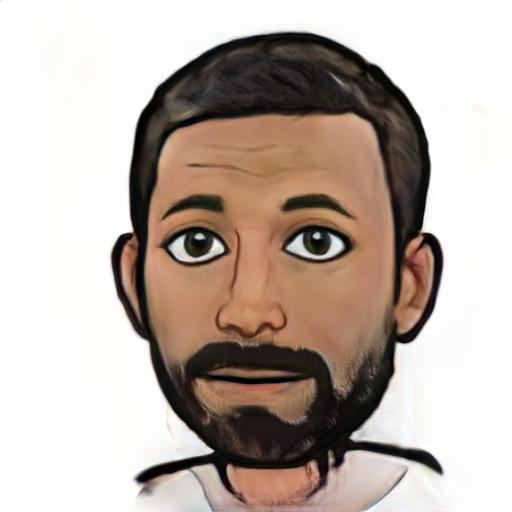}
    & \includegraphics[width=\www]{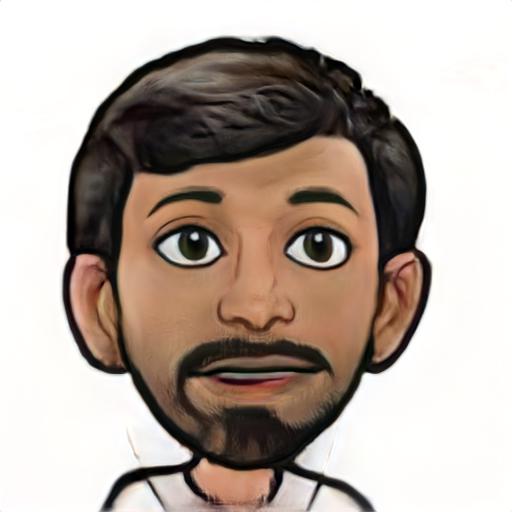}
    & \includegraphics[width=\www]{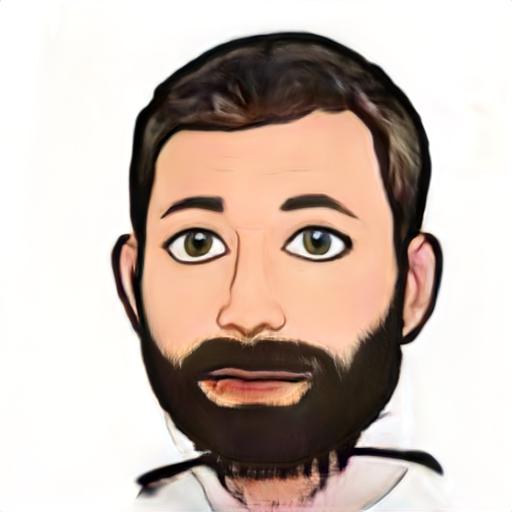}
    & \includegraphics[width=\www]{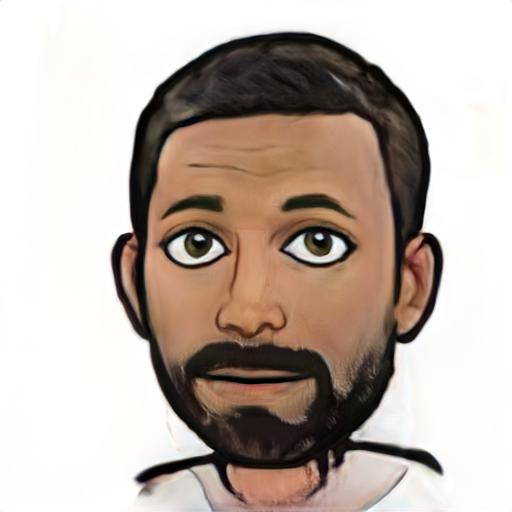}
    & \includegraphics[width=\www]{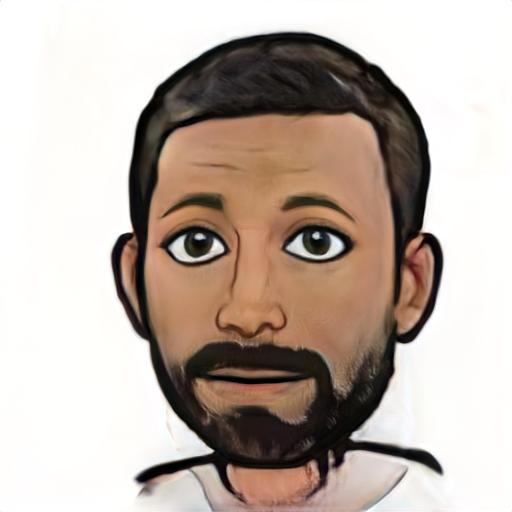}
    & \includegraphics[width=\www]{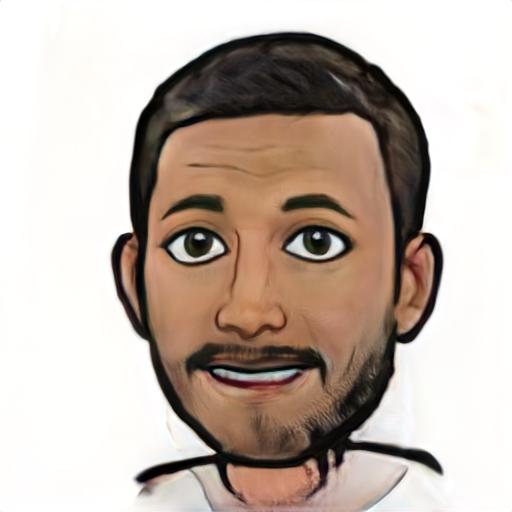}
    % & \includegraphics[width=\www]{fig_appendix/stylemix_bitmoji/ear/07.jpg}
    & \includegraphics[width=\www]{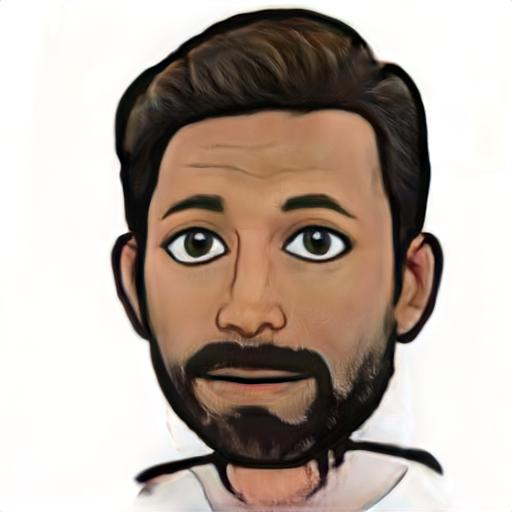} \\[0.2em]
    \includegraphics[width=\www]{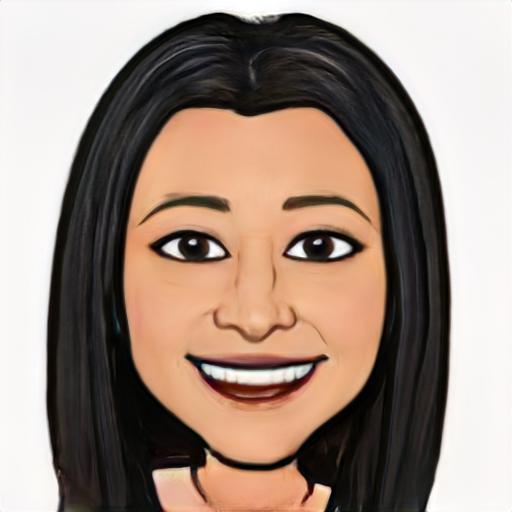}
    & \includegraphics[width=\www]{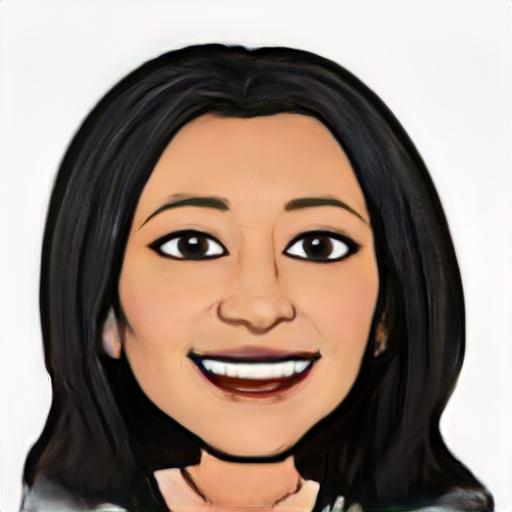}
    & \includegraphics[width=\www]{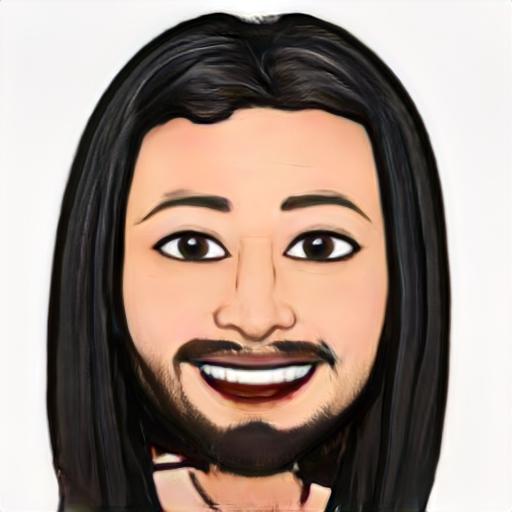}
    & \includegraphics[width=\www]{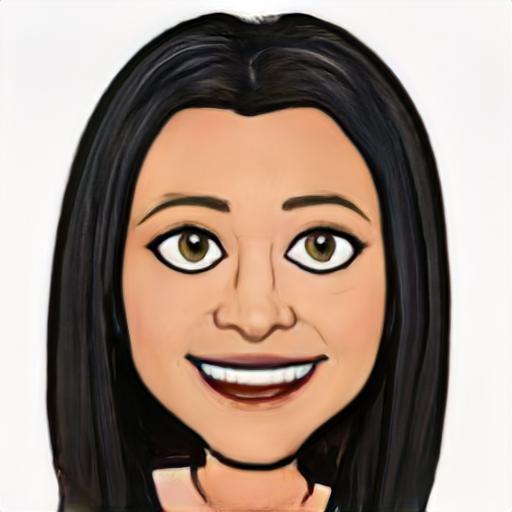}
    & \includegraphics[width=\www]{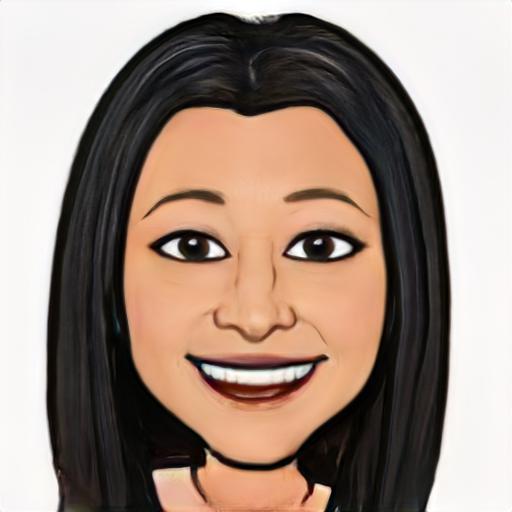}
    & \includegraphics[width=\www]{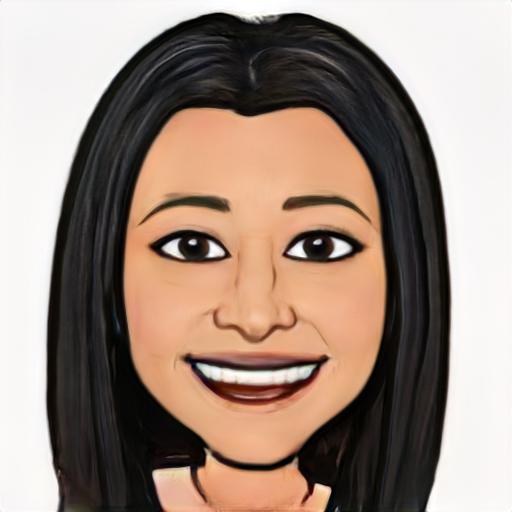}
    % & \includegraphics[width=\www]{fig_appendix/stylemix_bitmoji/ear/12.jpg}
    & \includegraphics[width=\www]{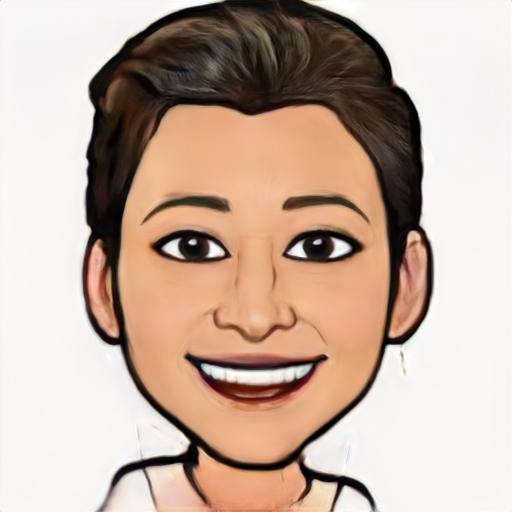} \\[0.2em]
    \includegraphics[width=\www]{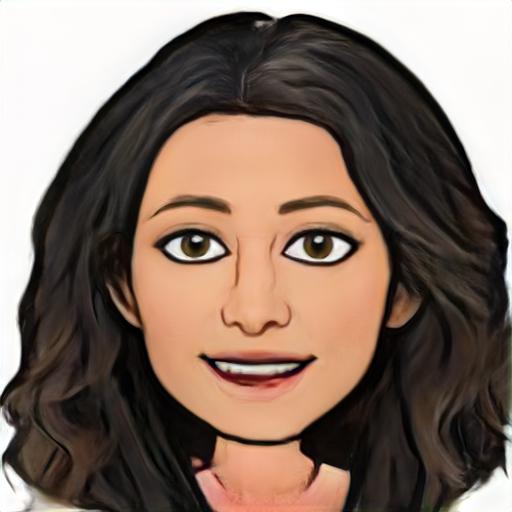}
    & \includegraphics[width=\www]{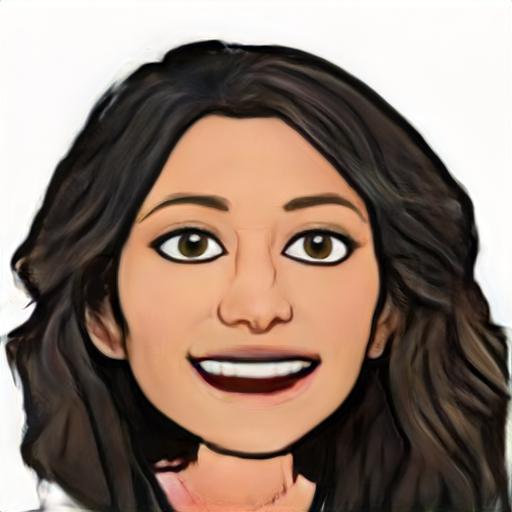}
    & \includegraphics[width=\www]{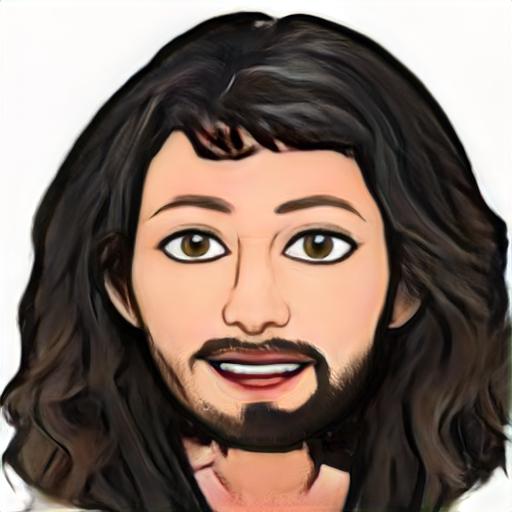}
    & \includegraphics[width=\www]{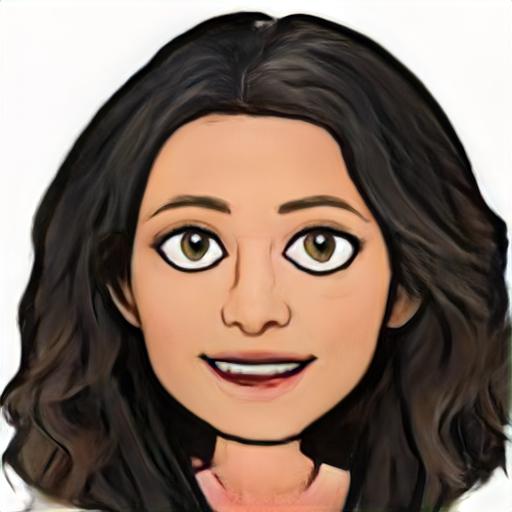}
    & \includegraphics[width=\www]{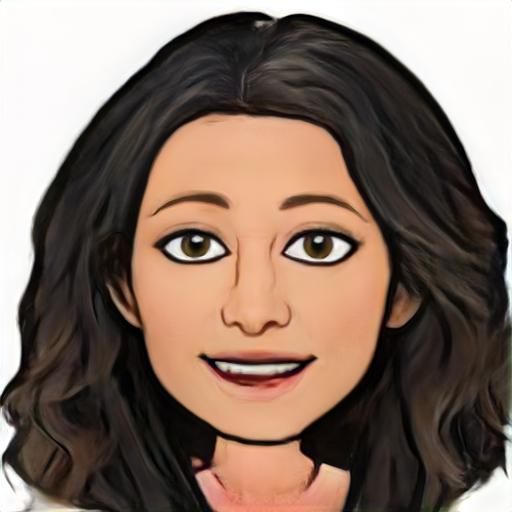}
    & \includegraphics[width=\www]{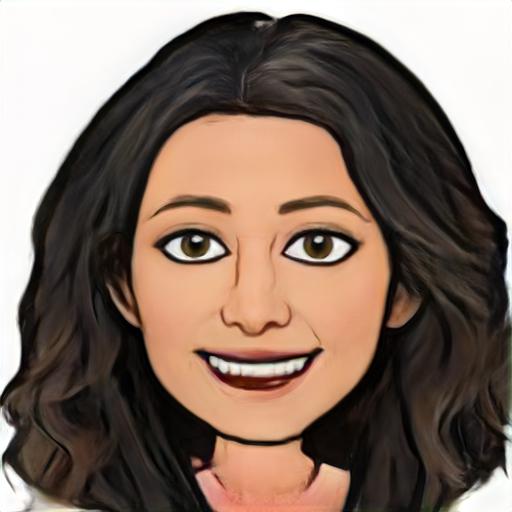}
    % & \includegraphics[width=\www]{fig_appendix/stylemix_bitmoji/ear/19.jpg}
    & \includegraphics[width=\www]{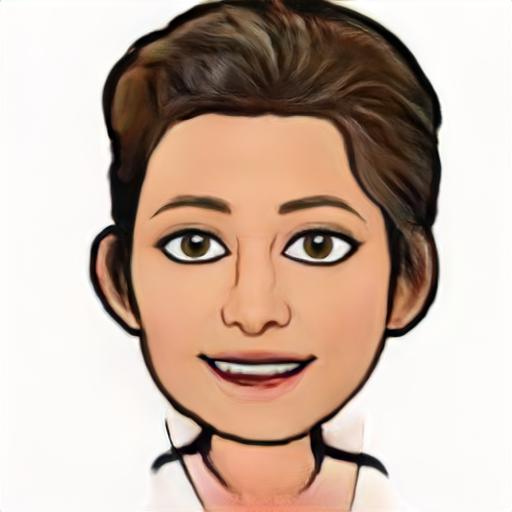} \\[0.2em]
\end{tabularx}
    \vspace{-1.0em}\caption{Local style mixing of the model fine-tuned on the BitMoji dataset. The first column shows randomly sampled images for editing. The remaining columns show the results of mixing local styles using the reference images in the first row. }\vspace{-1.0em}\vspace{-1.0em}
    \label{appendix:fig:stylemix_bitmoji}
\end{figure*}

\begin{figure*}[t]
\captionsetup{font=small}
\centering
\footnotesize
\setlength\tabcolsep{3px}
\newcommand{\www}{0.155\linewidth}
\renewcommand{\arraystretch}{0.5}
\newcolumntype{Y}{>{\centering\arraybackslash}X}
\begin{tabularx}{\linewidth}{ccccccccc}
    & Coarse Structure & Background & Hair & Up & Bottom\\
    & \includegraphics[width=\www]{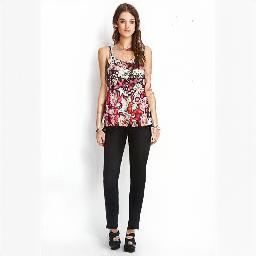}
    & \includegraphics[width=\www]{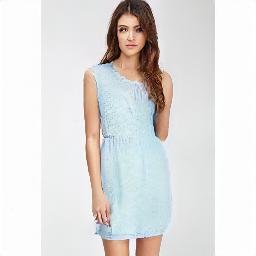}
    & \includegraphics[width=\www]{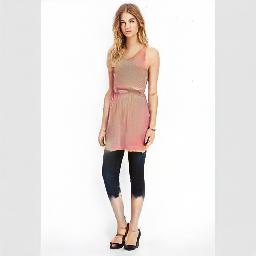}
    & \includegraphics[width=\www]{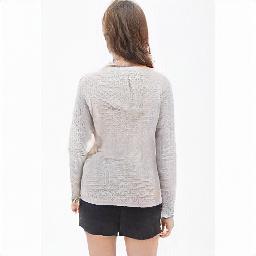}
    & \includegraphics[width=\www]{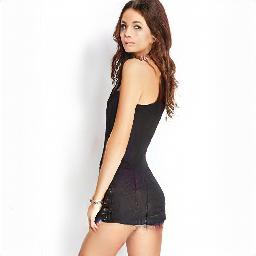}\\[0.5em]
    \includegraphics[width=\www]{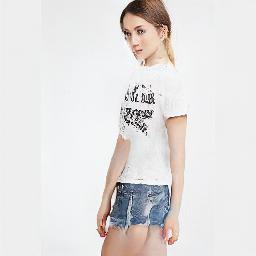}
    & \includegraphics[width=\www]{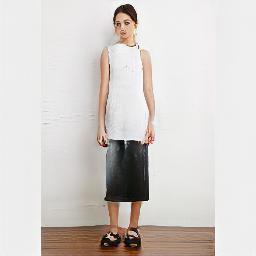}
    & \includegraphics[width=\www]{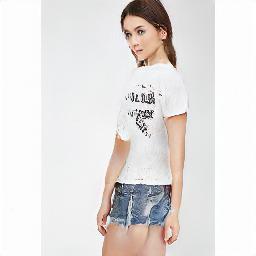}
    & \includegraphics[width=\www]{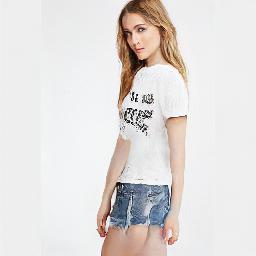}
    & \includegraphics[width=\www]{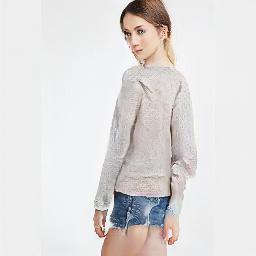}
    & \includegraphics[width=\www]{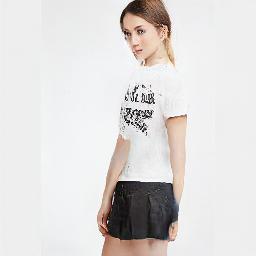}\\[0.5em]
    \includegraphics[width=\www]{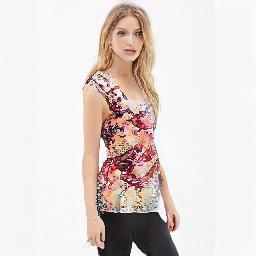}
    & \includegraphics[width=\www]{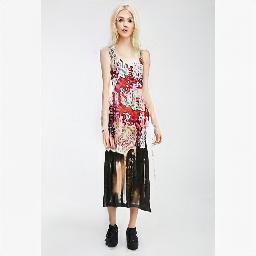}
    & \includegraphics[width=\www]{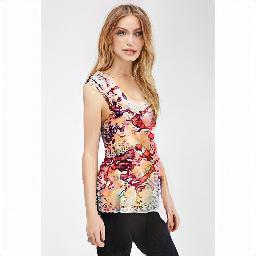}
    & \includegraphics[width=\www]{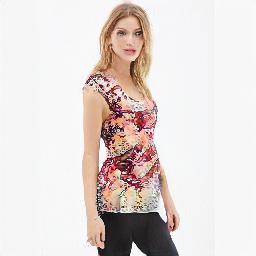}
    & \includegraphics[width=\www]{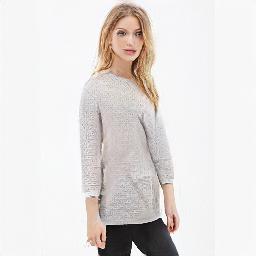}
    & \includegraphics[width=\www]{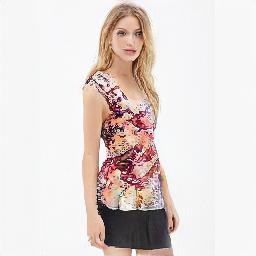}\\[0.5em]
    \includegraphics[width=\www]{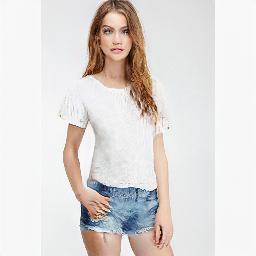}
    & \includegraphics[width=\www]{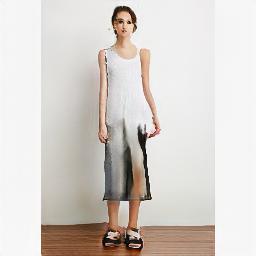}
    & \includegraphics[width=\www]{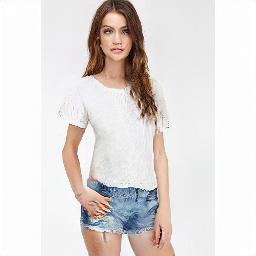}
    & \includegraphics[width=\www]{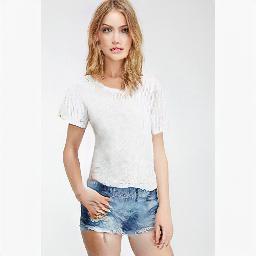}
    & \includegraphics[width=\www]{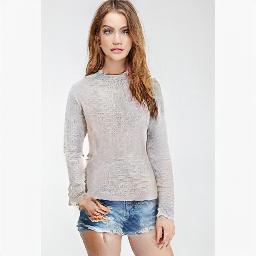}
    & \includegraphics[width=\www]{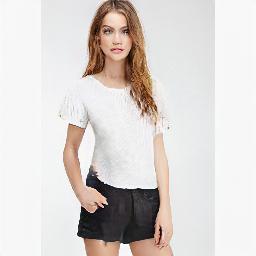}\\[0.5em]
    \includegraphics[width=\www]{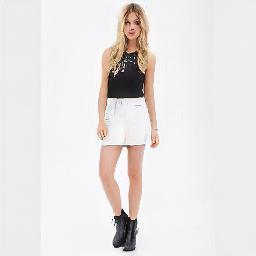}
    & \includegraphics[width=\www]{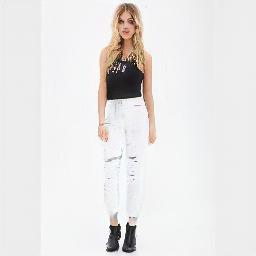}
    & \includegraphics[width=\www]{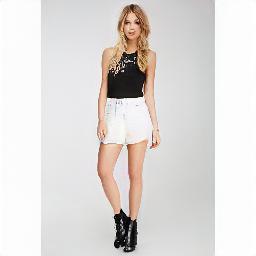}
    & \includegraphics[width=\www]{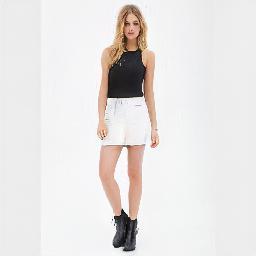}
    & \includegraphics[width=\www]{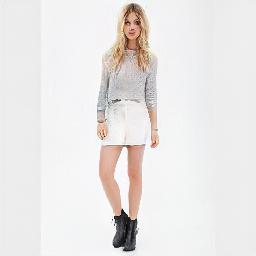}
    & \includegraphics[width=\www]{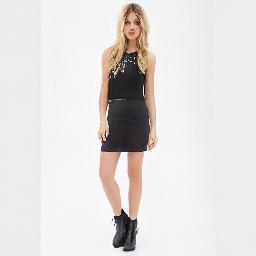}\\[0.5em]
    \includegraphics[width=\www]{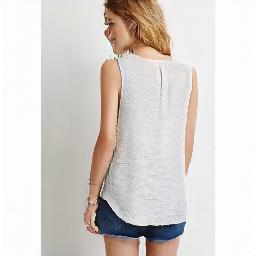}
    & \includegraphics[width=\www]{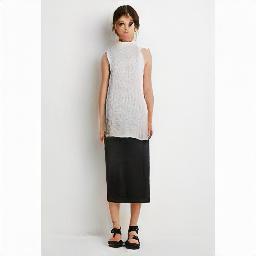}
    & \includegraphics[width=\www]{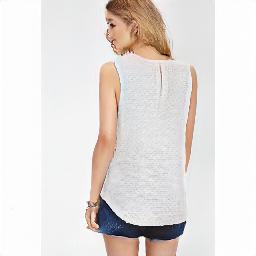}
    & \includegraphics[width=\www]{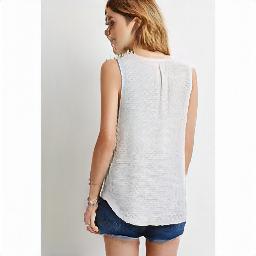}
    & \includegraphics[width=\www]{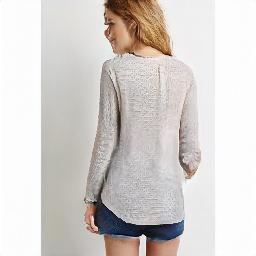}
    & \includegraphics[width=\www]{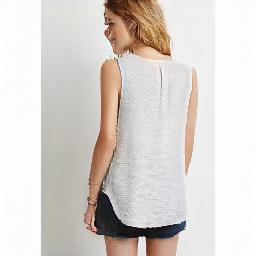}\\[0.5em]
    \includegraphics[width=\www]{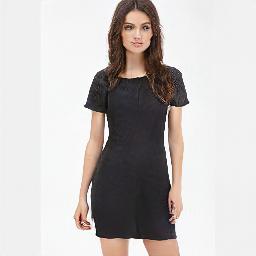}
    & \includegraphics[width=\www]{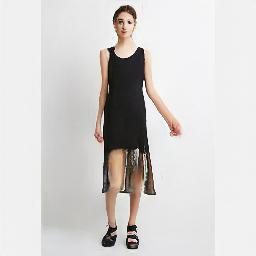}
    & \includegraphics[width=\www]{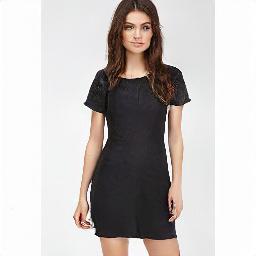}
    & \includegraphics[width=\www]{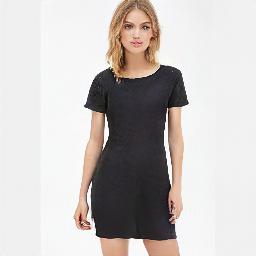}
    & \includegraphics[width=\www]{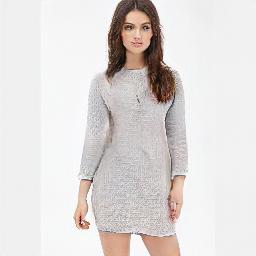}
    & \includegraphics[width=\www]{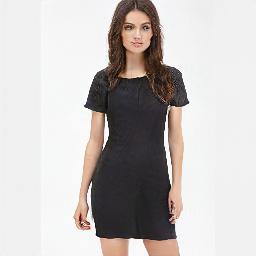}\\[0.5em]
\end{tabularx}
    \vspace{-1.0em}\caption{Local style mixing of the model trained on the DeepFashion dataset. The first column shows different randomly sampled images for editing. The remaining columns show the results of mixing local styles using the reference images in the first row. }\vspace{-1.0em}
    \label{appendix:fig:stylemix_deepfashion}
\end{figure*}

In the main paper, we showed that the proposed model can interpolate smoothly in a local latent space. Here, we show results on more fine-grained style mixing using our model. Different from StyleGANs~\cite{stylegan,stylegan2,karras2021alias}, we can conduct style mixing between local generators to transfer a certain semantic component from one image to another. This conducted by transferring both the shape code $\bw^k_s$ and texture code $\bw^k_t$. \cref{appendix:fig:stylemix} shows the results of semantic style mixing using our model trained on CelebAMask-HQ~\cite{CelebAMaskHQ}. Besides the local latent codes, we also show the transferring results of the coarse structure code $\bw_{base}$. It can be seen that our model is able to transfer most local component styles between images, including small components such as eyes and mouth. However, it is also observable that the coarse structure code is currently encoding some information about these local components, such as expression and hair. Although a user or developer is able to change the number of coarse structure codes dynamically during testing (and even manipulate all the layers in a local generator), we believe it would be beneficial to further regularize the information in the coarse code in the future.
 \cref{appendix:fig:stylemix_bitmoji} and  \cref{appendix:fig:stylemix_deepfashion} show the semantic style mixing results of a model after transfer learning (on BitMoji dataset~\cite{BitMoji}) and the model trained on DeepFashion dataset~\cite{liu2016deepfashion}. A similar effect can be seen on the DeepFashion that the coarse structure would affect certain components. Also, we see that sometimes the hair color is affected by the background on this dataset. Since the head in this dataset is rather small, we believe such entanglement is caused by the low-resolution (16$\times$16) feature map that was fed into render network for blending, which is originally selected for face datasets. Further tuning the hyper-parameters of the render net might alleviate such issues.
 
\begin{figure*}[t]
\captionsetup{font=small}
\centering
\scriptsize
\setlength\tabcolsep{1px}
\newcommand{\www}{0.123\linewidth}
\renewcommand{\arraystretch}{0.5}
\newcolumntype{Y}{>{\centering\arraybackslash}X}
\begin{tabularx}{\linewidth}{cccc}
    \includegraphics[width=\www]{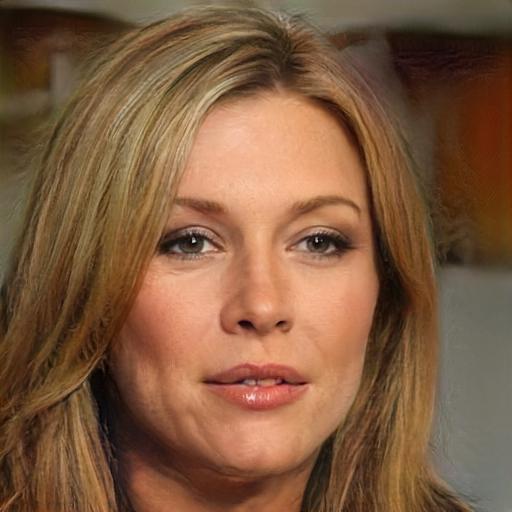}\hfill
    \includegraphics[width=\www]{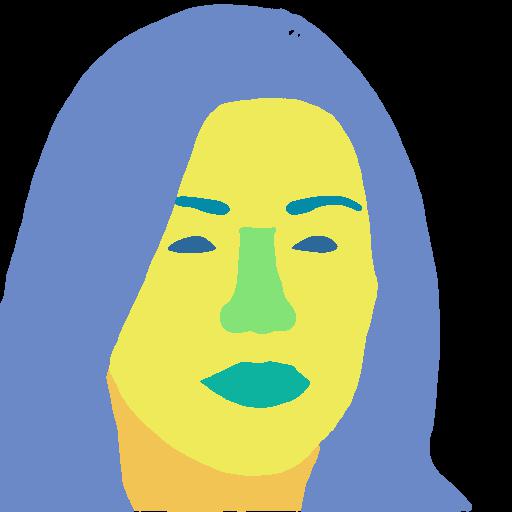} & 
    \includegraphics[width=\www]{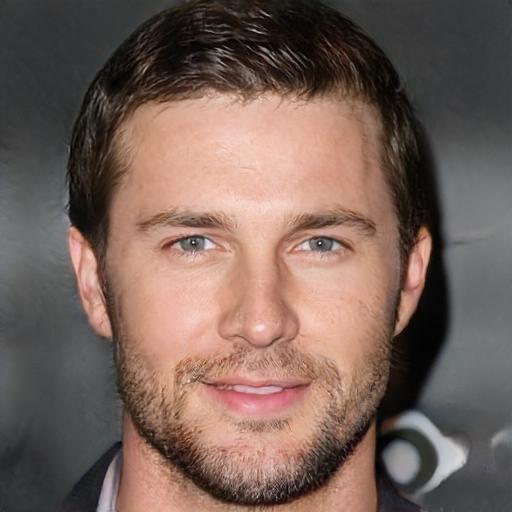}\hfill
    \includegraphics[width=\www]{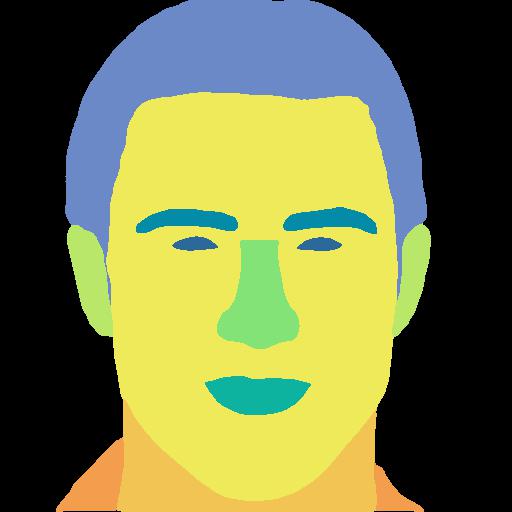} & 
    \includegraphics[width=\www]{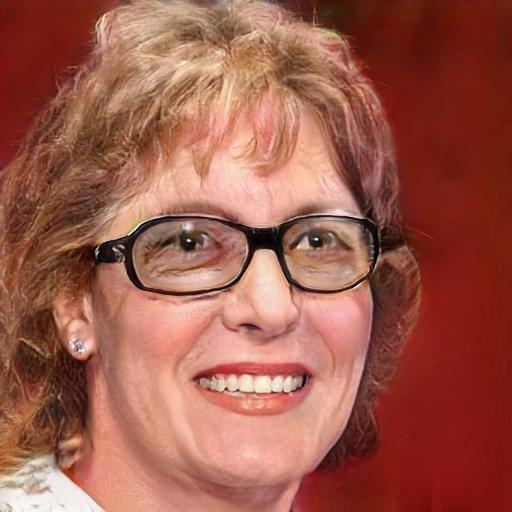}\hfill
    \includegraphics[width=\www]{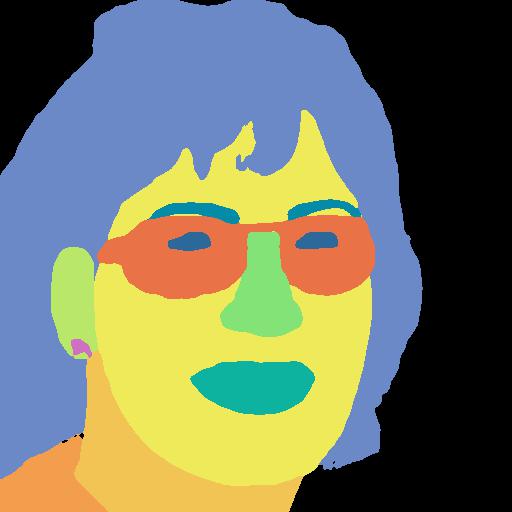} & 
    \includegraphics[width=\www]{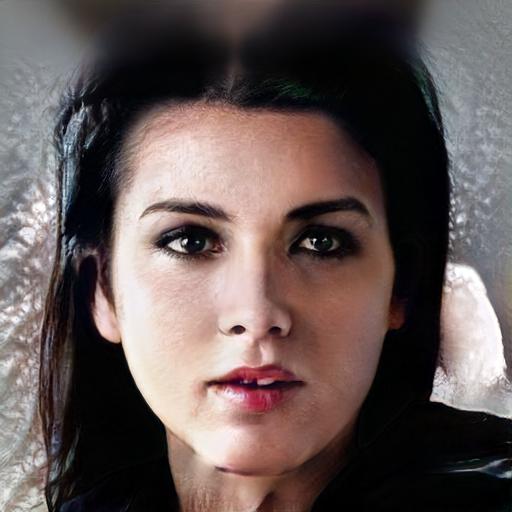}\hfill
    \includegraphics[width=\www]{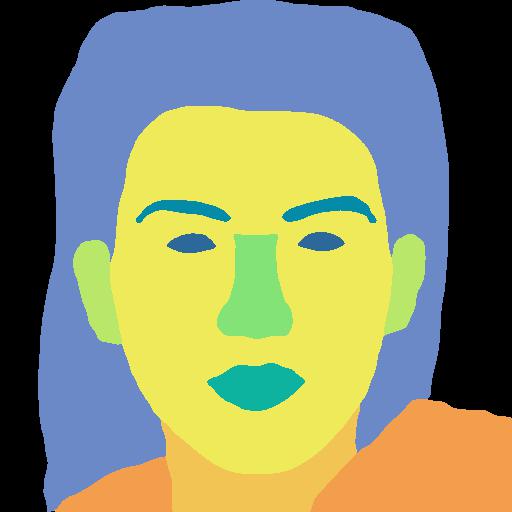} \\
    \includegraphics[width=\www]{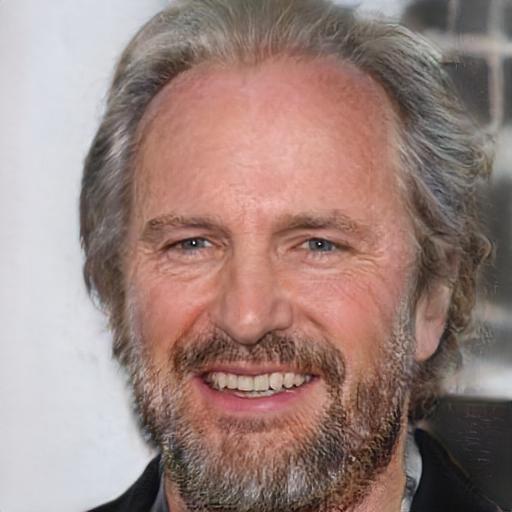}\hfill
    \includegraphics[width=\www]{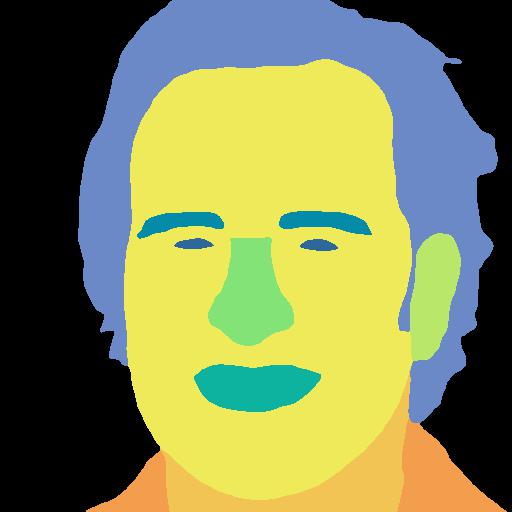} & 
    \includegraphics[width=\www]{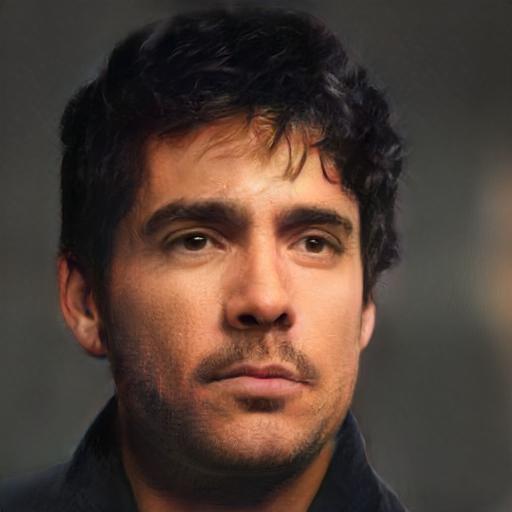}\hfill
    \includegraphics[width=\www]{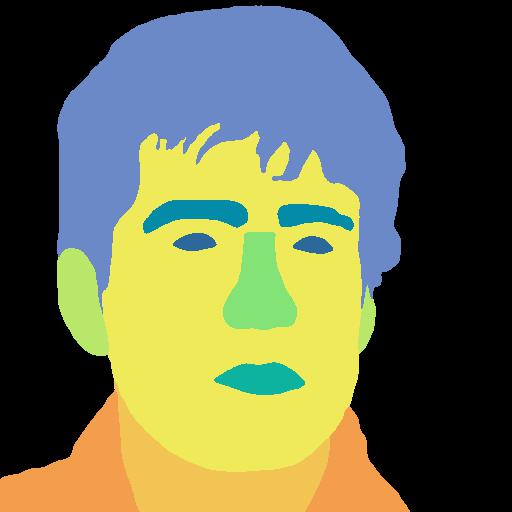} & 
    \includegraphics[width=\www]{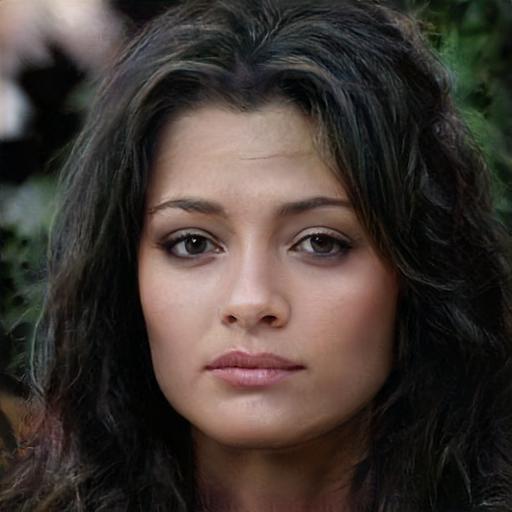}\hfill
    \includegraphics[width=\www]{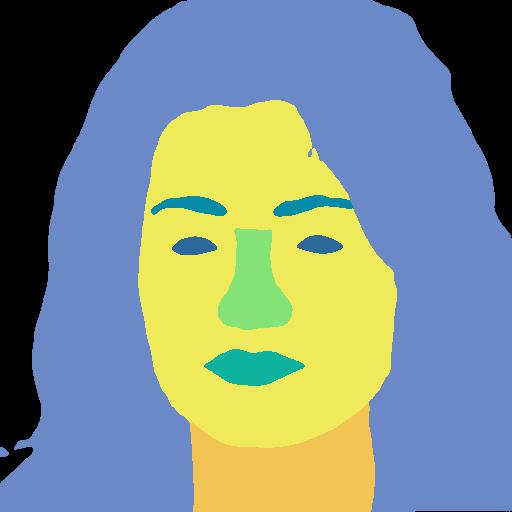} & 
    \includegraphics[width=\www]{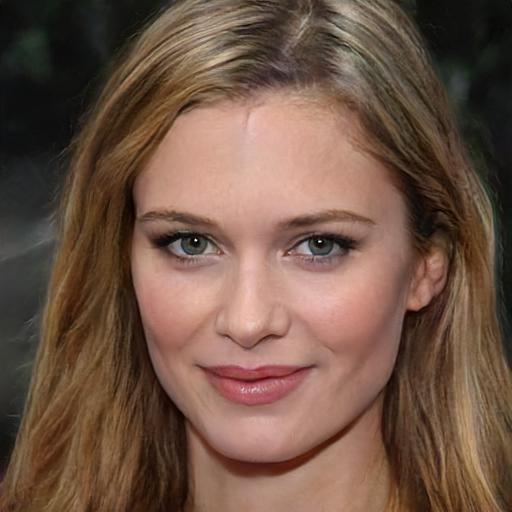}\hfill
    \includegraphics[width=\www]{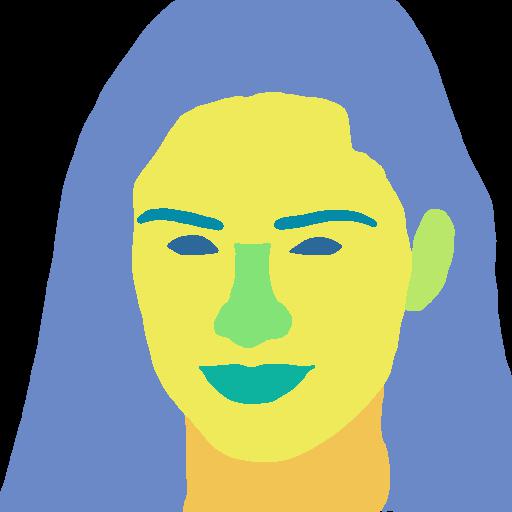} \\
    \includegraphics[width=\www]{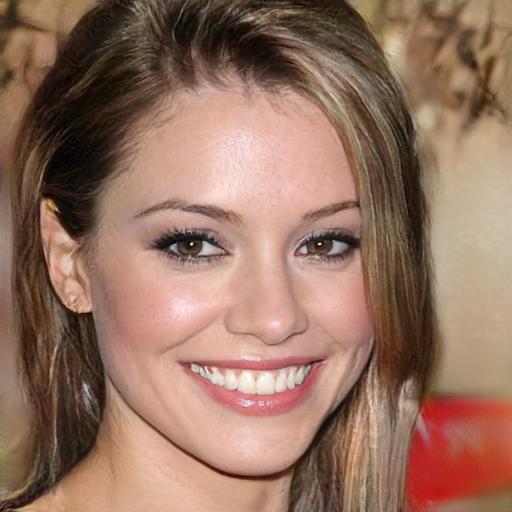}\hfill
    \includegraphics[width=\www]{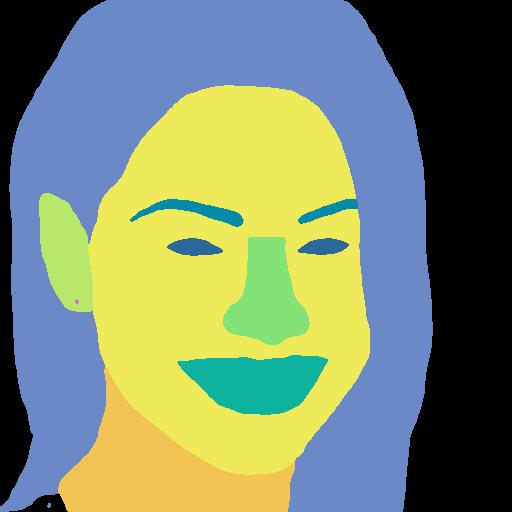} & 
    \includegraphics[width=\www]{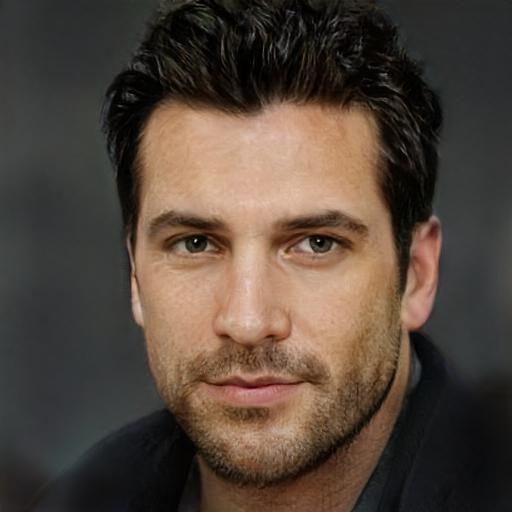}\hfill
    \includegraphics[width=\www]{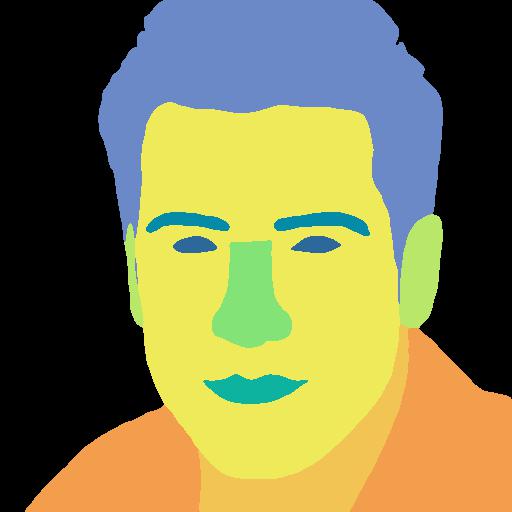} & 
    \includegraphics[width=\www]{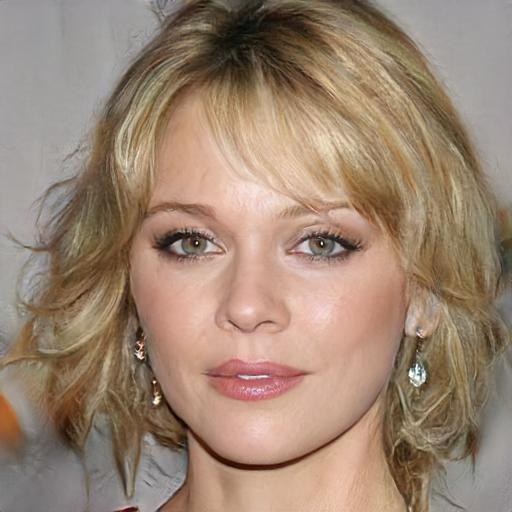}\hfill
    \includegraphics[width=\www]{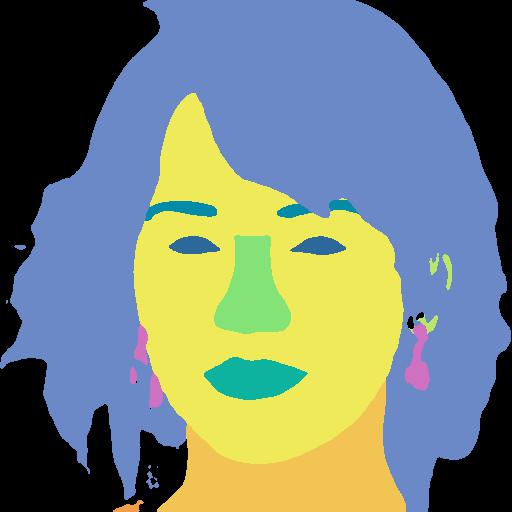} & 
    \includegraphics[width=\www]{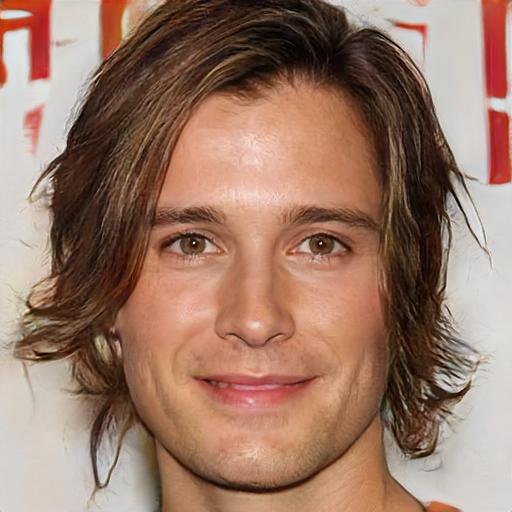}\hfill
    \includegraphics[width=\www]{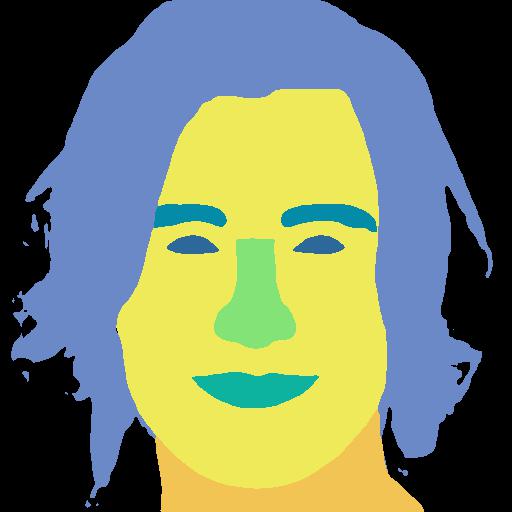} \\
    \includegraphics[width=\www]{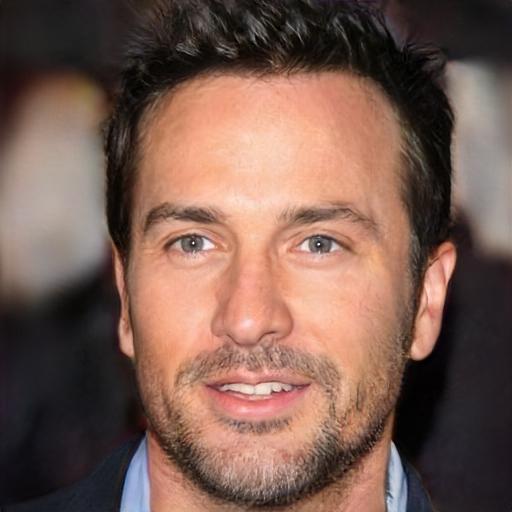}\hfill
    \includegraphics[width=\www]{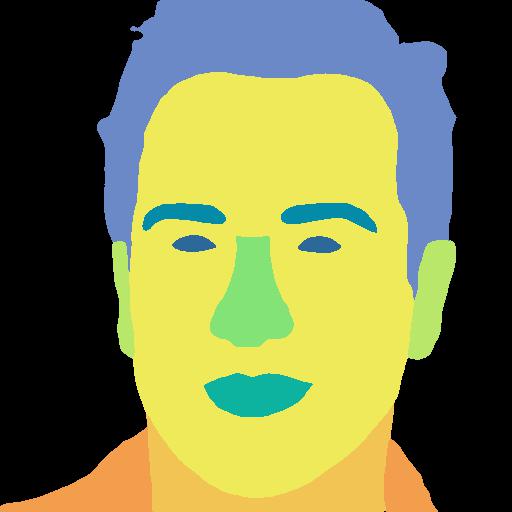} & 
    \includegraphics[width=\www]{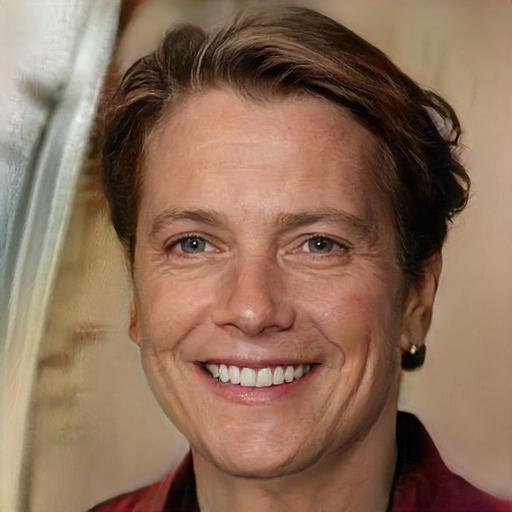}\hfill
    \includegraphics[width=\www]{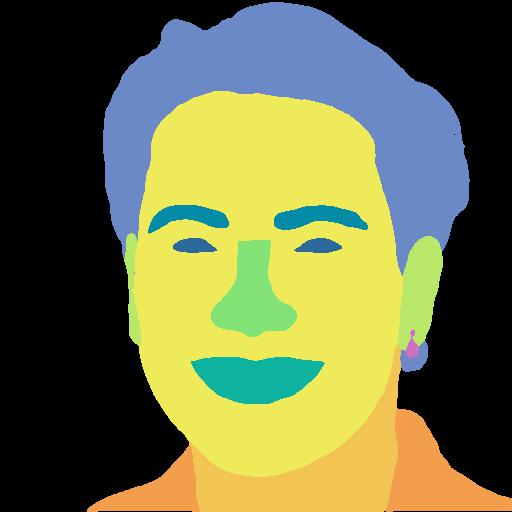} & 
    \includegraphics[width=\www]{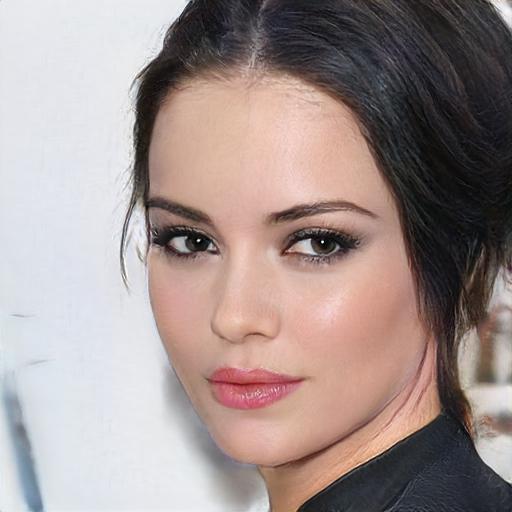}\hfill
    \includegraphics[width=\www]{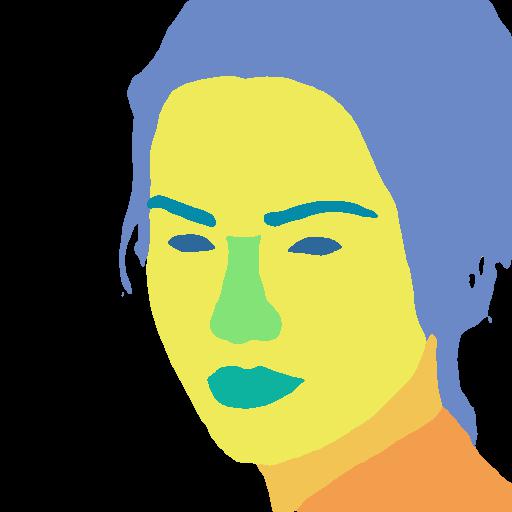} & 
    \includegraphics[width=\www]{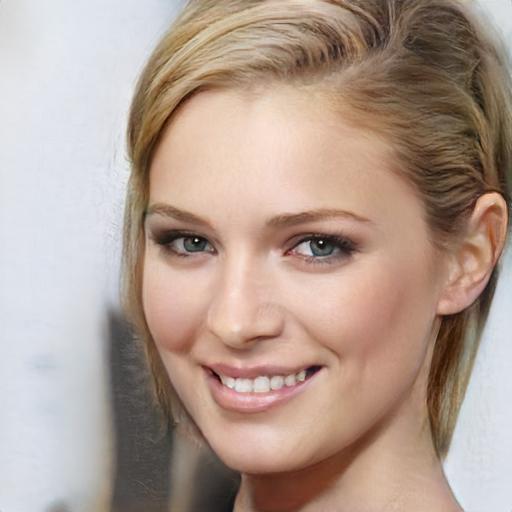}\hfill
    \includegraphics[width=\www]{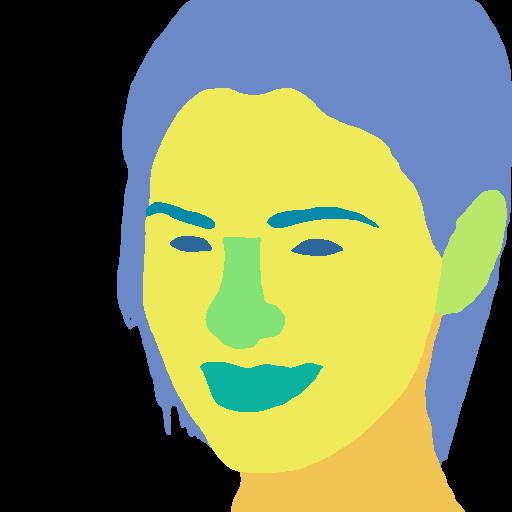} \\
    \includegraphics[width=\www]{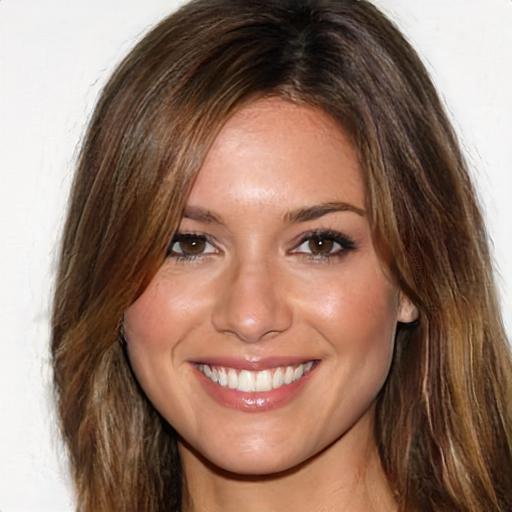}\hfill
    \includegraphics[width=\www]{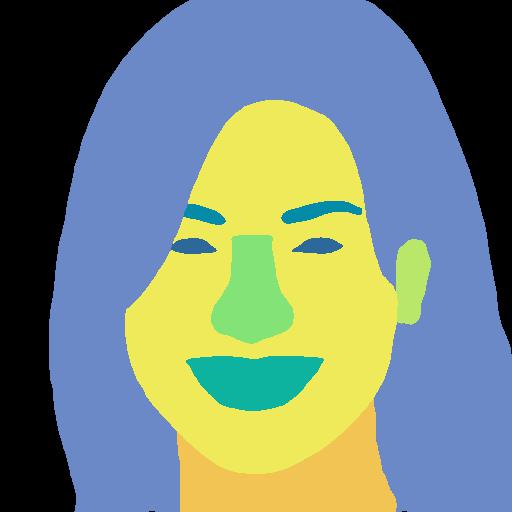} & 
    \includegraphics[width=\www]{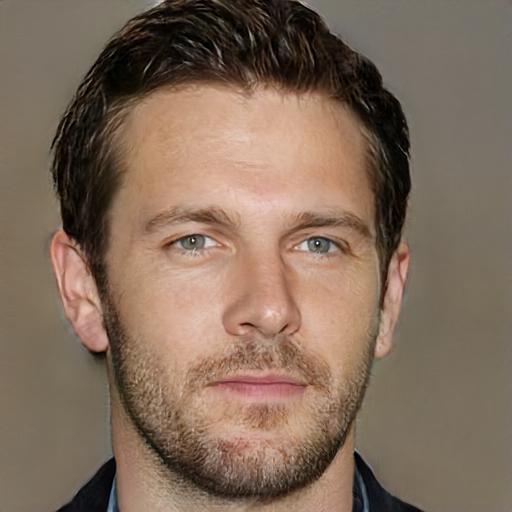}\hfill
    \includegraphics[width=\www]{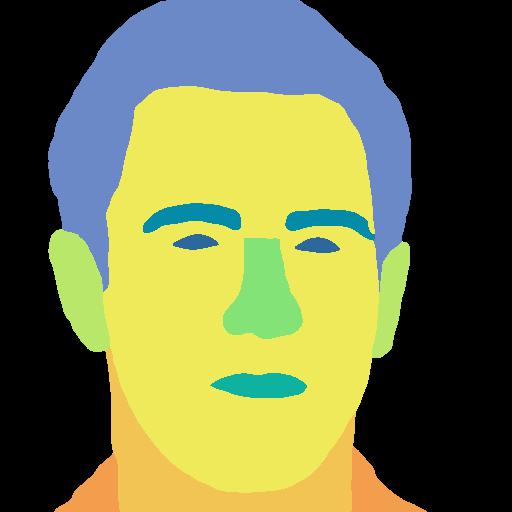} & 
    \includegraphics[width=\www]{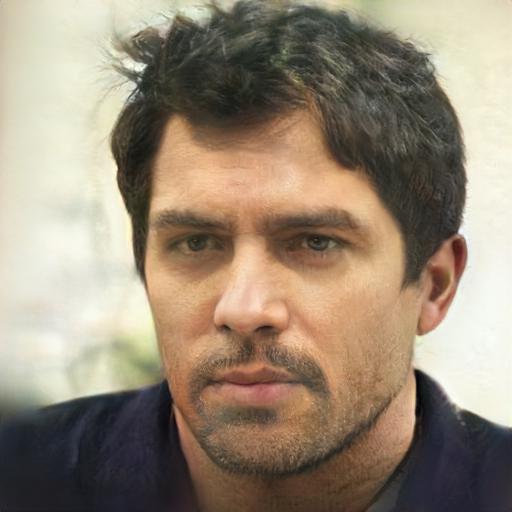}\hfill
    \includegraphics[width=\www]{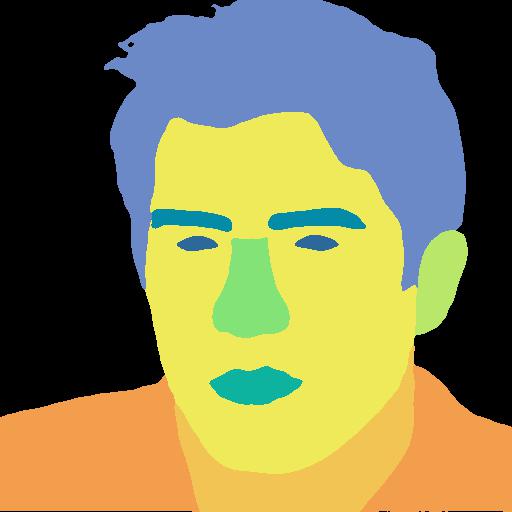} &
    \includegraphics[width=\www]{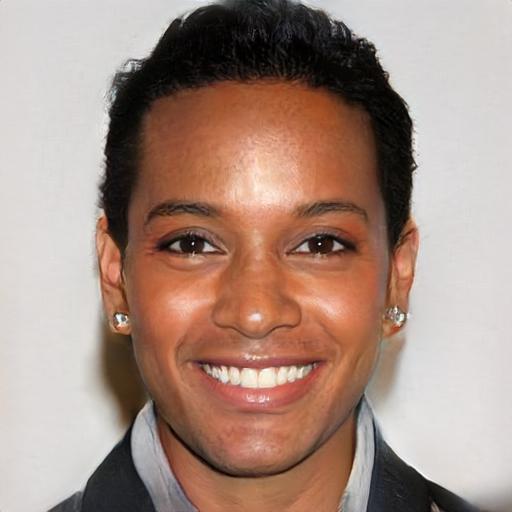}\hfill
    \includegraphics[width=\www]{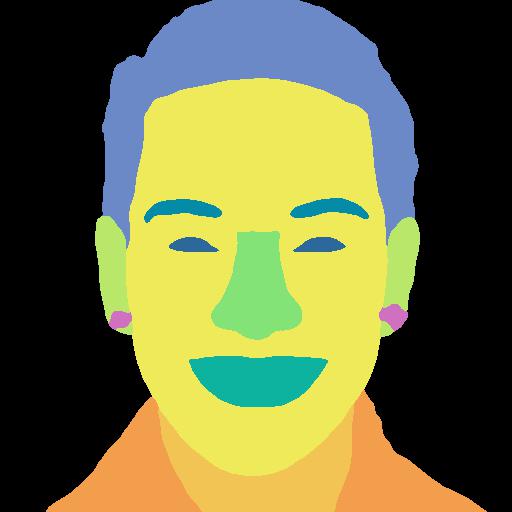} \\
    \includegraphics[width=\www]{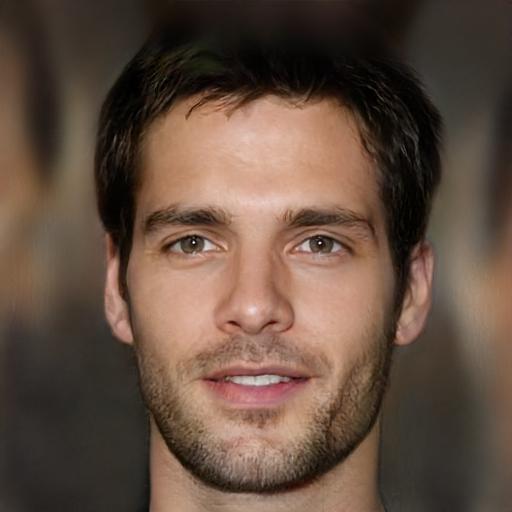}\hfill
    \includegraphics[width=\www]{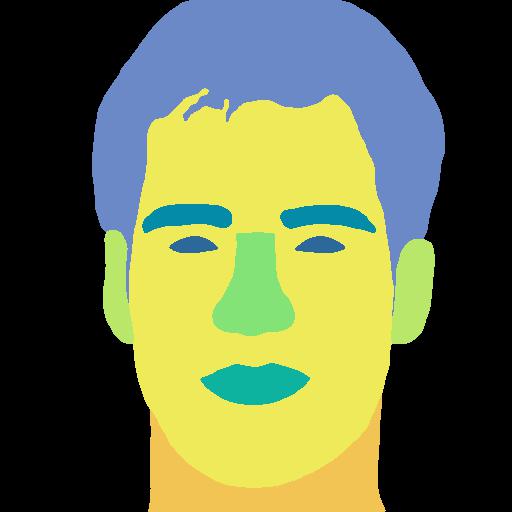} & 
    \includegraphics[width=\www]{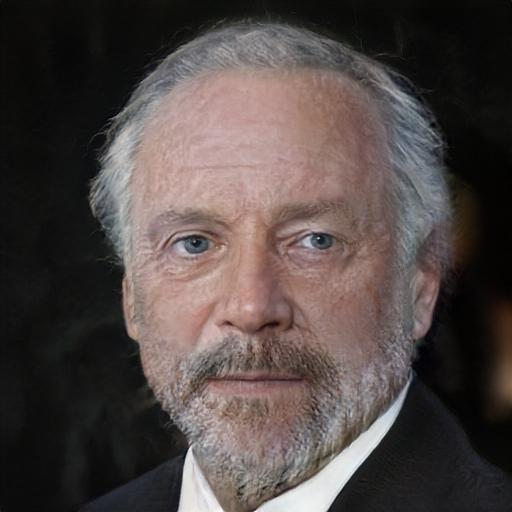}\hfill
    \includegraphics[width=\www]{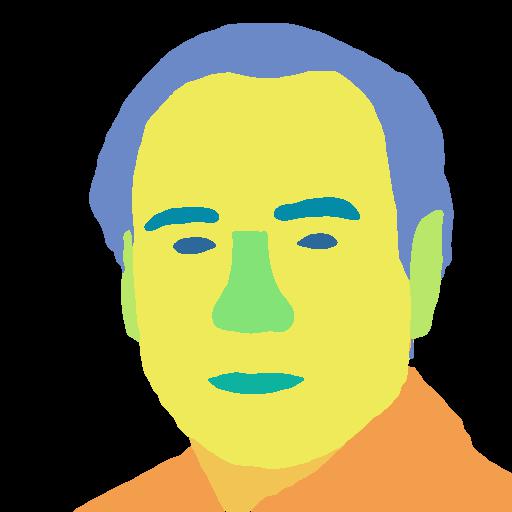} & 
    \includegraphics[width=\www]{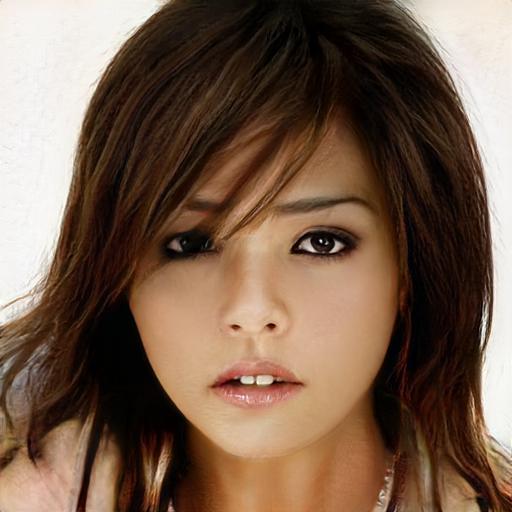}\hfill
    \includegraphics[width=\www]{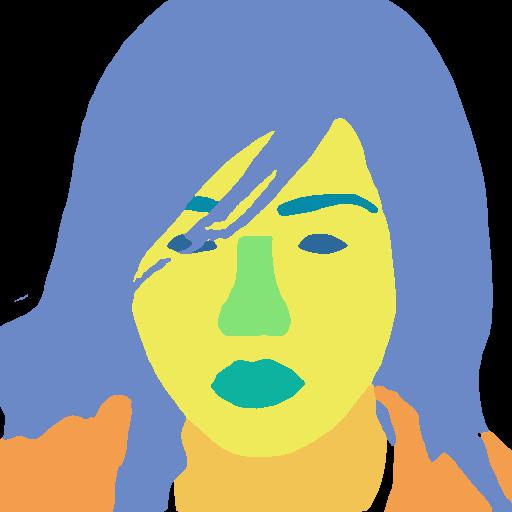} &
    \includegraphics[width=\www]{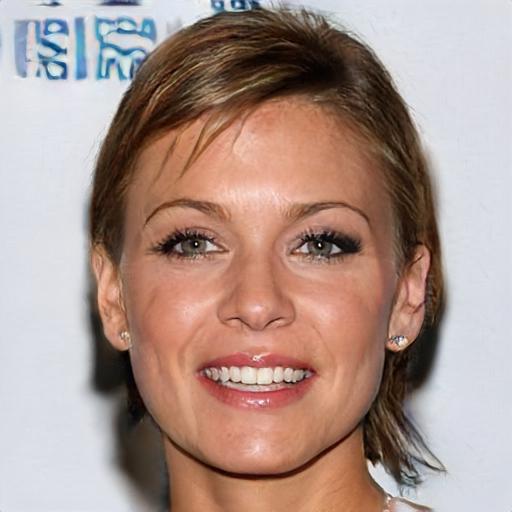}\hfill
    \includegraphics[width=\www]{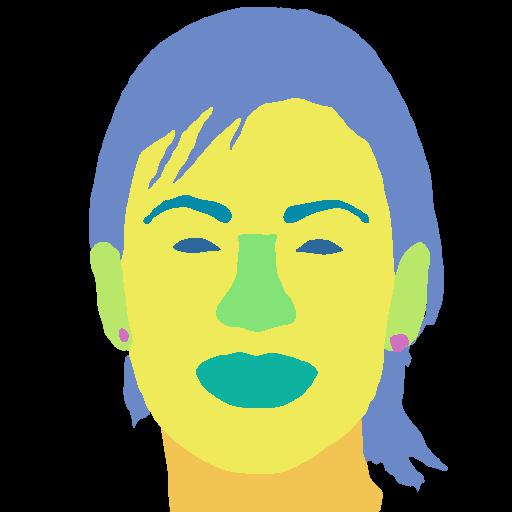} \\
    \includegraphics[width=\www]{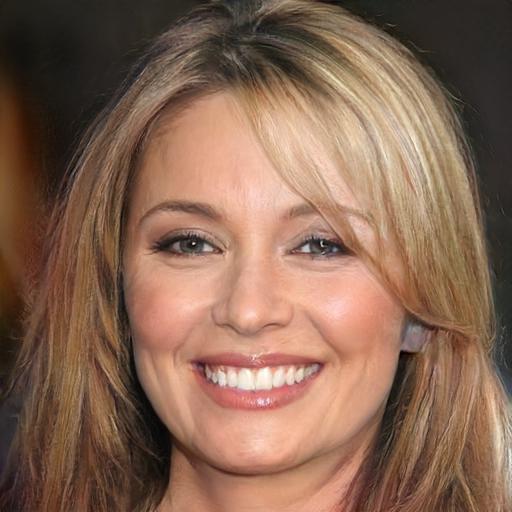}\hfill
    \includegraphics[width=\www]{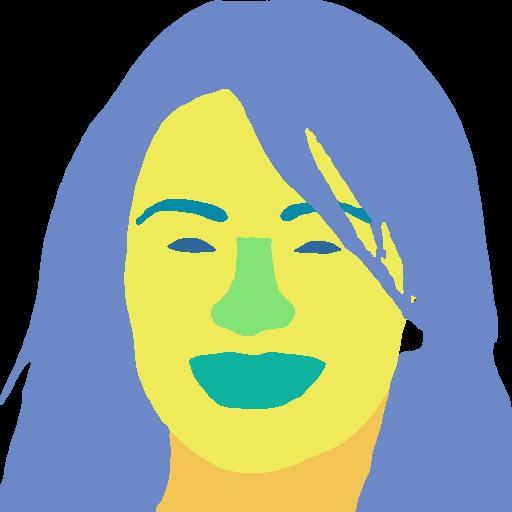} & 
    \includegraphics[width=\www]{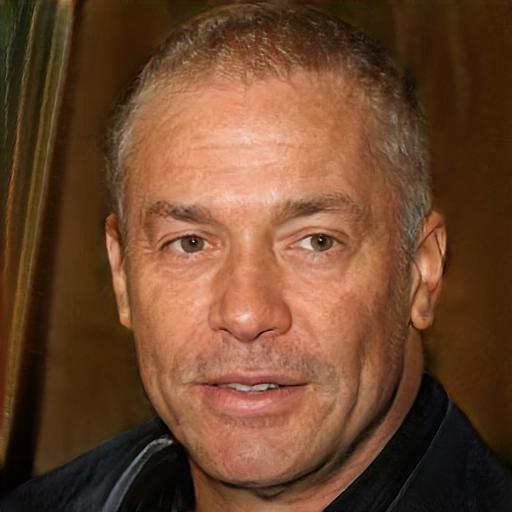}\hfill
    \includegraphics[width=\www]{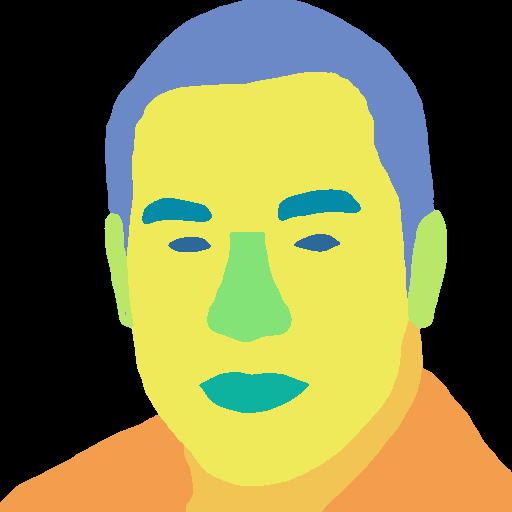} & 
    \includegraphics[width=\www]{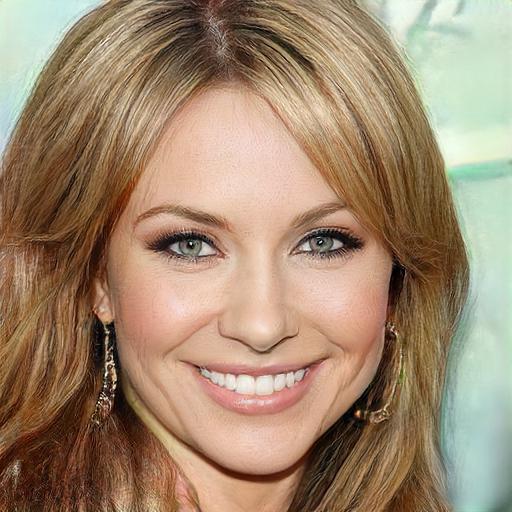}\hfill
    \includegraphics[width=\www]{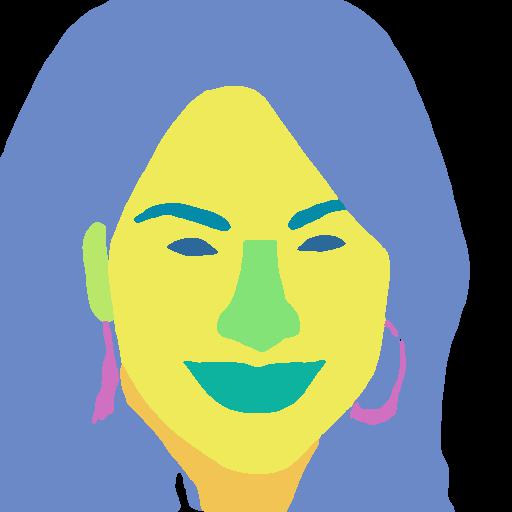} &
    \includegraphics[width=\www]{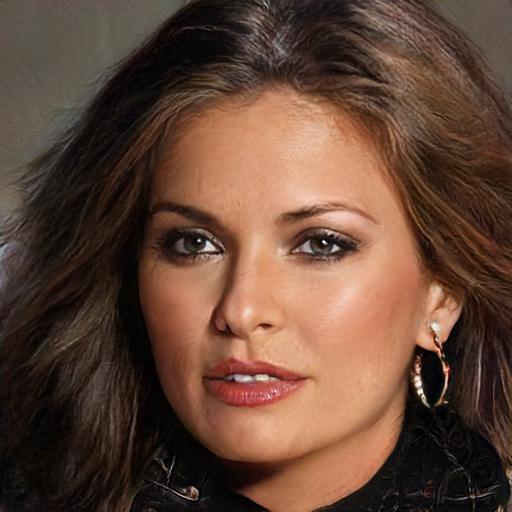}\hfill
    \includegraphics[width=\www]{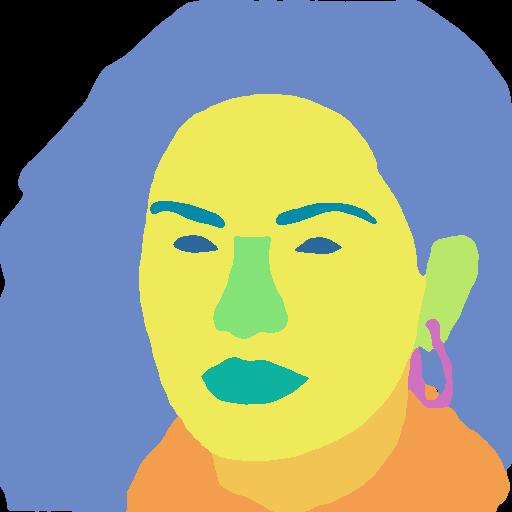} \\
    \includegraphics[width=\www]{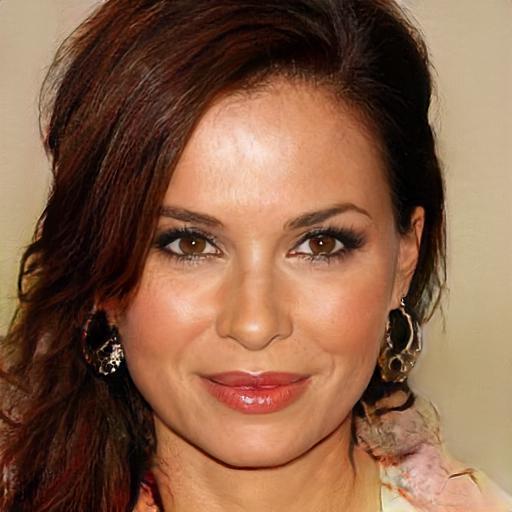}\hfill
    \includegraphics[width=\www]{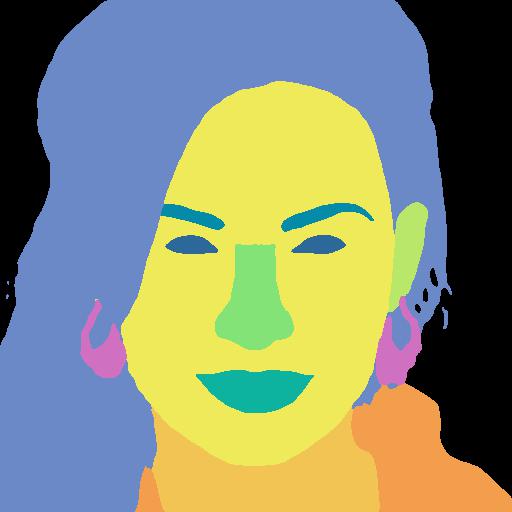} & 
    \includegraphics[width=\www]{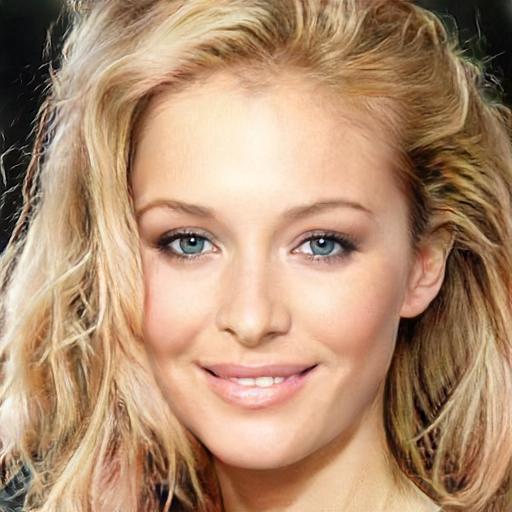}\hfill
    \includegraphics[width=\www]{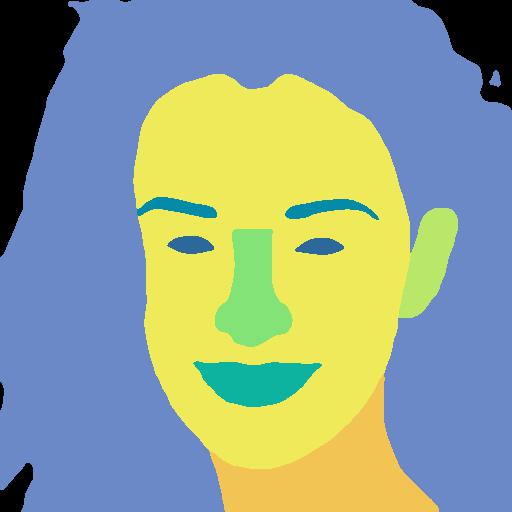} & 
    \includegraphics[width=\www]{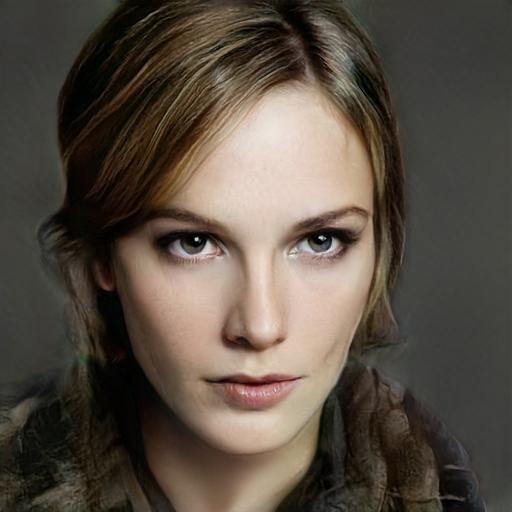}\hfill
    \includegraphics[width=\www]{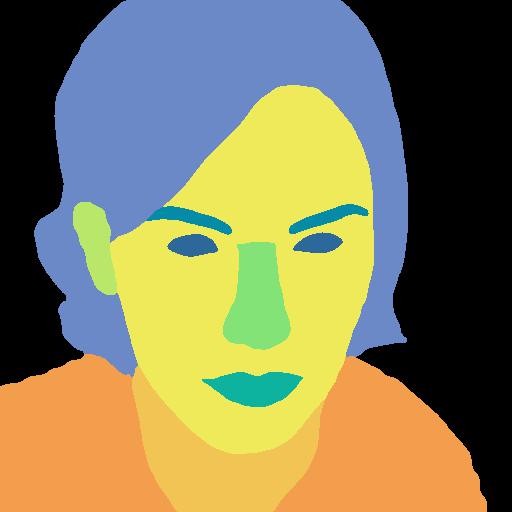} &
    \includegraphics[width=\www]{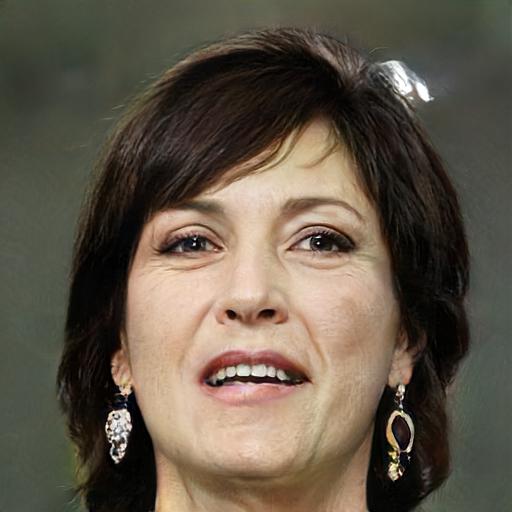}\hfill
    \includegraphics[width=\www]{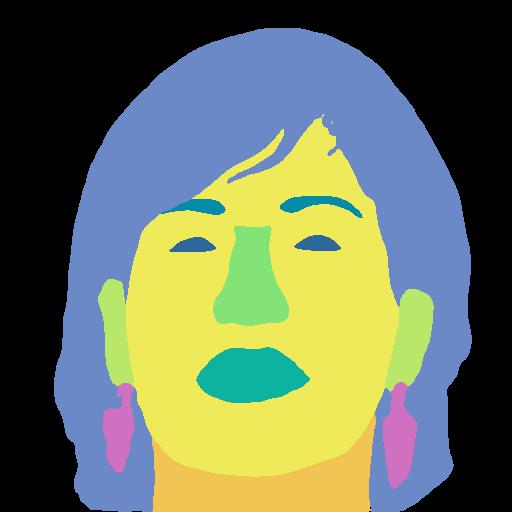} \\
    \includegraphics[width=\www]{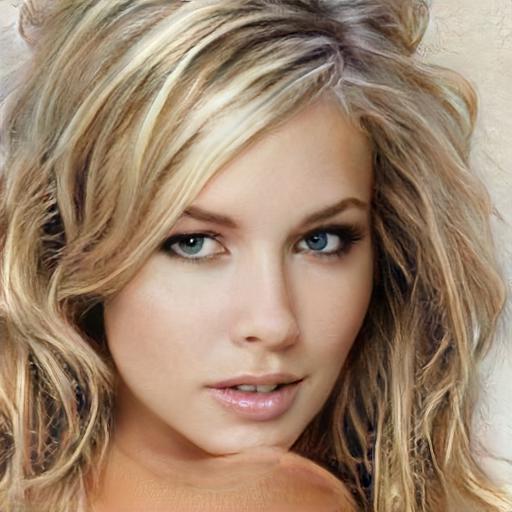}\hfill
    \includegraphics[width=\www]{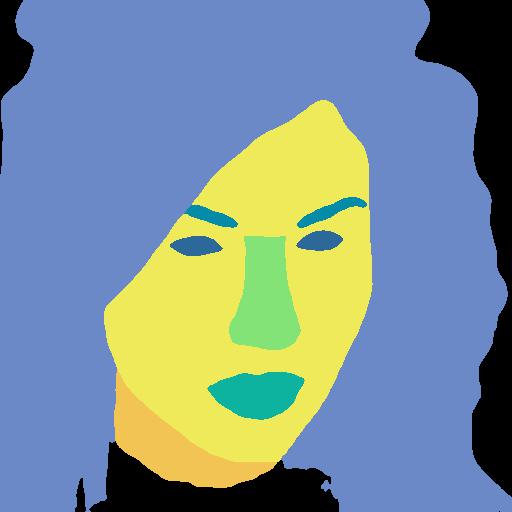} & 
    \includegraphics[width=\www]{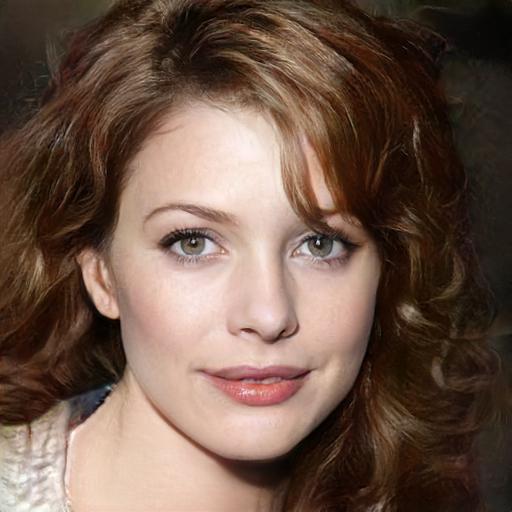}\hfill
    \includegraphics[width=\www]{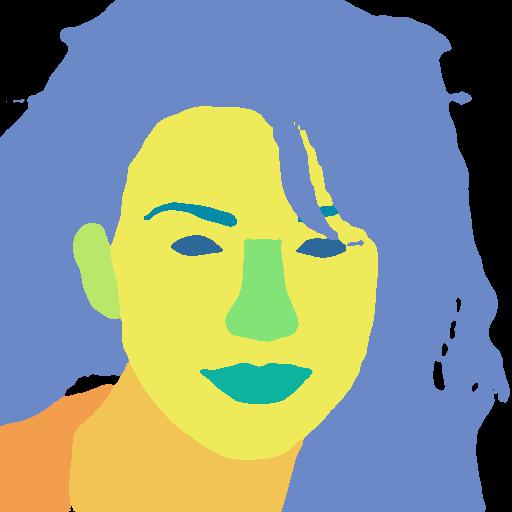} & 
    \includegraphics[width=\www]{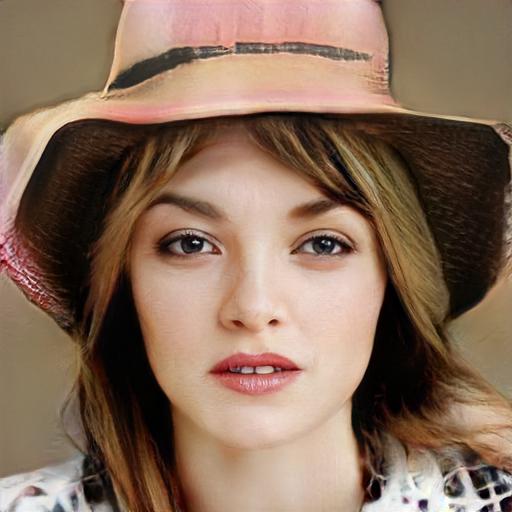}\hfill
    \includegraphics[width=\www]{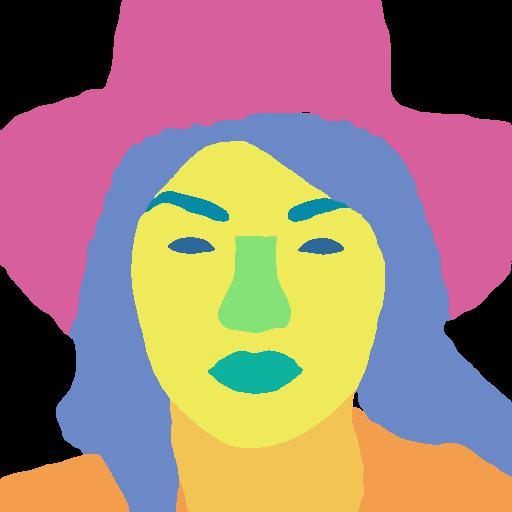} &
    \includegraphics[width=\www]{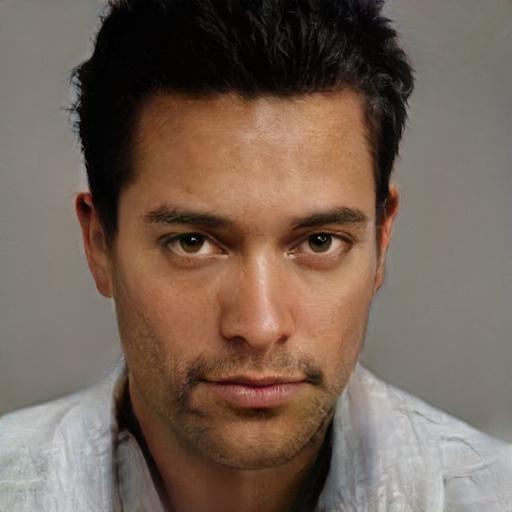}\hfill
    \includegraphics[width=\www]{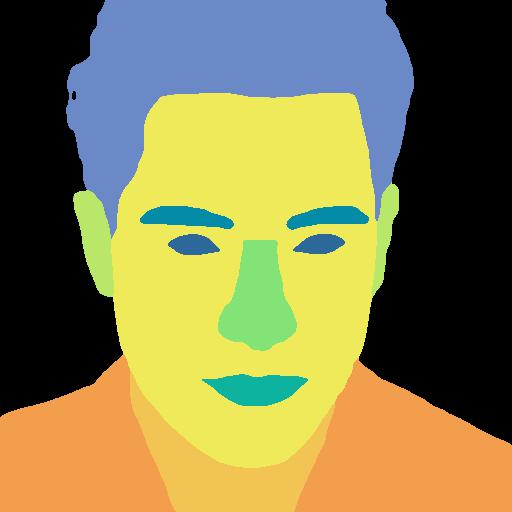} \\
\end{tabularx}
    \vspace{-1.0em}\caption{Example generated images and using our 512$\times$512 trained on CelebAMask-HQ. On the right of each generated photo is the refined segmentation mask output by the model.}\vspace{-1.0em}
    \label{appendix:fig:generation}
\end{figure*}

\begin{figure*}[t]
\captionsetup{font=small}
\centering
\scriptsize
\setlength\tabcolsep{1px}
\newcommand{\www}{0.1\linewidth}
\renewcommand{\arraystretch}{0.1}
\newcolumntype{Y}{>{\centering\arraybackslash}X}
\begin{tabularx}{\linewidth}{p{6pt} ccc ccc ccc c}
    \raisebox{2.2\height}{\rotatebox[origin=c]{90}{Image}} & 
    \includegraphics[width=\www]{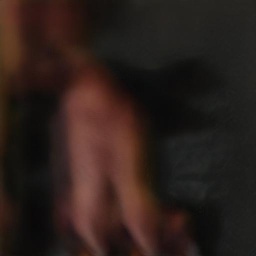}\hfill
    \includegraphics[width=\www]{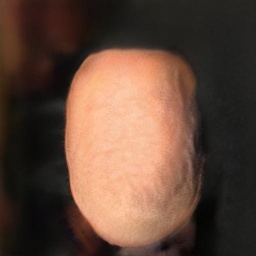}\hfill
    \includegraphics[width=\www]{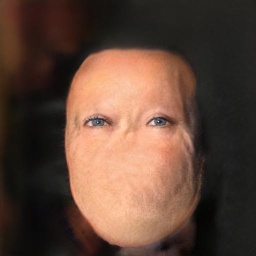}\hfill
    \includegraphics[width=\www]{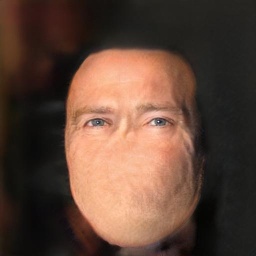}\hfill
    \includegraphics[width=\www]{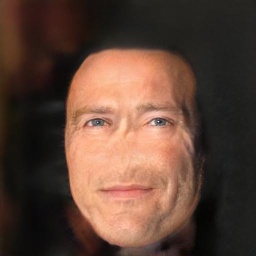}\hfill
    \includegraphics[width=\www]{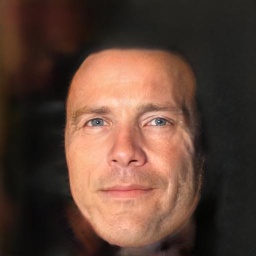}\hfill
    \includegraphics[width=\www]{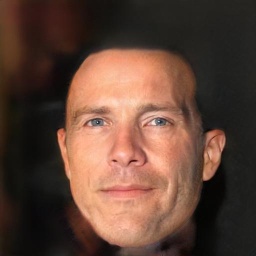}\hfill
    \includegraphics[width=\www]{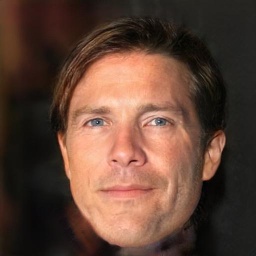}\hfill
    \includegraphics[width=\www]{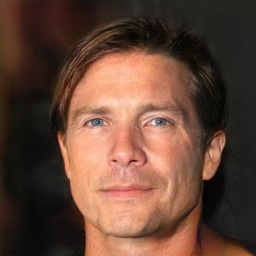}\hfill
    \includegraphics[width=\www]{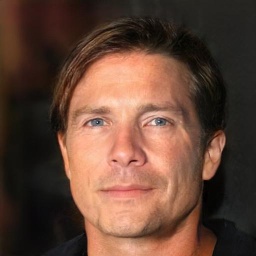}\\
    \raisebox{1.3\height}{\rotatebox[origin=c]{90}{Psuedo-depth}} & 
    \includegraphics[width=\www]{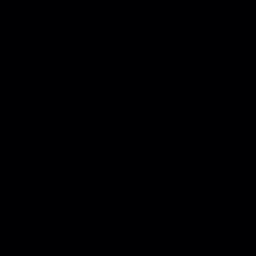}\hfill
    \includegraphics[width=\www]{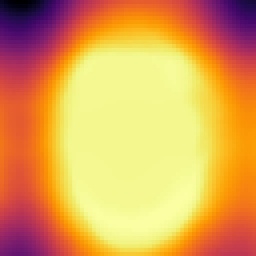}\hfill
    \includegraphics[width=\www]{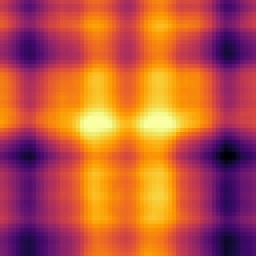}\hfill
    \includegraphics[width=\www]{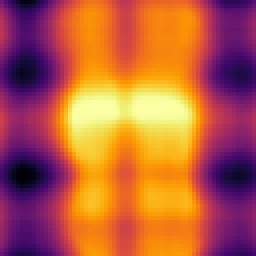}\hfill
    \includegraphics[width=\www]{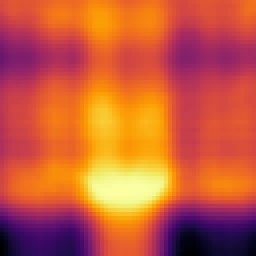}\hfill
    \includegraphics[width=\www]{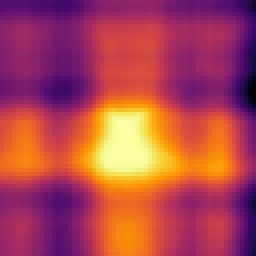}\hfill
    \includegraphics[width=\www]{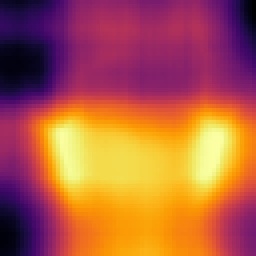}\hfill
    \includegraphics[width=\www]{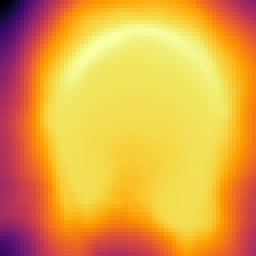}\hfill
    \includegraphics[width=\www]{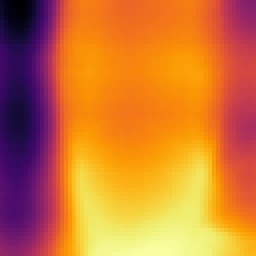}\hfill
    \includegraphics[width=\www]{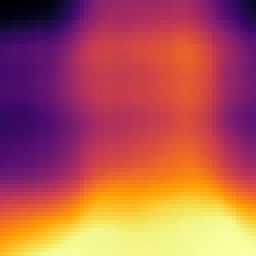}\\
    \raisebox{1.2\height}{\rotatebox[origin=c]{90}{Segmentation}} & 
    \includegraphics[width=\www]{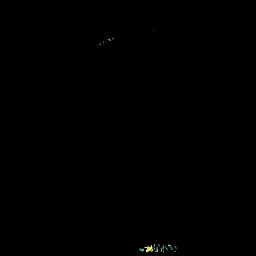}\hfill
    \includegraphics[width=\www]{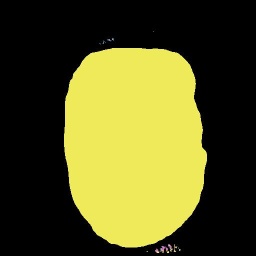}\hfill
    \includegraphics[width=\www]{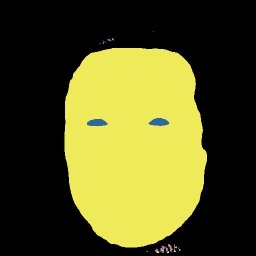}\hfill
    \includegraphics[width=\www]{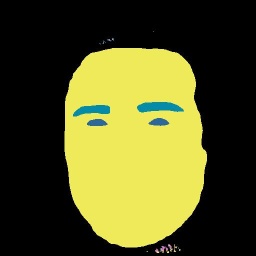}\hfill
    \includegraphics[width=\www]{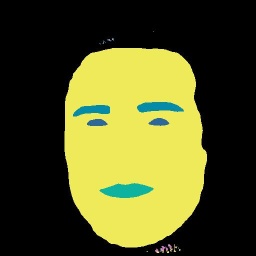}\hfill
    \includegraphics[width=\www]{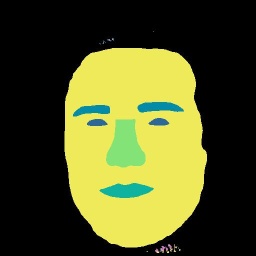}\hfill
    \includegraphics[width=\www]{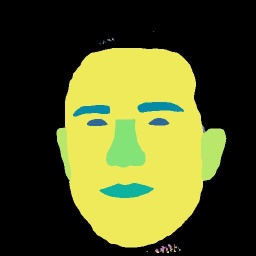}\hfill
    \includegraphics[width=\www]{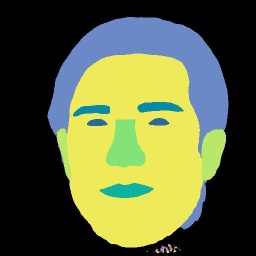}\hfill
    \includegraphics[width=\www]{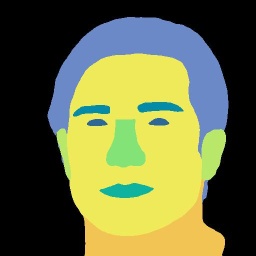}\hfill
    \includegraphics[width=\www]{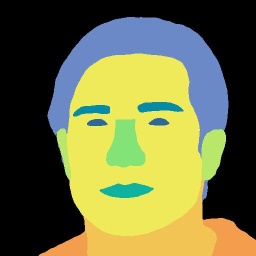}\\[0.5em]
    
    \raisebox{2.2\height}{\rotatebox[origin=c]{90}{Image}} & 
    \includegraphics[width=\www]{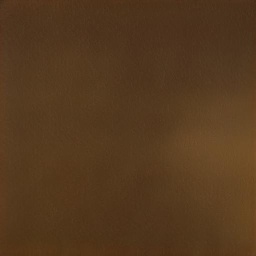}\hfill
    \includegraphics[width=\www]{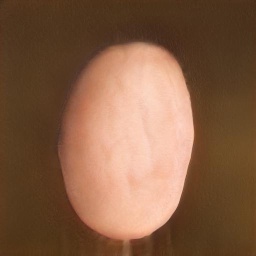}\hfill
    \includegraphics[width=\www]{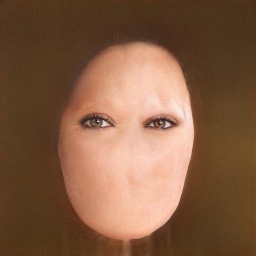}\hfill
    \includegraphics[width=\www]{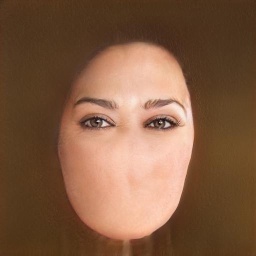}\hfill
    \includegraphics[width=\www]{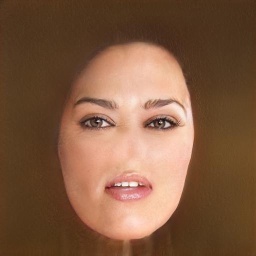}\hfill
    \includegraphics[width=\www]{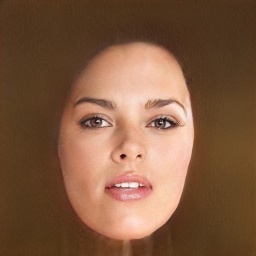}\hfill
    \includegraphics[width=\www]{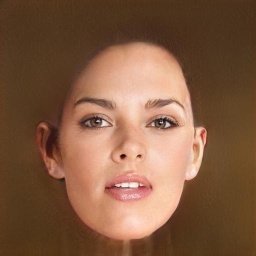}\hfill
    \includegraphics[width=\www]{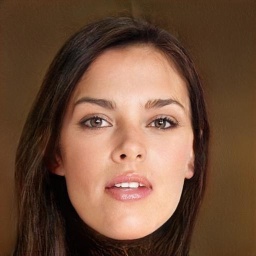}\hfill
    \includegraphics[width=\www]{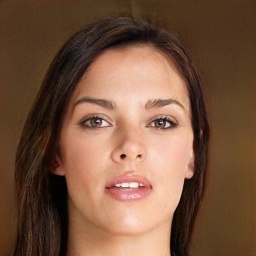}\hfill
    \includegraphics[width=\www]{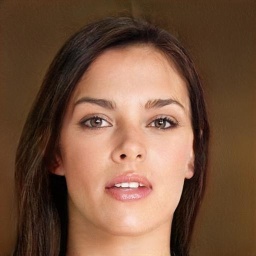}\\
    \raisebox{1.3\height}{\rotatebox[origin=c]{90}{Psuedo-depth}} & 
    \includegraphics[width=\www]{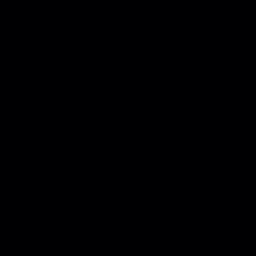}\hfill
    \includegraphics[width=\www]{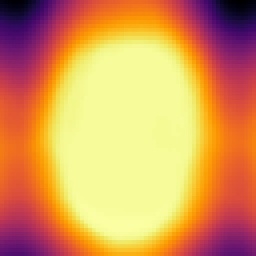}\hfill
    \includegraphics[width=\www]{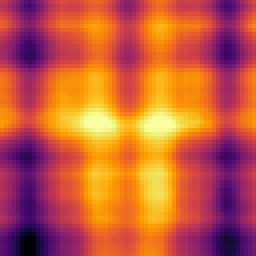}\hfill
    \includegraphics[width=\www]{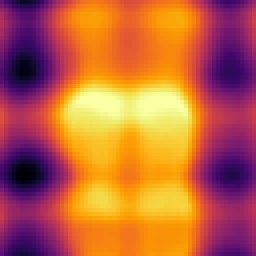}\hfill
    \includegraphics[width=\www]{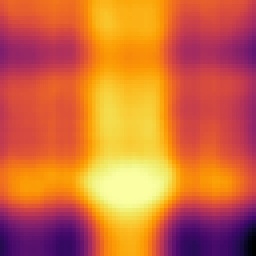}\hfill
    \includegraphics[width=\www]{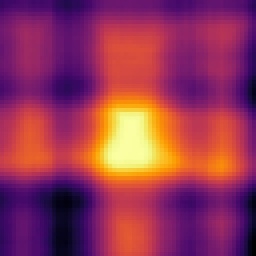}\hfill
    \includegraphics[width=\www]{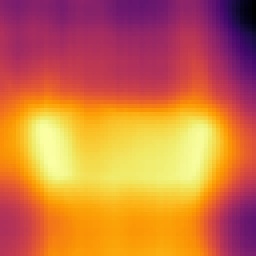}\hfill
    \includegraphics[width=\www]{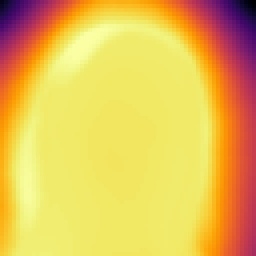}\hfill
    \includegraphics[width=\www]{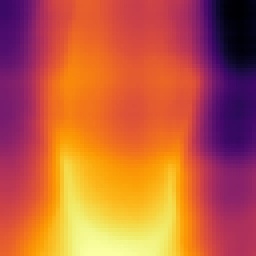}\hfill
    \includegraphics[width=\www]{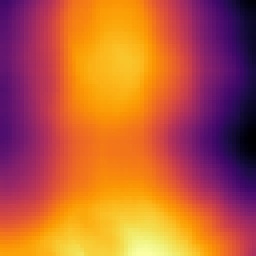}\\
    \raisebox{1.2\height}{\rotatebox[origin=c]{90}{Segmentation}} & 
    \includegraphics[width=\www]{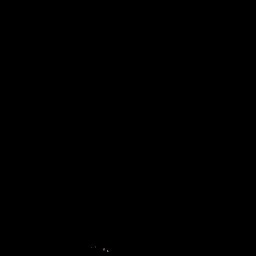}\hfill
    \includegraphics[width=\www]{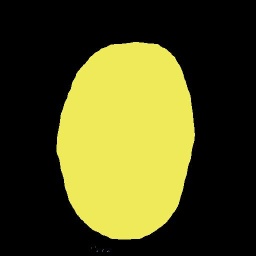}\hfill
    \includegraphics[width=\www]{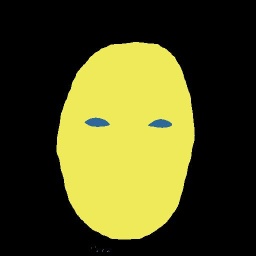}\hfill
    \includegraphics[width=\www]{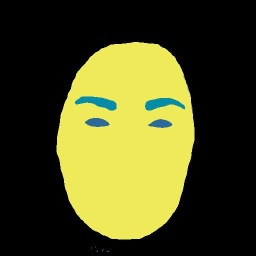}\hfill
    \includegraphics[width=\www]{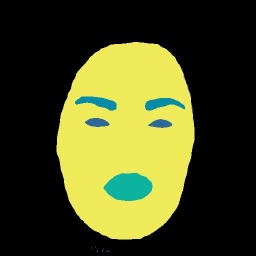}\hfill
    \includegraphics[width=\www]{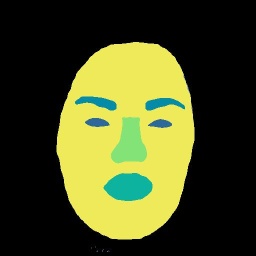}\hfill
    \includegraphics[width=\www]{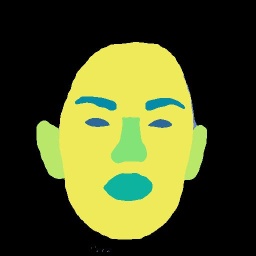}\hfill
    \includegraphics[width=\www]{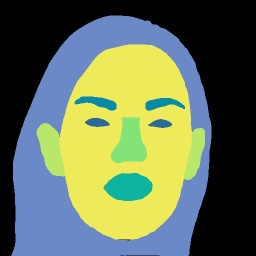}\hfill
    \includegraphics[width=\www]{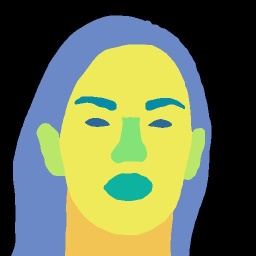}\hfill
    \includegraphics[width=\www]{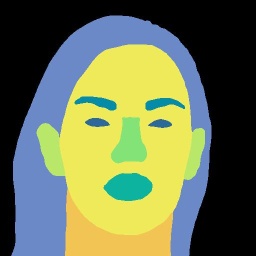}\\[0.5em]
    
    \raisebox{2.2\height}{\rotatebox[origin=c]{90}{Image}} & 
    \includegraphics[width=\www]{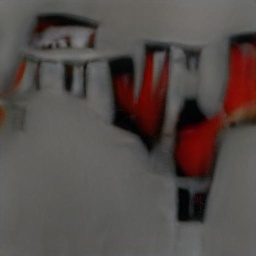}\hfill
    \includegraphics[width=\www]{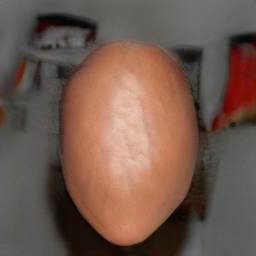}\hfill
    \includegraphics[width=\www]{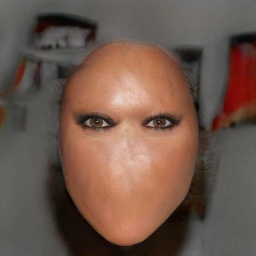}\hfill
    \includegraphics[width=\www]{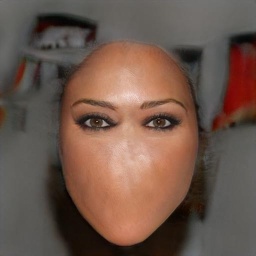}\hfill
    \includegraphics[width=\www]{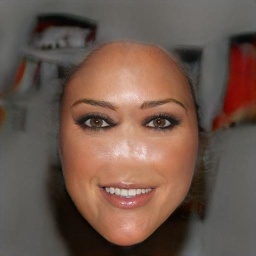}\hfill
    \includegraphics[width=\www]{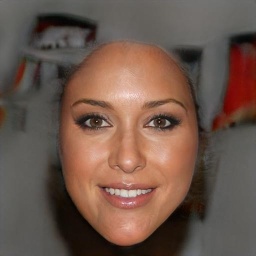}\hfill
    \includegraphics[width=\www]{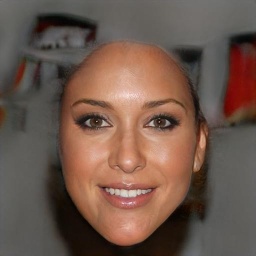}\hfill
    \includegraphics[width=\www]{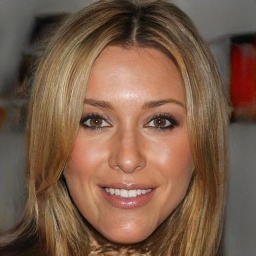}\hfill
    \includegraphics[width=\www]{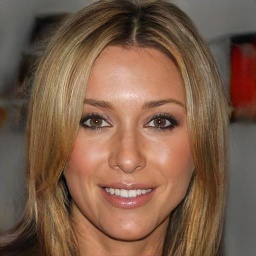}\hfill
    \includegraphics[width=\www]{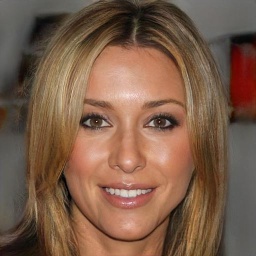}\\
    \raisebox{1.3\height}{\rotatebox[origin=c]{90}{Psuedo-depth}} & 
    \includegraphics[width=\www]{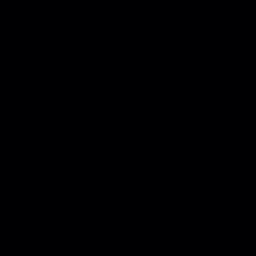}\hfill
    \includegraphics[width=\www]{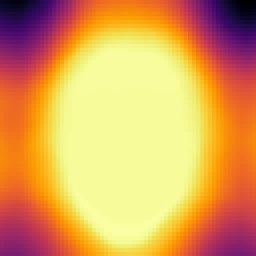}\hfill
    \includegraphics[width=\www]{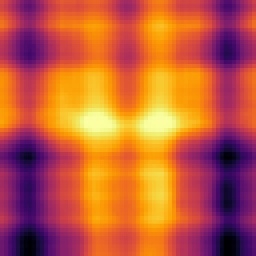}\hfill
    \includegraphics[width=\www]{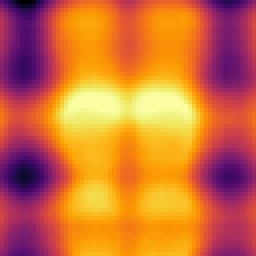}\hfill
    \includegraphics[width=\www]{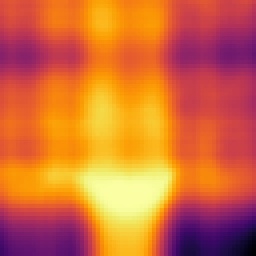}\hfill
    \includegraphics[width=\www]{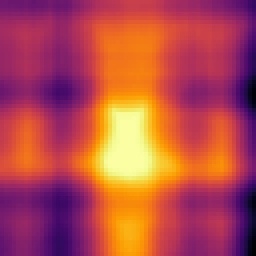}\hfill
    \includegraphics[width=\www]{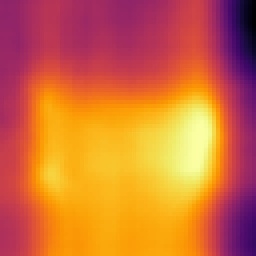}\hfill
    \includegraphics[width=\www]{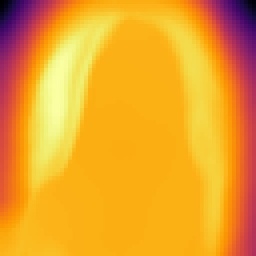}\hfill
    \includegraphics[width=\www]{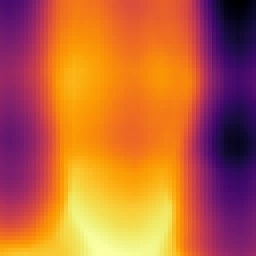}\hfill
    \includegraphics[width=\www]{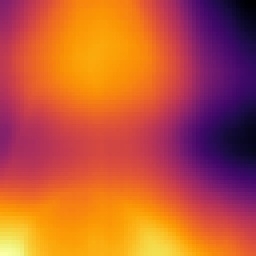}\\
    \raisebox{1.2\height}{\rotatebox[origin=c]{90}{Segmentation}} & 
    \includegraphics[width=\www]{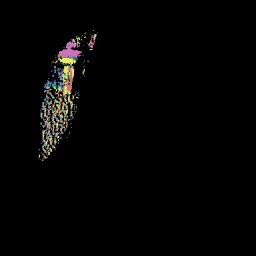}\hfill
    \includegraphics[width=\www]{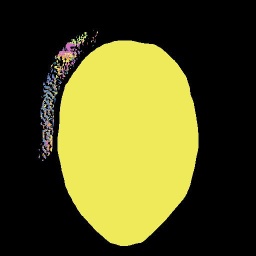}\hfill
    \includegraphics[width=\www]{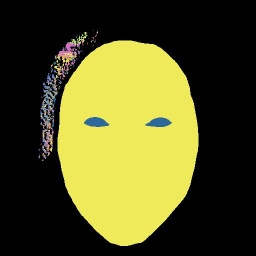}\hfill
    \includegraphics[width=\www]{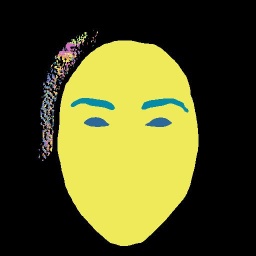}\hfill
    \includegraphics[width=\www]{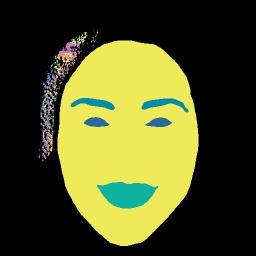}\hfill
    \includegraphics[width=\www]{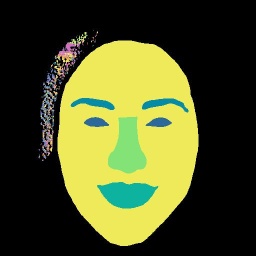}\hfill
    \includegraphics[width=\www]{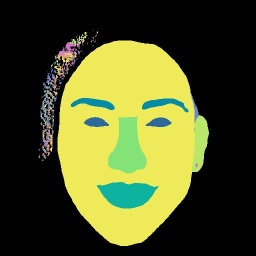}\hfill
    \includegraphics[width=\www]{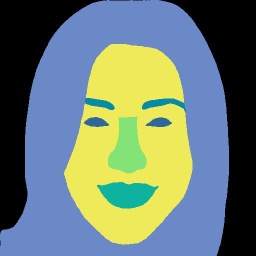}\hfill
    \includegraphics[width=\www]{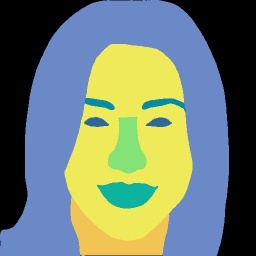}\hfill
    \includegraphics[width=\www]{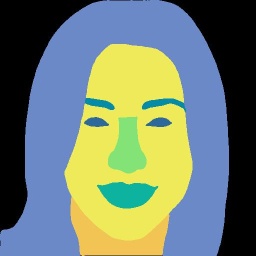}\\[0.5em]
    
    \raisebox{2.2\height}{\rotatebox[origin=c]{90}{Image}} & 
    \includegraphics[width=\www]{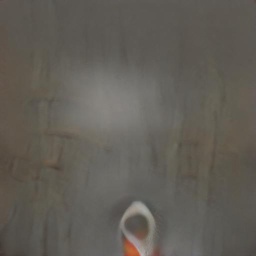}\hfill
    \includegraphics[width=\www]{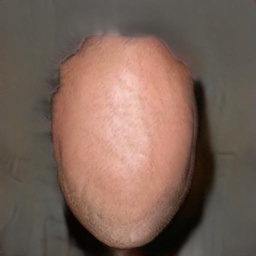}\hfill
    \includegraphics[width=\www]{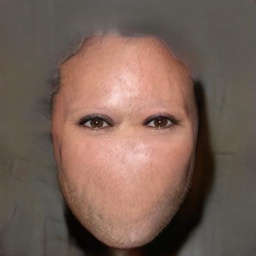}\hfill
    \includegraphics[width=\www]{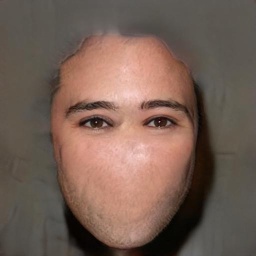}\hfill
    \includegraphics[width=\www]{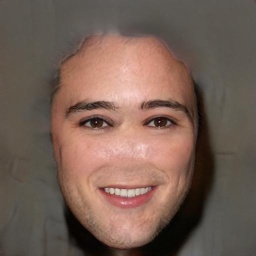}\hfill
    \includegraphics[width=\www]{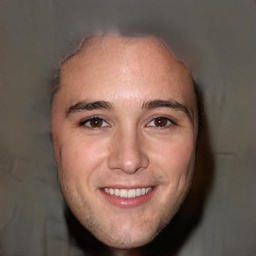}\hfill
    \includegraphics[width=\www]{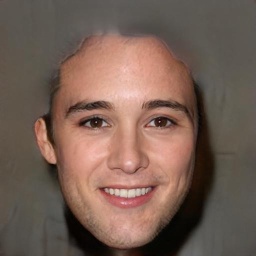}\hfill
    \includegraphics[width=\www]{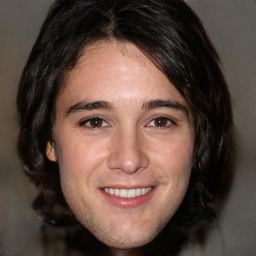}\hfill
    \includegraphics[width=\www]{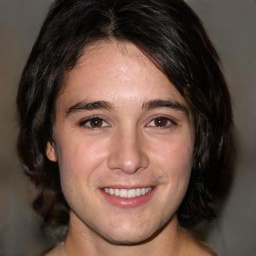}\hfill
    \includegraphics[width=\www]{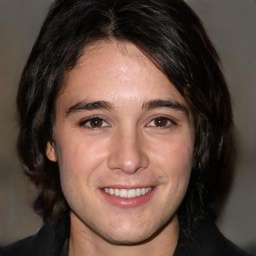}\\
    \raisebox{1.3\height}{\rotatebox[origin=c]{90}{Psuedo-depth}} & 
    \includegraphics[width=\www]{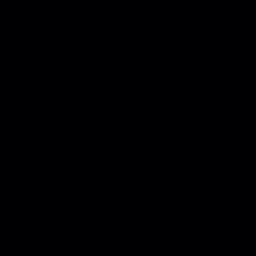}\hfill
    \includegraphics[width=\www]{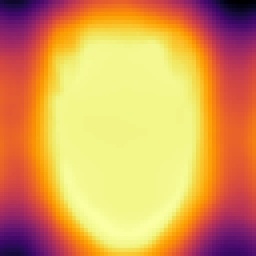}\hfill
    \includegraphics[width=\www]{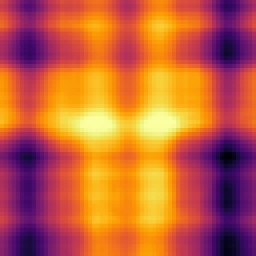}\hfill
    \includegraphics[width=\www]{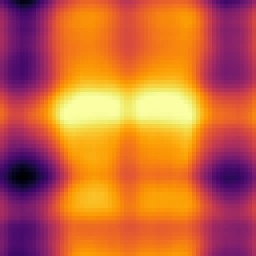}\hfill
    \includegraphics[width=\www]{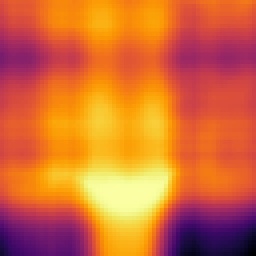}\hfill
    \includegraphics[width=\www]{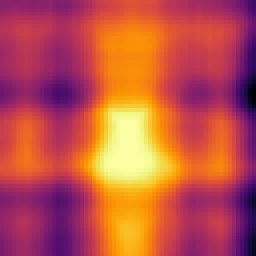}\hfill
    \includegraphics[width=\www]{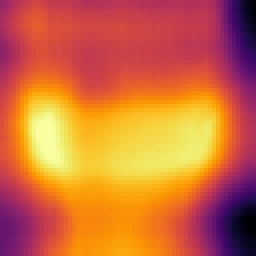}\hfill
    \includegraphics[width=\www]{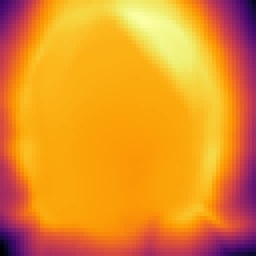}\hfill
    \includegraphics[width=\www]{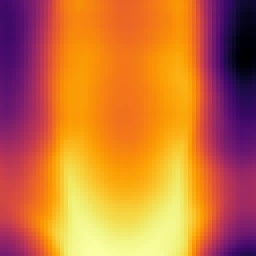}\hfill
    \includegraphics[width=\www]{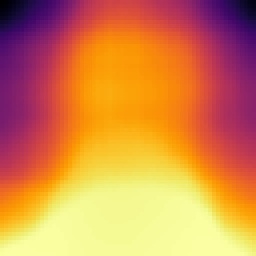}\\
    \raisebox{1.2\height}{\rotatebox[origin=c]{90}{Segmentation}} & 
    \includegraphics[width=\www]{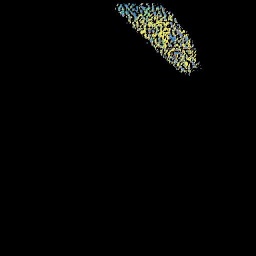}\hfill
    \includegraphics[width=\www]{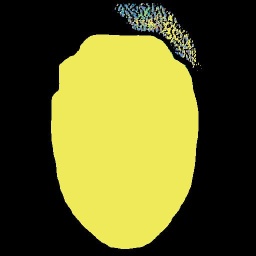}\hfill
    \includegraphics[width=\www]{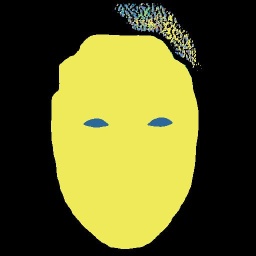}\hfill
    \includegraphics[width=\www]{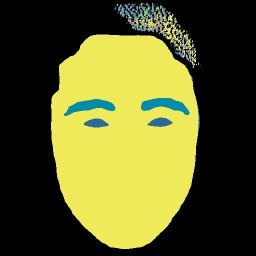}\hfill
    \includegraphics[width=\www]{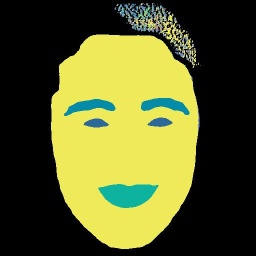}\hfill
    \includegraphics[width=\www]{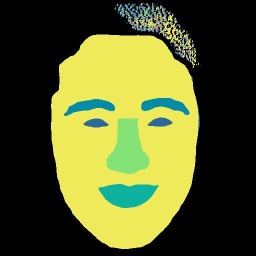}\hfill
    \includegraphics[width=\www]{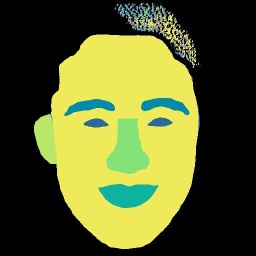}\hfill
    \includegraphics[width=\www]{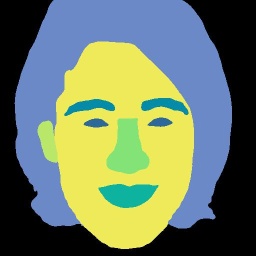}\hfill
    \includegraphics[width=\www]{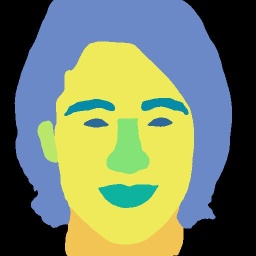}\hfill
    \includegraphics[width=\www]{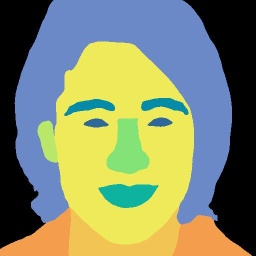}\\[0.5em]
\end{tabularx}
    \vspace{-1.0em}\caption{Illustration of compositional synthesis. Starting from background, we gradually add more components into the feature map. The second row of each sample shows the pseudo-depth map of each corresponding component used for fusion. During synthesis, all pseudo-depth maps are fused without an order.}\vspace{-1.0em}
    \label{appendix:fig:components}
\end{figure*}

\begin{figure*}[t]
\captionsetup{font=small}
\centering
\footnotesize
\setlength\tabcolsep{0.0pt}
\newcommand{\www}{0.089\linewidth}
\renewcommand{\arraystretch}{0.4}
\newcolumntype{Y}{>{\centering\arraybackslash}X}
\begin{tabularx}{\linewidth}{c @{\hskip 2pt}|@{\hskip 2pt} c @{\hskip 1.5pt} cc @{\hskip 1.5pt} cc @{\hskip 2pt}|@{\hskip 2pt} c @{\hskip 1.5pt} cc @{\hskip 1.5pt} cc}
    \includegraphics[width=\www]{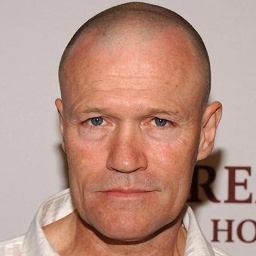} &
    \includegraphics[width=\www]{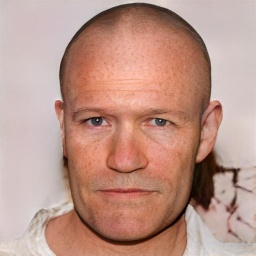} &
    \includegraphics[width=\www]{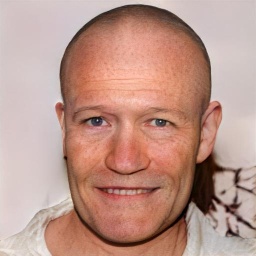} &
    \includegraphics[width=\www]{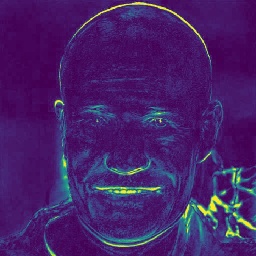} &
    \includegraphics[width=\www]{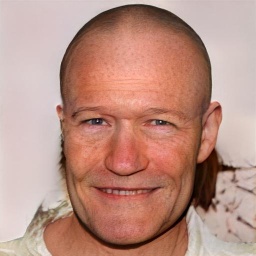} &
    \includegraphics[width=\www]{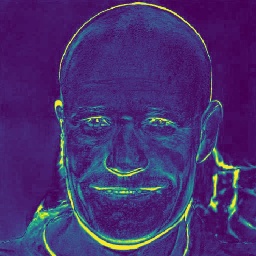} &
    \includegraphics[width=\www]{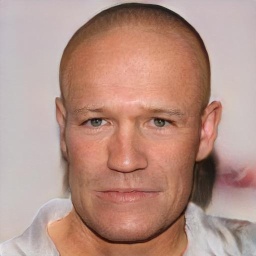} &
    \includegraphics[width=\www]{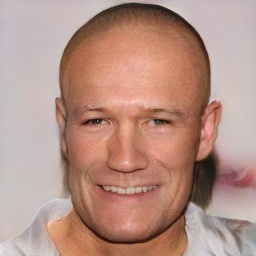} &
    \includegraphics[width=\www]{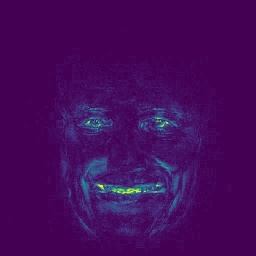} &
    \includegraphics[width=\www]{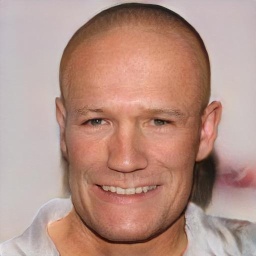} &
    \includegraphics[width=\www]{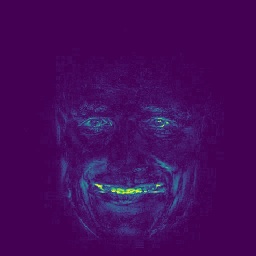} \\
    \includegraphics[width=\www]{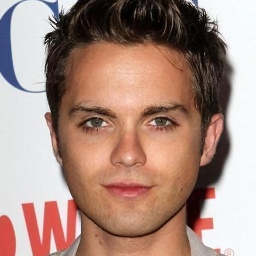} &
    \includegraphics[width=\www]{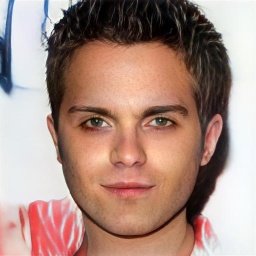} &
    \includegraphics[width=\www]{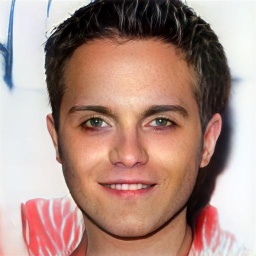} &
    \includegraphics[width=\www]{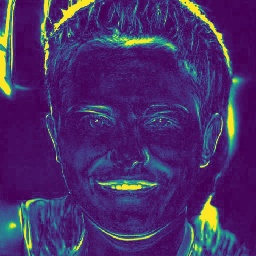} &
    \includegraphics[width=\www]{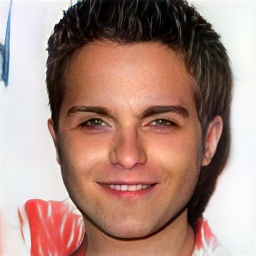} &
    \includegraphics[width=\www]{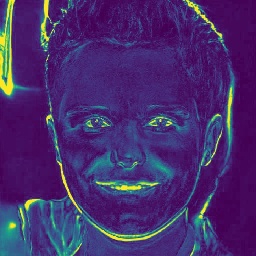} &
    \includegraphics[width=\www]{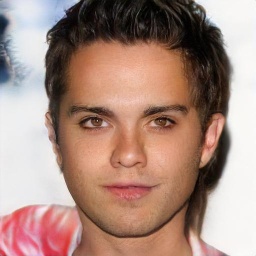} &
    \includegraphics[width=\www]{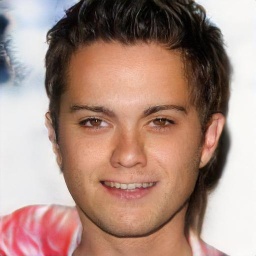} &
    \includegraphics[width=\www]{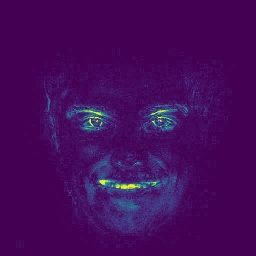} &
    \includegraphics[width=\www]{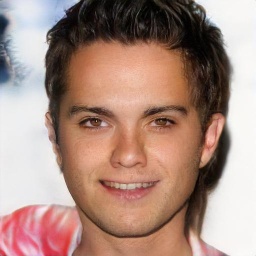} &
    \includegraphics[width=\www]{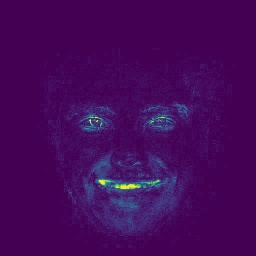} \\
    \includegraphics[width=\www]{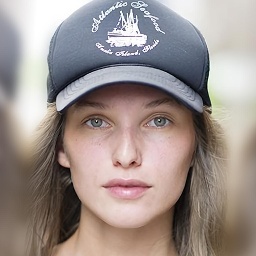} &
    \includegraphics[width=\www]{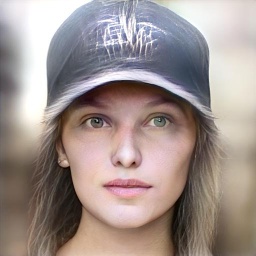} &
    \includegraphics[width=\www]{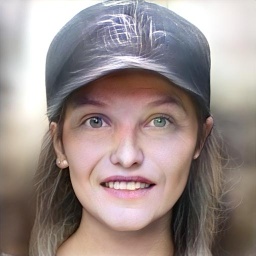} &
    \includegraphics[width=\www]{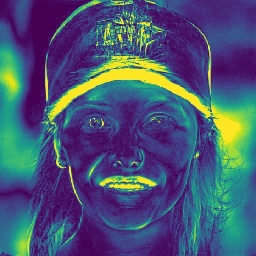} &
    \includegraphics[width=\www]{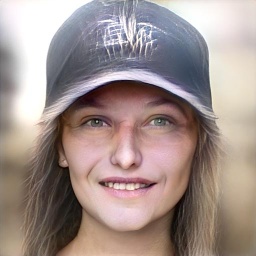} &
    \includegraphics[width=\www]{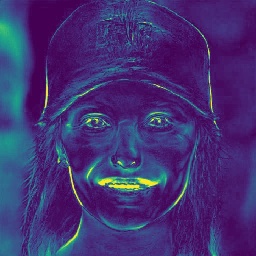} &
    \includegraphics[width=\www]{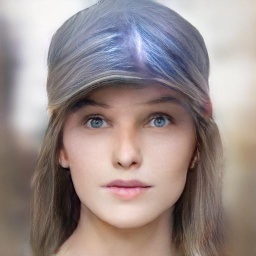} &
    \includegraphics[width=\www]{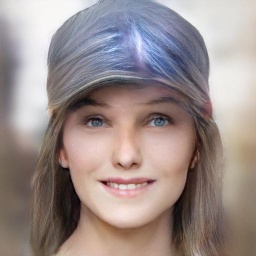} &
    \includegraphics[width=\www]{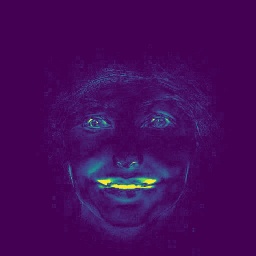} &
    \includegraphics[width=\www]{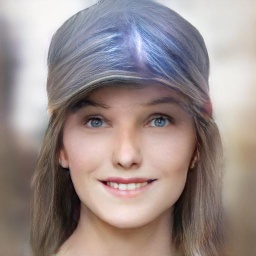} &
    \includegraphics[width=\www]{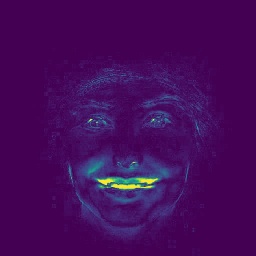} \\
    \includegraphics[width=\www]{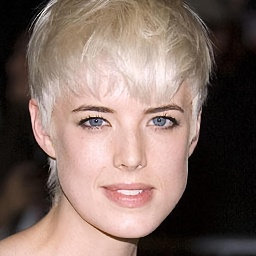} &
    \includegraphics[width=\www]{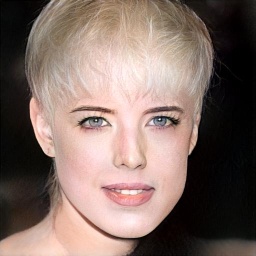} &
    \includegraphics[width=\www]{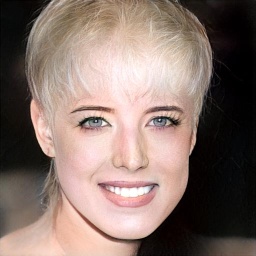} &
    \includegraphics[width=\www]{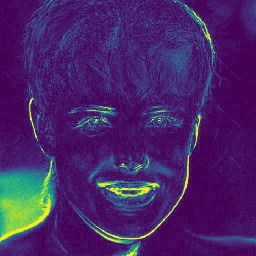} &
    \includegraphics[width=\www]{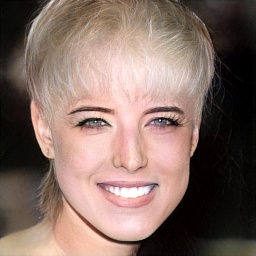} &
    \includegraphics[width=\www]{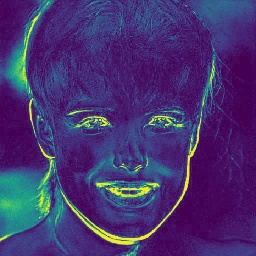} &
    \includegraphics[width=\www]{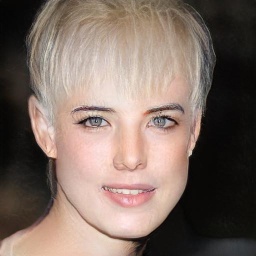} &
    \includegraphics[width=\www]{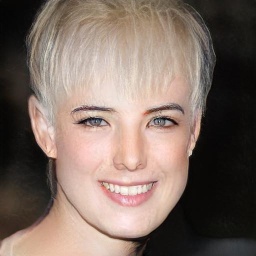} &
    \includegraphics[width=\www]{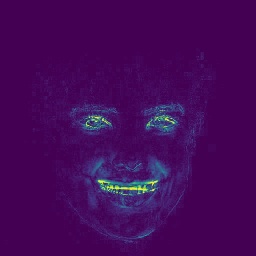} &
    \includegraphics[width=\www]{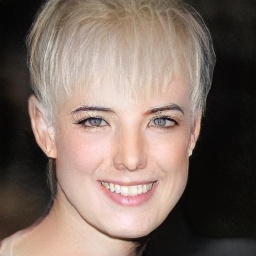} &
    \includegraphics[width=\www]{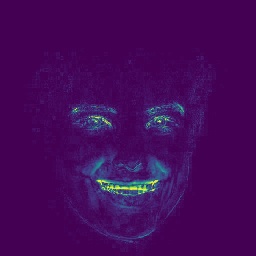} \\
    \includegraphics[width=\www]{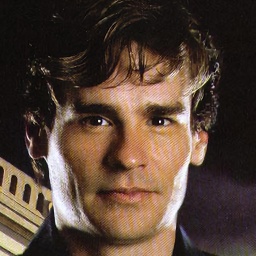} &
    \includegraphics[width=\www]{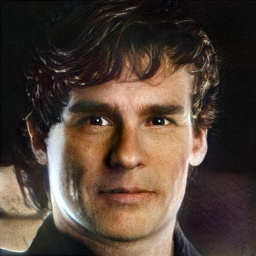} &
    \includegraphics[width=\www]{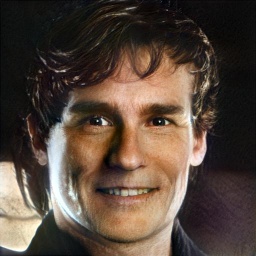} &
    \includegraphics[width=\www]{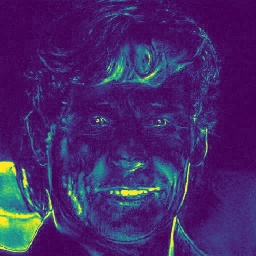} &
    \includegraphics[width=\www]{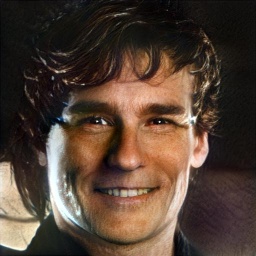} &
    \includegraphics[width=\www]{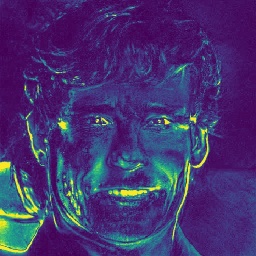} &
    \includegraphics[width=\www]{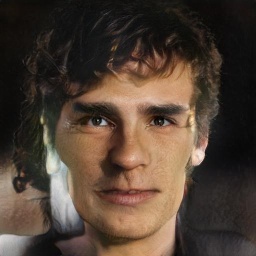} &
    \includegraphics[width=\www]{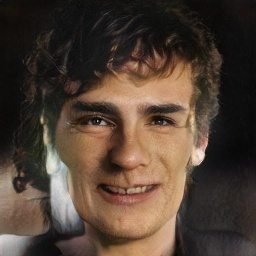} &
    \includegraphics[width=\www]{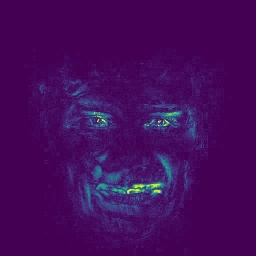} &
    \includegraphics[width=\www]{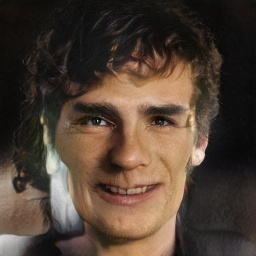} &
    \includegraphics[width=\www]{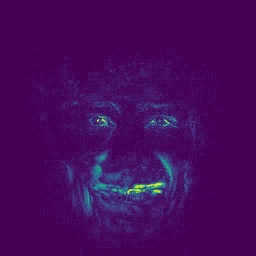} \\
    \includegraphics[width=\www]{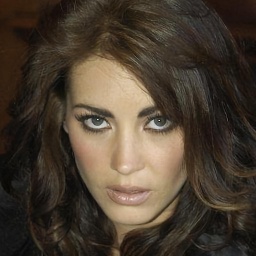} &
    \includegraphics[width=\www]{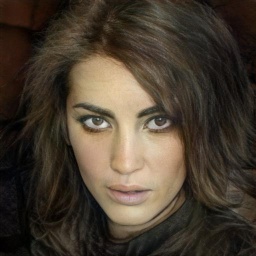} &
    \includegraphics[width=\www]{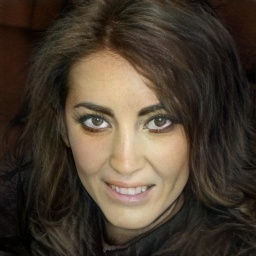} &
    \includegraphics[width=\www]{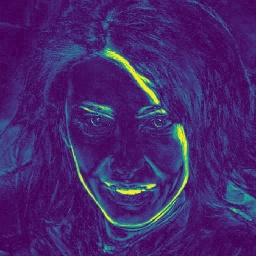} &
    \includegraphics[width=\www]{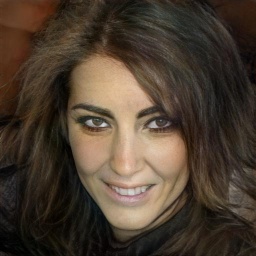} &
    \includegraphics[width=\www]{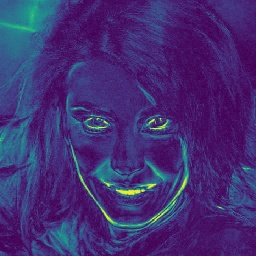} &
    \includegraphics[width=\www]{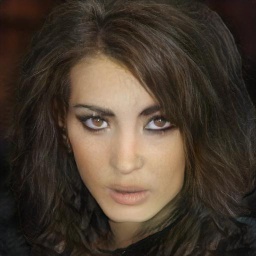} &
    \includegraphics[width=\www]{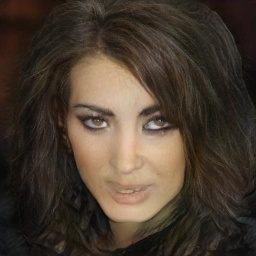} &
    \includegraphics[width=\www]{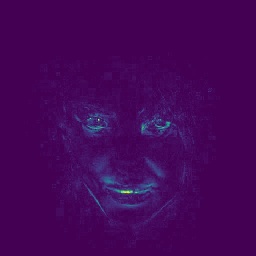} &
    \includegraphics[width=\www]{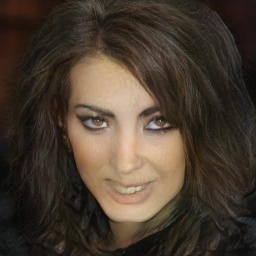} &
    \includegraphics[width=\www]{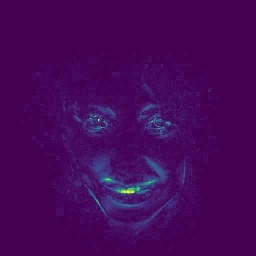} \\
    \multicolumn{1}{c}{Input} & \makecell{Inversion\\(StyleGAN2)} & \multicolumn{2}{c}{StyleFlow} & \multicolumn{2}{c}{InterFaceGAN}  & \makecell{Inversion\\(Ours)} & \multicolumn{2}{c}{StyleFlow+Ours} & \multicolumn{2}{c}{InterFaceGAN+Ours} \\
\end{tabularx}
    \vspace{-1.0em}\caption{Results of GAN inversion and editing for the \textbf{smile} attribute. For each  method, we show the inversion result of Restyle encoder, the edited image and the difference map between them.}\vspace{-0.4em}
    \label{appendix:fig:editing_smile}\vspace{-1.0em}
\end{figure*}

\begin{figure*}[t]
\captionsetup{font=small}
\centering
\footnotesize
\setlength\tabcolsep{0.0pt}
\newcommand{\www}{0.089\linewidth}
\renewcommand{\arraystretch}{0.4}
\newcolumntype{Y}{>{\centering\arraybackslash}X}
\begin{tabularx}{\linewidth}{c @{\hskip 2pt}|@{\hskip 2pt} c @{\hskip 1.5pt} cc @{\hskip 1.5pt} cc @{\hskip 2pt}|@{\hskip 2pt} c @{\hskip 1.5pt} cc @{\hskip 1.5pt} cc}
    \includegraphics[width=\www]{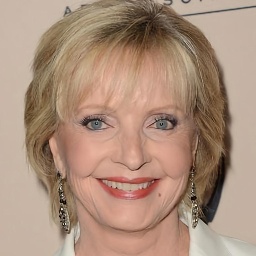} &
    \includegraphics[width=\www]{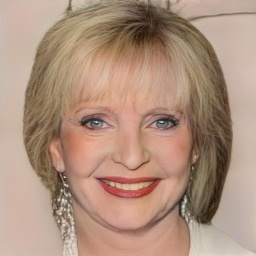} &
    \includegraphics[width=\www]{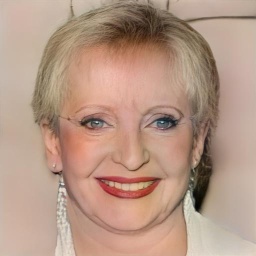} &
    \includegraphics[width=\www]{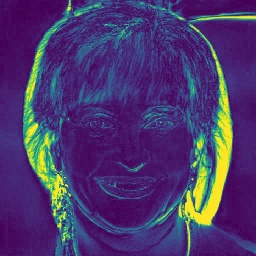} &
    \includegraphics[width=\www]{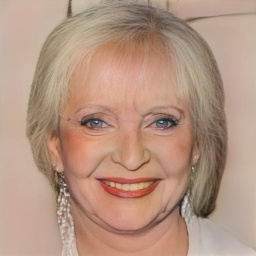} &
    \includegraphics[width=\www]{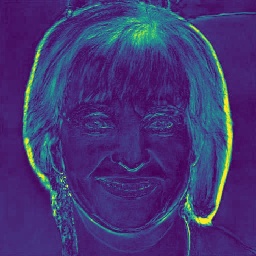} &
    \includegraphics[width=\www]{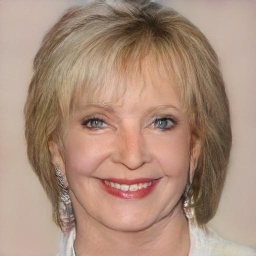} &
    \includegraphics[width=\www]{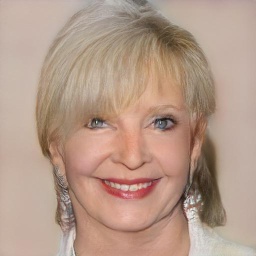} &
    \includegraphics[width=\www]{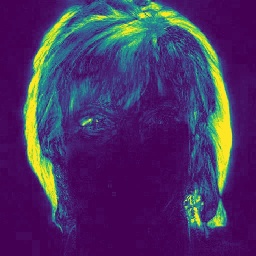} &
    \includegraphics[width=\www]{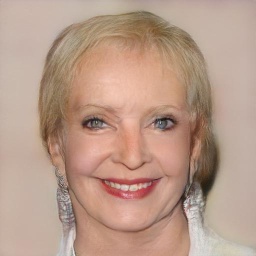} &
    \includegraphics[width=\www]{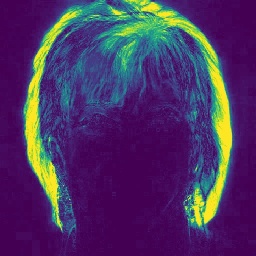} \\
    \includegraphics[width=\www]{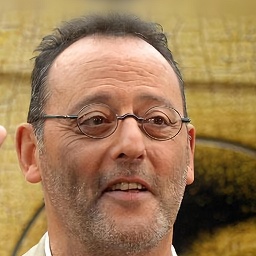} &
    \includegraphics[width=\www]{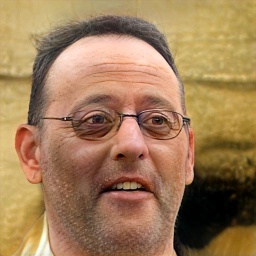} &
    \includegraphics[width=\www]{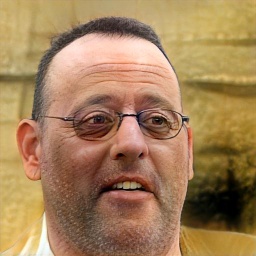} &
    \includegraphics[width=\www]{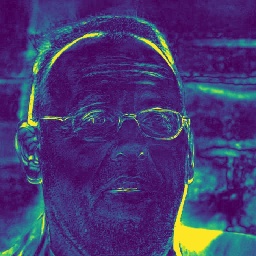} &
    \includegraphics[width=\www]{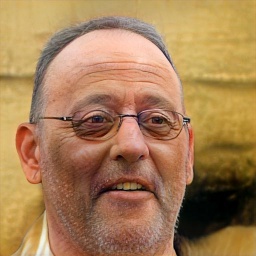} &
    \includegraphics[width=\www]{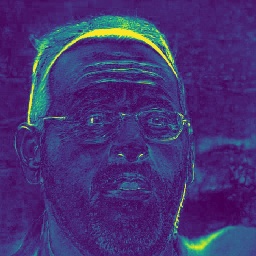} &
    \includegraphics[width=\www]{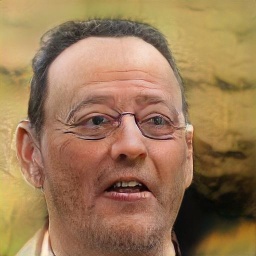} &
    \includegraphics[width=\www]{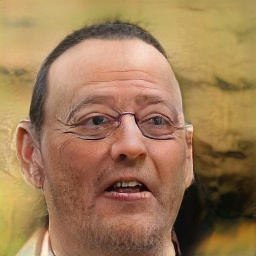} &
    \includegraphics[width=\www]{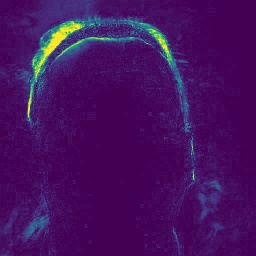} &
    \includegraphics[width=\www]{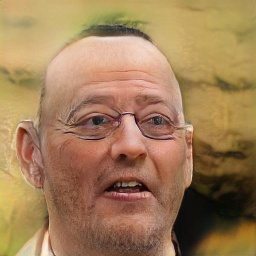} &
    \includegraphics[width=\www]{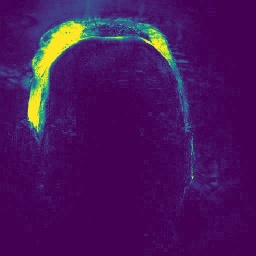} \\
    \includegraphics[width=\www]{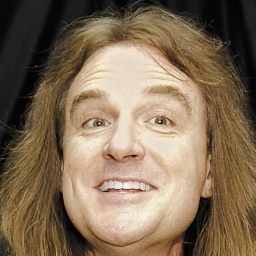} &
    \includegraphics[width=\www]{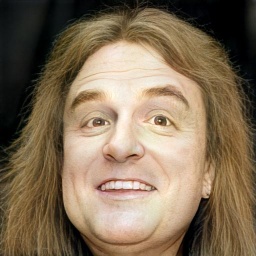} &
    \includegraphics[width=\www]{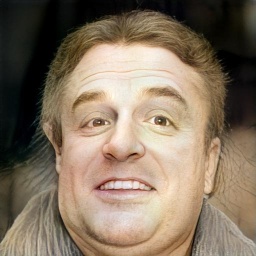} &
    \includegraphics[width=\www]{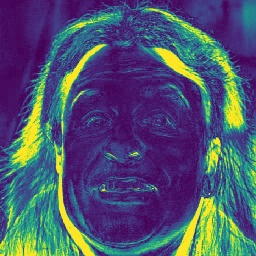} &
    \includegraphics[width=\www]{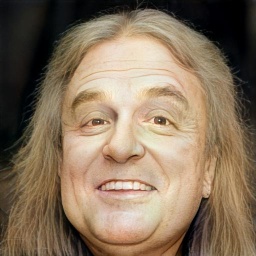} &
    \includegraphics[width=\www]{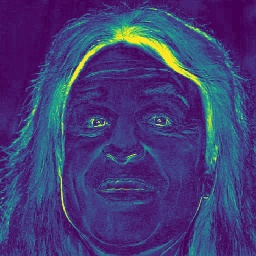} &
    \includegraphics[width=\www]{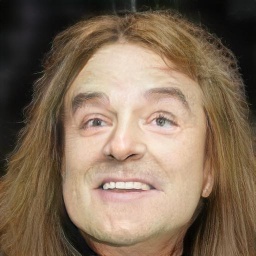} &
    \includegraphics[width=\www]{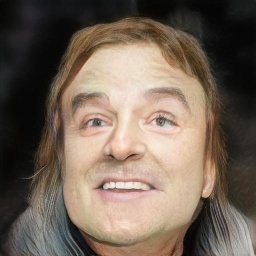} &
    \includegraphics[width=\www]{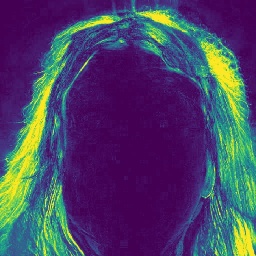} &
    \includegraphics[width=\www]{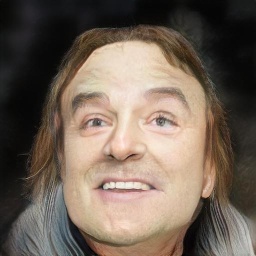} &
    \includegraphics[width=\www]{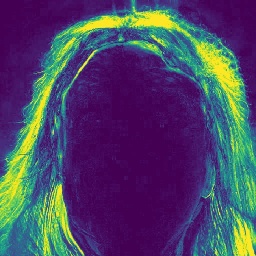} \\
    \includegraphics[width=\www]{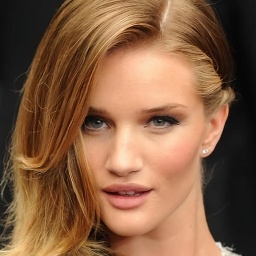} &
    \includegraphics[width=\www]{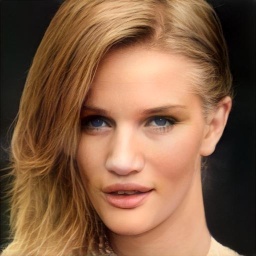} &
    \includegraphics[width=\www]{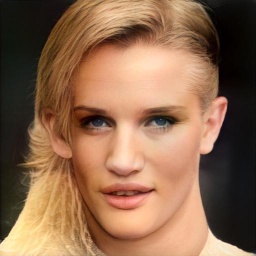} &
    \includegraphics[width=\www]{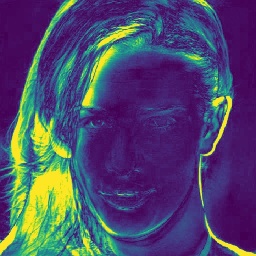} &
    \includegraphics[width=\www]{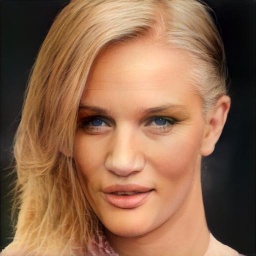} &
    \includegraphics[width=\www]{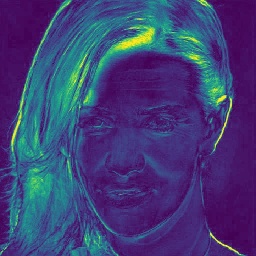} &
    \includegraphics[width=\www]{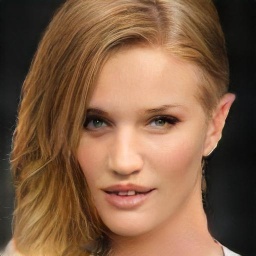} &
    \includegraphics[width=\www]{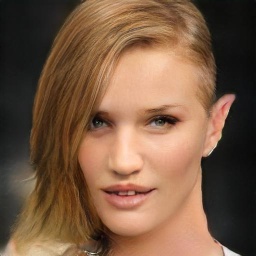} &
    \includegraphics[width=\www]{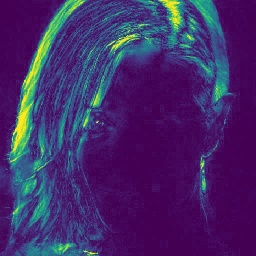} &
    \includegraphics[width=\www]{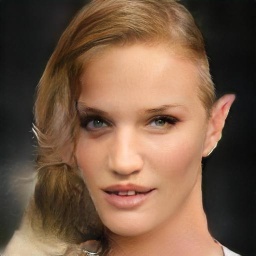} &
    \includegraphics[width=\www]{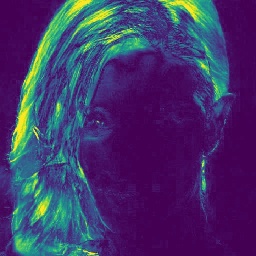} \\
    \includegraphics[width=\www]{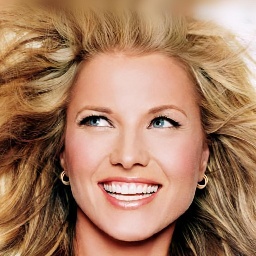} &
    \includegraphics[width=\www]{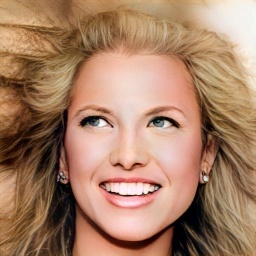} &
    \includegraphics[width=\www]{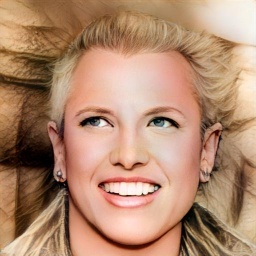} &
    \includegraphics[width=\www]{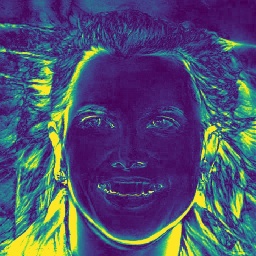} &
    \includegraphics[width=\www]{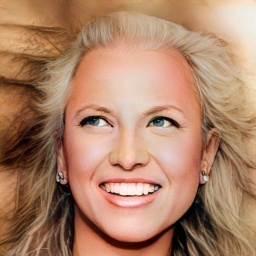} &
    \includegraphics[width=\www]{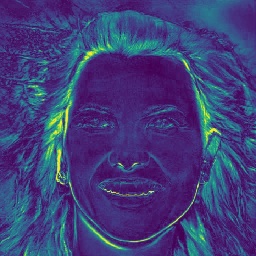} &
    \includegraphics[width=\www]{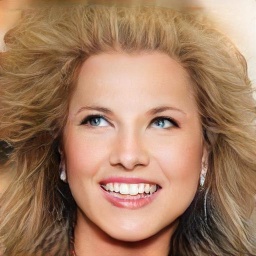} &
    \includegraphics[width=\www]{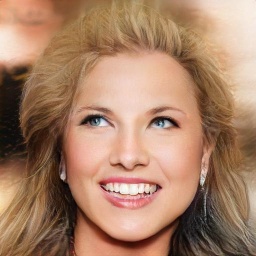} &
    \includegraphics[width=\www]{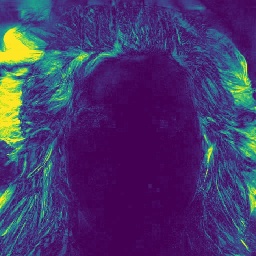} &
    \includegraphics[width=\www]{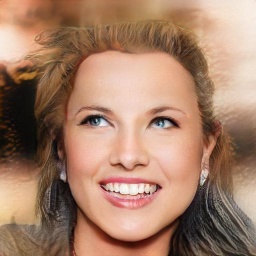} &
    \includegraphics[width=\www]{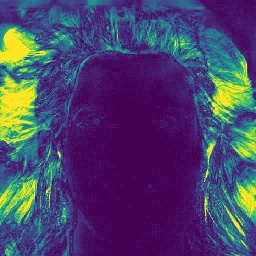} \\
    \includegraphics[width=\www]{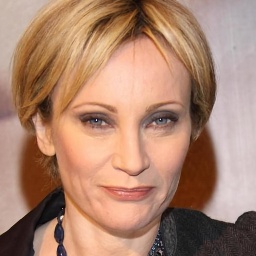} &
    \includegraphics[width=\www]{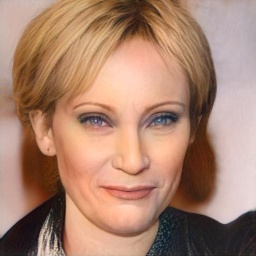} &
    \includegraphics[width=\www]{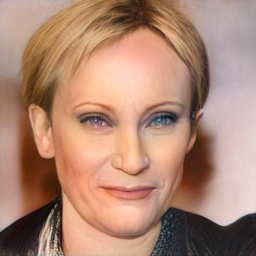} &
    \includegraphics[width=\www]{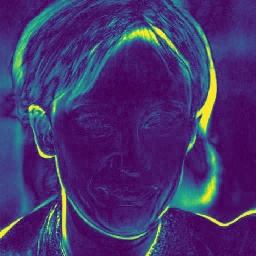} &
    \includegraphics[width=\www]{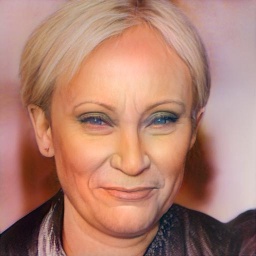} &
    \includegraphics[width=\www]{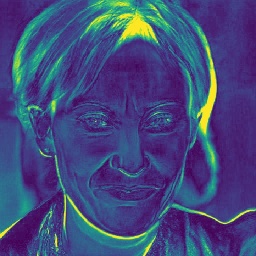} &
    \includegraphics[width=\www]{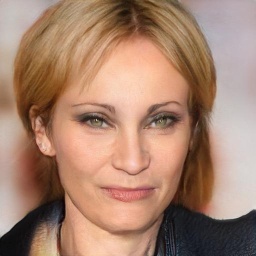} &
    \includegraphics[width=\www]{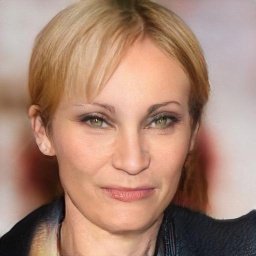} &
    \includegraphics[width=\www]{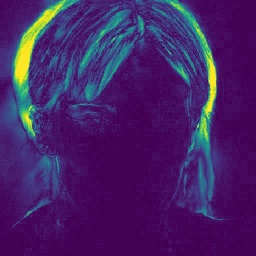} &
    \includegraphics[width=\www]{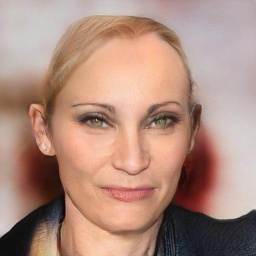} &
    \includegraphics[width=\www]{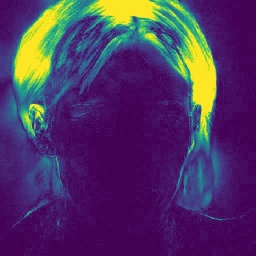} \\
    \multicolumn{1}{c}{Input} & \makecell{Inversion\\(StyleGAN2)} & \multicolumn{2}{c}{StyleFlow} & \multicolumn{2}{c}{InterFaceGAN}  & \makecell{Inversion\\(Ours)} & \multicolumn{2}{c}{StyleFlow+Ours} & \multicolumn{2}{c}{InterFaceGAN+Ours} \\
    
\end{tabularx}
    \vspace{-1.0em}\caption{Results of GAN inversion and editing for the \textbf{bald} attribute. For each  method, we show the inversion result of Restyle encoder, the edited image and the difference map between them.}\vspace{-0.4em}
    \label{appendix:fig:editing_bald}\vspace{-1.0em}
\end{figure*}

\begin{figure*}[t]
\captionsetup{font=small}
\centering
\footnotesize
\setlength\tabcolsep{0.0pt}
\newcommand{\www}{0.089\linewidth}
\renewcommand{\arraystretch}{0.4}
\newcolumntype{Y}{>{\centering\arraybackslash}X}
\begin{tabularx}{\linewidth}{c @{\hskip 2pt}|@{\hskip 2pt} c @{\hskip 1.5pt} cc @{\hskip 1.5pt} cc @{\hskip 2pt}|@{\hskip 2pt} c @{\hskip 1.5pt} cc @{\hskip 1.5pt} cc}
    \includegraphics[width=\www]{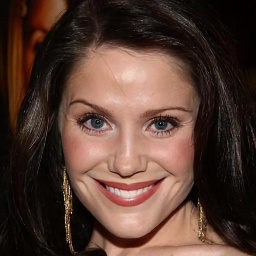} &
    \includegraphics[width=\www]{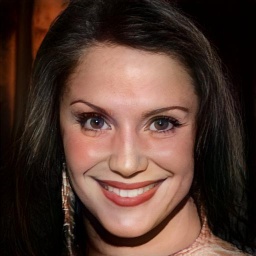} &
    \includegraphics[width=\www]{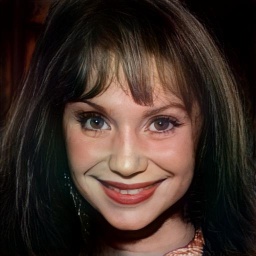} &
    \includegraphics[width=\www]{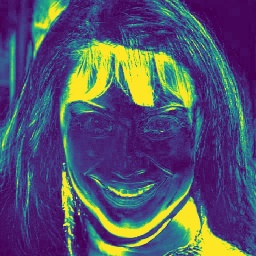} &
    \includegraphics[width=\www]{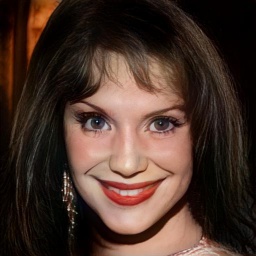} &
    \includegraphics[width=\www]{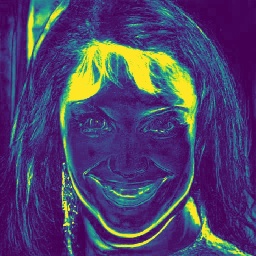} &
    \includegraphics[width=\www]{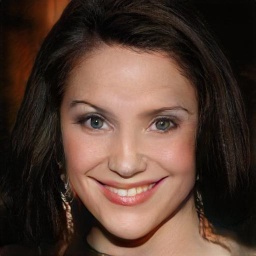} &
    \includegraphics[width=\www]{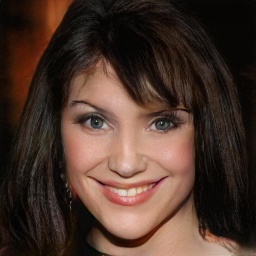} &
    \includegraphics[width=\www]{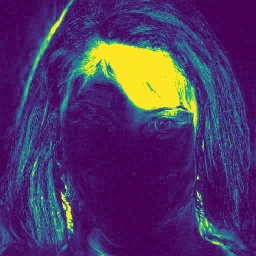} &
    \includegraphics[width=\www]{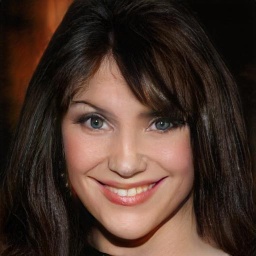} &
    \includegraphics[width=\www]{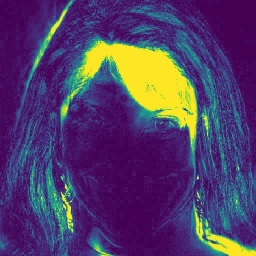} \\
    \includegraphics[width=\www]{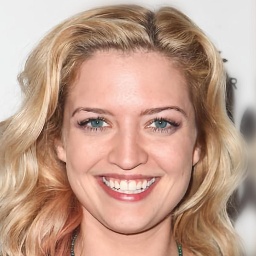} &
    \includegraphics[width=\www]{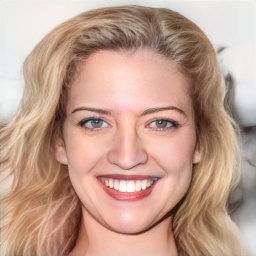} &
    \includegraphics[width=\www]{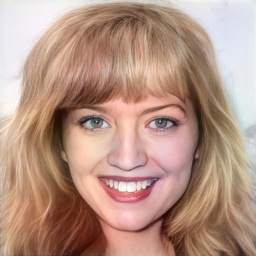} &
    \includegraphics[width=\www]{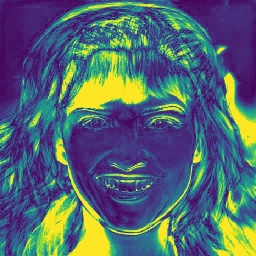} &
    \includegraphics[width=\www]{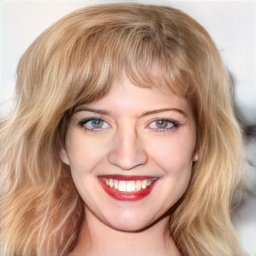} &
    \includegraphics[width=\www]{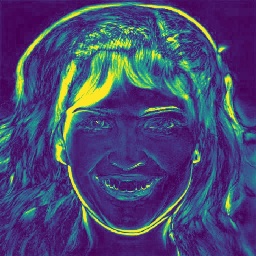} &
    \includegraphics[width=\www]{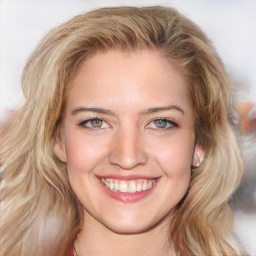} &
    \includegraphics[width=\www]{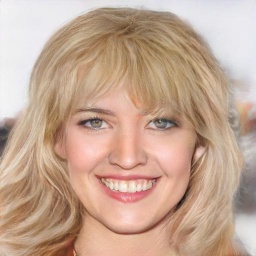} &
    \includegraphics[width=\www]{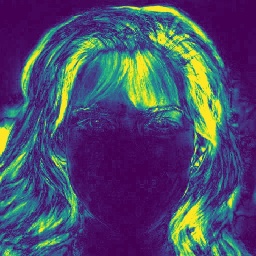} &
    \includegraphics[width=\www]{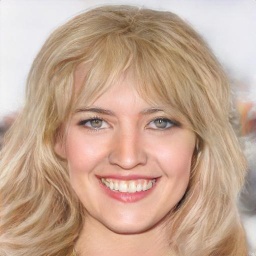} &
    \includegraphics[width=\www]{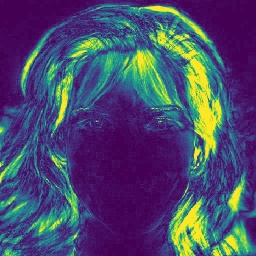} \\
    \includegraphics[width=\www]{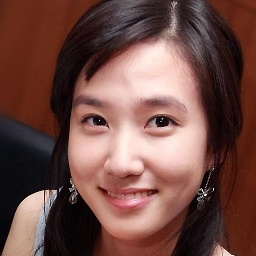} &
    \includegraphics[width=\www]{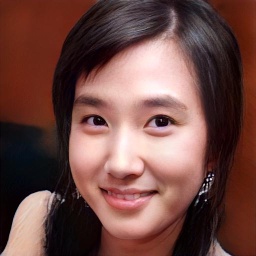} &
    \includegraphics[width=\www]{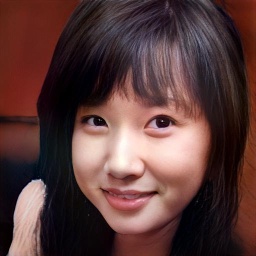} &
    \includegraphics[width=\www]{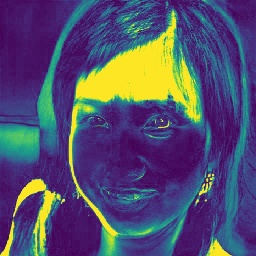} &
    \includegraphics[width=\www]{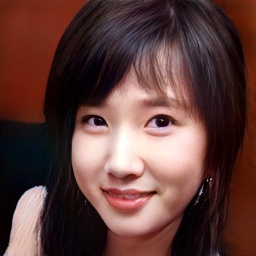} &
    \includegraphics[width=\www]{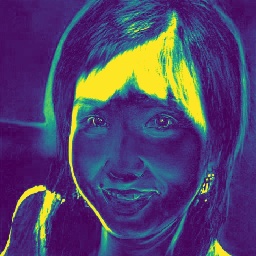} &
    \includegraphics[width=\www]{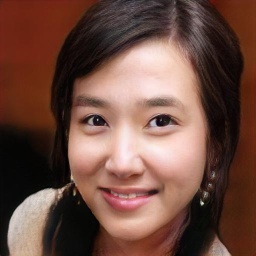} &
    \includegraphics[width=\www]{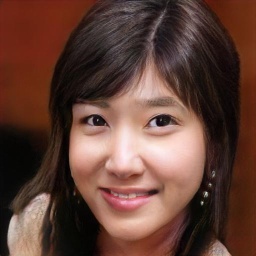} &
    \includegraphics[width=\www]{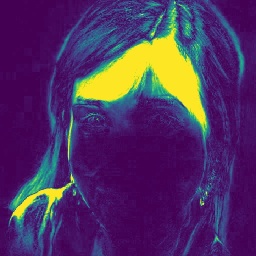} &
    \includegraphics[width=\www]{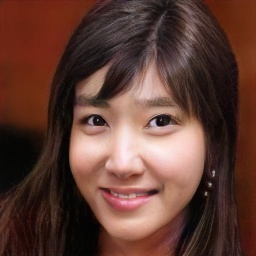} &
    \includegraphics[width=\www]{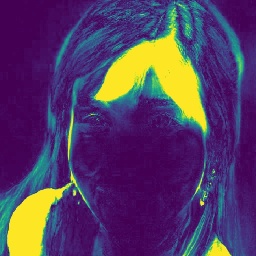} \\
    \includegraphics[width=\www]{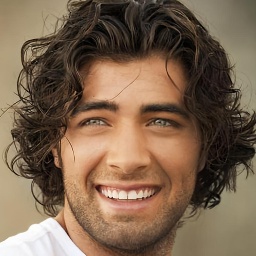} &
    \includegraphics[width=\www]{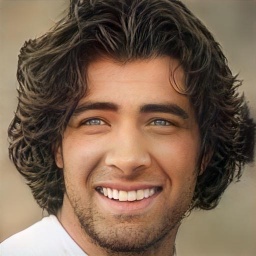} &
    \includegraphics[width=\www]{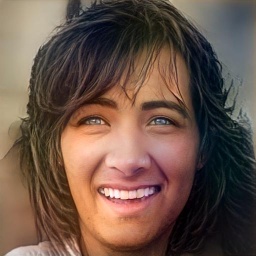} &
    \includegraphics[width=\www]{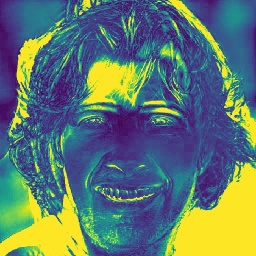} &
    \includegraphics[width=\www]{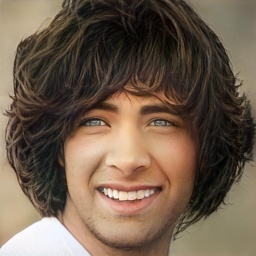} &
    \includegraphics[width=\www]{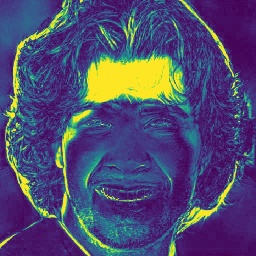} &
    \includegraphics[width=\www]{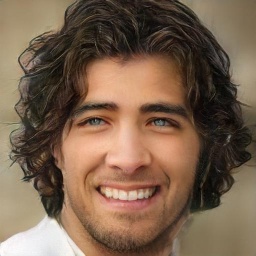} &
    \includegraphics[width=\www]{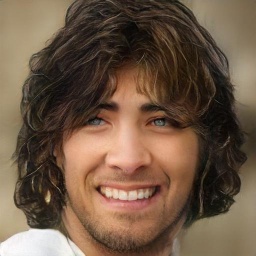} &
    \includegraphics[width=\www]{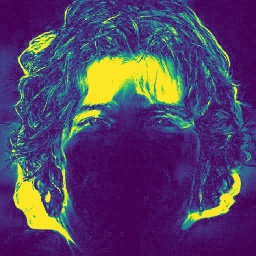} &
    \includegraphics[width=\www]{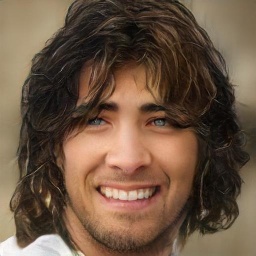} &
    \includegraphics[width=\www]{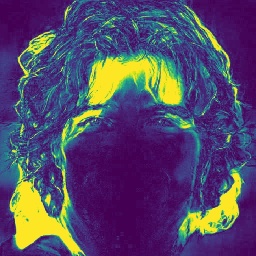} \\
    \includegraphics[width=\www]{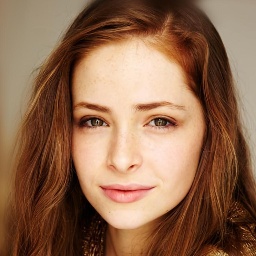} &
    \includegraphics[width=\www]{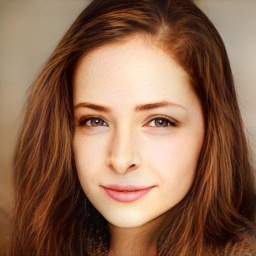} &
    \includegraphics[width=\www]{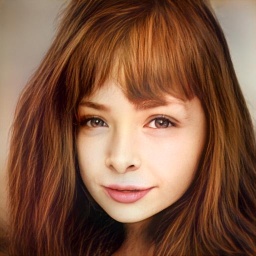} &
    \includegraphics[width=\www]{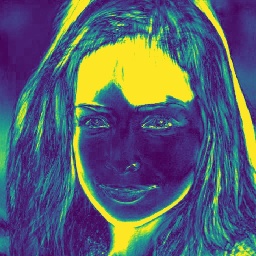} &
    \includegraphics[width=\www]{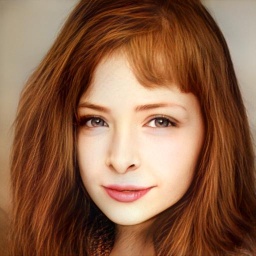} &
    \includegraphics[width=\www]{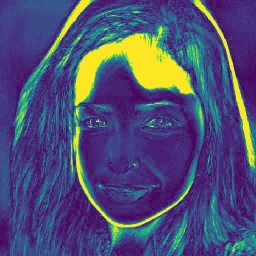} &
    \includegraphics[width=\www]{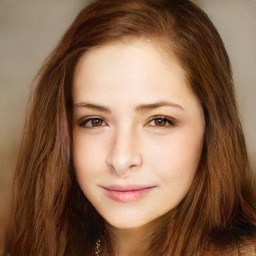} &
    \includegraphics[width=\www]{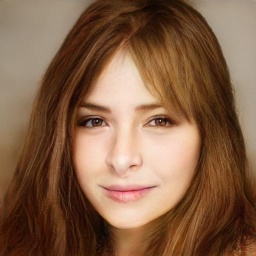} &
    \includegraphics[width=\www]{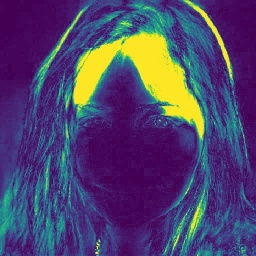} &
    \includegraphics[width=\www]{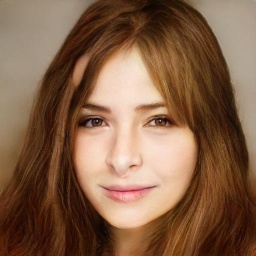} &
    \includegraphics[width=\www]{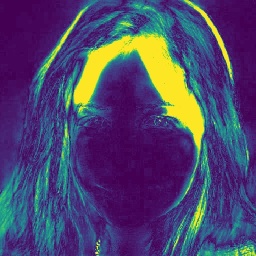} \\
    \includegraphics[width=\www]{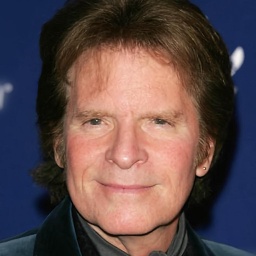} &
    \includegraphics[width=\www]{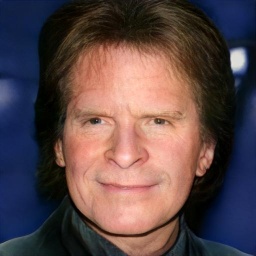} &
    \includegraphics[width=\www]{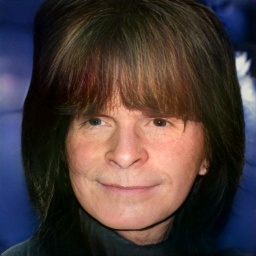} &
    \includegraphics[width=\www]{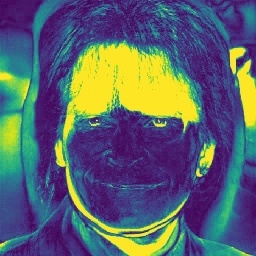} &
    \includegraphics[width=\www]{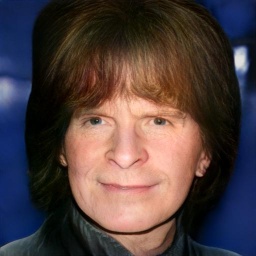} &
    \includegraphics[width=\www]{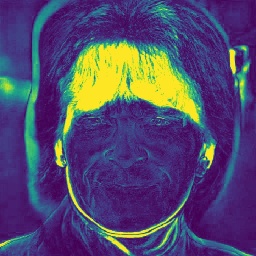} &
    \includegraphics[width=\www]{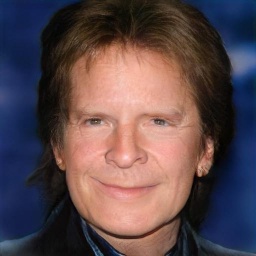} &
    \includegraphics[width=\www]{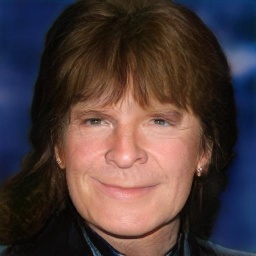} &
    \includegraphics[width=\www]{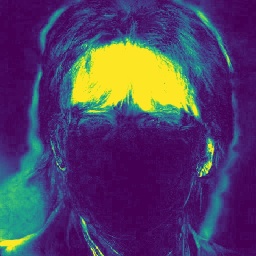} &
    \includegraphics[width=\www]{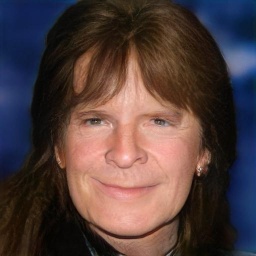} &
    \includegraphics[width=\www]{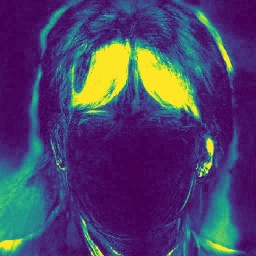} \\
    \multicolumn{1}{c}{Input} & \makecell{Inversion\\(StyleGAN2)} & \multicolumn{2}{c}{StyleFlow} & \multicolumn{2}{c}{InterFaceGAN}  & \makecell{Inversion\\(Ours)} & \multicolumn{2}{c}{StyleFlow+Ours} & \multicolumn{2}{c}{InterFaceGAN+Ours} \\
    
\end{tabularx}
    \vspace{-1.0em}\caption{Results of GAN inversion and editing for the \textbf{bangs} attribute. For each  method, we show the inversion result of Restyle encoder, the edited image and the difference map between them.}\vspace{-0.4em}
    \label{appendix:fig:editing_bangs}\vspace{-1.0em}
\end{figure*}

\begin{figure*}[t]
\captionsetup{font=small}
\centering
\footnotesize
\setlength\tabcolsep{0.0pt}
\newcommand{\www}{0.089\linewidth}
\renewcommand{\arraystretch}{0.4}
\newcolumntype{Y}{>{\centering\arraybackslash}X}
\begin{tabularx}{\linewidth}{c @{\hskip 2pt}|@{\hskip 2pt} c @{\hskip 1.5pt} cc @{\hskip 1.5pt} cc @{\hskip 2pt}|@{\hskip 2pt} c @{\hskip 1.5pt} cc @{\hskip 1.5pt} cc}
    \includegraphics[width=\www]{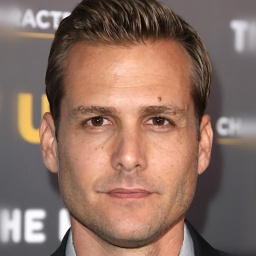} &
    \includegraphics[width=\www]{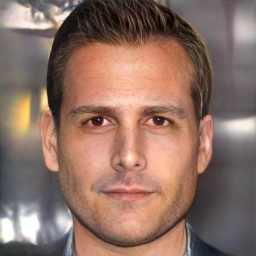} &
    \includegraphics[width=\www]{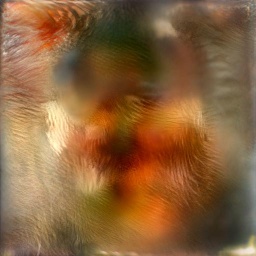} &
    \includegraphics[width=\www]{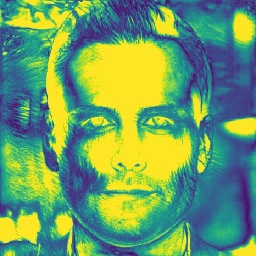} &
    \includegraphics[width=\www]{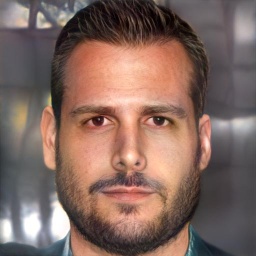} &
    \includegraphics[width=\www]{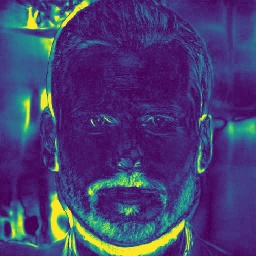} &
    \includegraphics[width=\www]{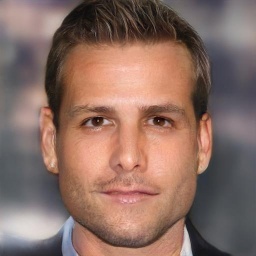} &
    \includegraphics[width=\www]{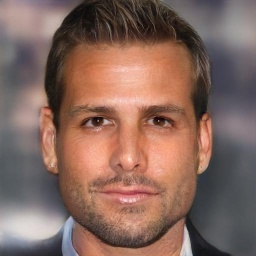} &
    \includegraphics[width=\www]{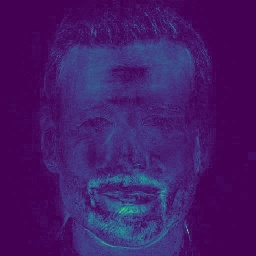} &
    \includegraphics[width=\www]{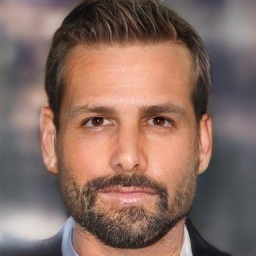} &
    \includegraphics[width=\www]{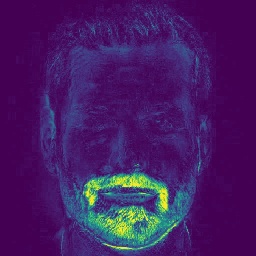} \\
    \includegraphics[width=\www]{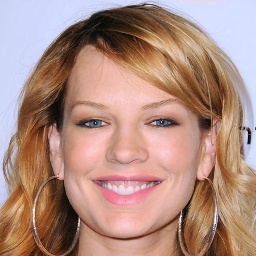} &
    \includegraphics[width=\www]{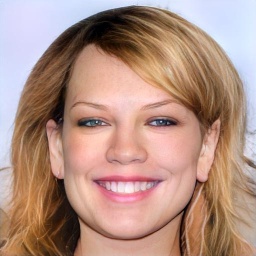} &
    \includegraphics[width=\www]{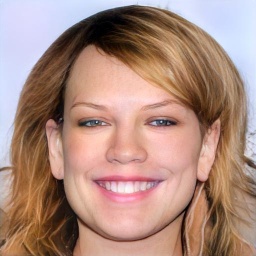} &
    \includegraphics[width=\www]{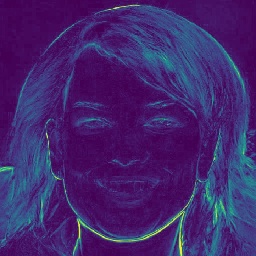} &
    \includegraphics[width=\www]{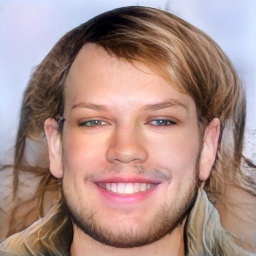} &
    \includegraphics[width=\www]{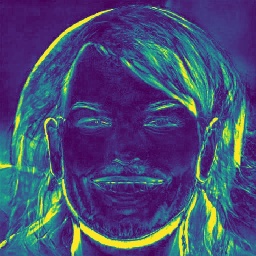} &
    \includegraphics[width=\www]{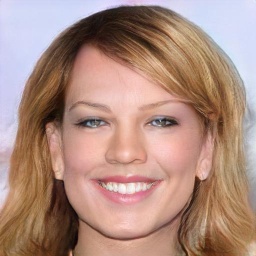} &
    \includegraphics[width=\www]{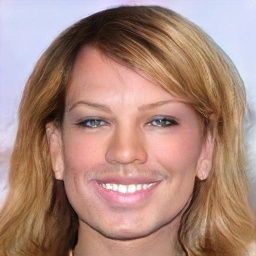} &
    \includegraphics[width=\www]{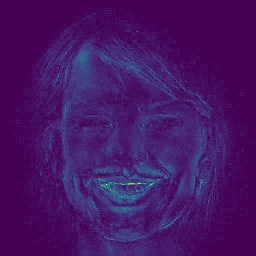} &
    \includegraphics[width=\www]{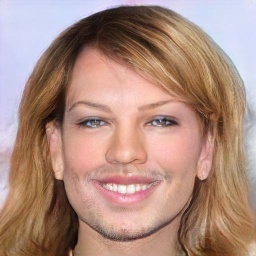} &
    \includegraphics[width=\www]{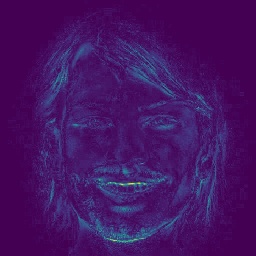} \\
    \includegraphics[width=\www]{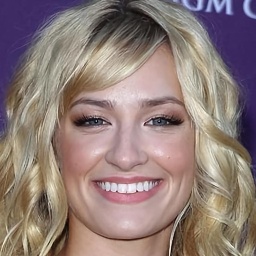} &
    \includegraphics[width=\www]{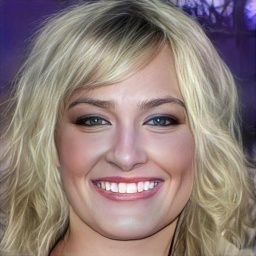} &
    \includegraphics[width=\www]{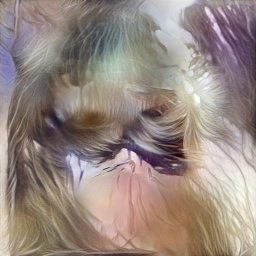} &
    \includegraphics[width=\www]{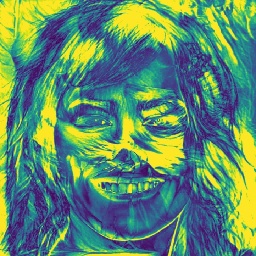} &
    \includegraphics[width=\www]{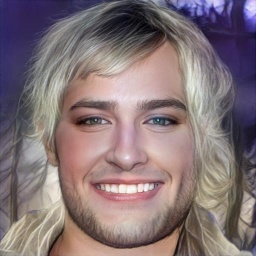} &
    \includegraphics[width=\www]{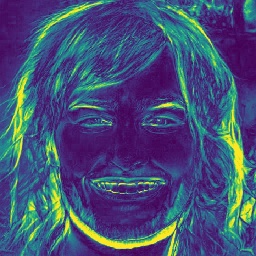} &
    \includegraphics[width=\www]{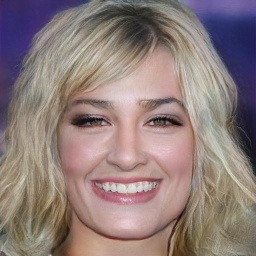} &
    \includegraphics[width=\www]{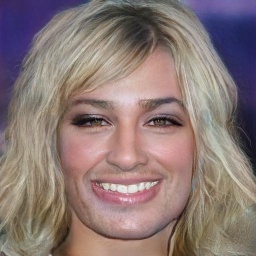} &
    \includegraphics[width=\www]{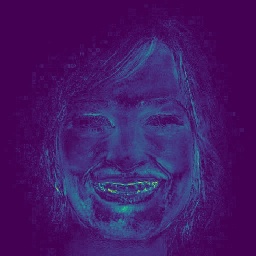} &
    \includegraphics[width=\www]{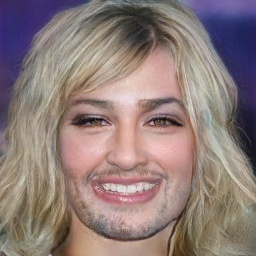} &
    \includegraphics[width=\www]{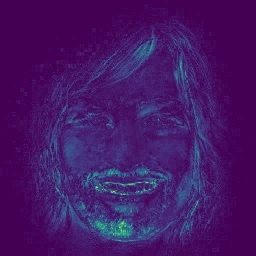} \\
    \includegraphics[width=\www]{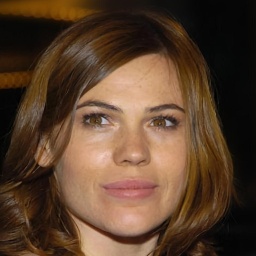} &
    \includegraphics[width=\www]{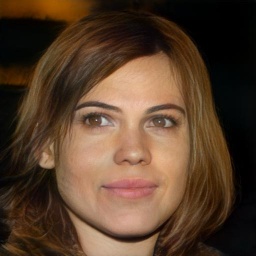} &
    \includegraphics[width=\www]{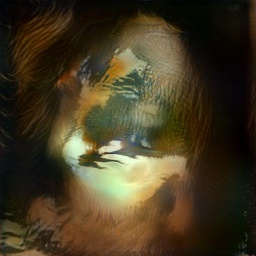} &
    \includegraphics[width=\www]{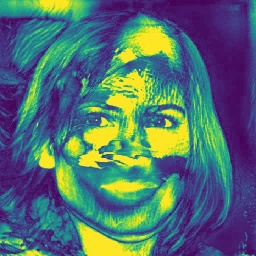} &
    \includegraphics[width=\www]{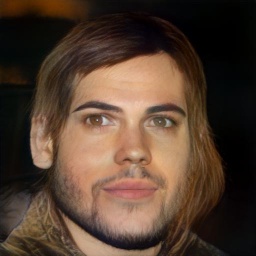} &
    \includegraphics[width=\www]{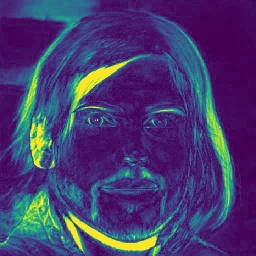} &
    \includegraphics[width=\www]{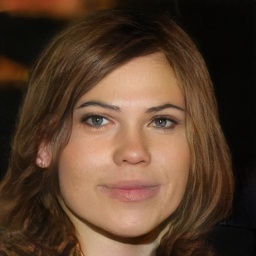} &
    \includegraphics[width=\www]{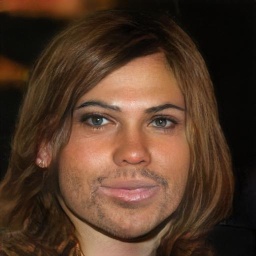} &
    \includegraphics[width=\www]{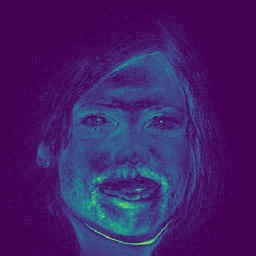} &
    \includegraphics[width=\www]{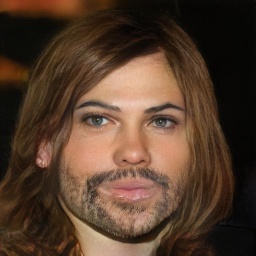} &
    \includegraphics[width=\www]{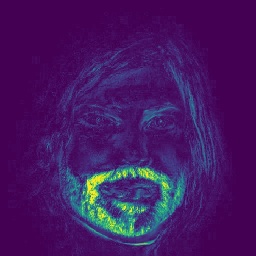} \\
    \includegraphics[width=\www]{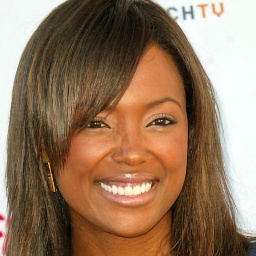} &
    \includegraphics[width=\www]{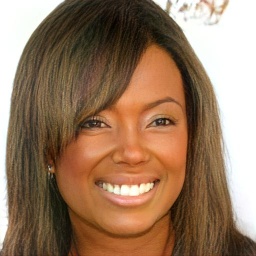} &
    \includegraphics[width=\www]{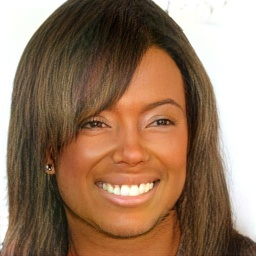} &
    \includegraphics[width=\www]{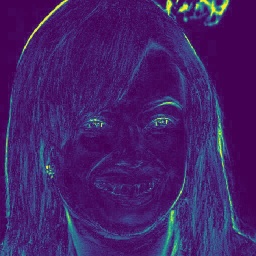} &
    \includegraphics[width=\www]{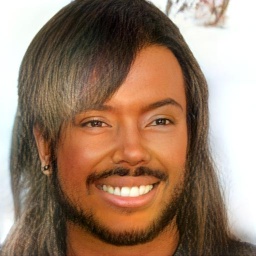} &
    \includegraphics[width=\www]{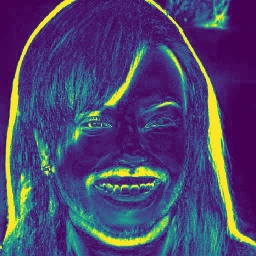} &
    \includegraphics[width=\www]{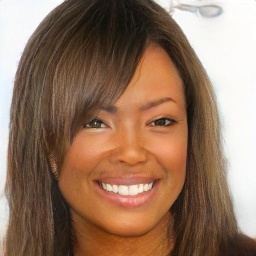} &
    \includegraphics[width=\www]{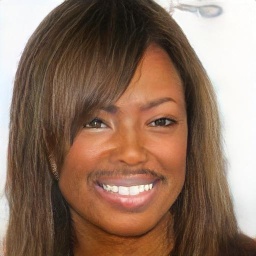} &
    \includegraphics[width=\www]{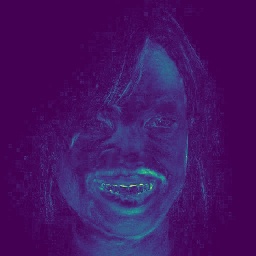} &
    \includegraphics[width=\www]{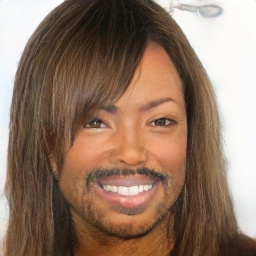} &
    \includegraphics[width=\www]{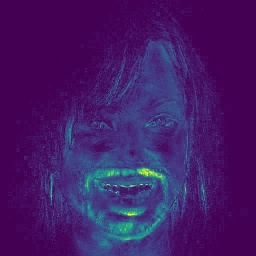} \\
    \includegraphics[width=\www]{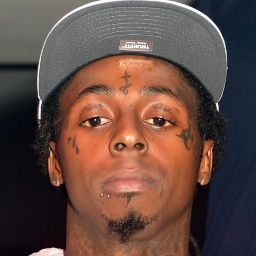} &
    \includegraphics[width=\www]{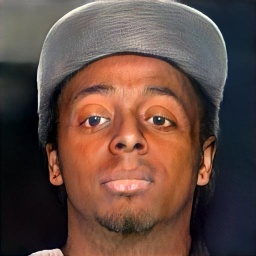} &
    \includegraphics[width=\www]{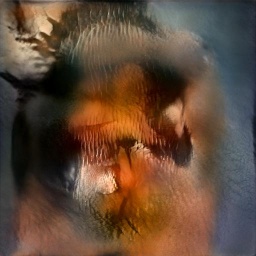} &
    \includegraphics[width=\www]{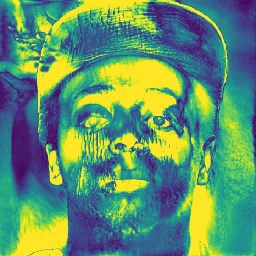} &
    \includegraphics[width=\www]{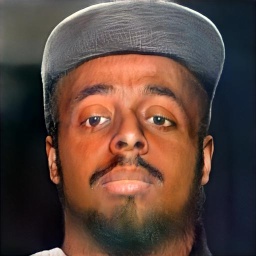} &
    \includegraphics[width=\www]{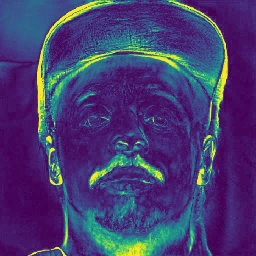} &
    \includegraphics[width=\www]{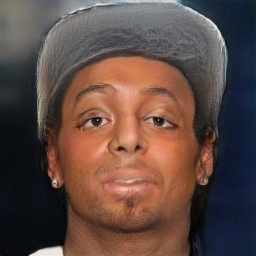} &
    \includegraphics[width=\www]{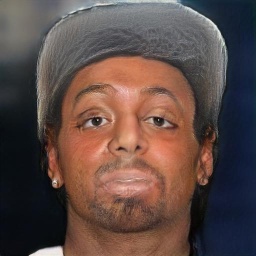} &
    \includegraphics[width=\www]{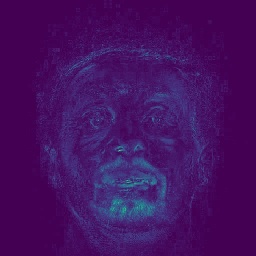} &
    \includegraphics[width=\www]{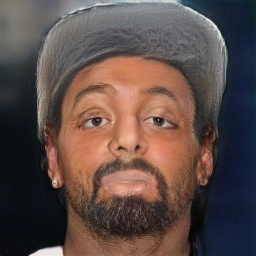} &
    \includegraphics[width=\www]{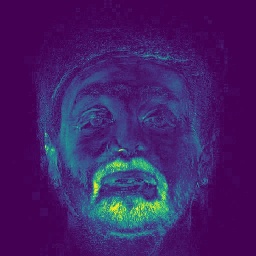} \\
    \multicolumn{1}{c}{Input} & \makecell{Inversion\\(StyleGAN2)} & \multicolumn{2}{c}{StyleFlow} & \multicolumn{2}{c}{InterFaceGAN}  & \makecell{Inversion\\(Ours)} & \multicolumn{2}{c}{StyleFlow+Ours} & \multicolumn{2}{c}{InterFaceGAN+Ours} \\
    
\end{tabularx}
    \vspace{-1.0em}\caption{Results of GAN inversion and editing for the \textbf{beard} attribute. For each  method, we show the inversion result of Restyle encoder, the edited image and the difference map between them.}\vspace{-0.4em}
    \label{appendix:fig:editing_beard}\vspace{-1.0em}
\end{figure*}

\cref{appendix:fig:generation} and  \cref{appendix:fig:components} show more results on randomly sampled images and pseudo-depth maps, respectively \cref{appendix:fig:editing_smile,appendix:fig:editing_bald,appendix:fig:editing_bangs,appendix:fig:editing_beard} show more results on real face editing using our model and original StyleGAN2. As mentioned in the main paper, we see that StyleFlow is more sensitive to data imbalance and less robust. Taking bangs for instance, it tries to reduce the hair on the side but not in the front for our model. For beard, it tries to make face skin look darker for our model while completely fails on the original StyleGAN2. Note that we re-train both StyleFlow and InterFaceGAN using newly sampled images and our own attribute prediction model. Overall, we can observe that our model achieves much more localized control when editing output images.

\end{document}